%% file: main.tex
\documentclass[twoside,11pt]{article}
\usepackage[nohyperref]{jmlr2e}
\usepackage{siunitx}

\usepackage[usenames,dvipsnames,table]{xcolor}
\definecolor{shadecolor}{gray}{0.9}
 \definecolor{mydarkblue}{rgb}{0,0.08,0.45}

\usepackage{microtype}
\usepackage{enumitem}
\usepackage{graphicx}
 \usepackage{subcaption}
\usepackage{multicol}
\usepackage{placeins}
\usepackage[para]{footmisc} 
\usepackage{multirow}

\usepackage{ragged2e}

\usepackage{natbib}
\usepackage[colorlinks,linktoc=all]{hyperref}
\usepackage[all]{hypcap}
\hypersetup{citecolor=mydarkblue}
\hypersetup{linkcolor=mydarkblue}
\hypersetup{urlcolor=mydarkblue}

\usepackage{booktabs}

\input{math_commands}

\def\google{1}
\def\oxford{2}

\jmlrheading{X}{2022}{XX-XX}{05/22}{XX/XX}{XX}{}

\ShortHeadings{\textsc{Plex}: Towards Reliability using Pretrained Large Model Extensions}{}
\firstpageno{1}

\makeatletter
\newcommand{\printfnsymbol}[1]{%
  \textsuperscript{\@fnsymbol{#1}}%
}
\makeatother

\begin{document}

\title{ 
  \textsc{Plex}: {\textbf{Towards Reliability Using\\Pretrained Large Model Extensions}}
 }

\author{%
\centering
 \name Dustin Tran$^{*\google}$, Jeremiah Liu$^\google$, Michael W. Dusenberry$^\google$, Du Phan$^\google$, \\ Mark Collier$^\google$, Jie Ren$^\google$, Kehang Han$^\google$, Zi Wang$^\google$, Zelda Mariet$^\google$, Huiyi Hu$^\google$, Neil Band$^\oxford$, Tim G. J. Rudner$^\oxford$, Karan Singhal$^\google$, Zachary Nado$^\google$, \\ Joost van Amersfoort$^\oxford$, Andreas Kirsch$^\oxford$, Rodolphe Jenatton$^\google$, Nithum Thain$^\google$, Honglin Yuan$^{\google\dagger}$, Kelly Buchanan$^{\google\dagger}$, Kevin Murphy$^\google$, D. Sculley$^\google$, Yarin Gal$^\oxford$, Zoubin Ghahramani$^\google$, Jasper Snoek$^\google$, Balaji Lakshminarayanan$^\google$ \\
 \addr 
$^\google$Google \quad $^\oxford$University of Oxford
 \\
\vspace{-5ex}
}

\editor{}

\maketitle

\begin{abstract}
A recent trend in artificial intelligence is the use of pretrained models for language and vision tasks, which have achieved extraordinary performance but also puzzling failures. Probing these models' abilities in diverse ways is therefore critical to the field. In this paper, we explore the \emph{reliability} of models, where we define a reliable model as one that not only achieves strong predictive performance but also performs well consistently over many decision-making tasks involving uncertainty (e.g., selective prediction, open set recognition), robust generalization (e.g., accuracy and proper scoring rules such as log-likelihood on in- and out-of-distribution datasets), and adaptation (e.g., active learning, few-shot uncertainty). We devise 10 types of tasks over 40 datasets in order to evaluate different aspects of reliability on both vision and language domains. To improve reliability, we developed ViT-Plex and T5-Plex, \emph{p}retrained \emph{l}arge model \emph{ex}tensions (\textsc{plex}) for vision and language modalities, respectively. Plex greatly improves the state-of-the-art across reliability tasks, and simplifies the traditional protocol as it improves the \emph{out-of-the-box} performance and does not require designing scores or tuning the model for each task. We demonstrate scaling effects over model sizes up to 1B parameters and pretraining dataset sizes up to 4B examples. We also demonstrate Plex's capabilities on challenging tasks including zero-shot open set recognition, active learning, and uncertainty in conversational language understanding.%
\blfootnote{%
\hspace{0.5em}$^*$ Corresponding author. Email: \url{trandustin@google.com}.  \qquad
\hspace{0.5em}$^\dagger$ Work done at Google. \\
}%
\footnote{%
Code for training and evaluation is at \url{https://goo.gle/plex-code} as part of Uncertainty Baselines \citep{nado2021uncertainty}. We also release model checkpoints for ViT-Plex and T5-Plex trained on public data.
Layer implementations use Edward2 \citep{tran2018simple}. Author contributions are detailed in \Cref{appendix:contributions}.
}
\end{abstract}


\section{Reliability as a Goal for Artificial Intelligence}
\label{sec:intro}

Over the past few years, the deep learning approach to artificial intelligence (AI) has made significant progress on benchmark tasks across domains such as computer vision \citep{dosovitskiy2020image} and natural language processing \citep{raffel2020exploring,brown2020language}. With this progress, there is unfettered excitement about the potential of AI to have a transformative impact
on applications ranging from medical diagnoses with a human-in-the loop, to using AI to detect online misinformation and toxicity, and to perhaps a pathway for artificial general intelligence.
While hypothesizing about this potential is important, we highlight that the typical tasks where deep learning has been most successful have been carefully devised to fit within narrow boundaries—for example, a focus on predictive performance with test inputs close to the data on which the model was trained.

\begin{figure}[!tb]
\centering
\includegraphics[width=0.95\linewidth]{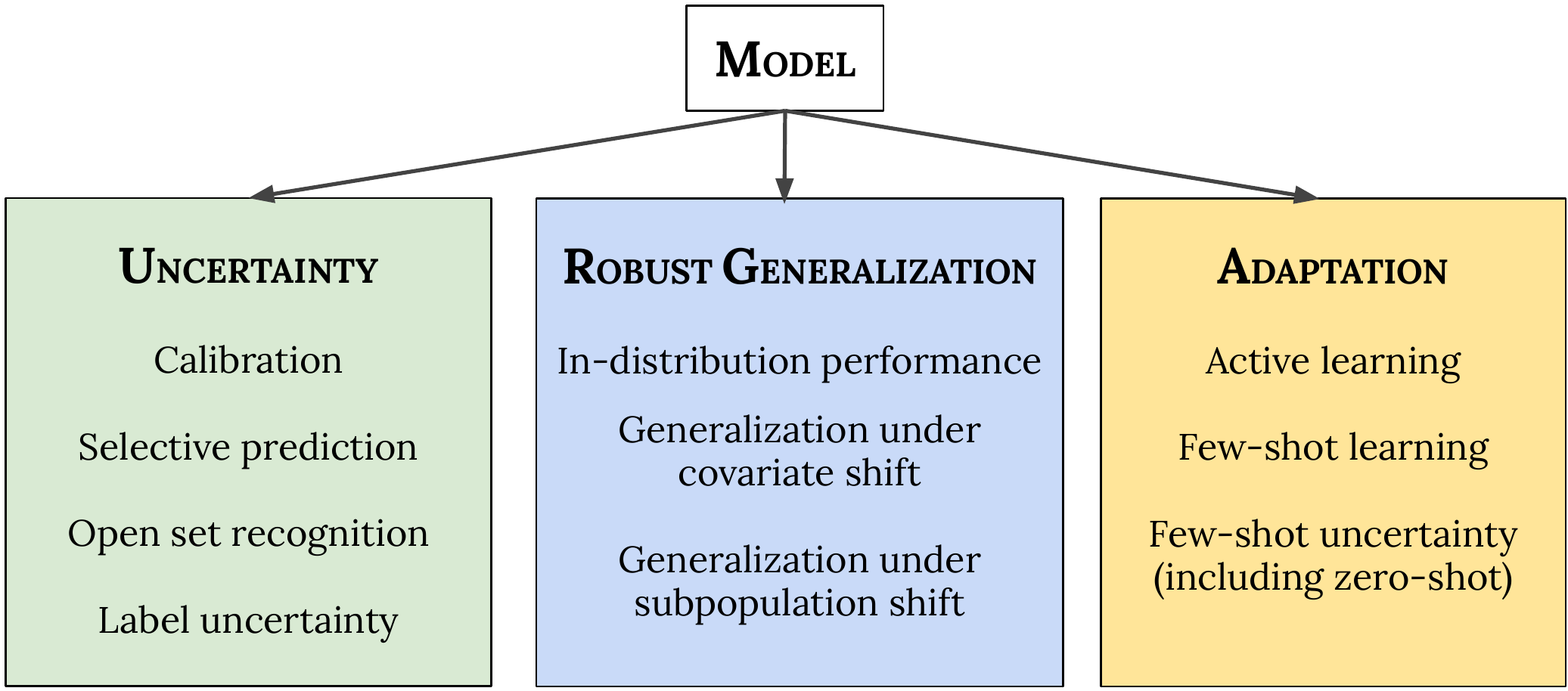}
\caption{%
\emph{Desiderata for a Reliable model}. We propose to simultaneously stress-test the   ``out-of-the-box'' model performance (i.e.~the predictive probability distribution $p(y|x)$) across a suite of uncertainty, robust generalization, and adaptation benchmarks, without any customization for individual tasks. 
}
\label{fig:desiderata}
\end{figure}

To go beyond these limitations, we argue that the ability of models to make \emph{reliable} decisions is critical to the deeper integration of AI in the real world. Here, we define reliability as the ability for a model to work consistently across real-world settings. 
We borrow the term from reliability engineering \citep{barlow1975statistical,o2012practical}, 
a discipline of engineering involving risk assessment, testability, and fault tolerance. Related nomenclature include robustness \citep{russell2015research}, safety \citep{amodei2016concrete,everitt2018agi,hendrycks2021unsolved},
calibration \citep{dawid1982well}, credibility \citep{d2020underspecification}, and trustworthiness \citep{avin2021filling}, each with their own broad and intersecting scopes.  

\subsection{Desiderata for Reliability}
\label{sub:desiderata}
It is common practice in machine learning research to focus on measures of performance based on the accuracy on a test set drawn from the same distribution as the training set, the so-called independent and identically distributed (i.i.d.) assumption. However, this does not capture the real-world deployment of AI systems, where often the testing environment is very different from the training environment, and where the tasks only indirectly involve accuracy measures. The emphasis in our paper is on how reliable an AI system is in broad array of scenarios. We posit three general categories of desiderata for reliable AI systems: they should represent their own uncertainty, they should generalize robustly to new scenarios, and they should be able to efficiently adapt to new data.

Importantly, the aim for a reliable model is 
to do well in \emph{all} of these areas simultaneously out-of-the-box without requiring any customization for individual tasks (\Cref{fig:desiderata}):

\begin{enumerate}[leftmargin=1em,itemsep=0em]
    \item 
\emph{Uncertainty} involves imperfect or unknown information where it is impossible to exactly describe an existing state \citep{ghahramani2015probabilistic}.
Predictive uncertainty quantification enables practitioners to know when to trust the model’s predictions, thereby enabling graceful failures when the model is likely to be wrong.
A variety of measurements can be used to quantify the quality of uncertainty such as expected calibration error \citep{naeini2015obtaining,nixon2019measuring}, which measures how well the model’s confidence
aligns with its accuracy.
Providing a quantification of uncertainty also enables better decision making \citep{parmigiani_decision_2009}; one popular setting is \textit{selective prediction} where a model may defer its prediction to human experts when it is not confident. 
Another popular task is \textit{open set recognition}%
\footnote{%
Earlier work
sometimes referred to this setting as out-of-distribution detection. However, in recent work, the term ``out-of-distribution'' is used as a unifying term for ``non I.I.D'' test data which includes any type of shift, e.g., covariate or subpopulation shift. We use the term open set recognition to denote detection for shifts where test inputs belong to semantic classes not encountered in the training data.
}%
, where the model encounters inputs from new classes at test time that were not seen during training, and the goal is to reliably detect that such inputs do not belong to any of the training classes.
    \item
\emph{Robust Generalization} involves an estimate or forecast about an unseen event \citep{abraham1983statistical,tran-2020-tutorial}. Prediction quality is typically measured using accuracy (e.g., top-1 error for classification problems and mean squared error for regression problems) and \emph{proper scoring rules} such as log likelihood and Brier score \citep{gneiting_strictly_2007}. In the real world, we care not only about metrics on new data obtained from the same distribution the model was trained on (i.i.d.), but also about \textit{robustness}, as measured by metrics on data under out-of-distribution shifts such as covariate or subpopulation shift. 
    \item
\emph{Adaptation} involves probing the model’s abilities over the course of its learning process. Benchmarks typically evaluate on static datasets with pre-defined train-test splits. However, in many applications, we are interested in models that can quickly adapt to new data and efficiently learn with as few labeled examples as possible.  Examples include \emph{few-shot learning} \citep{fei2006one}, where the model learns from a small set of examples; \emph{active learning} \citep{settles2009active}, where the model not only learns but also participates in acquiring the data to learn from; and lifelong learning \citep{thrun1998lifelong}, where the model learns over a sequence of tasks and must not forget about relevant information for previous tasks.
\end{enumerate}

\begin{figure}[t]
\centering
\includegraphics[width=\textwidth]{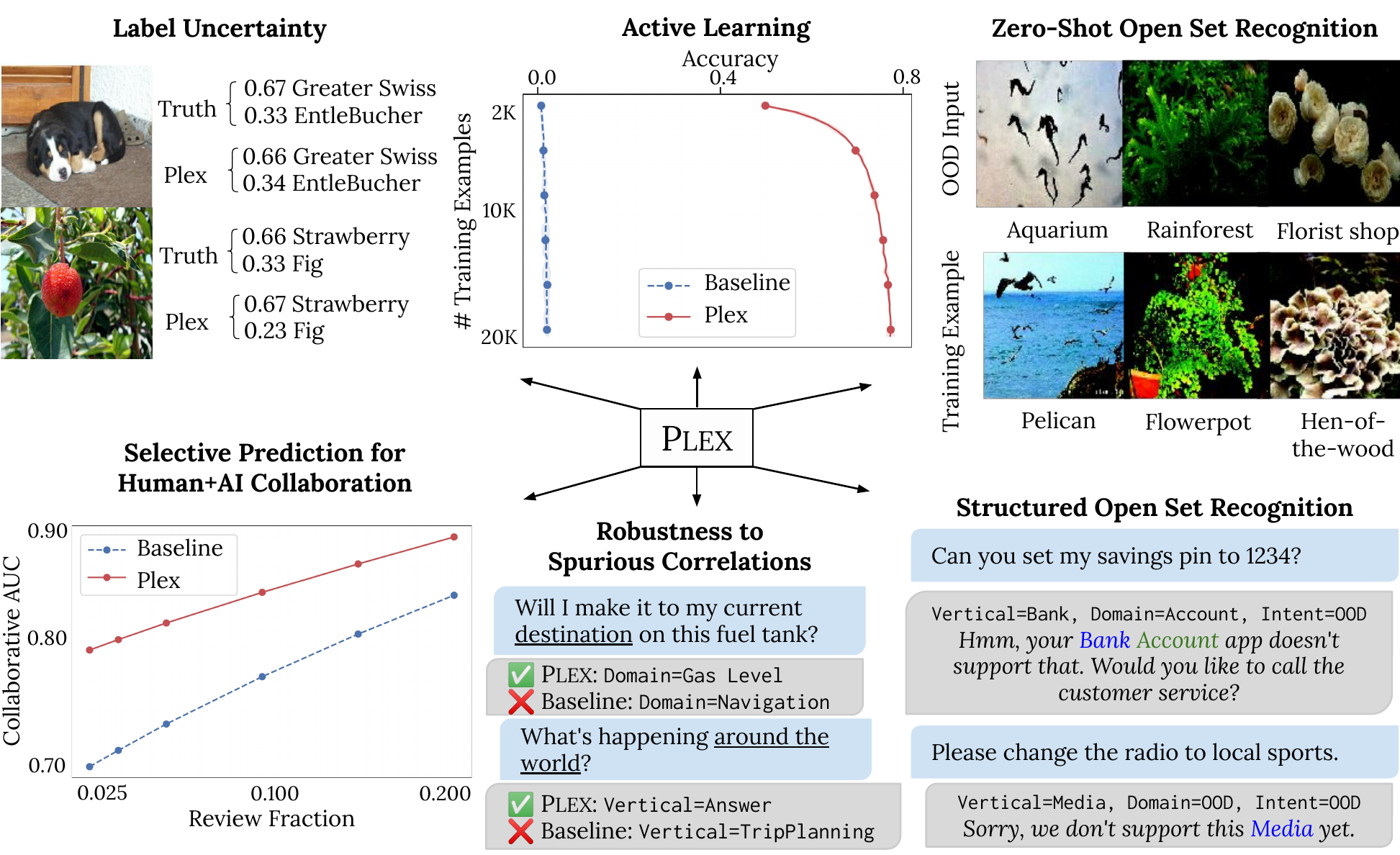}
\caption{%
\textbf{Top row: Examples of Plex's capabilities in vision}:
\textbf{(left)} Label uncertainty in ImageNet ReaL-H, demonstrating the ability to capture the inherent ambiguity of image labels assigned by humans. 
\textbf{(middle)} Active learning on ImageNet1K, displaying 
that Plex achieves significantly higher accuracy
compared to a non-pretrained baseline and with fewer labels.
\textbf{(right)} Zero-shot open set recognition on ImageNet1K vs Places365, showing that Plex can distinguish visually similar images without finetuning. 
\textbf{Bottom row: Examples of Plex's capabilities in language}:
\textbf{(left)} Plex enables human+AI collaboration with selective prediction, where the model is able to defer a fraction of test examples to humans. Plex is able to better identify cases where it is likely to be wrong than the baseline, and thus achieves higher Collaborative AUC. 
\textbf{(middle)} Plex is robust while a baseline latches onto spurious features such as ``destination'' and ``around the world''.
\textbf{(right)} Plex enables structured open set recognition. This provides nuanced clarifications, where Plex can distinguish cases where only part of a request is not supported.
}
\label{fig:plex-capabilities}
\end{figure}

While there has been initial progress on specific tasks within reliability, there remain critical limitations. First, prior work has typically focused on narrow settings. For example, much focus has been on accuracy and calibration error on ImageNet and its corrupted versions as a benchmark task
\citep{hendrycks_benchmarking_2019,ovadia19}.
Similarly, the literature on open set recognition has led to the development of methods that focus solely on improving open set recognition, often at the expense of other downstream tasks.
This has led to a fragmentation of research techniques, and it remains unknown how well advances on these individual tasks connect to a broader set of tasks and datasets. 
The goal of this work is to move towards a notion of \emph{general reliability} by seeking models that work well across many decision-making tasks.
rather than needing to design, train, and tune a model for each individual task. 
The approach is philosophically inspired by the trend in natural language processing where evaluating on multiple  tasks (cf. GLUE \citep{wang2018glue}) has encouraged the community to focus on developing general-purpose techniques.

\begin{figure*}[!tb]
\centering
\includegraphics[width=0.49\linewidth]{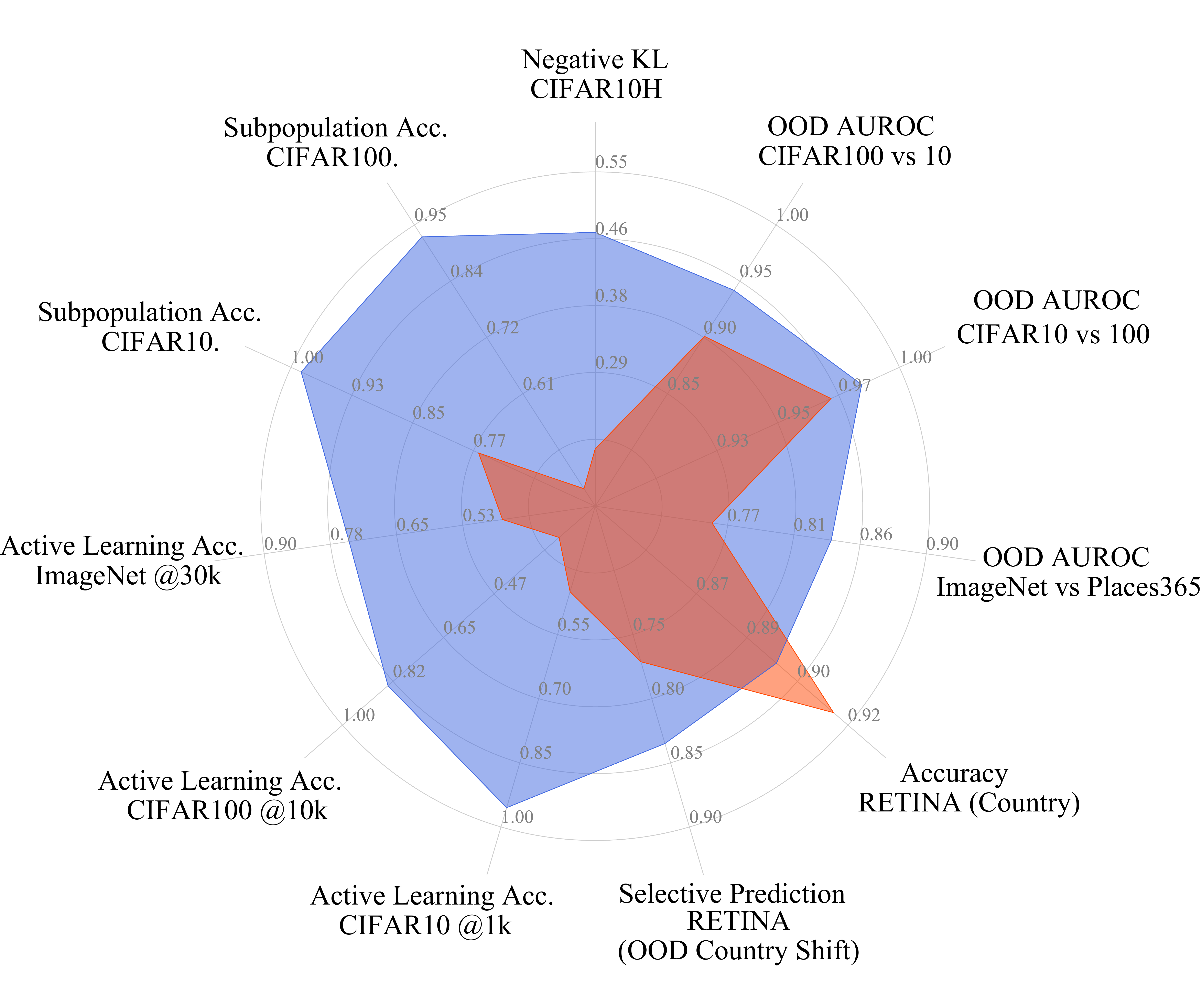}
\includegraphics[width=0.49\linewidth]{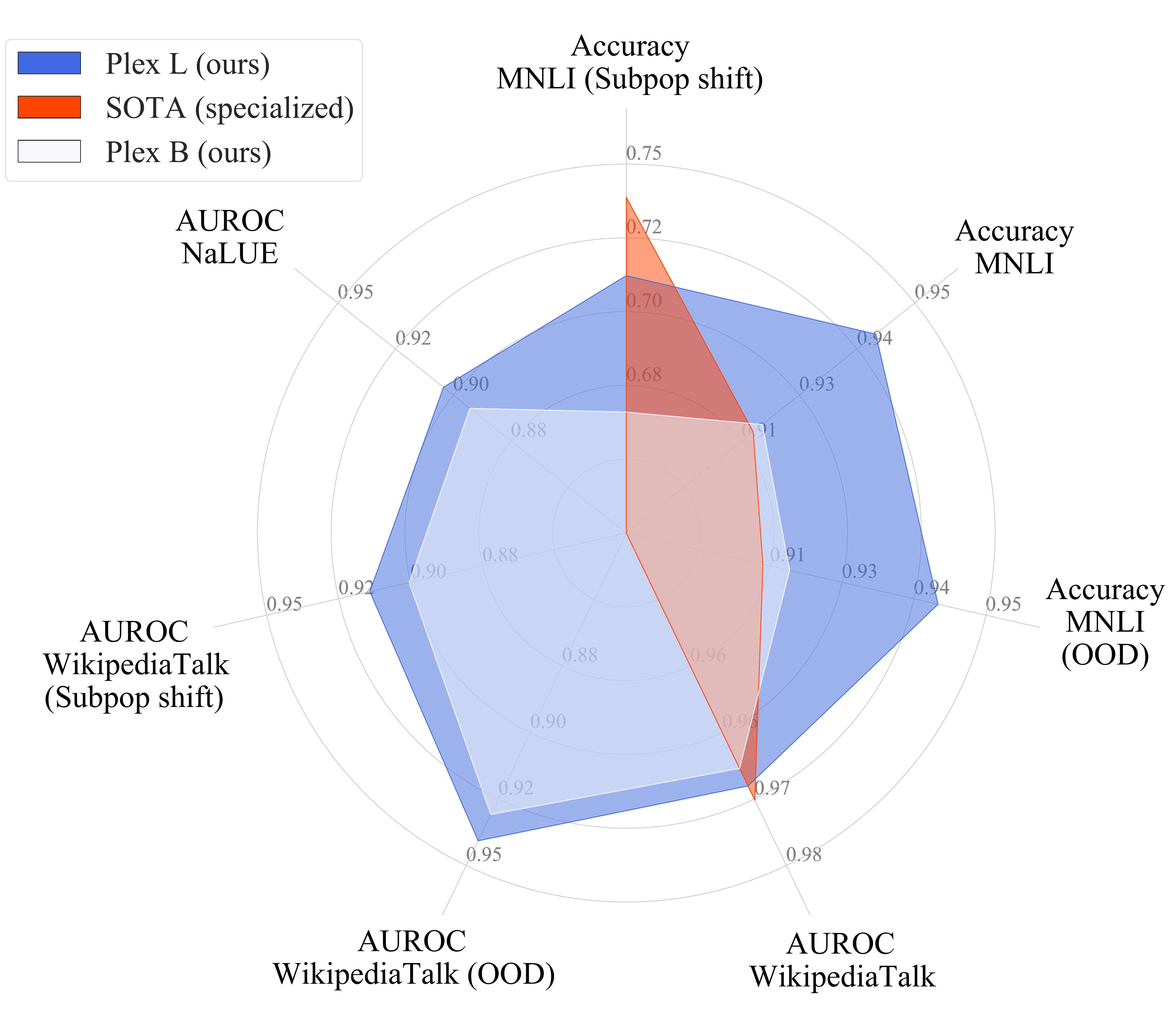}
\caption{%
ViT-Plex (\textbf{left}) and T5-Plex (\textbf{right}) evaluated on a highlighted set of reliability tasks. We also display the state-of-the-art for each task. ViT-Plex and T5-Plex significantly improve state-of-the-art across multiple tasks. Importantly, Plex unifies reliability performance under one general model for vision and language respectively as opposed to specific techniques for each downstream task.
References are listed in \Cref{sec:reliability:score}.}
\label{fig:radar}
\end{figure*}

\subsection{Approaches to Reliability}
Prior work has investigated a variety of approaches to improve narrower definitions of reliability. From the literature, several overarching dimensions arise \citep{tran-2020-tutorial}—such as the importance of model size; model inductive biases (e.g., architecture);
and the combination of multiple models (e.g., ensembles and Bayesian neural networks).
There is not yet an understanding of how these dimensions interact (and within current literature, it can be a surprise that combining certain methods is detrimental, e.g., \citet{wen2020combining}).
We investigate how to combine these dimensions to form an overall model for reliability.

Modern AI is trending towards training a single large model on a large and diverse data set, known as pretraining, and then applying the model to a wide variety of related downstream tasks \citep{brown2020language,kolesnikov2020big,clip,thoppilan2022lamda}. This often improves over task-specific state-of-the-art in predictive performance, with many considering such large scale models to represent a “paradigm shift” in ML \citep{bommasani2021opportunities}. Large-scale pretrained models have also significantly improved state-of-the-art on narrower tasks such as accuracy and calibration under covariate shift \citep{minderer2021revisiting}
and open set recognition \citep{fort2021exploring,ren2021simple}. Given these initial promising results, we use large pretrained models as a building block for reliability.
However, large models are compute-intensive and often operate at a different scale than has been studied in many uncertainty and robustness papers. This warrants revisiting existing recipes in this new context, where we focus on methods that scale efficiently to large models.

\subsection{Contributions}
First, we define and evaluate reliability in a comprehensive fashion. We use 10 types of tasks in order to capture the three reliability areas---uncertainty, robust generalization, and adaptation---and so that the tasks measure a diverse set of desirable properties in each area. Together the tasks comprise 40 downstream datasets across vision and natural language modalities: 14 datasets for finetuning (including few-shot and active learning-based adaptation) and 26 datasets for out-of-distribution evaluation. As part of the tasks, we build two new datasets in order to assess areas that we found missing in the literature: ImageNet Real-H for large-scale label uncertainty evaluation; and NaLUE for uncertainty in conversational language understanding.

To improve reliability, we develop ViT-Plex and T5-Plex, building on large pretrained models for vision (ViT; \citet{dosovitskiy2020image}) and language (T5; \citet{raffel2020exploring}), respectively.
We train Plex over model sizes up to 1 billion parameters and pretraining dataset sizes of up to 4 billion examples. 
\Cref{fig:radar} illustrates Plex’s performance on selected tasks comparing to existing state-of-the-art, which typically is a model specialized for that task. 
Plex achieves new state-of-the-art on many of the 40 datasets. Importantly, Plex achieves strong performance across all tasks using the out-of-the-box model output without requiring any custom designing or tuning for each task.

\section{Evaluating Reliability}

\begin{figure}[!tb]
\centering
\includegraphics[width=\linewidth]{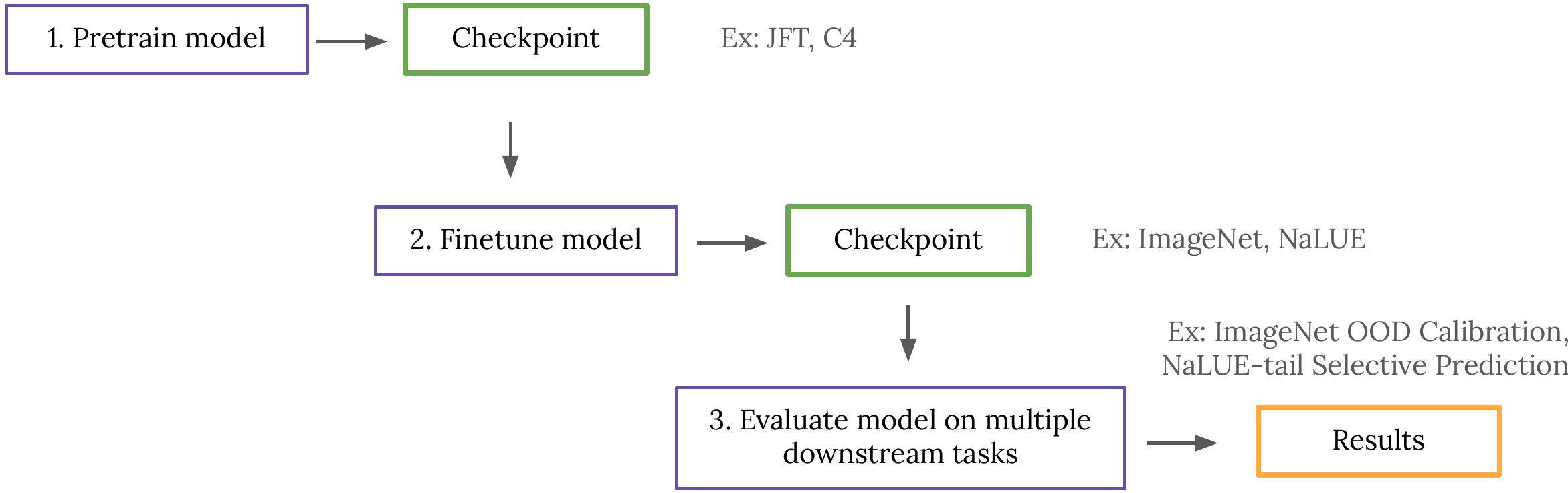}
\vspace{-4ex}
\caption{%
The model and task pipeline. We experiment with approaches for pretraining; given the pretrained model’s checkpoint, we experiment with finetuning and adaptation; finally, given the finetuned model's checkpoint, we evaluate downstream metrics.
}
\label{fig:setup}
\end{figure}

To experiment with a variety of models and tasks, we set up a pipeline for training and evaluation; see \Cref{fig:setup}. It involves three steps: pretrain on a large diverse dataset, finetune on a given dataset’s training split, and evaluate over both in-distribution and out-of-distribution datasets using the appropriate task metrics. 
We experiment with a variety of methods to improve reliability during both pretraining and finetuning.

\subsection{Tasks for Benchmarking Reliability}
\label{sub:tasks}

We evaluate a model’s reliability using 10 types of tasks, which we categorize as follows. More detailed descriptions of metrics in each task and datasets are provided in \Cref{appendix:datasets}.

\begin{figure}[!tb]
\centering
\includegraphics[width=\linewidth]{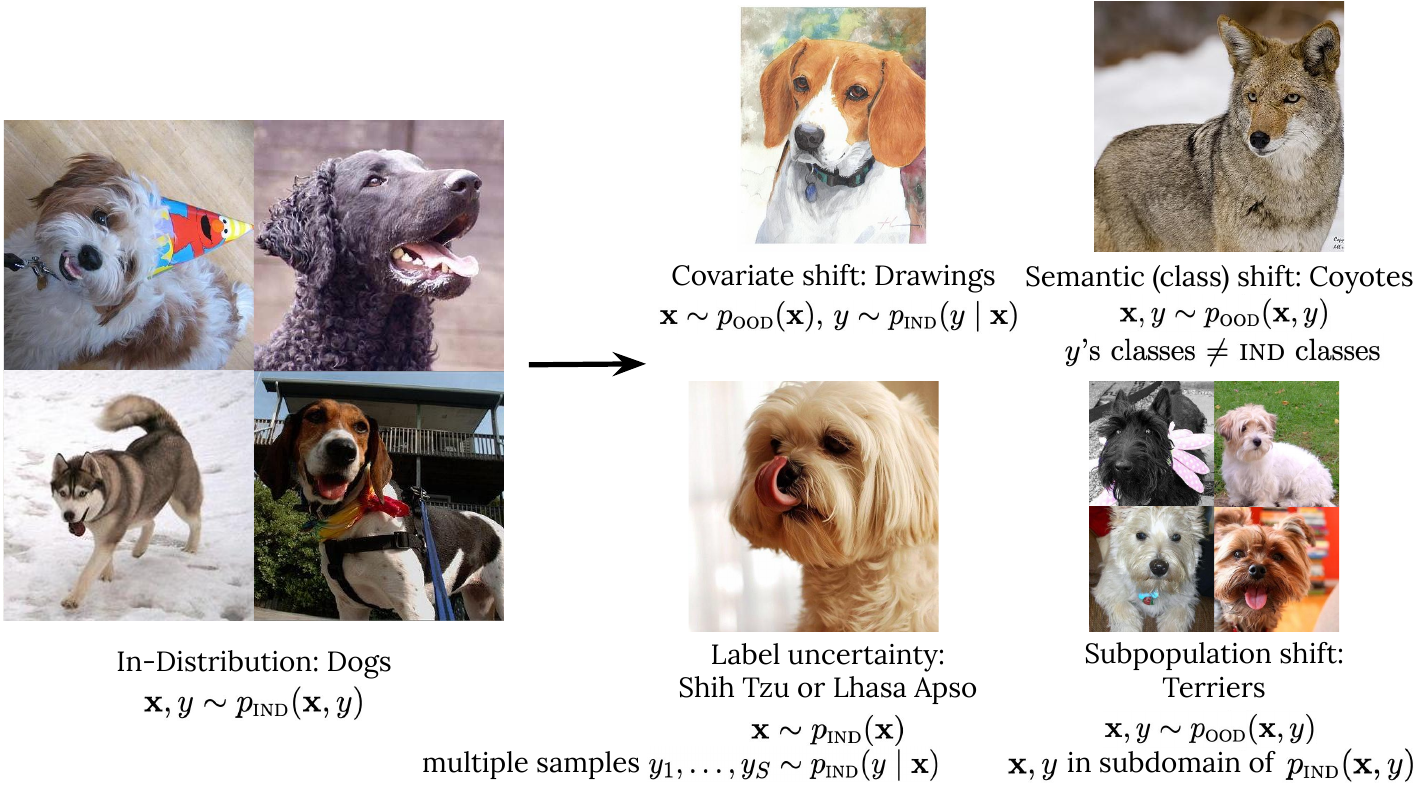}
\caption{%
Types of distribution shift
using an illustration of ImageNet dogs. $p_{\textrm{\textsc{ind}}}(\cdot)$ refers to in-distribution during training and $p_{\textrm{\textsc{ood}}}(\cdot)$ is out-of-distribution during testing.
}
\label{fig:shift}
\end{figure}

\subsubsection{Uncertainty}
\begin{itemize}[leftmargin=1em,itemsep=0em]
    \item 
\textbf{Calibration} assesses the quality of a model’s predicted confidence over a population \citep{dawid1982well}. It quantifies how well model confidence (the predictive probability of correctness) aligns with model accuracy (the observed probability of correctness). We compute expected calibration error \citep{naeini2015obtaining} and calibration AUROC \citep{kivlichan2021measuring}
on 14 image and 10 text datasets.
    \item 
\textbf{Selective prediction} jointly assesses a model's predictive performance and quality of uncertainty estimates, by abstaining from making predictions when the model is uncertain \citep{el2010foundations}.
There is no popular metric that's the de facto standard in the literature, so we experiment with two: 
Oracle Collaborative Accuracy measures accuracy where any deferred predictions are sent to an oracle as a proxy for human$+$AI collaboration \citep{kivlichan2021measuring}; and
selective prediction by rejection rate traces a curve of predictive accuracies over the percentage of examples that can be rejected \citep{band2021benchmarking}.
We examine selective prediction on 4 image and 10 text datasets.
    \item 
\textbf{Open set recognition} assesses how well a model can detect examples belonging to none of the training classes \citep{geng2020recent}. 
This happens in the case of semantic (class) shift, where the test input and label distributions both change, particularly in a structured manner where
the output classes change from train to test.
For an overview of distribution shifts, see \Cref{fig:shift}, where we train on a distribution $p_{\textsc{ind}}(\mathbf{x},y)$ and we evaluate on another distribution.
For example, the training set may consist only of dog images and the new input is a coyote image.
We compute AUROC and use maximum softmax probability as a simple and general detection score.
    \item 
\textbf{Label uncertainty} is a type of uncertainty inherent in the data labels. This is a form of data uncertainty---irreducible output noise---which is distinct from uncertainty arising from the choice of model~\citep{dusenberry2020analyzing}. Label uncertainty can arise when, for example, human raters disagree about the label for an ambiguous input.  If human disagreement is encoded as a label distribution, we can directly compare our model's predictive distribution to it.  We analyze label uncertainty on two image datasets.%
\footnote{We focus on label uncertainty in an OOD evaluation-only context, as collecting label distributions can be expensive and therefore the datasets are small relative to single-label training sets. It is not exclusive to the distribution shift setting as training sets may also carry a label distribution per input.}
\end{itemize}

\subsubsection{Robust Generalization}
 \begin{itemize}[leftmargin=1em,itemsep=0em]
    \item 
\textbf{In-distribution generalization} assesses how well a model can make predictions after finetuning on a downstream dataset. In particular, we examine accuracy, negative log-likelihood, and Brier score on the in-distribution test splits of 5 image and 3 text datasets. For binary classification tasks, we also look at AUROC and AUPRC.
\item
\textbf{Covariate shift} refers to scenarios where the distribution of inputs changes while the conditional distribution of outputs is unchanged \citep{sugiyama2012machine}.
For example, the training set may include natural dog images and the new input is a drawing of a dog.
We use the same metrics as those used for assessing in-distribution generalization.
%
\item
\textbf{Subpopulation shift} refers to scenarios where the distribution of interest is only part of the full distribution seen during training. A common assumption is that data is sampled from individual subpopulation distributions, which are themselves sampled from a meta population distribution \citep{santurkar2020breeds,yuan2022we}. For example, the training set may include natural dog images and the test set only includes terriers.
In this setting, we aim to improve predictive performance on unseen or long-tail subpopulations.
\end{itemize}

\subsubsection{Adaptation}
 \begin{itemize}[leftmargin=1em,itemsep=0em]
    \item 
\textbf{Active learning} assesses a model’s ability to not only learn over a fixed set of data points, but also participate in knowing which data points to learn from in the first place.
This procedure assesses a model’s label efficiency, where label annotations may be scarce, and so we would like to maximize performance while minimizing the number of labeled data points used
\citep{settles2009active}. We assess accuracy over the total number of acquired examples and apply margin sampling for multi-class uncertainty sampling.
    \item 
\textbf{Few-shot learning} assesses how well a model can make predictions downstream with only a few training examples \citep{fei2006one}.
We use 9 datasets and evaluate the settings of 1-shot, 5-shot, 10-shot, and 25-shot ($x$-shot means $x$ examples per class).
    \item 
With \textbf{few-shot uncertainty}, we examine calibration
and open set recognition in the few-shot regime. 
We use all 9 datasets for few-shot learning in order to evaluate calibration,
and we use those with OOD datasets
for open set recognition. We also
perform \emph{zero-shot open set recognition} by using Mahalanobis distance scoring to detect whether an input is out-of-distribution based on the model's representation layer \citep{lee2018simple}.
\end{itemize}

\subsection{Datasets}
\label{ssec:datasets}
We selected a broad suite of 40 downstream datasets under the tasks, each ranging from several thousand to a million examples.
We outline the datasets for each modality.

\subsubsection{Images}
%
We're motivated to capture datasets spanning natural web images, specialized domains that are likely rare or unseen in large pretrained models, and with both small and large sizes.
To do so, we use 11 datasets
which we describe below.

\begin{itemize}[leftmargin=1em,itemsep=0em]
    \item 
CIFAR-10 is a dataset of web images with a training set of 50,000 examples and a test set of 10,000 examples \citep{krizhevsky2009learning}. Following \citet{dosovitskiy2020image}, we use 99\% of the training set for training and 1\% for validation.
\item
CIFAR-100 is a dataset of web images with a training set of 50,000 examples and a test set of 10,000 examples. Following \citet{dosovitskiy2020image}, we use 99\% of the training set for training and 1\% for validation.
\item
ImageNet1K is an image dataset organized according to the WordNet hierarchy, with a training set of roughly 1.2 million examples and a test set of 50,000 examples \citep{deng2009imagenet}. Following \citet{dosovitskiy2020image}, we use 98\% of the training set for training and 2\% for validation.
\item
RETINA is a set of benchmarking datasets containing retina scans exhibiting varying degrees of diabetic retinopathy, a medical condition that can result in a loss of eyesight \citep{band2021benchmarking}.
We chose RETINA as an example of difficult transfer, since retina images are quite different from natural web images used for pretraining.
RETINA includes two types of splits: a ``Country Shift'' split with an in-distribution test set and a test set exhibiting covariate shift;
and a ``Severity Shift'' split with an in-distribution test set and a test set containing labels not included in the training data, representing more severe types of diabetic retinopathy.
\item
We use 7 datasets with a range from 1,880 to 8,144 training examples: Describable Textures Dataset \citep{cimpoi14describing}, UC Merced \citep{Nilsback08}, Caltech 101 \citep{FeiFei2004LearningGV}, Oxford-IIIT Pets \citep{parkhi12a}, Colorectal Histology \citep{kather2016multi}, Caltech-UCSD Birds 200 \citep{welinder2010caltech}, and Cars196 \citep{KrauseStarkDengFei-Fei_3DRR2013}.
\end{itemize}

Distribution shift is a common challenge for image problems, and so we cover multiple types for a total of 19 datasets. \Cref{table:ood-vision-datasets} provides an outline. Most notable, there is little work for evaluating label uncertainty, so we propose a large-scale dataset which we call \emph{ImageNet ReaL-H}.  ImageNet ReaL recollects human ratings for the original ImageNet test set \citep{beyer2020we}, and we use its raw data of individual ratings to construct a label distribution representing rater uncertainty for each image. For ImageNet1K, we use 7 datasets for different covariate shifts: ImageNet-A, ImageNet-C, ImageNet-R, ImageNetV2, ImageNet-Vid-Robust, ObjectNet, and YTBB Robust.

\begin{table*}[tb]
\centering
\begin{tabular}{lllll}
\toprule
& Covariate shift & Semantic shift & Label uncertainty & Subpopl. shift\\
\midrule
CIFAR-10 & CIFAR-10-C & CIFAR-100, SVHN & CIFAR-10H & SP-CIFAR-10\\
CIFAR-100 & CIFAR-100-C & CIFAR-10, SVHN & --- & SP-CIFAR-100\\
ImageNet1K & 7 datasets & Places365 & ImageNet ReaL-H & --- \\
RETINA & Country Shift & Severity Shift & --- & --- \\
\bottomrule
\end{tabular}
\caption{%
Vision datasets for evaluation on distribution shift.
}
\label{table:ood-vision-datasets}
\end{table*}

\subsubsection{Text}
For text, we consider real-world decision making tasks that are known to deploy machine learning models: natural language inference, toxic comments detection, and conversational language understanding.
Natural language inference and toxic comments are binary classification tasks that map a (pair of) natural language sentences to a binary category: entailment or no entailment, and toxic or non-toxic, respectively. Conversational language understanding is a task common in chatbot design, where the model maps a natural language query to a multi-token prediction of user intents: for example, ``I want to order dinner using Uber Eats'' $\to$ 3-token prediction of (FoodDelivery, Uber, Order).

\begin{itemize}[leftmargin=1em,itemsep=0em]
    \item 
For natural language inference, we use the Multi-Genre Natural Language Inference (MNLI) corpus which consists of 433k sentence pairs from a diverse collection of genres (fiction, government report, news magazine articles, etc.) \citep{williams2017broad}. 
    \item 
For toxic comments detection, we use the WikipediaTalk corpus \citep{Wulczyn} which is composed of roughly 200k English Wikipedia talk page comments between Wikipedia editors across the world. 
    \item 
For conversational language understanding, a large-scale corpus for evaluating uncertainty quantification is lacking. We propose a new dataset \textit{Natural Language understanding Uncertainty Evaluation} (NaLUE) that is a relabelled and aggregated version of three large NLU corpuses: CLINC150 \citep{larson_evaluation_2019}, Banks77 \citep{zhang2021pretrained} and HWU64 \citep{liu2021benchmarking}. NaLUE contains 50k+ utterances spanning 18 verticals, 77 domains, and roughly 260 intents. For this task, the model needs to map each utterance to a 3-token sequence of (vertical name, domain name, intent name).
\end{itemize}

In terms of data distribution, MNLI has a balanced distribution both across the genre and across the label class. NaLUE exhibits a slight skewness toward some popular domains for chatbot development (e.g, banking customer service requests). On the other hand, the toxic comments datasets often exhibit extreme label imbalance. For example, $\sim$10\% of the examples in Wikipedia Talk Corpus examples have positive labels, since most online content is not toxic \citep{kivlichan2021measuring}.

\begin{table*}[tb]
\centering
\begin{tabular}{llll}
\toprule
& Covariate shift & Subpopulation shift & Semantic (class) shift \\
\midrule
MNLI & MNLI-mismatched & HANS & --- \\
WikipediaTalk & CivilComments & CivilCommentsIdentity & --- \\
NaLUE & --- & NaLUE-tail & Standard-OOS, Near-OOS \\
\bottomrule
\end{tabular}
\caption{%
Language datasets for evaluation on distribution shift.
}
\label{table:ood-language-datasets}
\end{table*}

Natural language is diverse, fast evolving, and rich in long-tail linguistic phenomena. Therefore out-of-distribution examples, particularly long-tail subpopulations, are pervasive in the real-world deployment environment.
In \Cref{table:ood-language-datasets}, we outline a total of 7 out-of-distribution challenge sets.
Most notably, we construct three new out-of-distribution shifts for NaLUE. \emph{NaLUE-tail} contains utterances from 28 low-frequency intents categories in NaLUE. \emph{NaLUE Standard-OOS} and \emph{NaLUE Near-OOS} contain utterances that describe out of the scope services, differing in their closeness in distribution to NaLUE.

\section{\textsc{Plex}: \emph{P}retrained \emph{L}arge model \emph{Ex}tensions}
\label{sub:plex}


Plex is the result of an extensive study of the reliability of large pretrained models and their complementarity with existing reliability methods. ViT-Plex and T5-Plex use several key ingredients:
\begin{itemize}[leftmargin=1em,itemsep=-0.1em]
    \item 
\textbf{Base Transformer architecture.} We adopt the Transformer standard of an alternating sequence of attention and feedforward layers. We build on T5 1.1 \citep{raffel2020exploring} for text as a Transformer in an encoder-decoder setup where the raw text is tokenized with SentencePiece, and on Vision Transformer \citep{dosovitskiy2020image} for images in an encoder-only setup where the raw images are effectively tokenized into patches.
    \item
\textbf{Model size.} We investigate 3 scales of the model size in ViT-Plex: \textbf{S}mall (ViT-Plex S; $\sim$22 million parameters) has patch size 32, 384 embedding size, 1536 feedforward size, 12 residual blocks, and 6-headed attention; \textbf{B}ase (ViT-Plex B; $\sim$87 million parameters) has patch size 32, 768 embedding size, 3072 feedforward size, 12 residual blocks, and 12-headed attention; and \textbf{L}arge (ViT-Plex L; $\sim$325 million parameters) has patch size 32, 1024 embedding size, 4096 feedforward size, 24 residual blocks, and 16-headed attention.

For T5-Plex, we consider three model sizes: \textbf{S}mall (T5-Plex S; $\sim$77 million parameters) has 512 embedding size, 8 encoder / decoder blocks, and 6-headed attention; \textbf{B}ase (T5-Plex B; $\sim$250 million parameters) has 768 embedding size, 12 encoder / decoder blocks, and 12-headed attention; and \textbf{L}arge (T5-Plex L; $\sim$880 million parameters) has 1024 embedding size, 24 encoder / decoder blocks, and 16-headed attention.
    \item
\textbf{Pretraining dataset size.} For vision, we scale pretraining from ImageNet-21K to the JFT web dataset on up to 4B images.
This mirrors recent work on scaling vision models \citep{zhai2021scaling,pham2021combined}. For language, we use the C4 dataset which consists of hundreds of gigabytes of English text scraped from the web \citep{raffel2020exploring}.
    \item
\textbf{Efficient ensembling.} Ensembles and Bayesian neural nets have shown to be very effective for uncertainty and robustness \citep{ovadia19,dusenberry2020efficient,band2021benchmarking}. To apply ensembling scalably, we use BatchEnsemble (BE) \citep{wen_batchensemble_2020-1} and experiment with its use on both the attention and feedforward layers or on only the feedforward layer. For faster training, we only apply BatchEnsemble at a select number of later layers, similar to mixture of experts models \citep{riquelme2021scaling}. In both ViT-Plex and T5-Plex, we apply no dropout.
    \item
\textbf{Last layer changes.}
We experiment with two approaches that modify the model's final layer to improve reliability, given a
fixed representation (a.k.a.~\emph{deterministic uncertainty quantification} setting \citep{van2020uncertainty}).
First, we use a Gaussian process (GP) last-layer, which improves distance-awareness of the decision surface by increasing uncertainty far away from the training representations. We use the GP layer implementation proposed by \citet{liu2020simple}.
In addition, to model input-dependent label noise in datasets with many output classes we apply the Heteroscedastic (Het) method of \citet{collier2021correlated}.
For detailed background, see \Cref{appendix:ingredients}.
    \item
\textbf{What to apply in pretraining versus finetuning.}
We apply efficient ensembling during both pretraining and finetuning.
For last-layer methods, we find pretraining benefits can be obtained primarily from only applying the method during finetuning, so we restrict them to that setting (detailed in \Cref{sub:effect}).
In addition, due to compute constraints, we exclusively focus on the finetuning-only setting for T5-Plex. That is, T5-Plex models are initialized from the official pretrained T5 checkpoints, and we apply efficient ensembling and last layer changes during finetuning.
    \item
\textbf{Few-shot protocol.} As an alternative to logistic regression on the final layer of frozen representations, we experiment with gradient descent over all parameters; on 5-shot and higher, training the full model
with gradient descent
can lead to significant gains (for example, 2-3\% accuracy boost on ImageNet).
We also experiment with a GP or Heteroscedastic last layer as an alternative to a linear last layer.
\end{itemize}

\section{Summary of Results and Scaling Trends}
\label{sub:trends}

\Cref{fig:radar} displays the largest variants of Plex's performance compared to existing specialized state-of-the-art on a diverse collection of reliability tasks. Plex not only sets new state-of-the-art on many tasks
but Plex also unifies reliability performance under one general model for vision and language respectively. Here, we validate several takeaways as we ablate to understand Plex's ingredients.

\begin{figure}[!tb]
\centering
\includegraphics[width=0.95\linewidth]{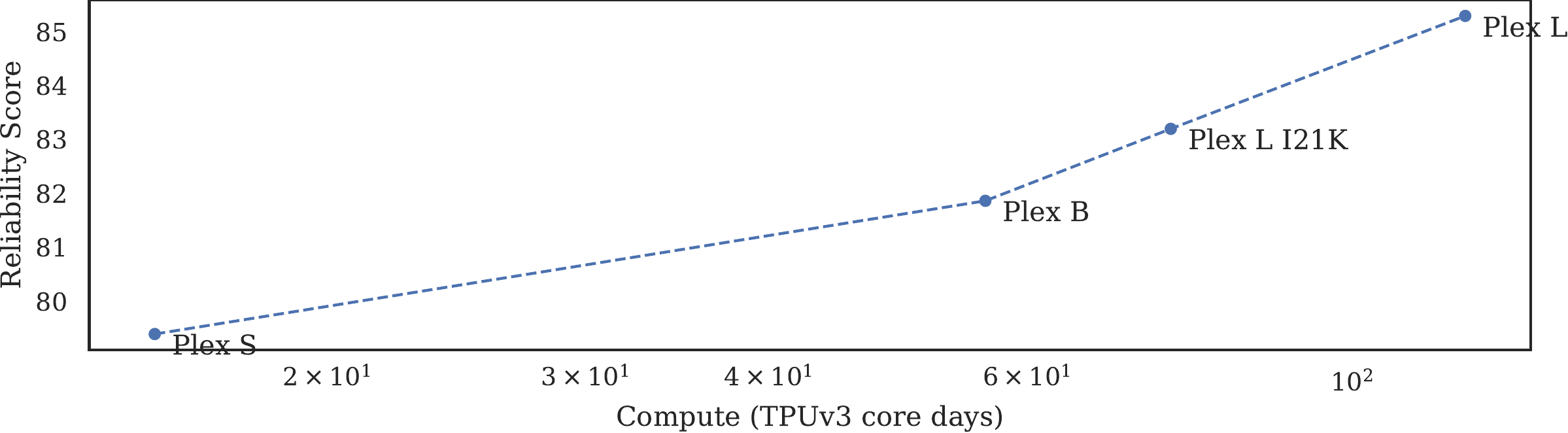} \\[10pt]
\vspace{-2.125ex}
\includegraphics[width=0.95\linewidth]{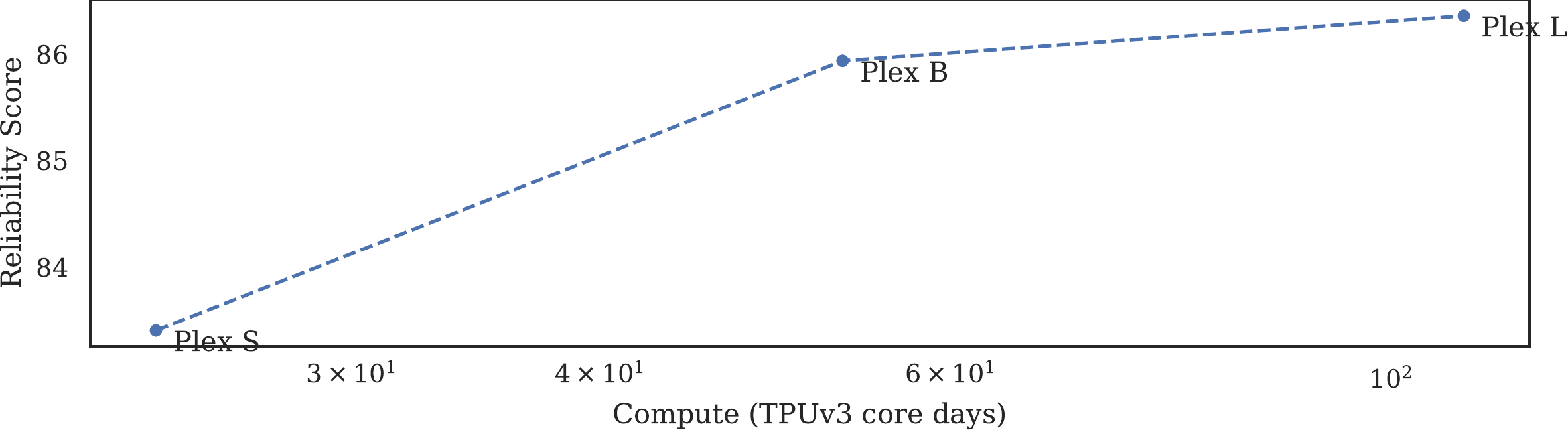} \\[10pt]
\includegraphics[width=0.59\linewidth]{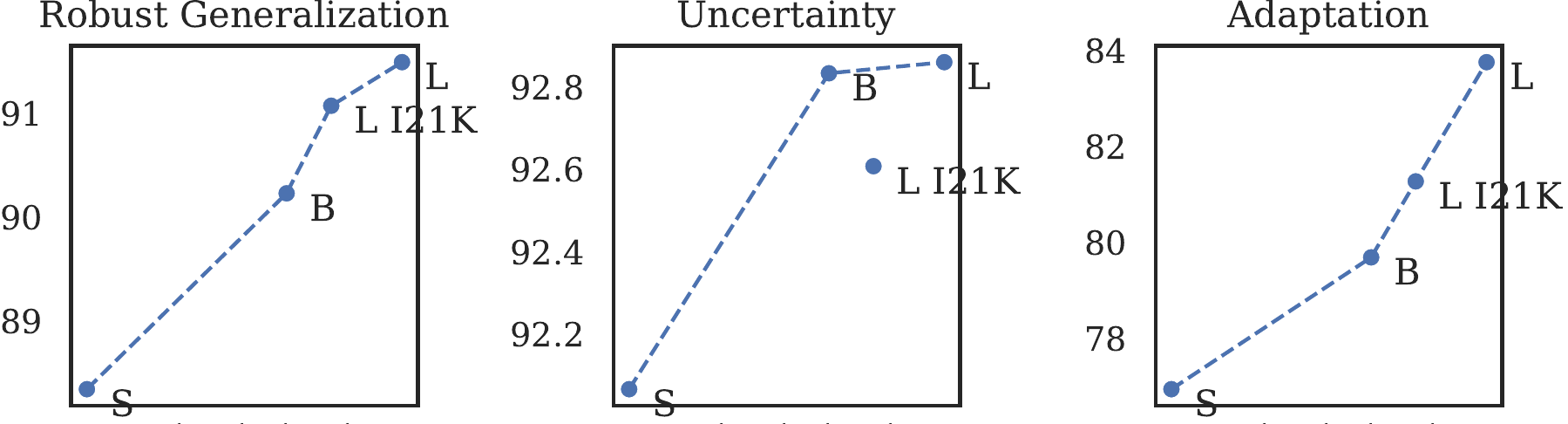}
\includegraphics[width=0.39\linewidth]{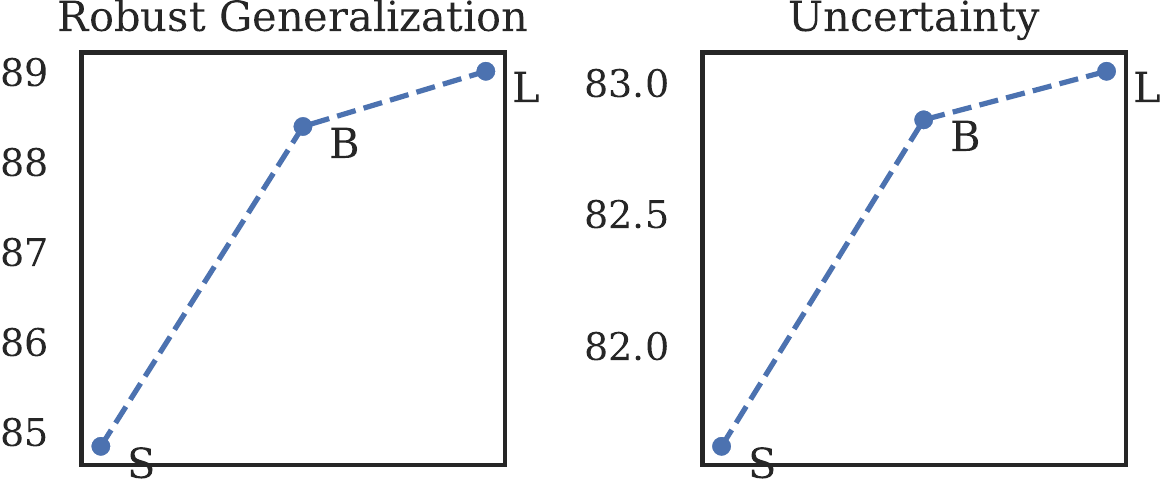} \\
\caption{%
Plex's performance aggregated across \textbf{(top)} 139 vision task metrics and \textbf{(middle)} 54 language task metrics.
Subplots \textbf{(bottom)} display average performance for individual reliability areas for vision (first 3) and language (last 2).
Compute is the total \# of training days for a single TPUv3 core.
Dots represent size (\textbf{S}mall, \textbf{B}ase, \textbf{L}arge), pretrained on JFT for images and C4 for text; I21K denotes pretraining on ImageNet-21K.}
\label{fig:scale}
\end{figure}

\textbf{Scaling model size improves reliability.} 
\Cref{fig:scale} displays ViT-Plex and T5-Plex over varying model sizes.
We compute a reliability score which is a normalized average over all task metrics: 139 for vision and 54 for language (see \Cref{sec:reliability:score} for details).
We also display reliability scores for individual reliability areas, which are the averages separately over the uncertainty, robust generalization, and adaptation tasks.
Classical machine learning theory would suggest that a larger model translates to more overfitting and might therefore be less reliable as it may be overconfident and less robust. However, we find that scale improves overall performance across tasks.
%

\textbf{Scaling pretraining dataset size improves reliability.}
ViT-Plex L with JFT performs better than ViT-Plex L with ImageNet-21K
(\Cref{fig:scale}).
In \Cref{table:pretraining-dataset-size}, we also perform an ablation by comparing pretraining on JFT with 300M examples to JFT with 4B examples.
We pretrain on up to 8M steps with batch size 4096, which is up to 8X more steps than we typically use for pretraining; each result is a separate run using a tuned learning rate schedule.
ImageNet 10-shot accuracy is always better on JFT 4B under the same number of training steps. The models also converge faster with the smaller JFT 300M, reaching a performance limit, whereas JFT 4B keeps improving.

\begin{table*}[!htb]
\centering
\begin{tabular}{lllll}
\toprule
Pretraining Dataset & 1M steps & 2M steps & 4M steps & 8M steps \\
\midrule
JFT 300M  & 72.1\% & 72.9\% & 73.0\% & 73.0\% \\
JFT 4B & \textbf{72.7\%} & \textbf{74.4\%} & \textbf{74.6\%} & --- \\
\bottomrule
\end{tabular}
\caption{%
ImageNet 10-shot accuracy consistently improves with a larger pretraining dataset.
}
\vspace{-2ex}
\label{table:pretraining-dataset-size}
\end{table*}

\begin{figure}[!tb]
\centering
\includegraphics[width=0.49\linewidth]{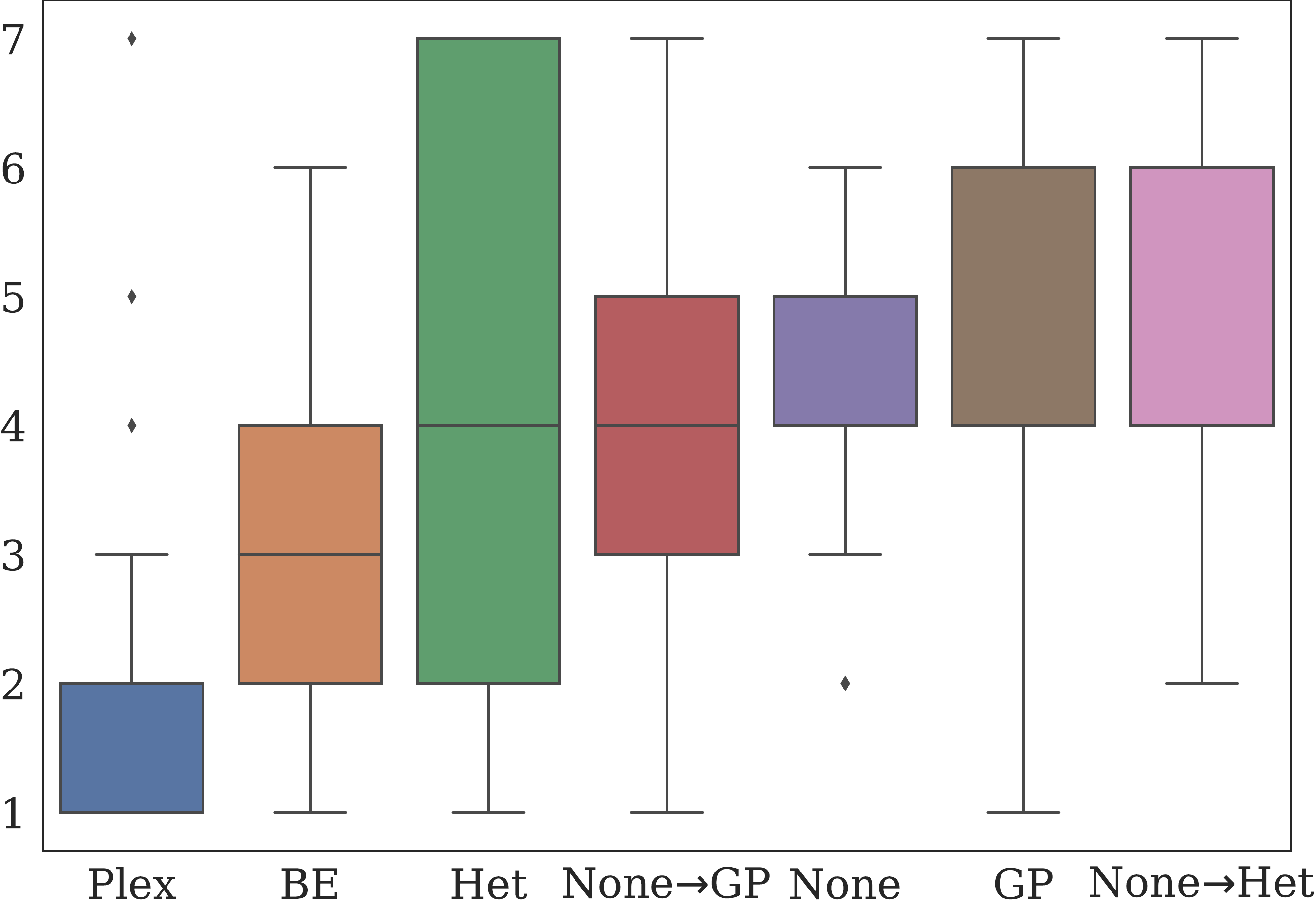}
\includegraphics[width=0.49\linewidth]{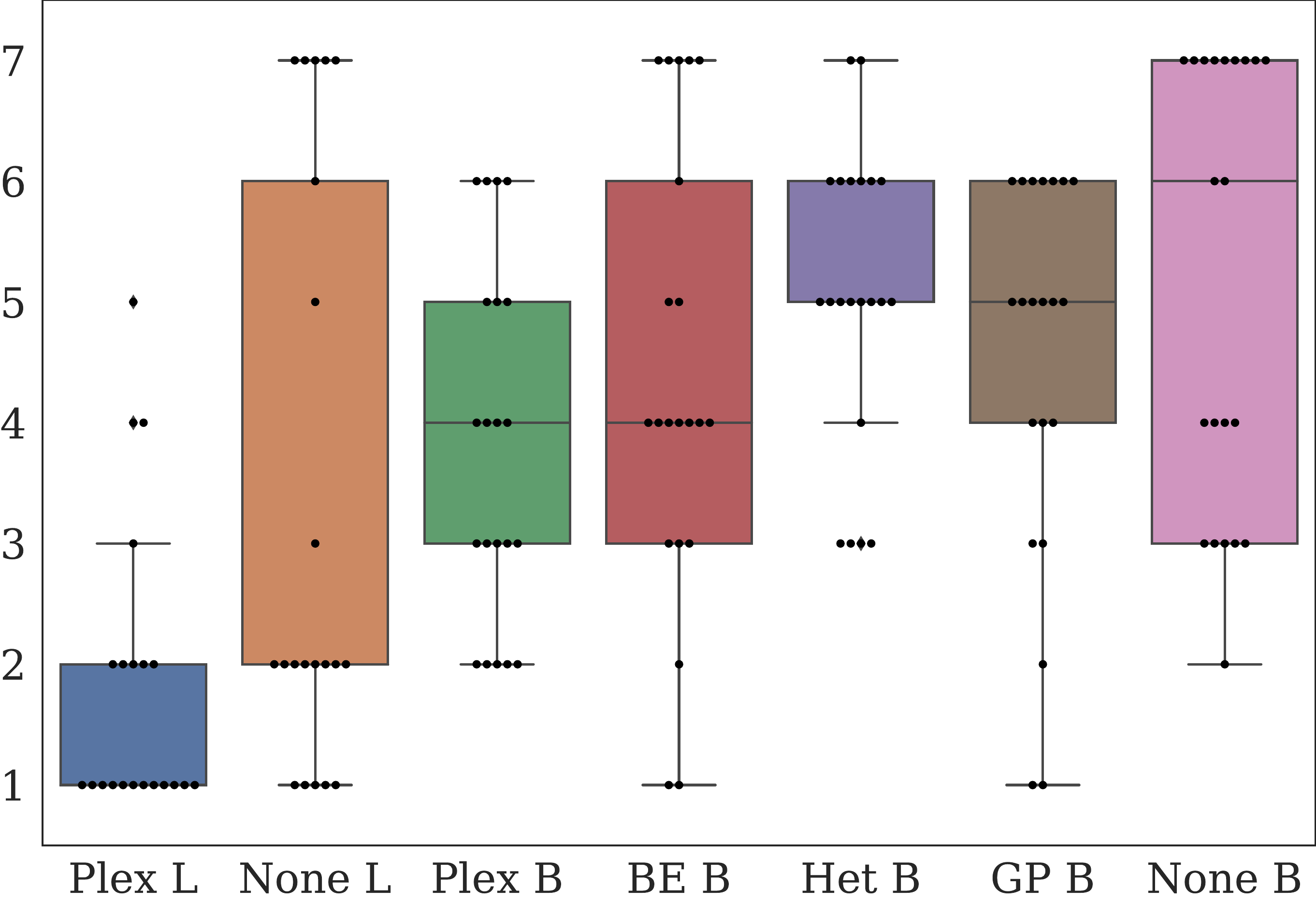}
\caption{%
\textbf{(left)} Ranking of method ablations over 139 metrics on vision tasks and \textbf{(right)} 54 metrics on language tasks. Each model has a box plot of rankings (lower is better).
Plex's combination of efficient ensembling and last-layer changes ranks best on average.
}
\label{fig:ranking}
\end{figure}


\textbf{BatchEnsemble improves pretraining.} 
For vision, we run ablations at the fixed setting of L pretrained with JFT, and we use both B and L sizes for text, which are highly competitive settings. \Cref{fig:ranking} displays the ranking across tasks for each model. Methods are applied either during both pretraining and finetuning, or only during finetuning given a pretrained model checkpoint (``None$\to$Het'' denotes pretraining with no changes and finetuning
with Het on top).
All the methods displayed improve over a baseline without ensembling or last layer changes (``None''). BatchEnsemble is consistently the best for pretraining.

\textbf{Last-layer methods improve finetuning.}
The best ranked models for the vision and language tasks use all of Plex's ingredients: Het on top of a pretrained BE for vision and GP on top of a BE for language.
In particular, for T5-Plex ablations, 
BE$+$GP and BE tend to have the strongest performance.
From more detailed per-dataset analysis in \Cref{appendix:language}, BE$+$GP and BE perform well on MNLI and NaLUE with BE$+$GP performing slightly better; notably, they outperform even an expensive deep ensemble baseline which also performs well on MNLI and NaLUE. BE$+$GP outperforms None on Toxic Comments while a Monte Carlo Dropout baseline performs best on that task. T5-Plex L also outperforms T5-Plex B, which indicates the benefit of scale not only in \Cref{fig:scale}'s normalized average score but in their average ranking.

\begin{figure*}[!tb]
\centering
\includegraphics[width=\linewidth]{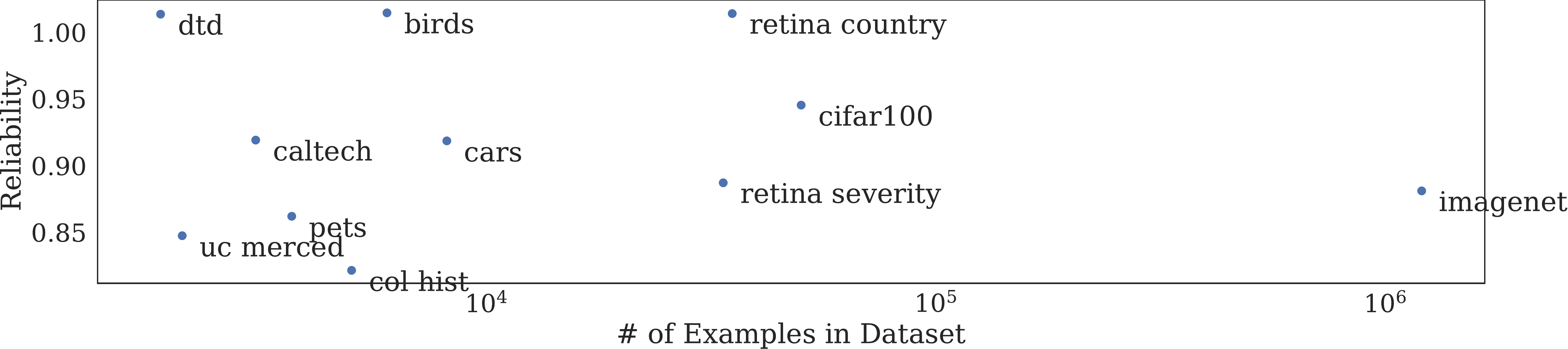}
\caption{%
Reliability performance of ViT-Plex L, aggregated by each downstream dataset.
}
\label{fig:reliablity:datasets}
\end{figure*}

\textbf{Downstream dataset size has no obvious pattern with reliability.} \Cref{fig:reliablity:datasets} decomposes the performance of ViT-Plex L by analyzing it as a function of the downstream dataset’s size of training set. That is, we aggregate reliability performance for each training dataset separately, ranging from a size of 1,880 examples with Describable Textures Dataset (dtd) to 1.2 million examples with ImageNet. There is no clear pattern with respect to size. On the other hand, the datasets with lower reliability tend to be more different from the distribution of natural images in JFT: UC Merced is a remote sensing dataset of map areas, and Colectoral Histology is a histology dataset of human colorectal cancer. Pet images are not uncommon in JFT, but most breeds in Oxford IIT Pets aren't in JFT.
This suggests reliability performance is possibly more connected to the distribution shift between pretraining and downstream dataset than the downstream dataset’s size.

\subsection{Relationship Between Reliability Tasks}

Given that there are many tasks that make up reliability, here we analyze relationships between the individual tasks. We’re motivated by the following question: Is there an inherent underlying evaluation metric that is indicative of reliability? Specifically, can reliability be predicted from pretraining performance, without any downstream finetuning or adaptation?


\begin{figure*}[!tb]
\centering
\includegraphics[width=\linewidth]{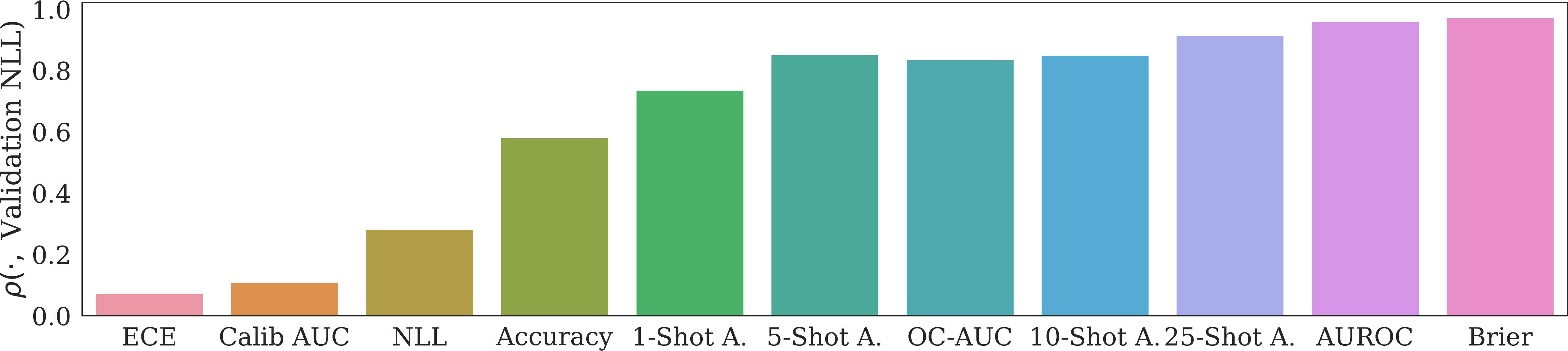}
\caption{%
Correlation of pretraining validation loss with each task (evaluation metric), averaging over all datasets for that task. Most of the tasks correlate highly, meaning upstream performance is a key signal for downstream performance.
}
\label{fig:reliablity:correlation}
\end{figure*}

In  \Cref{fig:reliablity:correlation}, we analyze the correlation of validation loss on the pretraining dataset with each downstream metric. Most of the tasks correlate highly, meaning that the pretraining performance is an important predictor of downstream performance. The one exception are two uncertainty metrics---calibration error and calibration AUROC---which may make sense intuitively given that calibration as a property about models is not tied to predictive performance.
Interestingly, AUROC, which refers to the metric for open set detection (an uncertainty task), is strongly coupled with predictive performance (Pearson correlation of 0.99!), more than, say, selective prediction performance (OC-AUC); this suggests that for open set recognition, prediction may be more important than uncertainty.
Few-shot accuracy also correlates strongly with pretraining performance, and more examples leads to higher correlation ($25 > 10 \approx 5 > 1$-shot).
Accuracy and NLL correlate less strongly than few-shot: this is likely because they are measured on both in- and out-of-distribution datasets whereas few-shot is only evaluated on in-distribution test splits.

Surprisingly, we can find an even stronger answer to the question: training loss on pretraining data is predictive of reliability. \Cref{fig:reliablity:correlation-train} shows Pearson correlation of upwards of 0.97 for the reliability score, prediction, and adaptation areas. Uncertainty is the least correlated but still quite strong at 0.76. This suggests that to perform well for reliability, simply fitting the data---that is, having high model capacity and the ability to efficiently train to utilize that capacity---is one of the most essential ingredients.

\begin{table}[t]
\centering
\begin{tabular}{ccccc}
\toprule
{} &  \textsc{Score} & \textsc{Score robustness} & \textsc{Score uncertainty} & \textsc{Score adaptation} \\
\midrule
JFT & \textbf{-0.977} & \textbf{-0.973} & \textbf{-0.762} & \textbf{-0.982} \\
CIFAR-10 & -0.494 & -0.909 & 0.173 & -0.376 \\
CIFAR-100 & -0.517 & -0.965 & -0.271 & -0.360 \\
ImageNet & -0.566 & -0.957 & 0.038 & 0.440 \\
\bottomrule
\end{tabular}
\caption{Correlation matrix between the training loss on each dataset and the reliability over all vision tasks as well as separately in the three categories. A value of \textbf{-1.0} means a lower training loss has a perfect linear relationship with a higher score. Training loss on the JFT pretraining dataset has a surprisingly high correlation with reliability. Performance on downstream datasets is predictive of overall robustness but otherwise has less correlation.}
\label{fig:reliablity:correlation-train}
\end{table}

\subsection{The Effect of Pretraining vs Finetuning}
\label{sub:effect}

How do the effects we find above differ as model ingredients are applied during the pretraining phase vs finetuning phase?
Here, we analyze the performance of models depending on choices made separately during the two phases.
%
For example, can the benefits of BatchEnsemble be obtained when only applied during finetuning given a model pretrained without any changes?

To broadcast the weights of a pretrained model $U$ into those of a downstream model $D$, we follow the initialization scheme for $D$ and replace any weights common to $U$ and $D$ by those in $U$. 
For example, a BatchEnsemble obtained by broadcasting a pretrained vanilla model will inherit its slow weights from the model, but fast weights will be initialized randomly.
Consistent with all finetuning experiments, the head layer is also reinitialized rather than inherited from the pretrained model.

\begin{figure}[ht]
\centering
\includegraphics[width=\textwidth]{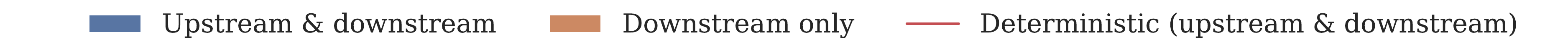} 

\begin{subfigure}{.32\textwidth}
\centering
\includegraphics[width=\textwidth]{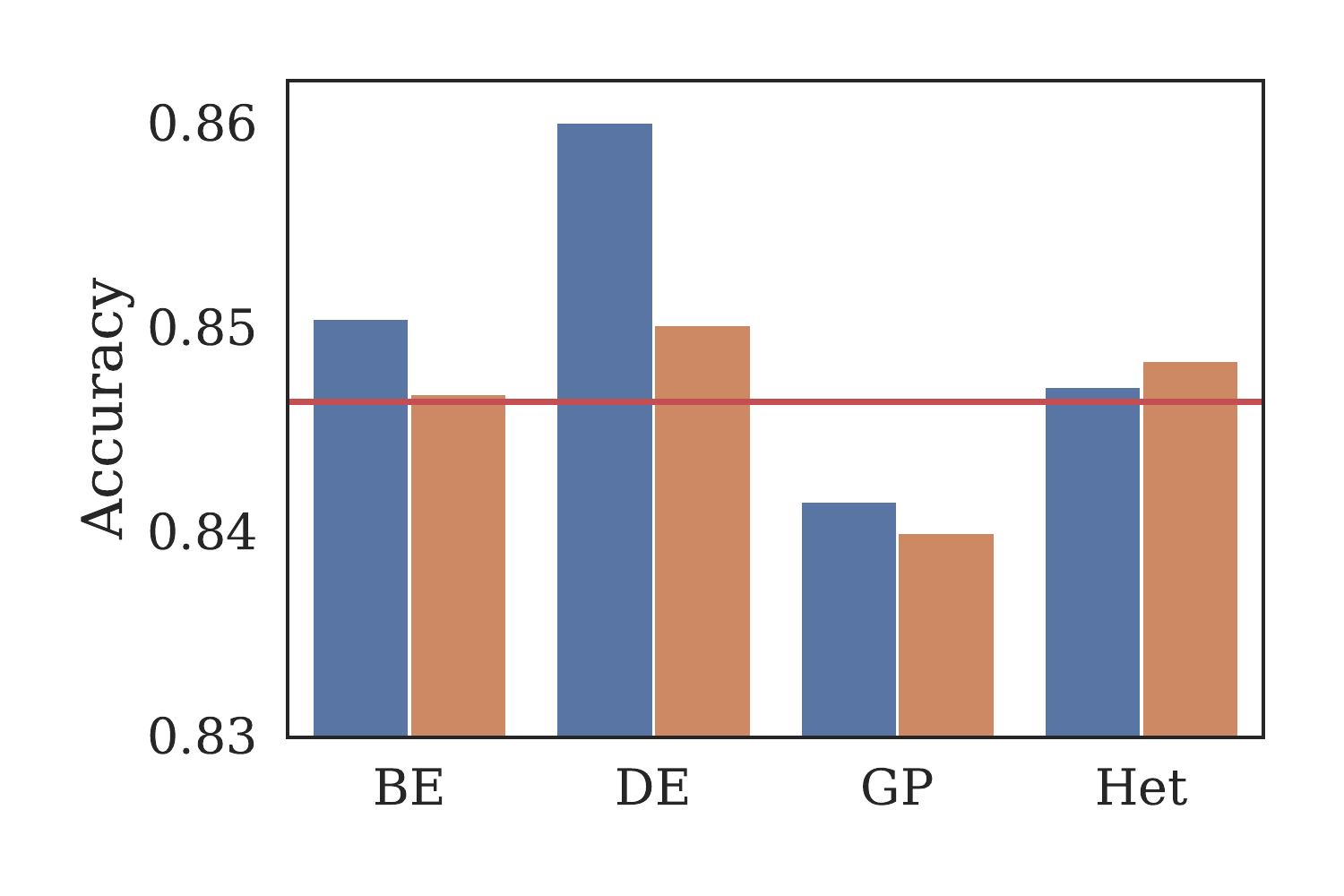}
\end{subfigure}
\begin{subfigure}{.32\textwidth}
\centering
\includegraphics[width=\textwidth]{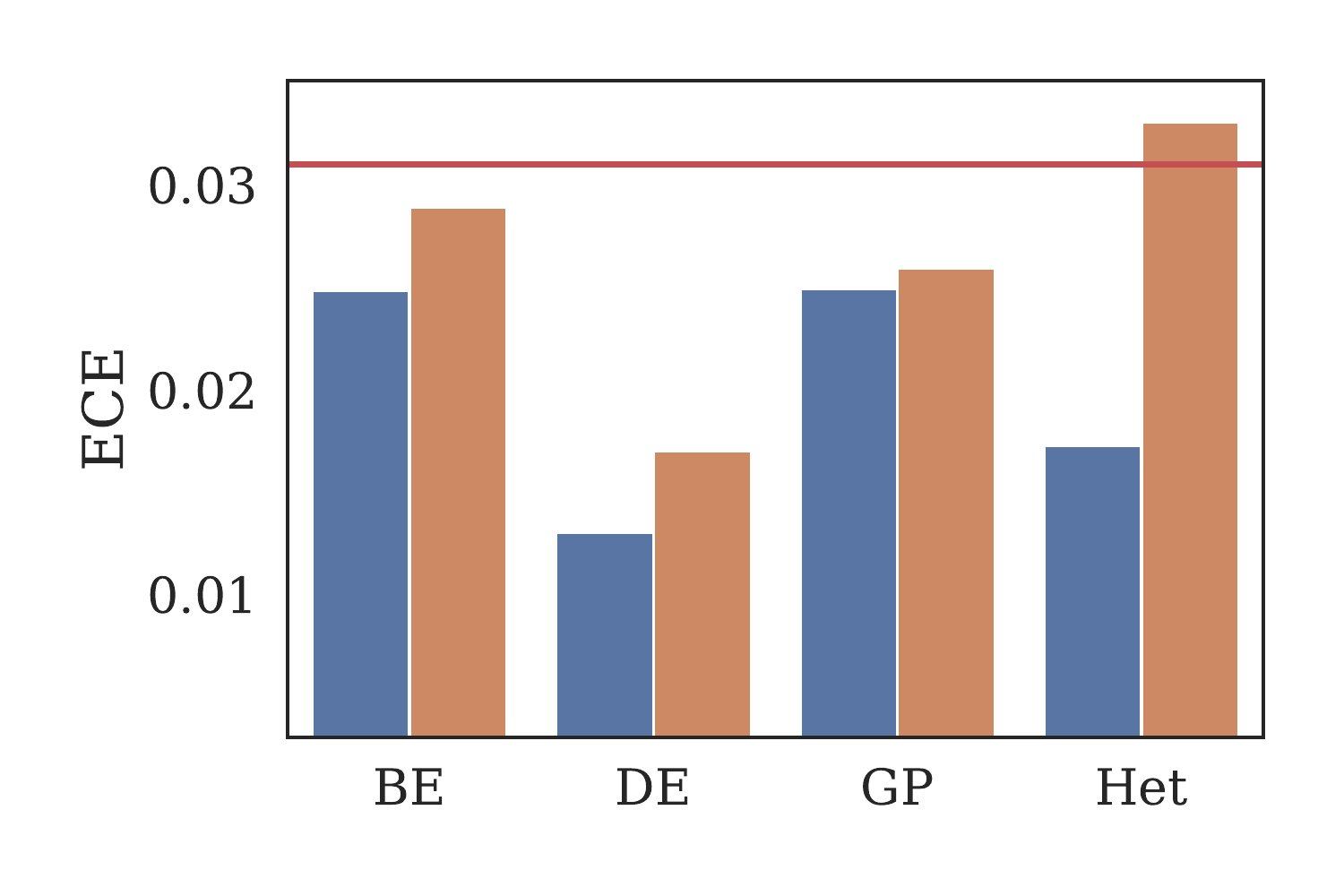}
\end{subfigure}
\begin{subfigure}{.32\textwidth}
\centering
\includegraphics[width=\textwidth]{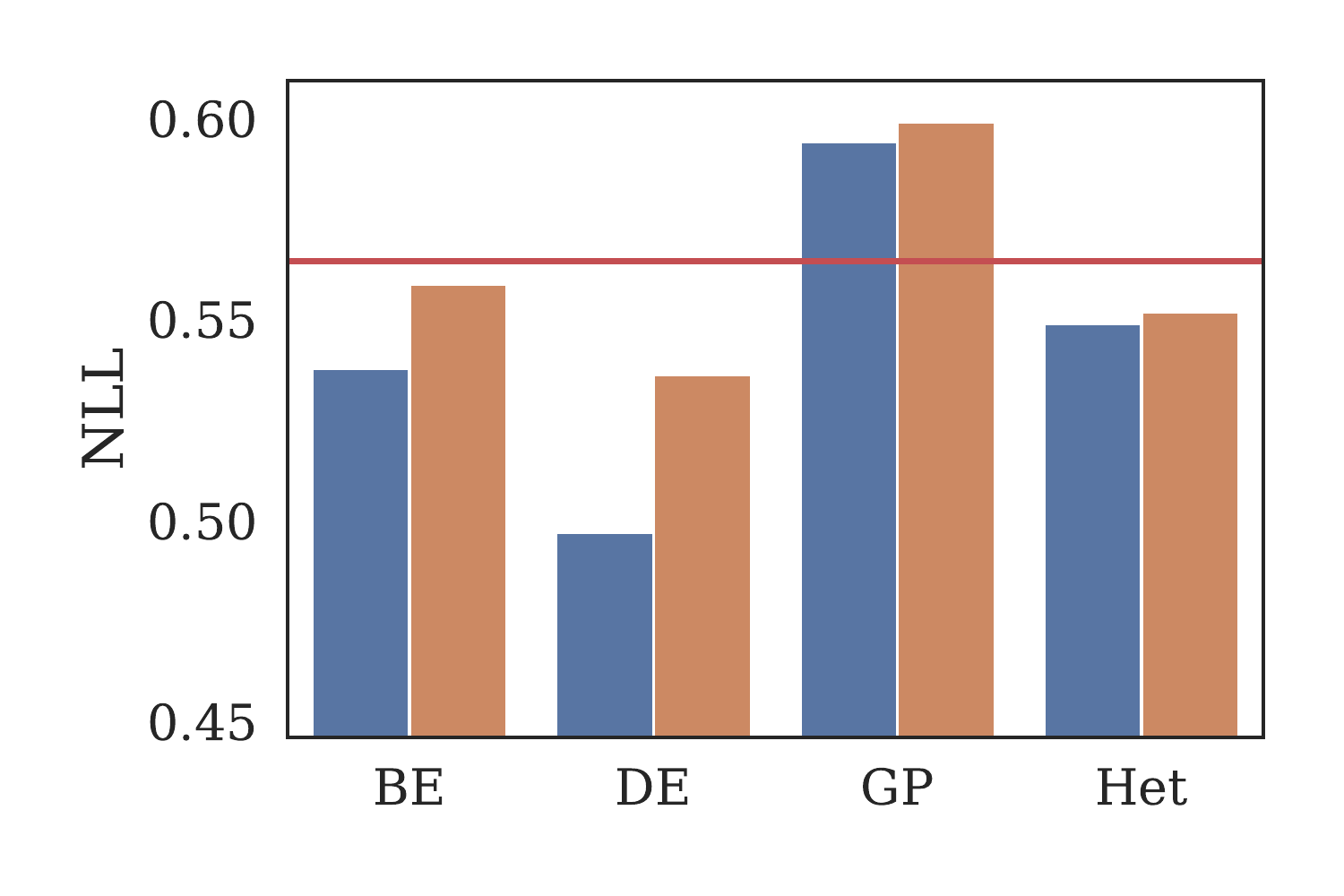}
\end{subfigure}
\caption{Upstream and downstream versus downstream-only use of specific architectures on Imagenet. We compare the extent to which a single pretrained deterministic model can be cast into alternative architectures during finetuning.
}
\label{fig:up_vs_down_inet}
\end{figure}

\Cref{fig:up_vs_down_inet} displays results on ImageNet. BatchEnsemble when applied during both pretraining and finetuning strictly outperforms the finetuning-only variant across the 3 metrics. We also ran a naive deep ensemble baseline and find that this is similarly the case. On the other hand, for both GP and Heteroscedastic, applying the method only during finetuning roughly matches the performance of the method when applied during both pretraining and finetuning. Therefore we apply BatchEnsemble for both pretraining and finetuning, and we restrict using last-layer methods for finetuning. See \Cref{app:heteroscedastic} for more analysis of the heteroscedastic last layer for vision tasks.


\subsection{Ensemble Scaling}
\label{sub:ensembles}
Ensembles of neural networks~\citep{hansen90,deep-ensembles}, which aggregate the predictions of several instances of the same model class, provide a remarkably simple, yet effective, way to improve the performance of the base model. This holds true not only for predictive performance
but also, crucially, for robustness and uncertainty quantification \citep{ovadia19, gustafsson20,wen_batchensemble_2020-1}. However, ensembles require an increasing amount of compute as one increases the size of the ensemble. We're motivated to understand this axis of model scaling, ensemble size, as it compares to an alternative, popular axis of model scaling: increasing width and depth.

\begin{figure*}[ht]
\begin{subfigure}{.49\textwidth}
\centering
\includegraphics[width=\textwidth]
{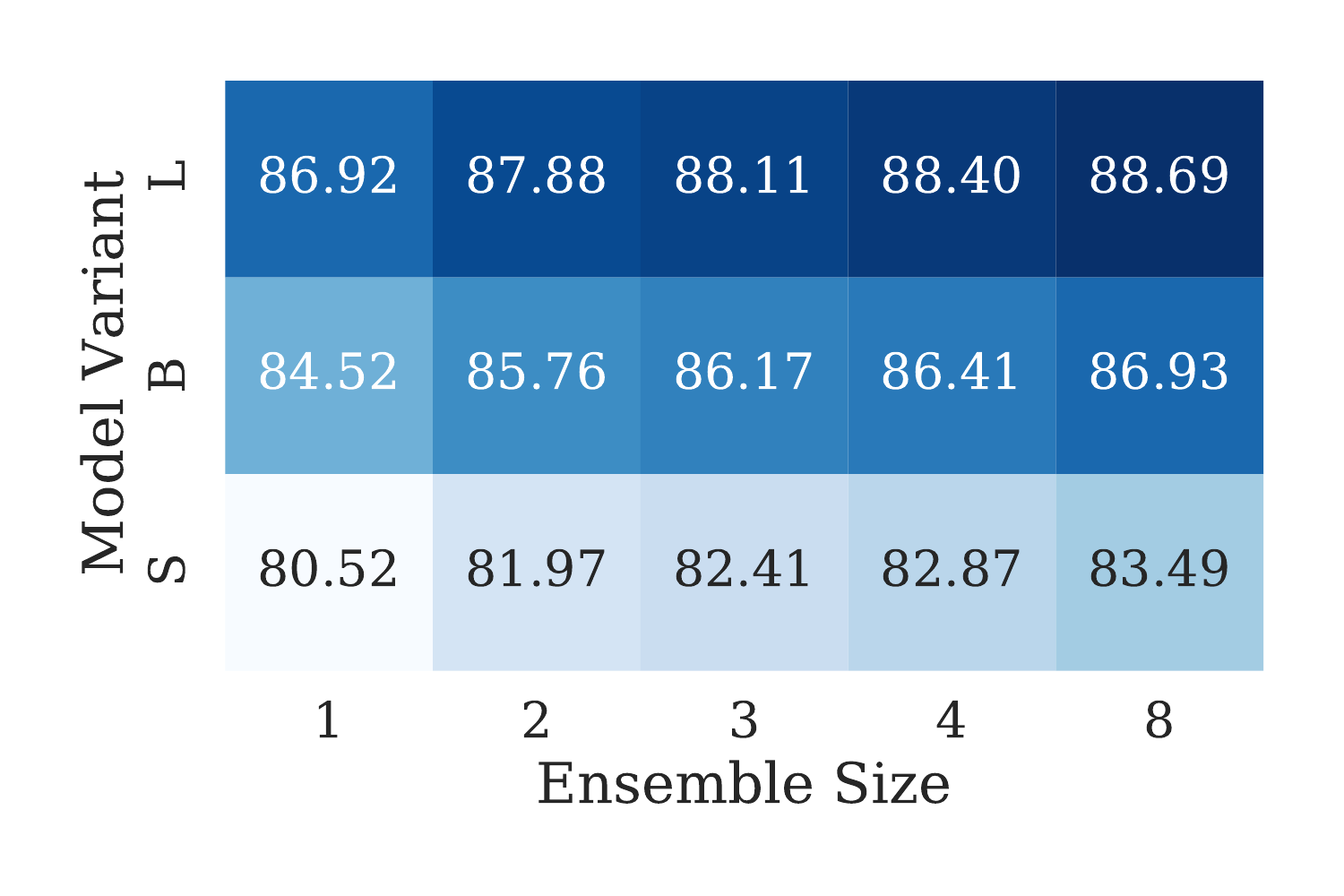}
\caption{Aggregated reliability score}
\end{subfigure}
\begin{subfigure}{.49\textwidth}
\centering
\includegraphics[width=\textwidth]
{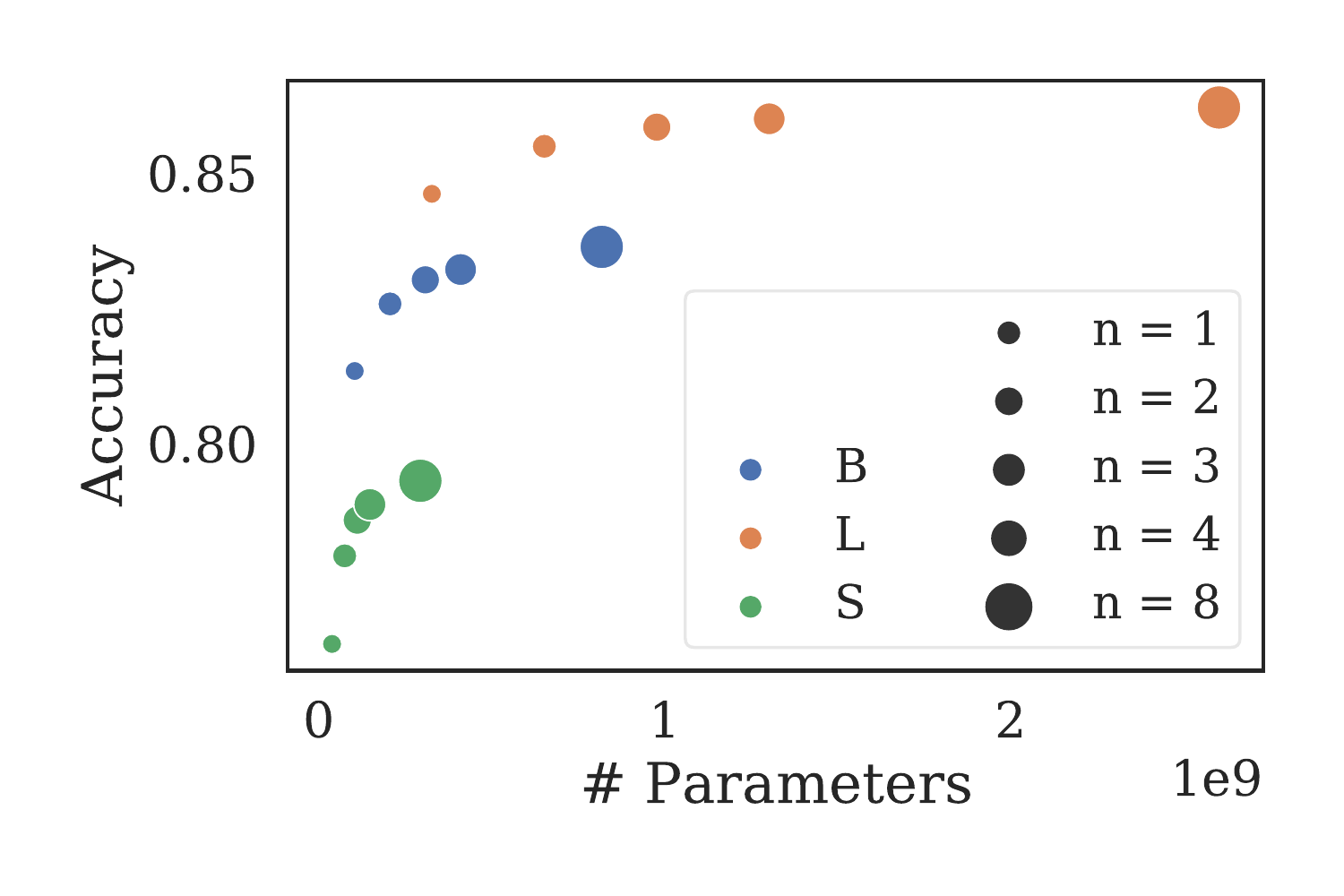}
\caption{ImageNet test accuracy}
\end{subfigure}
\vspace{-1.5ex}
\caption{Trading off model size with deep ensemble size. Overall, scaling the single model from Small to Large has a greater influence on the overall performance.
}
\label{fig:ensembles/model-vs-ensemble-size}
\end{figure*}

\Cref{fig:ensembles/model-vs-ensemble-size} shows that naive ensembling consistently provides better performance as the number of ensemble members increases. However, this comes at significant computational cost. Scaling up the model size from S to B or from B to L leads to an $\sim$4x increase in compute, and larger single models tend to outperform the ensembles of 4 smaller models. This motivates the importance of efficient ensembles in Plex: BatchEnsemble adds minimal extra compute and outperforms the baseline without ensembles consistently in the ranking across tasks (\Cref{fig:ranking}).

\section{Reliability Task Results}
\label{sec:reliability}

In this section, we examine performance on individual tasks across the areas of uncertainty, robust generalization, and adaptation.


\subsection{Uncertainty}

To make an accurate assessment of the quality of a model's predictive uncertainty, we consider a set of tasks, each highlighting a different property of models' predictive uncertainty estimates.

\subsubsection{Calibration}
\label{sub:calibration}

\begin{figure*}[!tb]
\centering
\includegraphics[width=\linewidth]{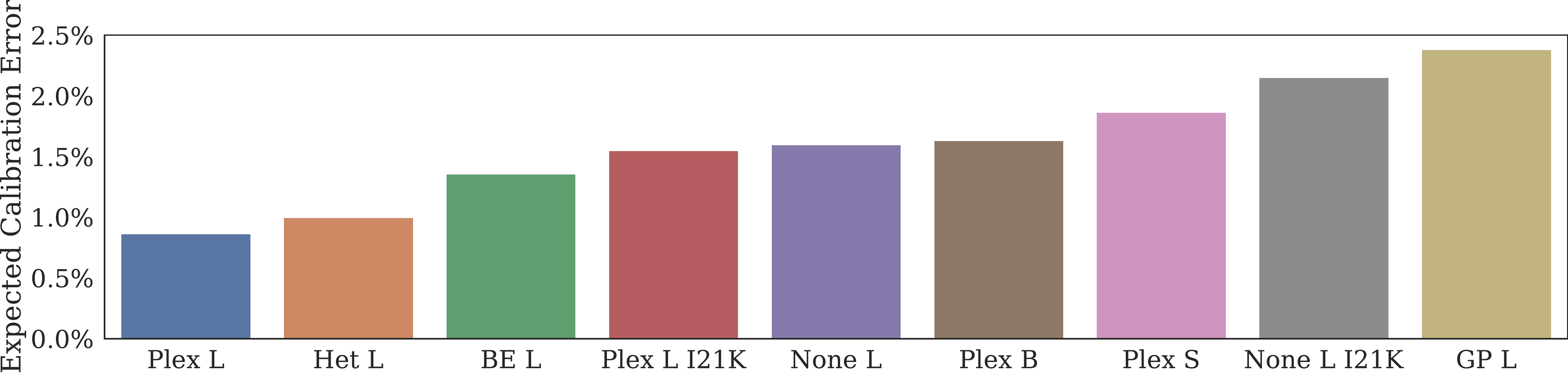} \\[10pt]
\includegraphics[width=0.8\linewidth]{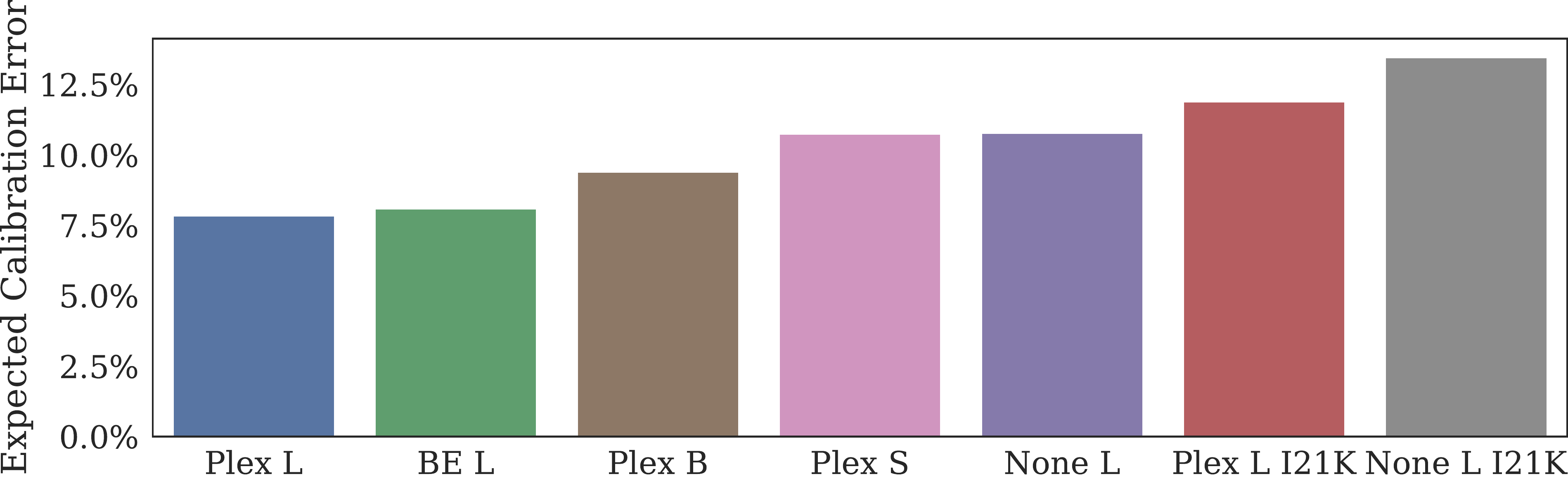}
\caption{%
    Expected calibration error averaged over three in-distribution vision evaluation datasets \textbf{(top)} and six out-of-distribution vision evaluation datasets \textbf{(bottom)}.
    Model types are shown on the $x$-axis. 
}
\label{fig:calibration-vision}
\end{figure*}

\begin{figure*}[!htb]
\centering
\includegraphics[width=0.95\linewidth]{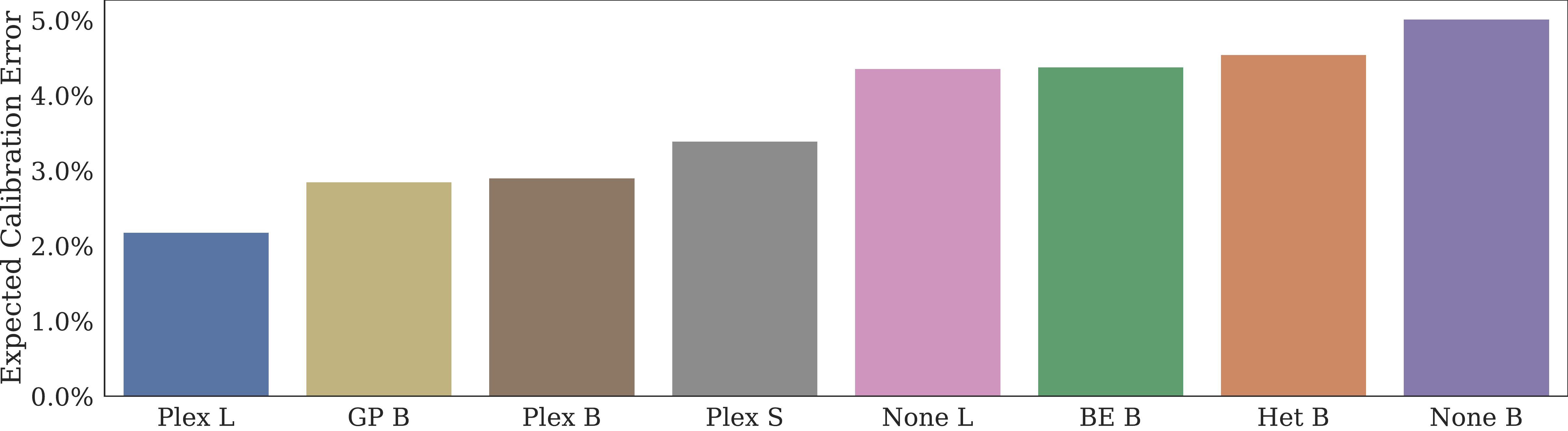} \\[10pt]
\includegraphics[width=0.95\linewidth]{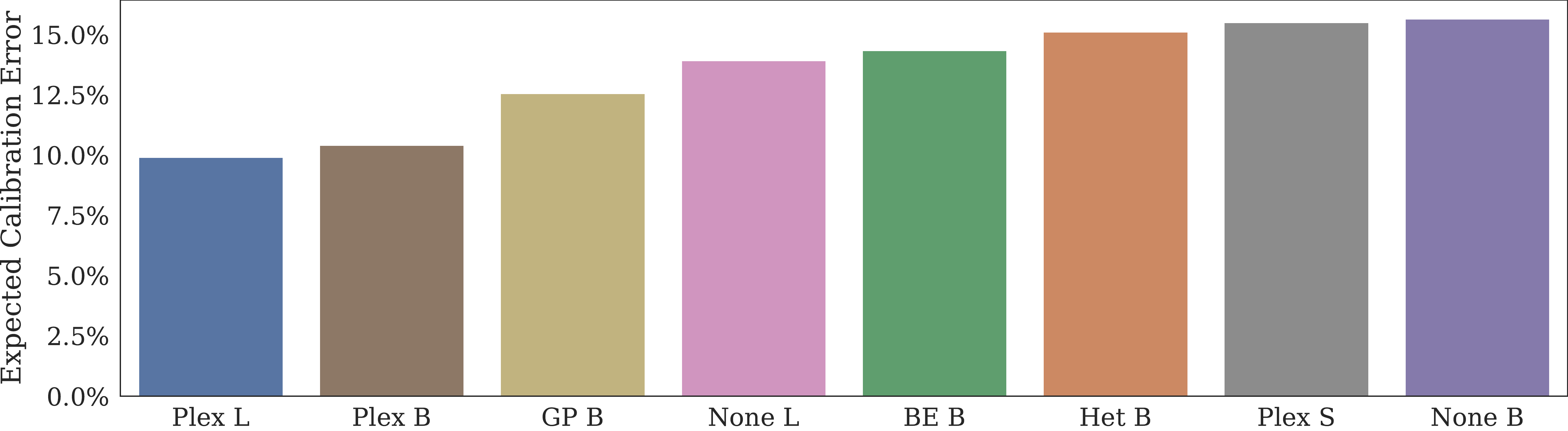}
\caption{%
    Expected calibration error averaged over two in-distribution language evaluation datasets \textbf{(top)} and three out-of-distribution language evaluation datasets \textbf{(bottom)}.
    Model types are shown on the $x$-axis. 
}
\label{fig:calibration-language}
\end{figure*}


\begin{tldr}
Plex improves calibration over a baseline without any changes, especially on out-of-distribution. Scaling model size and pretraining dataset size can improve calibration.
\end{tldr}

Intuitively, calibration is about the ``accuracy'' of a model's uncertainty estimates. 
It reflects how well the model’s confidence, which quantifies the predictive probability of correctness, aligns with the model's accuracy, which is the observed probability of correctness \citep{dawid1982well}. We investigate two different measures of calibration.

\paragraph{Expected Calibration Error}
Expected calibration error (ECE) \citep{naeini2015obtaining} is a binning metric that computes the average difference in confidence and accuracy within different confidence bins.
We evaluate ECE of vision models on 3 in-distribution datasets---CIFAR-10, CIFAR-100, ImageNet---and 6 out-of-distribution---CIFAR-10H, ImageNet ReaL-H, ImageNet-A, ImageNet-C, ImageNet-R, and ImageNet-V2. We also evaluate ECE on T5-Plex over 2 in-distribution datasets---MNLI-matched and NaLUE---and 3 out-of-distribution datasets---MNLI-mismatched, HANS, and NaLUE-tail\footnote{We excluded toxicity detection datasets which exhibits severe label imbalance. For these datasets, ECE as a population-average metric is not a suitable measure. For example, a naive model that always predict the majority label will trivially achieve high accuracy and low ECE.}.

\Cref{fig:calibration-vision} displays the average calibration error of vision models on both in- and out-of-distribution evaluation datasets.
Calibration errors on in-distribution are relatively low in general (less than 2.5\% ECE).
Plex improves calibration error over using no changes (None) by roughly a reduction in half on average; on out-of-distribution, it is roughly a 3\% improvement whether the models 
both use JFT or both use ImageNet21K.
Regarding model size, we find calibration error improves over S, B, and L. Another significant difference is in pretraining dataset size from ImageNet-21K to JFT, where we find JFT models consistently outperform ImageNet-21K pretrained models.

A similar trend can be observed in the language domain.
As shown in \Cref{fig:calibration-language}, compared to the baseline without changes (None), Plex improves the model's calibration performance by roughly 3\% on in- and out-of-distribution.
On the other hand, a model's uncertainty performance is impacted more by the type of uncertainty methods than the model size. For example, the Plex B model is on average stronger than the None L model.

\begin{table}[t]
\centering
\begin{minipage}{.5\linewidth}
    \centering
    \begin{tabular}{cccc}
    \toprule
    Models &  C10 & C100 & ImageNet \\
    \midrule
    Plex S & \bf 93.6\% & 90.9\% & 86.7\% \\
    Plex B & \bf 93.6\% & 92.4\% & 87.4\% \\
    Plex L & 89.2\% & \bf 93.7\% & \bf 87.7\% \\
    \bottomrule
    \end{tabular}
\end{minipage}%
\begin{minipage}{.5\linewidth}
    \centering
    \begin{tabular}{cccc}
    \toprule
    Models &  MNLI & NaLUE & Toxic Comments \\
    \midrule
    Plex S & 76.4\% & 92.9\% & 90.5\% \\
    Plex B & 82.7\% & \bf 93.4\% & 90.4\% \\
    Plex L & \bf 86.0\% & 88.9\% & \bf 91.3\% \\
    \bottomrule
    \end{tabular}
\end{minipage}%
\caption{%
    Calibration AUROC, measured on the in-distribution test split of each vision dataset \textbf{(left)} and text dataset \textbf{(right)}.
}
\label{fig:calibration_auc}
\end{table}

\paragraph{Calibration AUROC.}
Calibration AUROC considers the binary classification problem of predicting from a model's uncertainty on a given input whether its prediction for the same input (i.e., the class associated with the highest class probability) is correct.
It is the area under the ROC curve for this binary classification task, and it
aggregates the model's predictive performance over all possible confidence thresholds.
Comparing to ECE, Calibration AUROC only evaluates the uncertainty score’s ranking performance, that is, whether a model consistently assigns higher uncertainty to incorrect predictions rather than correct predictions, and such an uncertainty score can be used as a signal for prediction correctness with good precision-recall and ROC performance~\citep{krishnan2020improving,kivlichan2021measuring}. \Cref{fig:calibration_auc} presents Calibration AUROC on vision and text datasets. We see a similar pattern as ECE where larger models generally perform better, but the pattern is slightly weaker. For example, Plex L performs best on only 4/6 datasets.

\subsubsection{Selective Prediction}

\begin{tldr}
The model extensions in Plex provide a significant improvement for selective prediction, enabling the ability to predict at much lower error rates when deferring just a small fraction of predictions.
\end{tldr}

Deep learning models are typically evaluated under the lens of average predictive accuracy. However, this doesn't account for the real-world cost of mistakes in deployed models today.  Often the risk associated with a mistake can outweigh the benefit of being correct and thus, in expectation, it can be better to not predict at all under some confidence level or defer to a more expensive procedure (e.g., a human expert). 
This motivates selective prediction, which includes predictive performance as part of a larger decision-making scenario in which the model may abstain from making certain predictions \citep{el2010foundations}.

Selective prediction performance can be evaluated using several approaches, and there has not been much standardization in the literature. We investigate two that use model uncertainty, through joint human-model collaboration and through rejection rates.

\begin{table}[t]
\centering
\begin{subfigure}{.5\linewidth}
    \centering
    \begin{tabular}{cccc}
    \toprule
    Models &  C10 & C100 & ImageNet \\
    \midrule
    Plex S & 0.986 & 0.960 & 0.940 \\
    Plex B & 0.991 & 0.960 & 0.960 \\
    Plex L & \bf 0.999 & \bf 0.980 & \bf 0.980 \\
    Plex L I21K & 0.997 & 0.970 & 0.960 \\
    \bottomrule
    \end{tabular}
\end{subfigure}%
\caption{%
    Oracle Collaborative AUROC with a review fraction of 0.5\% of all predictions.
    Default pretraining dataset is JFT and I21K signifies ImageNet21K.
}
\label{table:ocauc-vision}
\end{table}

\paragraph{Selective Prediction by Model Uncertainty using Human--Model Collaboration.}
%
%
Many real world use cases of AI allow for a model to defer a subset of predictions to a human expert.
Oracle Collaborative Accuracy and AUROC measure the performance of an oracle--model collaboration system, where the oracle acts a proxy for human experts in order to automate evaluation. 
The model sends predictions with high predictive uncertainty to an oracle subject to a fixed referral budget (e.g., only 1\% of all queries can be referred to the oracle).
We compute Oracle Collaborative Accuracy and AUROC over a range of budgets and evaluate on both vision and language modalities.

For vision, \Cref{table:ocauc-vision} displays Oracle Collaborative AUROC with a review budget of 0.5\% on three vision datasets over different variants of Plex.
Surprisingly, the AUCs are quite high: on ImageNet for example, Plex L attains 0.98, and the models all achieve greater than 0.9 across the datasets.
This implies that the ability to defer can enable lower error, which can be useful for higher-risk applications.  Plex consistently performs best, even when taking into account different model and dataset sizes.  We also examined the metric over review budgets of 1\%, 2\%, and 5\%, and found the results to be unchanged. This suggests that the results on 0.5\% are the model's limiting performance with an oracle in the loop.

\begin{figure*}[t]
\begin{subfigure}{.3\textwidth}
\centering
\includegraphics[width=\textwidth]{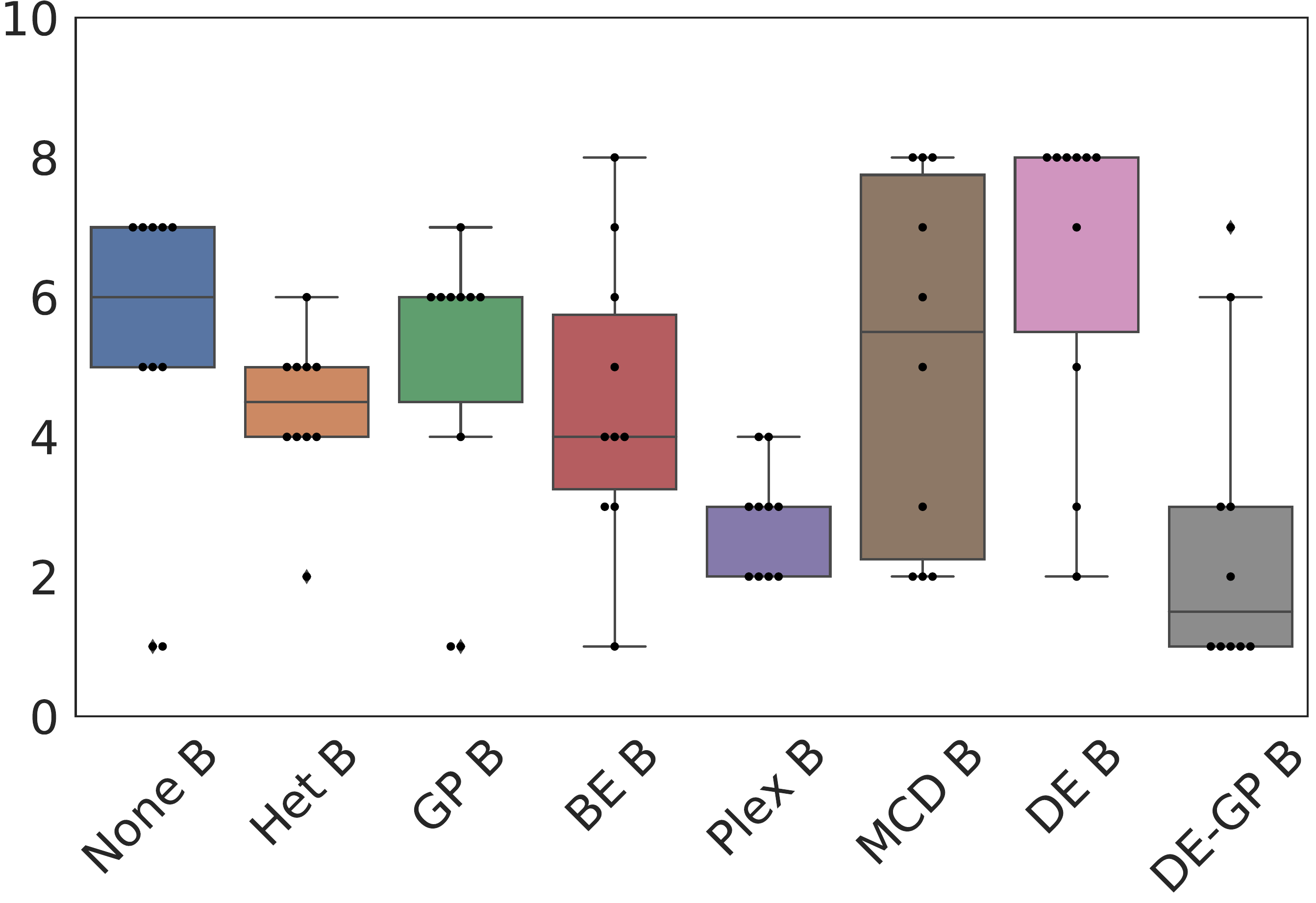}
\caption{In-domain, by method}
\label{fig:t5-collaboration-in-domain-methods}
\end{subfigure}
\begin{subfigure}{.3\textwidth}
\centering
\includegraphics[width=\textwidth]{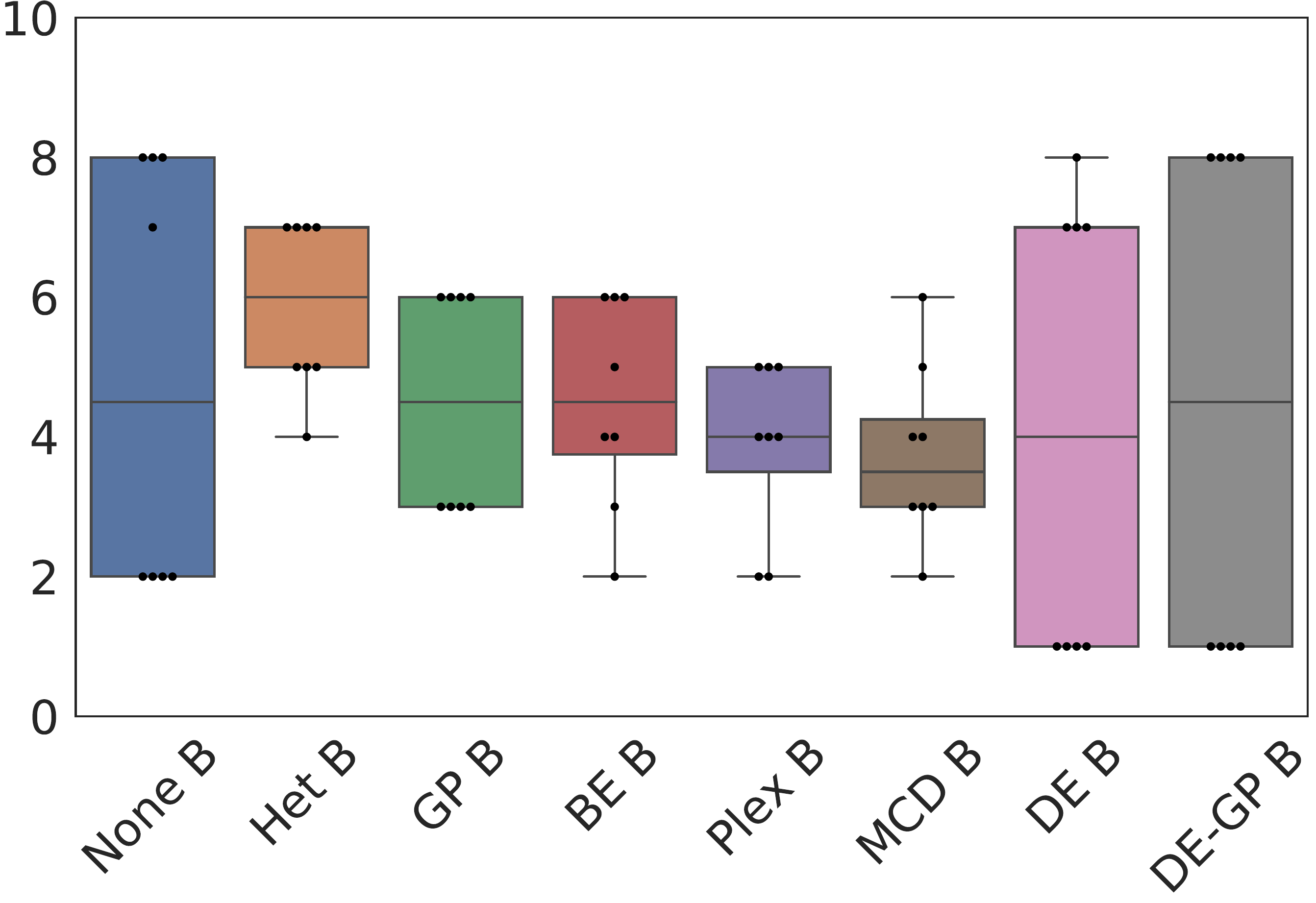}
\caption{OOD, by method}
\label{fig:t5-collaboration-ood-methods}
\end{subfigure}
\begin{subfigure}{.3\textwidth}
\centering
\includegraphics[width=\textwidth]{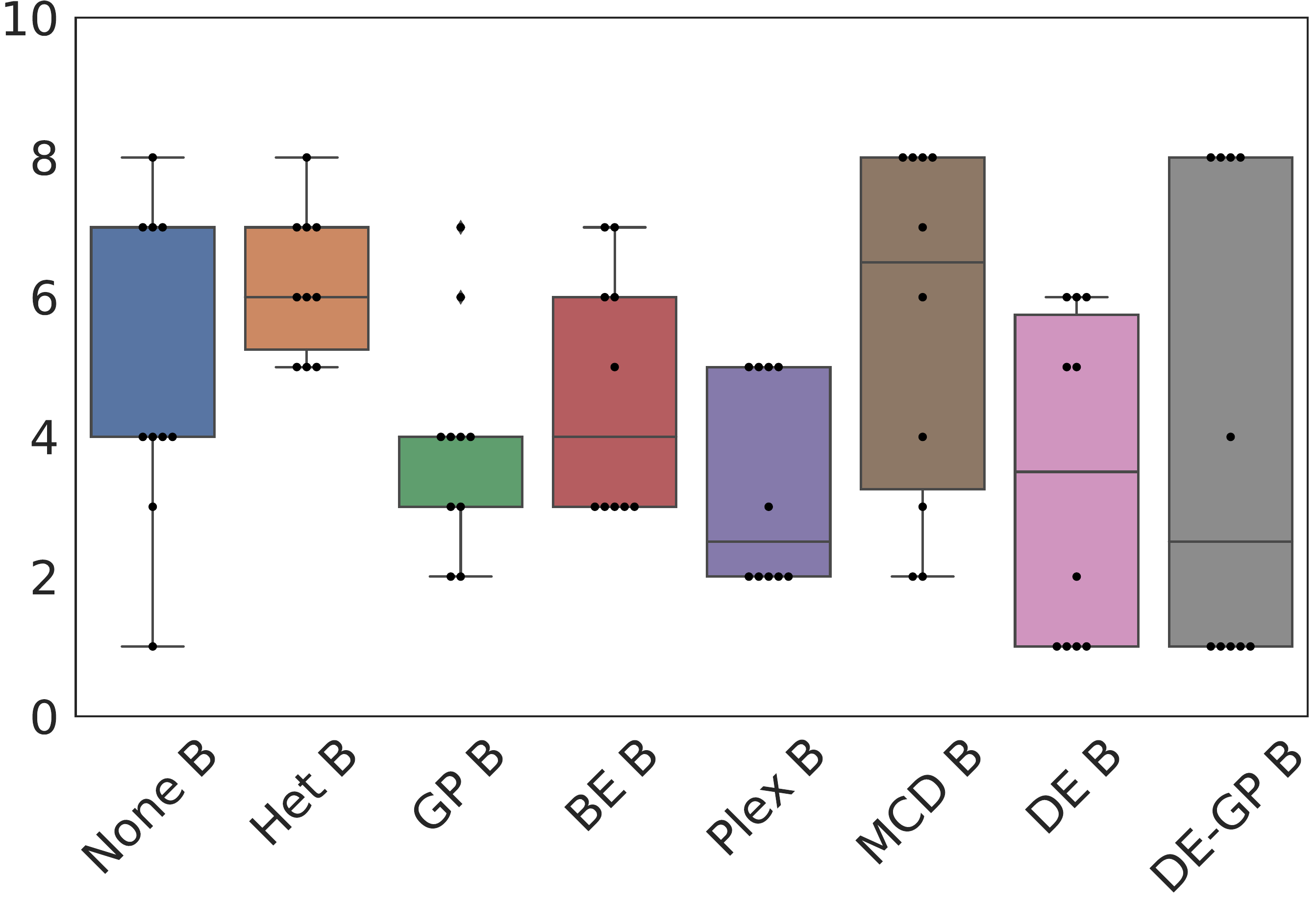}
\caption{Subpopulation, by method}
\label{fig:t5-collaboration-subpop-methods}
\end{subfigure}
\begin{subfigure}{.3\textwidth}
\centering
\includegraphics[width=\textwidth]{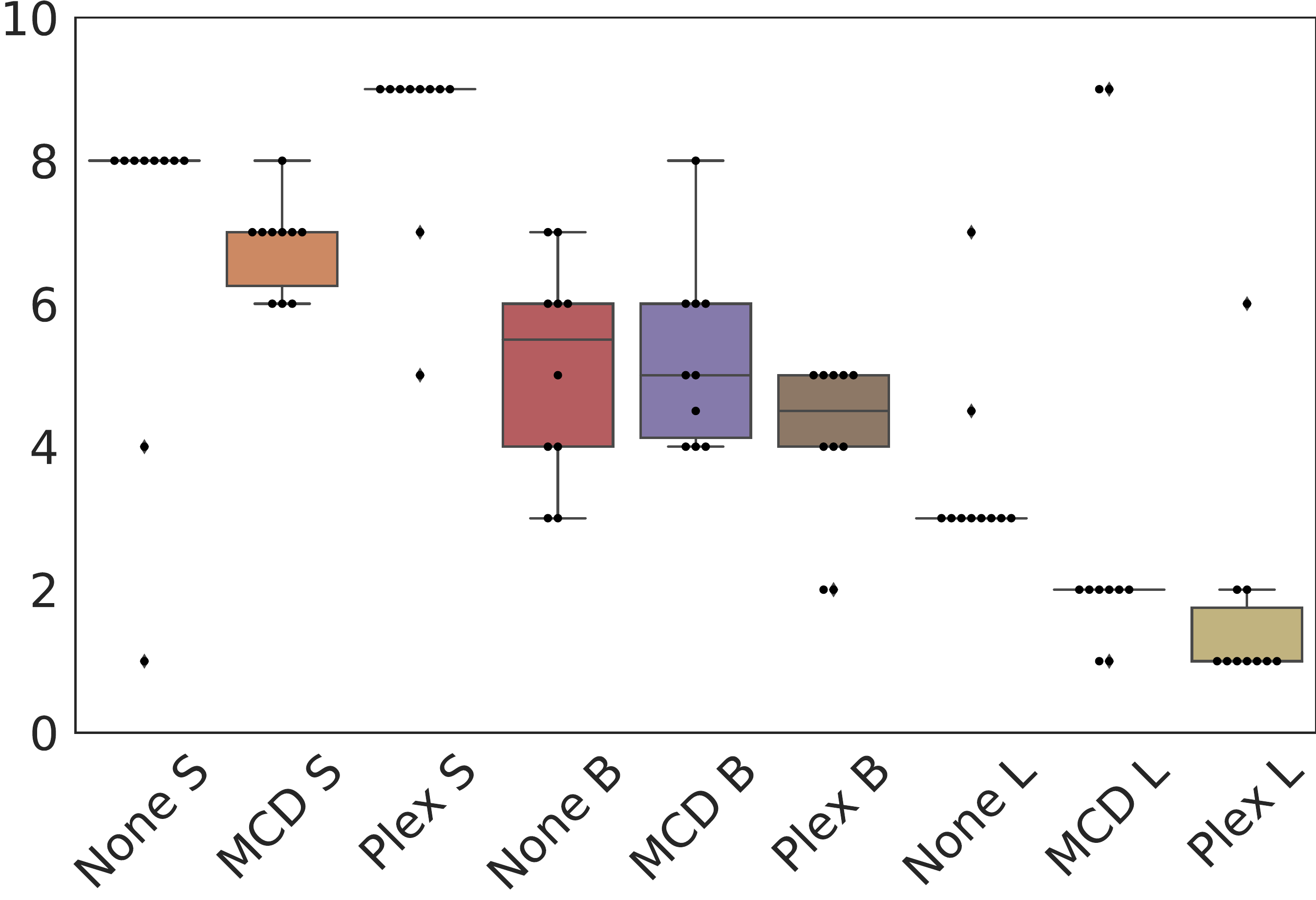}
\caption{In-domain, by size}
\label{fig:t5-collaboration-in-domain-size}
\end{subfigure}
\hfill
\begin{subfigure}{.3\textwidth}
\centering
\includegraphics[width=\textwidth]{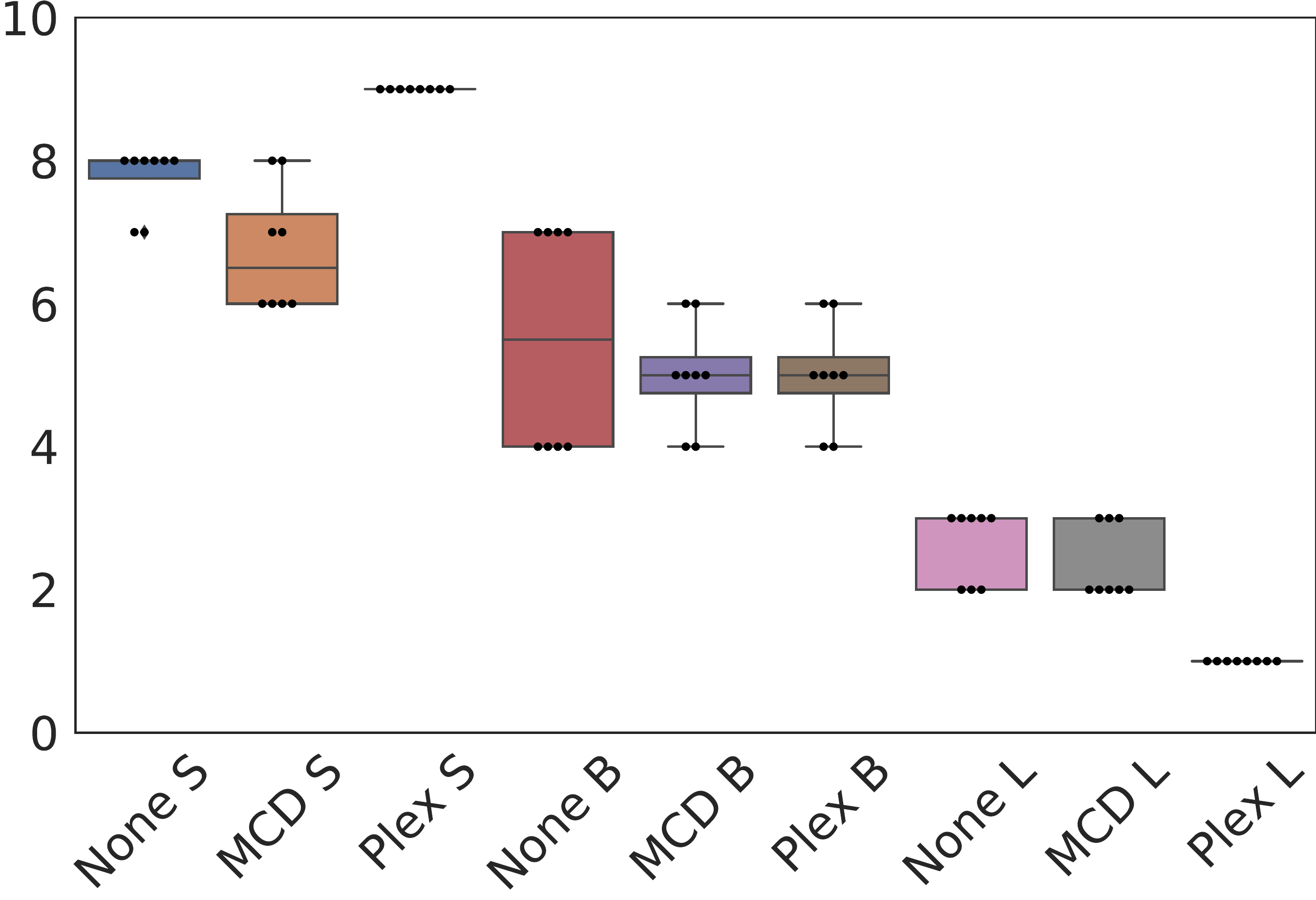}
\caption{OOD, by size}
\label{fig:t5-collaboration-ood-size}
\end{subfigure}
\hfill
\begin{subfigure}{.3\textwidth}
\centering
\includegraphics[width=\textwidth]{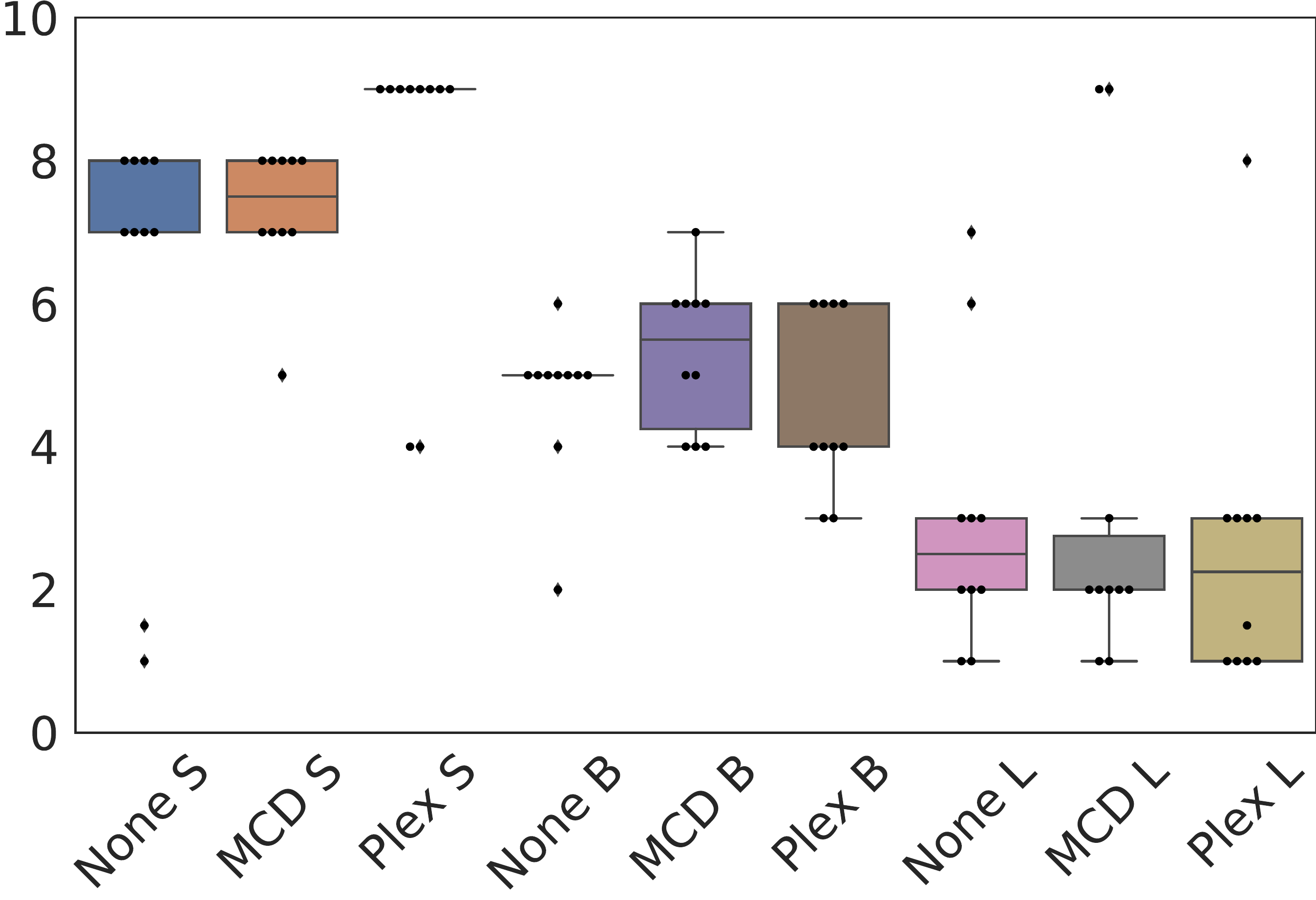}
\caption{Subpopulation, by size}
\label{fig:t5-collaboration-subpop-size}
\end{subfigure}
\vspace{-1.5ex}
\caption{Ranking performance in selective prediction (lower is better) across different methods (\ref{fig:t5-collaboration-in-domain-methods}-\ref{fig:t5-collaboration-subpop-methods}) and model sizes (\ref{fig:t5-collaboration-in-domain-size}-\ref{fig:t5-collaboration-subpop-size}). Each model's box plot is its ranking over multiple datasets: ranking is done by AUROC on MNLI and Toxic Comments, and by Accuracy on NaLUE (AUROC is not well-defined for multi-class classification).
}
\label{fig:t5-collaboration-rank-compare}
\end{figure*}

For language, \Cref{fig:t5-collaboration-rank-compare} reports results across three different model sizes (S, B, L) and for eight methods: a baseline without changes (None), MC Dropout (MCD), Deep Ensemble (DE), Gaussian process (GP), Batch Ensemble (BE), Heterostochastic (Het) and two method combinations Gaussian process ensemble (DE+GP) and BE-GP (Plex) (see \Cref{fig:t5-method-compare,fig:t5-arch-compare} in the Appendix for detailed results). We show the rankings of these methods under different types of test data distribution (i.e. in-domain, OOD and tail-population). We first see that across different methods, DE$+$GP, BE$+$GP (Plex), BE, and MCD tend to have the strongest performance. In particular, DE$+$GP almost always dominates the other methods on MNLI and NaLUE, and remains competitive in the case of label imbalance (i.e., Toxic Comments). However, DE$+$GP is an expensive method that costs 10x more in memory and compute and therefore is not competitive in scale (a more thorough analysis is in \Cref{sub:ensembles}). On the other hand, among the more efficient, single-model methods (i.e., Plex, BE and MCD), BE and Plex perform well on MNLI and NaLUE (notably, outperform the most expensive DE), while MCD stands out in the Toxic Comments. The above observations suggest that, when the training examples are drawn from a relatively simple distribution, quantifying output-layer uncertainty alone is sufficient to attain strong performance. However, when there are pathologies in the data distribution (e.g., extreme label imbalance and high label noise), quantifying the uncertainty within the model's intermediate representations (e.g., via some form of perturbation like BE) becomes important.

Next we investigate how a model's performance is impacted by the model size. 
For model size scaling, we evaluate BE$+$GP (Plex), MCD, and None.
We evaluate the performance of each method under progressively larger architectures, S, B, and L, and observe how the behavior changes across the method and with respect to the architecture size.
\Cref{fig:t5-collaboration-in-domain-size}-\ref{fig:t5-collaboration-subpop-size}
summarizes the rankings of uncertainty methods organized by the sizes of the architecture. 
As shown, comparing across architecture sizes, we see a larger architecture almost always leads to stronger performance in collaboration. This trend remains largely consistent even under distributional shifts and in tail groups. On the other hand, a model's uncertainty performance is impacted more by the type of uncertainty methods. That is, the MCD and BE$+$GP models are on average stronger than None models, regardless of architecture size. 
Finally, within larger architectures (i.e., T5 B and T5 L), BE$+$GP generally outperforms MCD, and MCD outperforms None. This trend is broken in two situations, (1) in Toxic Comments, MCD strongly outperforms all other architectures, (2) in small architecture, BE-GP is often the poorer performing model, and None achieves the best performance in multi-token prediction problems (NaLUE). 
Finally, between the larger architectures (i.e., T5 B v.s. T5 L), Plex (BE$+$GP) generally outperforms MCD, and MCD outperforms None. This trend is broken in two situations, (1) in Toxic Comments, MCD strongly outperforms all other architectures, (2) among the small architectures, the None baseline achieves the best performance in multi-token prediction problems (i.e., NaLUE). 

\begin{figure*}[t!]
\centering
\includegraphics[width=\linewidth]{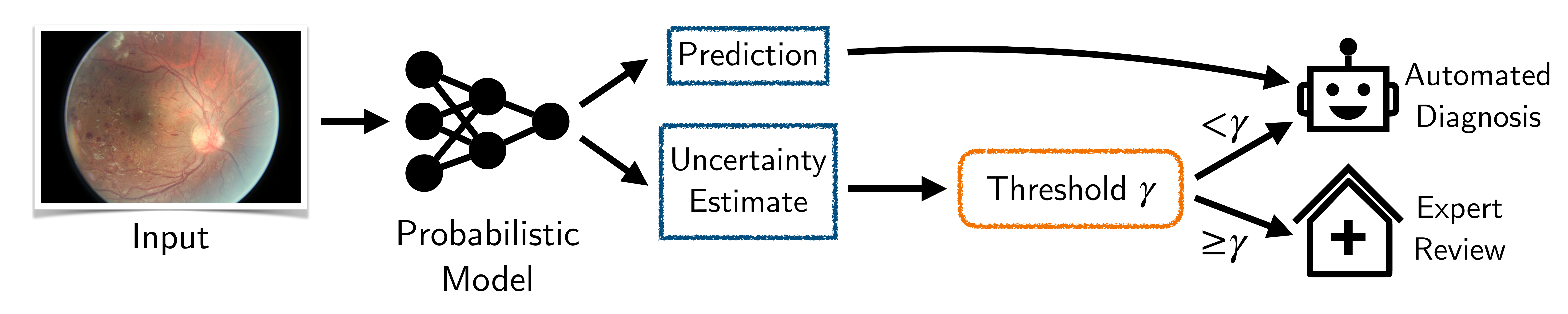}
\vspace{-4ex}
\caption{%
    \textbf{Selective Prediction Pipeline.} Given an input, a model returns a prediction and uncertainty estimate.
    If the uncertainty estimate exceeds a threshold $\gamma$ (indicating a level of uncertainty corresponding to a rejection rate $\tau$ that reflects the capacity for expert reviews), the diagnosis is referred to a human expert; otherwise, it is processed with no further review.
    The schematic is based on Figure 2 in \citet{band2021benchmarking}.
}
\label{fig:retina_sel_pred}
\vspace*{-5pt}
\end{figure*}

\paragraph{Selective Prediction by Model Uncertainty using Rejection Rates.}
Rejection AUC measures a model's performance in the scenario where it is permitted to not predict on some fraction of the data for which the model is most uncertain, e.g., up to 10\% of all queries.  Unlike the collaboration metric, it does not assume the query is passed to an oracle as rejected inputs may be subject to further review for broader decision making such as to improve data collection.
Rejection AUC is computed from predictive uncertainty (e.g., predictive entropy, predictive variance, confidence) and a chosen predictive performance evaluation metric (e.g., accuracy, AUROC, AUPRC) at different rejection rates.
For a rejection rate of $\tau \in [0, 1]$, the $\tau \times 100$\% of data points in the evaluation set for which the model is most uncertain are identified.
Those inputs are then rejected, and we assess the predictions on the remaining $(1 - \tau) \times 100$\% of data points.
Accuracy--Rejection AUC is given by computing the area under the Accuracy--Rejection Rate curve.
Additional results for AUROC and AUPRC---Rejection AUCs can be found in \Cref{appendix:selective-prediction}.

\begin{figure*}[!bt]
\centering
\begin{subfigure}{\linewidth}
    \includegraphics[width=\linewidth]{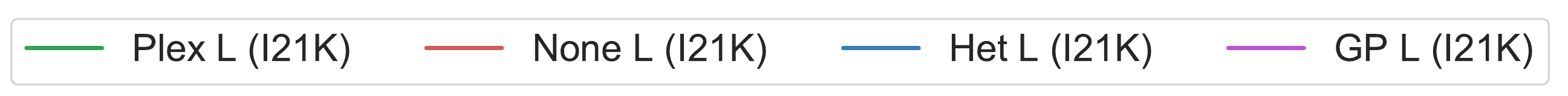}
\end{subfigure}
\hspace*{10pt}\begin{subfigure}[l]{0.465\linewidth}
    \includegraphics[width=\linewidth]{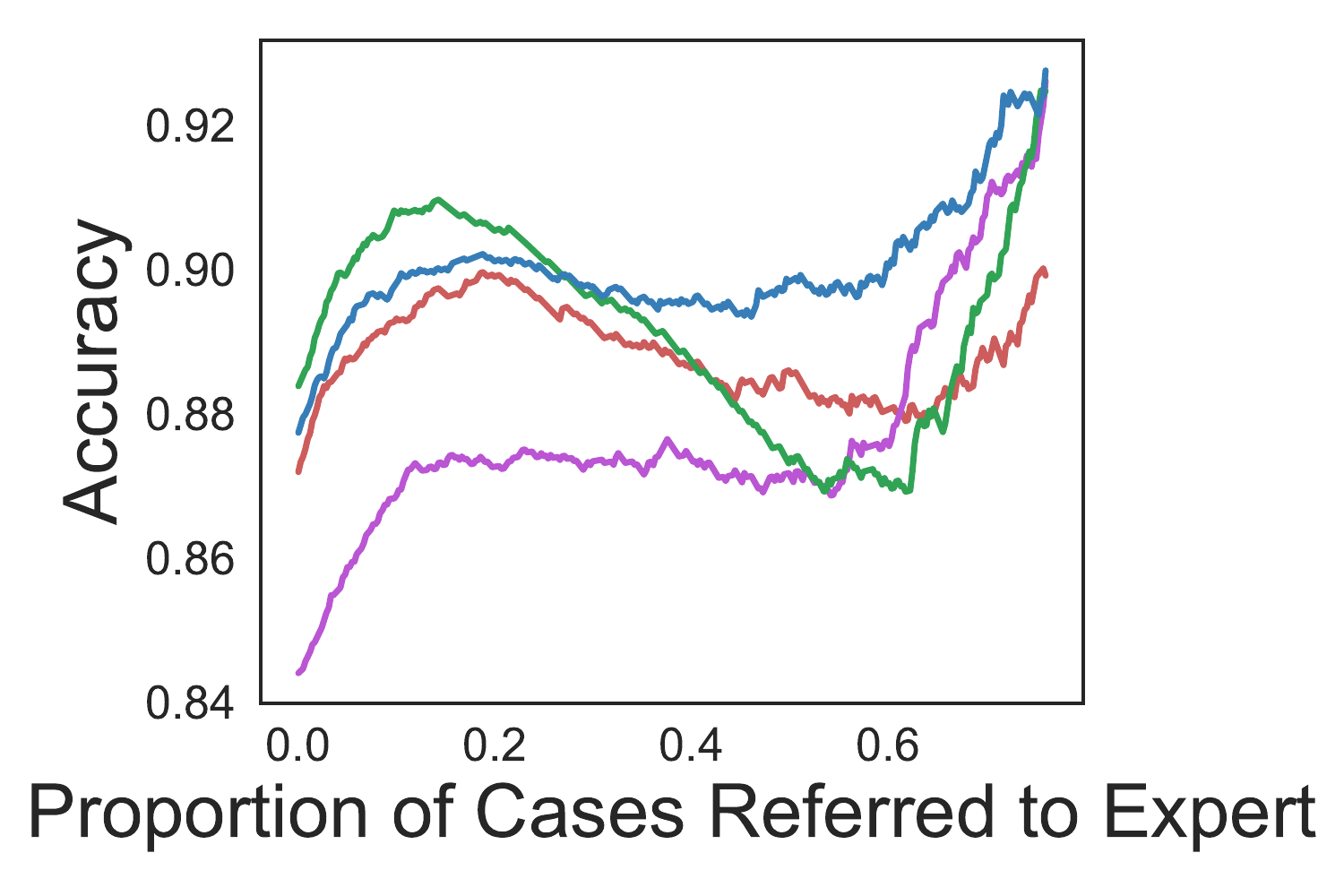}
    \caption{Country Shift: Out-of-Distribution\\$~$}
\end{subfigure}
\hspace*{10pt}
\begin{subfigure}[l]{0.465\linewidth}
    \includegraphics[width=\linewidth]{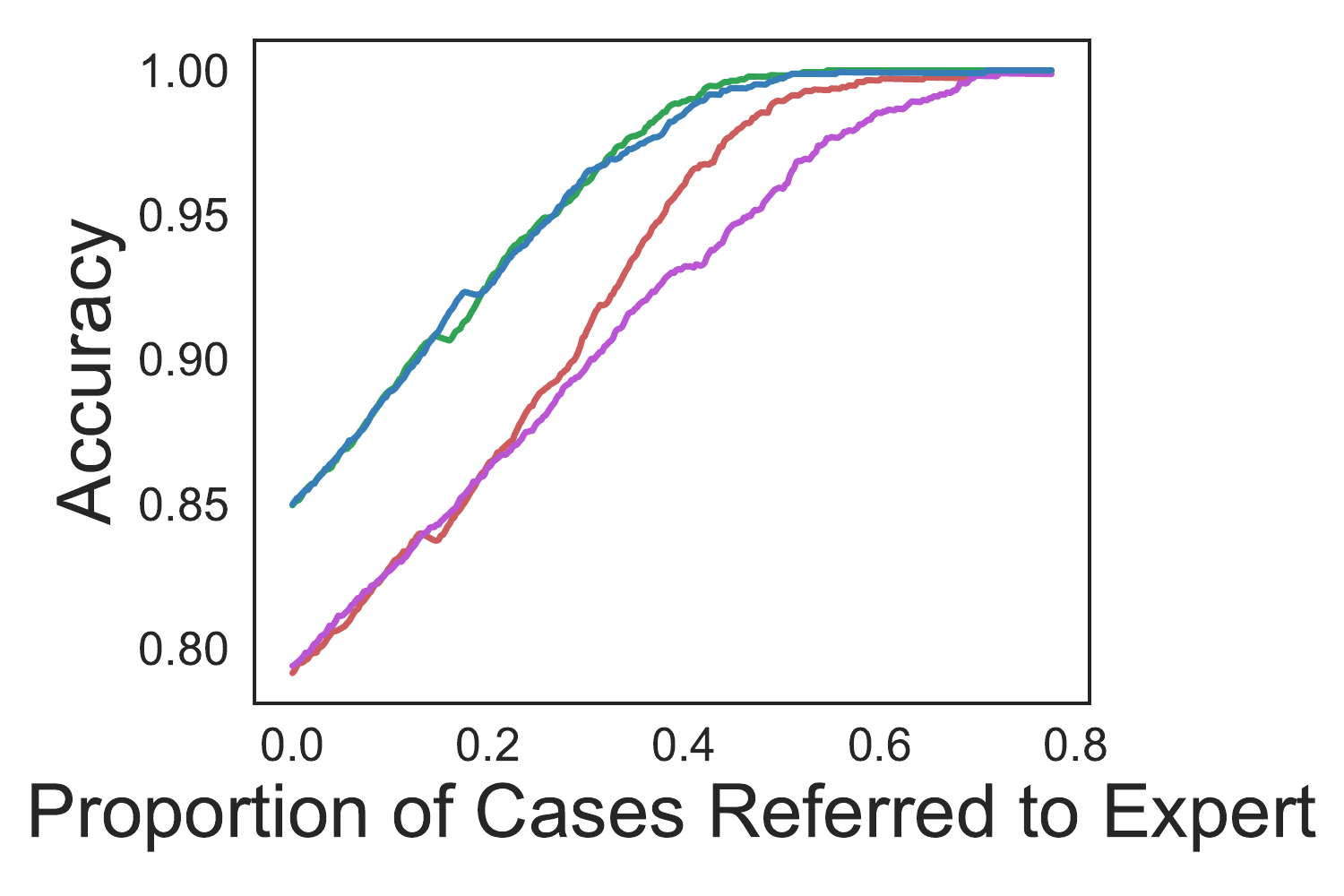}
    \caption{Severity Shift: Out-of-Distribution\\$~$}
\end{subfigure}
\vspace{-3ex}
\caption{%
    \textbf{Selective Prediction on RETINA (Out-of-Distribution).}
    \textbf{(left)} 
    The Country Shift task is concerned with diagnosing diabetic retinopathy from retina scans
    obtained using different medical equipment and a different patient population.
    \textbf{(right)}
    The Severity Shift task is concerned with detecting levels of severity of diabetic retinopathy more severe than those included in the training data.
    Accuracy is used as the evaluation metric and model predictive entropy is used as the per-example uncertainty estimate.
}
\label{fig:retina_sel_pred_ood}
\end{figure*}
\begin{table*}[!bt]
\centering
\begin{tabular}{llcccc}
\toprule
 & & Plex L  &
None L  & 
Het L & 
GP L \\
&  & (I21K) &
(I21K) & 
(I21K) & 
 (I21K)\\
\toprule
\multirow{2}{*}{Country Shift} &
In-Distribution &
\bf 96.3\% &
96.2\% &
\bf 96.3\% &
95.1\% \\
& Out-of-Distribution &
90.4\% &
89.5\% &
\bf 91.2\% &
89.4\% \\
\multirow{2}{*}{Severity Shift} &
In-Distribution &
95.4\% &
\bf 96.3\% &
96.0\% &
96.0\% \\
& Out-of-Distribution &
\bf 96.9\% &
94.4\% &
\bf 96.9\% &
93.6\% \\
\bottomrule
\end{tabular}
\caption{%
    Accuracy--Rejection AUC on RETINA Country and Severity Shift datasets.
}
\label{table:retina-sel-pred-country-accuracy}
\vspace*{-5pt}
\end{table*}

\Cref{table:retina-sel-pred-country-accuracy} displays overall performance for a selection of finetuned models that were pretrained on ImageNet-21K, and \Cref{fig:retina_sel_pred_ood} displays the full accuracy--rejection curves for the Country and Severity Shift out-of-distribution tasks.
We find that that Plex generally improves performance over using no changes on both datasets on three out of four evaluation tasks and that Het L performs as well or better than Plex L on all four tasks.

Examining the quality of different models' predictive uncertainty estimates for out-of-distribution evaluation more carefully, we find that on the Country Shift out-of-distribution (covariate shift) task, all models exhibit non-monotonically increasing accuracy--rejection rate curves.
Since the rejected examples are selected based on the models' predictive uncertainty estimates, this behavior indicates that beyond certain rejection-rate thresholds, the models tend to be underconfident about correct predictions and overconfident about mistakes.
We do not observe this pattern on the Severity Shift out-of-distribution (semantic shift) task or on the Country and Severity Shift in-distribution tasks (see \Cref{appendix:selective-prediction}).
\clearpage\newpage
\subsubsection{Open-set recognition (OSR)}
\label{sub:open-set-recognition}

\begin{tldr}
A simple scoring technique (maximum softmax probability) works well across open set recognition problems. Over image and text datasets, Plex provides a consistent improvement for new state-of-the-art, and without custom changes per dataset.
\end{tldr}

Open-set recognition, also called out-of-distribution detection or anomaly detection, aims to detect samples from new classes that are not included in training. In contrast to OOD generalization where the test example belongs to the same in-distribution training classes $(x, y)$, $y \in Y_\textsc{ind}$, OSR aims to detect test example $(x, y)$, where $y \in Y_\textsc{ood}$.

To evaluate OSR performance, we design an uncertainty score $U_{\theta}(x)$ that indicates the likelihood of an input $x$ being OOD, and $\theta$ denotes the classification model's parameters. 
As a default for all OSR problems, we apply the most commonly used uncertainty score: $1-\text{MSP}$, where MSP is the maximum over predicted softmax probabilities \citep{hendrycks_baseline_2017}.
A mixture of in-distribution examples and OOD examples are used as the test set. For each test example, the model generates an uncertainty score. 
Comparing the uncertainty score with its ground truth OOD label (0 indicates the example is from in-distribution data, and 1 indicates the example is from OOD data), we use AUROC to measure how well the uncertainty score separates the two groups. Note that the model is not finetuned using any OOD examples; test OOD examples are only used for evaluation. 
We discuss results on alternative uncertainty scores in \Cref{appendix:osr}.

\begin{table*}[tb]
\centering
\begin{tabular}{llll}
                          &                         & \textbf{In-Distribution}                                 & \textbf{Out-of-Distribution}                                 \\ \toprule
\multirow{3}{*}{Far-OOD}  & \multirow{2}{*}{Vision} & CIFAR-10                                                 & SVHN                                         \\ 
                          &                         & CIFAR-100                                                & SVHN                                         \\
                          & Language                & \multicolumn{2}{l}{NaLUE Standard Out-of-Scope} \\ \midrule
\multirow{4}{*}{Near-OOD} & \multirow{3}{*}{Vision} & CIFAR-10                                                 & CIFAR-100                                    \\ 
                          &                         & CIFAR-100                                                & CIFAR-10                                     \\
                          &                         & ImageNet2012                                             & Places365                                    \\
                          & Language                & \multicolumn{2}{l}{NaLUE Near Out-of-Scope}  \\ \bottomrule   
\end{tabular}
\caption{%
The open set recognition tasks studied in this section.
}
\label{table:osr-task}
\end{table*}

\textbf{Comparing models’ OSR performance across tasks}. 
We evaluate the performance of Plex and multiple baseline models.
Based on the similarity between the in- and out-of-distribution data, we can classify OSR tasks under two categories (\Cref{table:osr-task}): (a) far-OOD detection and (b) near-OOD detection.
In \Cref{table:osr-msp}, we observe that first, larger models have better performance. 
Second, among all the models, Plex performs the best for most takes.
Third, for ImageNet2012 vs Places365, models pretrained with ImageNet-21K (I21K) outperform the same models but pretrained on JFT.

\begin{table*}[tb]
\centering
\small
\begin{tabular}{ccc|ccc}
\multirow{2}{*}{MSP} & \multicolumn{2}{c|}{Far OOD}                              & \multicolumn{3}{c}{Near-OOD}                                     \\
                     & C-10 vs SVHN & C-100 vs SVHN & C-10 vs C-100 & C-100 vs C-10 & I2012 vs Places \\ \toprule
None                  & 0.996                      & 0.932                       & 0.986                          & 0.929                          & 0.828                                \\
None (I21K)             & 0.996                      & 0.913                       & 0.981                          & 0.924                          & \underline{0.838}                                \\
None$\to$GP           & \textbf{0.997}                      & 0.931                       & 0.986                          & 0.927                          & 0.824                                \\
None$\to$Het          & 0.996                      & 0.925                       & 0.984                          & \underline{0.948}                          & 0.827                                \\
None$\to$DE      & 0.996                      & 0.923                       & 0.986                          & 0.940                          & 0.831                                \\
None$\to$BE           & 0.996                      & 0.934                       & 0.982                          & 0.924                          & 0.822                                \\
GP                   & 0.995                      & \underline{0.938}                       & 0.981                          & 0.927                          & 0.828                                \\
Het                  & 0.996                      & \textbf{0.943}                       & 0.979                          & \underline{0.948}                          & 0.828                                \\
DE L              & 0.996                      & 0.937                       & \underline{0.987}                          & 0.946                          & 0.834                                \\
BE L              & 0.996                      & 0.927                       & 0.987                          & 0.940                          & 0.827                                \\
Plex L (I21K)       & 0.996                      & 0.932                       & 0.983                          & 0.934                          & \textbf{0.842}                                \\
Plex L       & \textbf{0.997}                      & \underline{0.938}                       & \textbf{0.988}                          & \textbf{0.954}                          & 0.831                                \\
Plex B              & 0.995                      & 0.925                       & 0.975                          & 0.906                          & 0.801                                \\
Plex S              & 0.990                      & 0.882                       & 0.957                          & 0.864                          & 0.773     \\ \bottomrule
\end{tabular}
\caption{%
The MSP based OOD AUROC of all types of models across the five tasks. 
}
\label{table:osr-msp}
\end{table*}

\textbf{Open-set intent detection based on large language pretrained models}.
To study OSR performance of different models in the language domain, we design an intent detection task for detecting natural utterances that are out of the scope (OOS) services. For the in-distribution dataset NaLUE, the model needs to map each utterance input $x$ to a 3-token sequence of $y=(y_1,y_2,y_3) = (\text{vertical name}, \text{domain name}, \text{intent name})$. To study both far-OOD performance and near-OOD performance, we construct two out-of-the-scope datasets for NaLUE: 
\begin{itemize}
    \item \textbf{NaLUE Standard-OOS}: completely out-of-domain queries, based on CLINC150-OOS. The standard-OOS queries do not share any of the (vertical name, domain name, intent name) with the in-domain query. 
    \item \textbf{NaLUE Near-OOS}: in-domain, out-of-scope queries. It is created by sampling 20\% intent from each known domain as OOD data. The near-OOS examples can share the same vertical name / domain name (but definitely different intent name) with the in-domain query.
\end{itemize}

The model outputs a sequence, and we compute the uncertainty score based on the conditional softmax probability $p( y_{l} | y_{<l} ,x, \theta)$, using the conditional entropy. The conditional entropy is a standard uncertainty score for sequences \citep{malinin2020uncertainty},
\begin{align}
\operatorname{uncertainty}(y|x,\theta)= - \frac{1}{L} \sum_{l=1}^L \mathbb{H}( y_{l} | y_{<l} ,x, \theta),
\end{align}
where
$\mathbb{H}( y_{l} | y_{<l} ,x, \theta)= - \sum_{k=1}^K p(y_l=k|y_{<l}, x, \theta) \log\left(p(y_l=l | y_{<l}, x, \theta) \right)$ is the conditional entropy based on the conditional distribution $p(y_l=k|y_{<l},x, \theta)$, and  $k=1\ldots K$ is the total number of intent names. 
The uncertainty score is compared with the ground truth OOD label. An input $(x, y_1,y_2,y_3)$ is labeled as OOD if any of $y_l$ belongs to the OOD labels.

In \Cref{tab:oos-intent-detection}, models including None, GP, Het, MCD, BE, BE-GP, Deep Ensemble (DE), DE-GP are evaluated for their detection performance on Standard-OOS and Near-OOS. BE and DE-GP outperform the other models, followed by MCD.

\begin{table*}[tb]
\centering
\begin{tabular}{ccccc}
        & \multicolumn{2}{c}{Standard-OOS} & \multicolumn{2}{c}{Near-OOS} \\
Model   & AUROC           & AUPRC          & AUROC         & AUPRC        \\ \toprule
None   & 0.987           & 0.941          & 0.877         & 0.535        \\
Het   & 0.981           & 0.932          & 0.854         & 0.491        \\
GP    & 0.997           & 0.987          & 0.884         & 0.554        \\
MCD   & \underline{0.998}           & \underline{0.989}          & \textbf{0.917}         & 0.616        \\
BE    & \textbf{0.999}           & \textbf{0.997}          & 0.909         & \textbf{0.627}        \\
Plex & 0.997           & 0.986          & 0.890         & 0.543        \\
DE    & 0.991           & 0.955          & 0.884         & 0.558        \\
DE-GP & \textbf{0.999}           & \textbf{0.997}          & \underline{0.916}         & \textbf{0.625} \\ \bottomrule 
\end{tabular}
\caption{
OOS intent detection performance with AUROC and AUPRC for different models.
}
\label{tab:oos-intent-detection}
\end{table*}


\subsubsection{Label Uncertainty}
\label{sub:label-uncertainty}

\begin{tldr}
We propose label uncertainty as an important challenge, with a new large-scale dataset and propose an evaluation metric. The heteroscedastic last-layer particularly helps for label uncertainty.
\end{tldr}

It is typical to encounter label noise in non-academic datasets. However, benchmark datasets are often carefully curated to reduce label noise due to unreliable or disagreeing annotators. 
We use the existing CIFAR-10H dataset, which is a relabelled version of the CIFAR-10 test set with on average $>$50 crowd-sourced noisy labels per input \citep{peterson2019human}.

There is a notable lack of label uncertainty datasets in the literature, and we found limiting performance on CIFAR-10 as our models achieve $>$99\% accuracy on the original CIFAR-10 test set.
Therefore we build a larger dataset called ImageNet ReaL-H, which leverages the raw annotations of ImageNet ReaL \citep{beyer2020we}. When raw labels were available for an image we followed the same averaging procedure as for CIFAR-10H to produce a soft label. There are some cases where no raw annotations were available. These cases correspond to an image where a set of ML models all agreed with the original ImageNet label for the image and thus the image was not sent to human annotators for re-labelling. In these cases we took the one-hot ImageNet label as the soft label (equivalent to all human annotators agreeing the original ImageNet label was correct).

\paragraph{Metric} 
To evaluate a model's ability to capture label uncertainty, consider a divergence measure $\mathcal{D}$ between probability distributions,
\begin{align}
    \mathbb{E}_{x\sim p_{\textsc{data}}(\mathbf{x})} \Big[ \mathcal{D}\Big( p_{\textsc{data}}(y\mid\mathbf{x})~\|~ p_{\textsc{model}}(y\mid\mathbf{x}) \Big) \Big]
    \approx
    \frac{1}{N} \sum_{n=1}^N \mathcal{D}\Big( p_{\textsc{data}}(y\mid\mathbf{x}_n)~\|~ p_{\textsc{model}}(y\mid\mathbf{x}_n) \Big),
\end{align}
where $p_{\textsc{data}}(\cdot)$ is the true label distribution for the data point and $p_{\textsc{model}}(\cdot)$ is the model's distribution. This evaluation metric only reaches 0 when the model captures the true label distribution for every example. We use KL divergence and approximate the label distribution using an empirical distribution over the multiple label samples per input.

\begin{figure*}[!tb]
\centering
\begin{subfigure}{.49\textwidth}
\includegraphics[width=\linewidth]{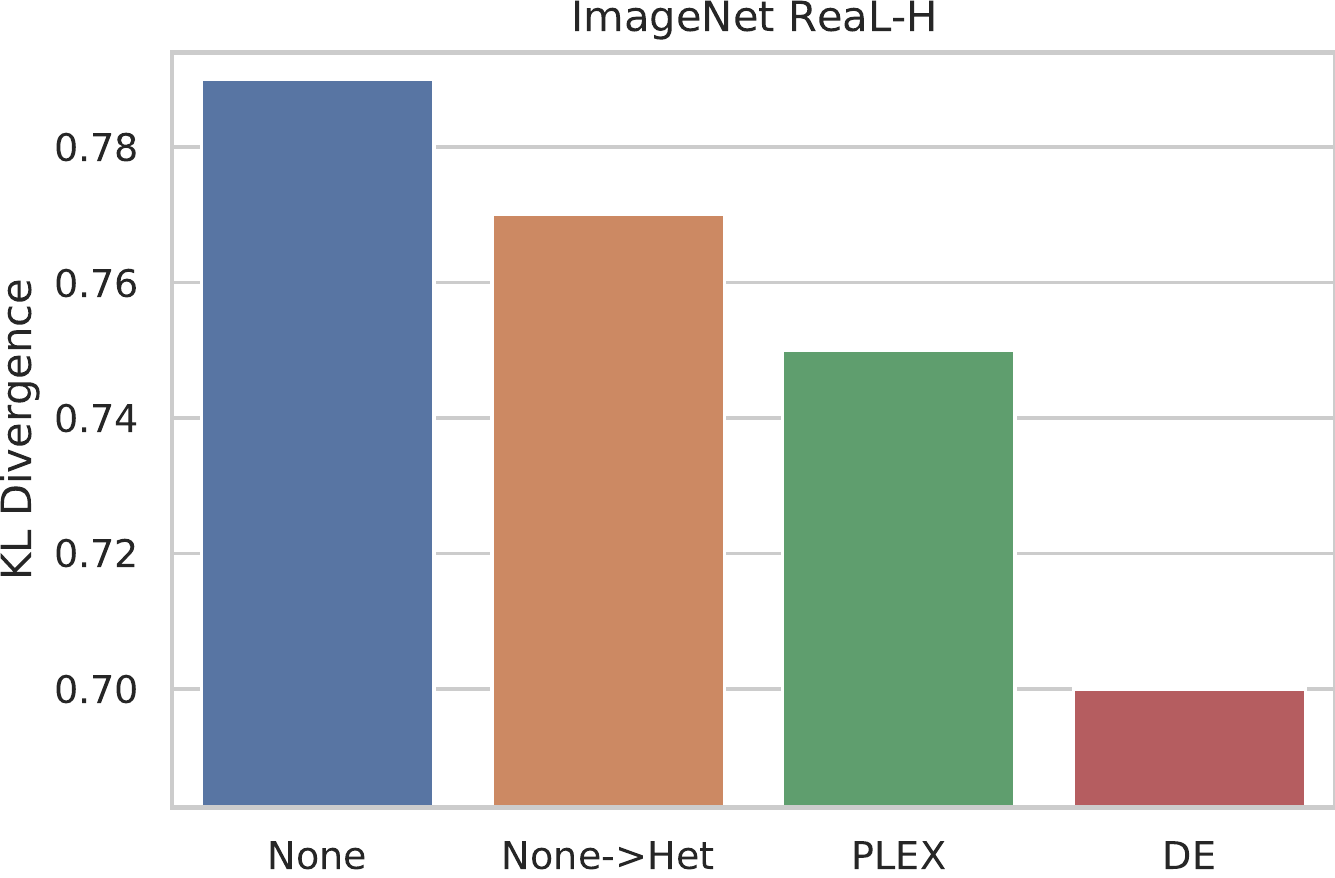}
\label{fig:imagenet_real}
\end{subfigure}
\begin{subfigure}{.49\textwidth}
\includegraphics[width=\linewidth]{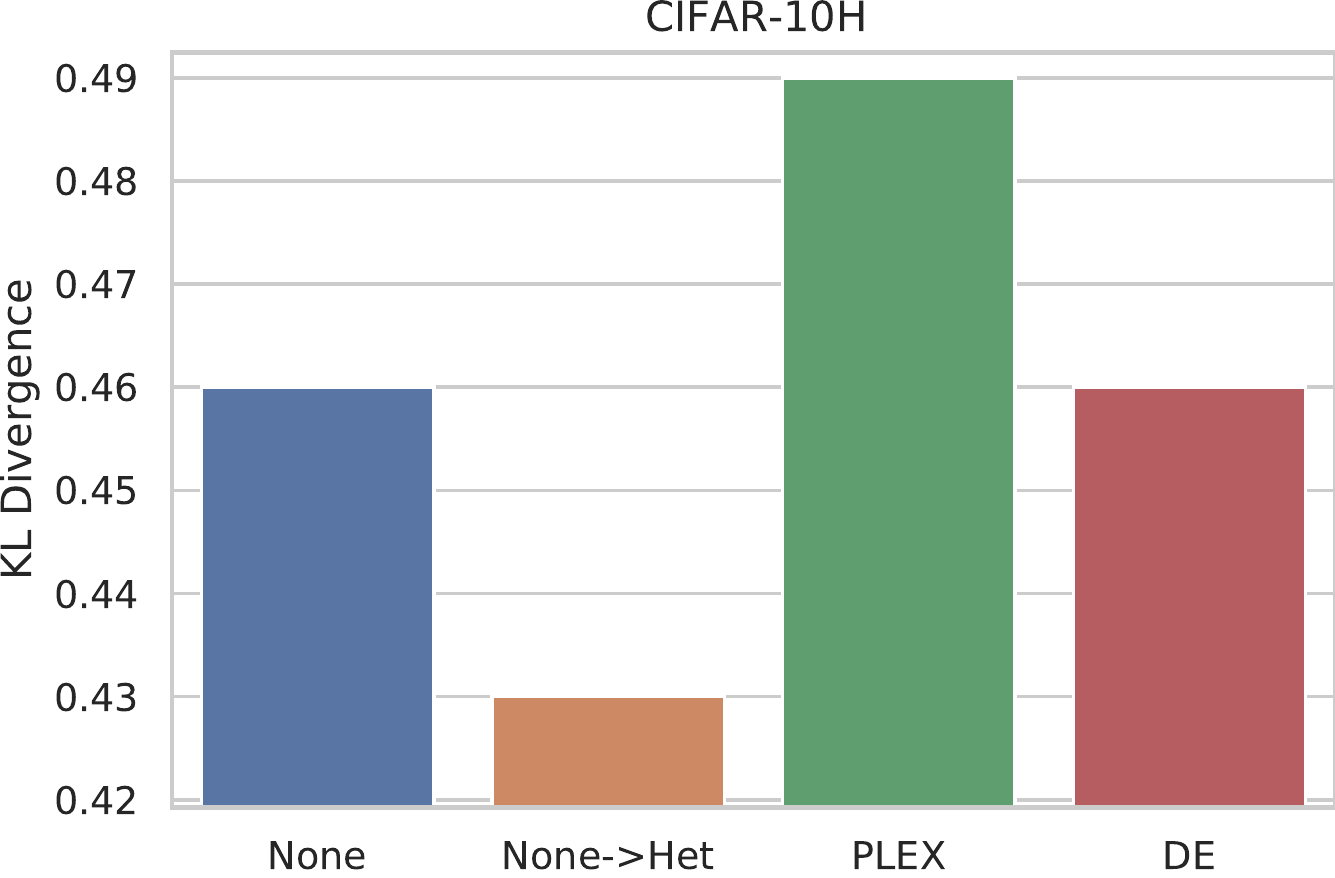}
\label{fig:cifar10h}
\end{subfigure}
\vspace{-4ex}
\caption{
KL divergence on ImageNet ReaL-H \textbf{(left)} and CIFAR-10H \textbf{(right)} for None, None upstream$\to$heteroscedastic downstream, BatchEnsemble upstream$\to$BatchEnsemble L + heteroscedastic downstream, and Deep Ensemble models.}
\label{fig:label_uncertainty}
\end{figure*}

\paragraph{Results} 
We observe in \Cref{fig:label_uncertainty} that the heteroscedastic models (Het in finetuning-only and Plex) outperform the None model.
Plex does best on ImageNet ReaL-H excluding DE which uses more compute, but Plex performs worse on CIFAR-10H.

\subsection{Robust Generalization}
\label{sub:robust-generalization}

An important aspect of model reliability is it's ability to make accurate predictions when the test data distribution changes.  In this section, we examine model robustness to different forms of data distribution shift.  We look at performance both in-distribution and under two out-of-distribution shifts: covariate shift and subpopulation shift.\footnote{%
There are 2 distribution shift types which we do not cover here: semantic (class) shift and label uncertainty. Semantic (class) shift is not possible to measure predictive performance under without an open vocabulary model such as recent image-text models \citep{clip}. Therefore we restrict studies of class shift to the open set recognition task (\Cref{sub:open-set-recognition}). Label uncertainty is covered in \Cref{sub:label-uncertainty}.
}

\subsubsection{In-Distribution Generalization}
\label{sub:ind-generalization}

\begin{tldr}
Ablation experiments show that pretraining does not help for diabetic retinopathy diagnosis on images, but the Plex model extensions do give a benefit.  Across language tasks, pretraining, model scale, and the Plex extensions all contribute to better in-domain generalization.
\end{tldr}

\begin{figure*}[!tb]
\centering
\includegraphics[width=0.85\linewidth]{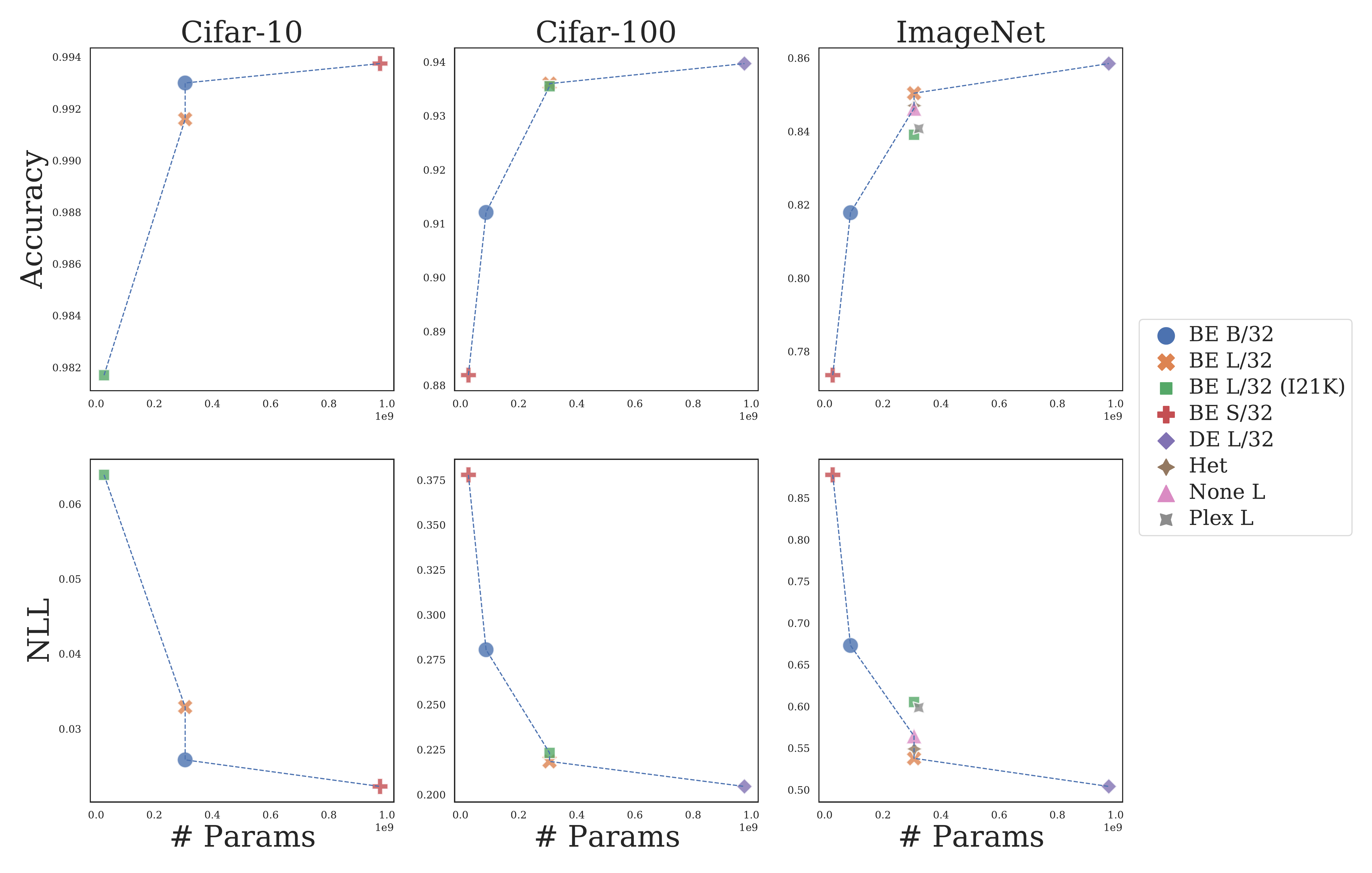}
\caption{%
Predictive performance on downstream datasets, with \textbf{(top)} accuracy and \textbf{(bottom)} negative log-likelihood over CIFAR-10, CIFAR-100, and ImageNet respectively as a function of the model scale (number of model parameters).  The blue dotted line is the Pareto frontier.  Note, while number of parameters is a useful comparison of model scale (e.g. memory footprint) it doesn't capture all aspects of scale such as computational overhead, training time, etc.
}
\label{fig:in-distribution-vision}
\end{figure*}

We look at accuracy and NLL across the in-distribution test splits of each dataset after finetuning on their respective training splits.

\paragraph{Vision Tasks}
For RETINA in \Cref{table:retina-sel-pred_id}, we compare Plex to the best-performing ResNet-50 baseline results based on a wide array of uncertainty quantification methods \citep{gal2016dropout,blundell2015mfvi,Rudner2021fsvi,Rudner2021sfsvi,dusenberry2020efficient,farquhar2020radial} reported in \citet{band2021benchmarking} and find that pretraining on neither I21K or JFT results in an improvement in predictive performance across all evaluation metrics compared to the state-of-the-art ResNet-50 trained from scratch.
This failure may be due to the fact that the retina scans used for the diagnosis tasks differ significantly in appearance from the images in the pretraining datasets and as such may be too dissimilar to convey meaningful inductive biases into the finetuned neural network.  Nevertheless, the Plex extensions do still provide a benefit over the pretrained model.
See \Cref{fig:in-distribution-vision} for CIFAR and ImageNet.

\begin{table*}[!tb]
\centering
\begin{tabular}{lcccccc}
\toprule
& SoTA   & 
SoTA   &
Plex L  &
Plex L  & 
None L & 
None L \\
&  (Ensemble) & 
 (Single Model) &
 (JFT) &
(I21K) & 
(JFT) & 
 (I21K)\\
\toprule
Accuracy & 
\textbf{91.6\%} &
90.9\% & 
90.0\% &
89.8\% &
88.7\% &
89.8\%  \\
NLL &
\textbf{0.25} &
0.29 &
0.28 &
0.28 &
0.30 &
0.32  \\
AUROC &
\textbf{92.5\%} &
91.4\% &
-- &
88.9\% &
-- &
89.8\%  \\
AUPRC &
\textbf{84.4\%} &
82.6\% &
-- &
78.1\% &
-- &
79.8\% \\
\bottomrule
\end{tabular}
\caption{%
    Results on the RETINA Country Shift task (for in-distribution evaluation). 
    The SoTA (Ensemble) outperforms Plex L pretrained on both ImageNet-21K or JFT on all evaluation metrics but ECE, where Plex L pretrained on ImageNet-21K performs best.
}
\label{table:retina-sel-pred_id}
\end{table*}

\paragraph{Language Tasks} In the Appendix, \Cref{fig:t5-method-compare} summarizes the performance of different uncertainty methods, and \Cref{fig:t5-arch-compare} evaluates the models' performance across three different scales (i.e., T5$_{\texttt{small}}$, T5$_{\texttt{base}}$, T5$_{\texttt{large}}$). Finally, \Cref{fig:t5-rank-method,fig:t5-rank-arch} in the Appendix visually summarize the relative performance ranking across methods and scales. Comparing across methods, uncertainty methods (specifically, BE, DE, GP, and their combinations) provides improved generalization performance when compared to the None baseline.
Specifically, DE+GP provides the strongest performance across almost all tasks. Among the more efficient methods, Plex (BE+GP) consistently ranked among the top performing methods, while the performance of GP, MCD and BE varies across the datasets (\Cref{fig:t5-method-compare}). Comparing across model scale, the methods' generalization performance generally improves as the model scale increases, with Plex (BE+GP) the method benefiting the most strongly from scaling, delivering on average the strongest performance at T5$_{\texttt{large}}$.

\subsubsection{Covariate shift}
\label{sub:covariate-shift}

\begin{tldr}
In general, Plex significantly improves performance across metrics under covariate shift, even when it doesn't outperform on in-distribution test data. Model size (larger models perform better under covariate shift) and pretraining data are consistently major factors.
\end{tldr}

In order to be robust under covariate shift, a model should be able to reliably make correct predictions on noisy, corrupted, and otherwise distribution-shifted inputs. In this section, we evaluate robustness using a variety of datasets that exhibit different types of covariate shift.

\begin{table*}[!tb]
\centering
\begin{tabular}{lcccccc}
\toprule
& SoTA   & 
SoTA   &
Plex L  &
Plex L  & 
None L & 
None L \\
&  (Ensemble) & 
 (Single Model) &
 (JFT) &
(I21K) & 
(JFT) & 
 (I21K)\\
\toprule
Accuracy  &
87.6\%  &
86.8\% &
\textbf{88.8\%} &
88.4\% &
85.5\% &
87.2\% \\
NLL &
0.92 &
1.07 &
0.45 &
\textbf{0.44} &
0.49 &
0.59 \\
AUROC &
94.1\% &
94.0\% &
-- &
\textbf{95.2\%} &
-- &
94.7\% \\
AUPRC &
88.3\% &
88.8\% &
-- &
\textbf{90.8\%} &
-- &
88.9\% \\
\bottomrule
\end{tabular}
\caption{%
    Predictive performance on the RETINA Country Shift task (for out-of-distribution evaluation). 
    Plex L pretrained on either ImageNet-21K or JFT outperforms the SoTA Ensemble on all evaluation metrics.
}
\label{table:retina-sel-pred_ood}
\end{table*}

\paragraph{RETINA under Country Shift}
RETINA's Country Shift task is concerned with diagnosing diabetic retinopathy from retina scans obtained using different medical equipment and a different patient population.
In \Cref{table:retina-sel-pred_ood}, we compare Plex to the best-performing ResNet-50 baseline results based on a wide array of methods to improve uncertainty quantification~\citep{gal2016dropout,blundell2015mfvi,Rudner2021fsvi,Rudner2021sfsvi,dusenberry2020efficient,farquhar2020radial} reported in \citet{band2021benchmarking} and find that pretraining on either I21K or JFT results in an improvement in predictive performance across all evaluation metrics compared to the state-of-the-art ResNet-50 trained from scratch.
This result is in contrast with corresponding results on the Country Shift \emph{in-distribution} task, for which the best-performing models trained from scratch outperform all pretrained models.  In this case, pretraining does improve generalization under covariate shift even when training from scratch yields better predictive performance on in-distribution evaluation tasks.  The Plex entensions also yield a significant improvement both in accuracy and log-likelihood.

\begin{figure*}[!tb]
\centering
\includegraphics[width=0.85\linewidth]{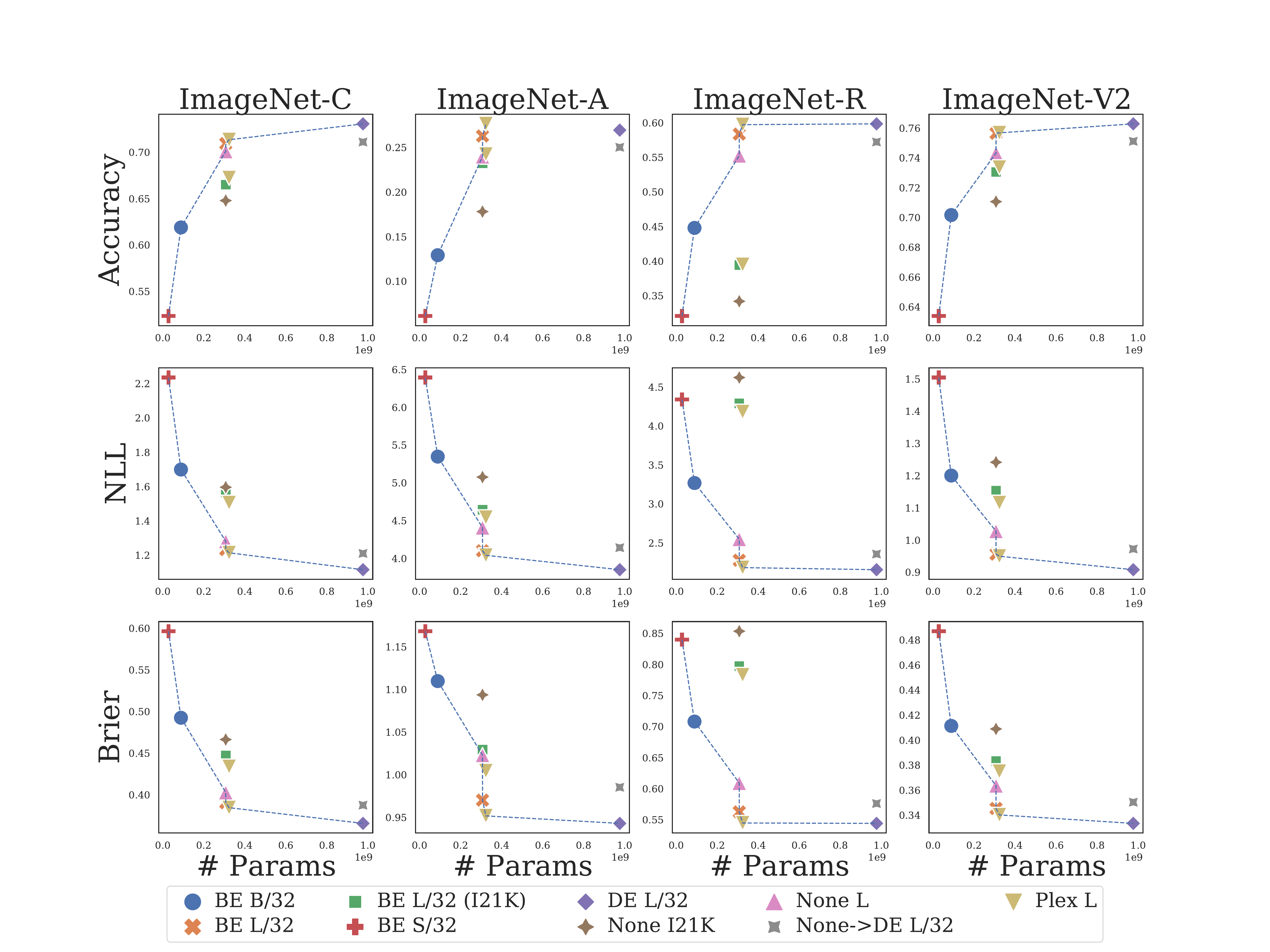}
\caption{%
Predictive performance with, starting from top row, \textbf{(1)} accuracy,
\textbf{(2)} NLL, and \textbf{(3)} Brier score
across 4 out-of-distribution datasets (columns): ImageNet-C, ImageNet-A, ImageNet-R, and ImageNet-V2 respectively as a function of model scale (number of parameters).
Plex L trained on JFT consistently outperforms other methods and pretraining datasets, and increasing model scale consistently leads to improved metric performance under corrupted shifts. 
}
\label{fig:imagenet-c}
\end{figure*}

\paragraph{ImageNet Covariate Shift} \Cref{fig:imagenet-c} displays performance over multiple types of covariate shift as a function of model scale (number of parameters): 
image corruptions (ImageNet-C \citep{hendrycks_benchmarking_2019}),
natural adversarial examples (ImageNet-A \citep{hendrycks2021natural}),
artistic renditions (ImageNet-R \citep{hendrycks2021many}),
and ``natural'' shift obtained by following ImageNet's data collection procedure to obtain a new test set (ImageNet-V2 \citep{recht2019imagenet,taori2020measuring}).
%
%
We find that BE ViT L outperforms None methods, that pretraining on JFT outperforms models trained on other pretraining datasets, and that increasing model scale consistently leads to improved performance under shifts across metrics. Unlike in some sections, for performance under shifts, we consistently find that pretraining on JFT outperforms that on ImageNet-21K.

\begin{figure*}[!tb]
\centering
\includegraphics[width=0.7\linewidth]{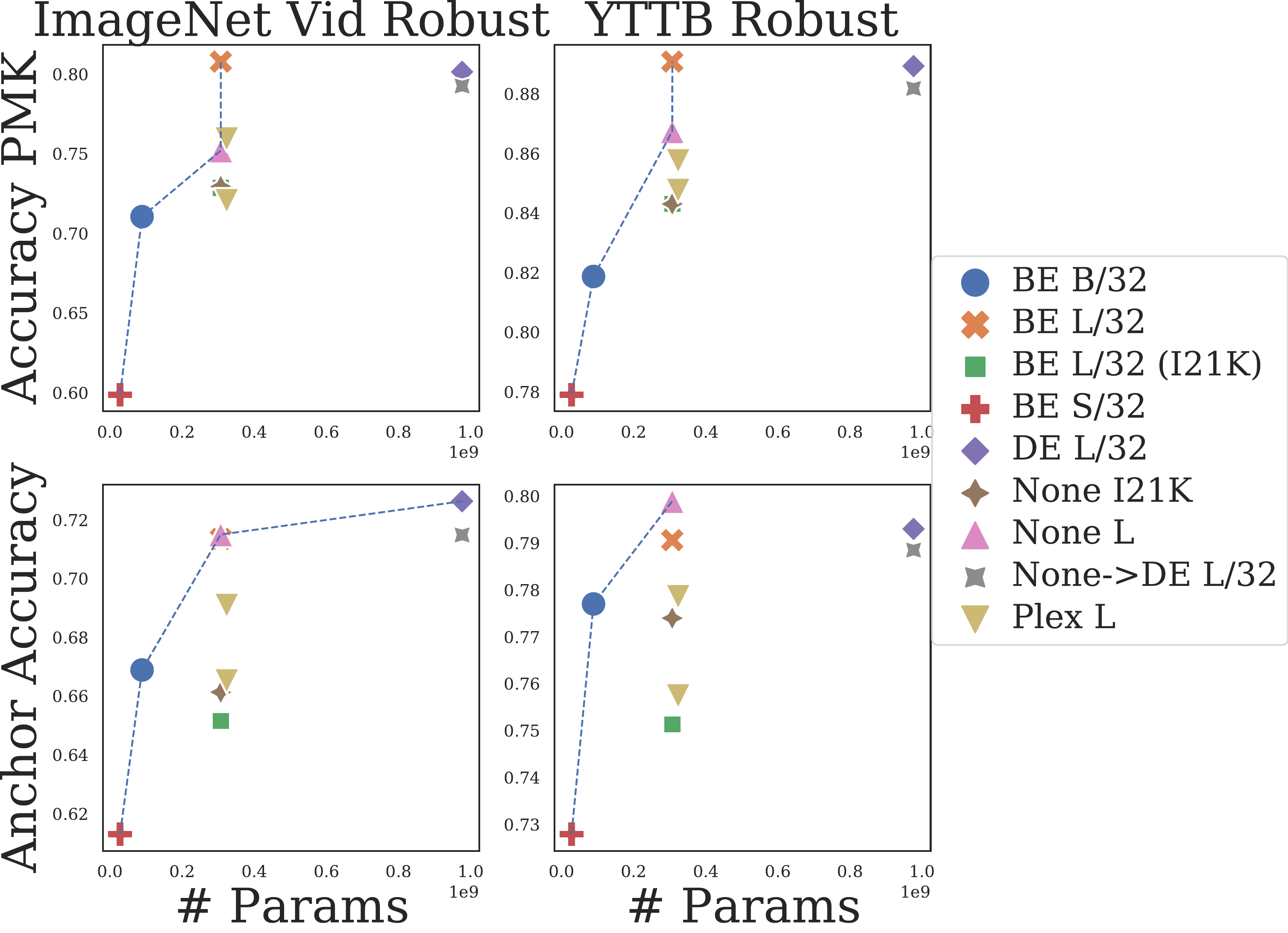}
\caption{%
Stability measured with Anchor Accuracy and Accuracy PMK on ImageNet-Vid-Robust and YTBB-Robust. We find that on the main pm-k accuracy, None, Plex L and BE L models perform similarly, although BE L outperforms slightly.  On anchor accuracy, None and BE perform similarly.  We also see that performance improves with model size.
}
\label{fig:imagenet-vid-robust-ytbb}
\end{figure*}

\paragraph{Stability} ImageNet-Vid-Robust and YTBB-Robust \citep{shankar2021image} are benchmark datasets for measuring robustness to natural perturbations in images by using subsequent frames from video sources with human curation. On the robust accuracy (pm-k accuracy) metric, in \Cref{fig:imagenet-vid-robust-ytbb}, None and BE L models perform equally well, and performance improves with model size. Models pretrained on JFT outperform those pretrained on I21K, with BE B JFT even outperforming None L I21K and BE L I21K.

\begin{figure*}[t]
\begin{subfigure}{.3\textwidth}
\centering
\includegraphics[width=\textwidth]{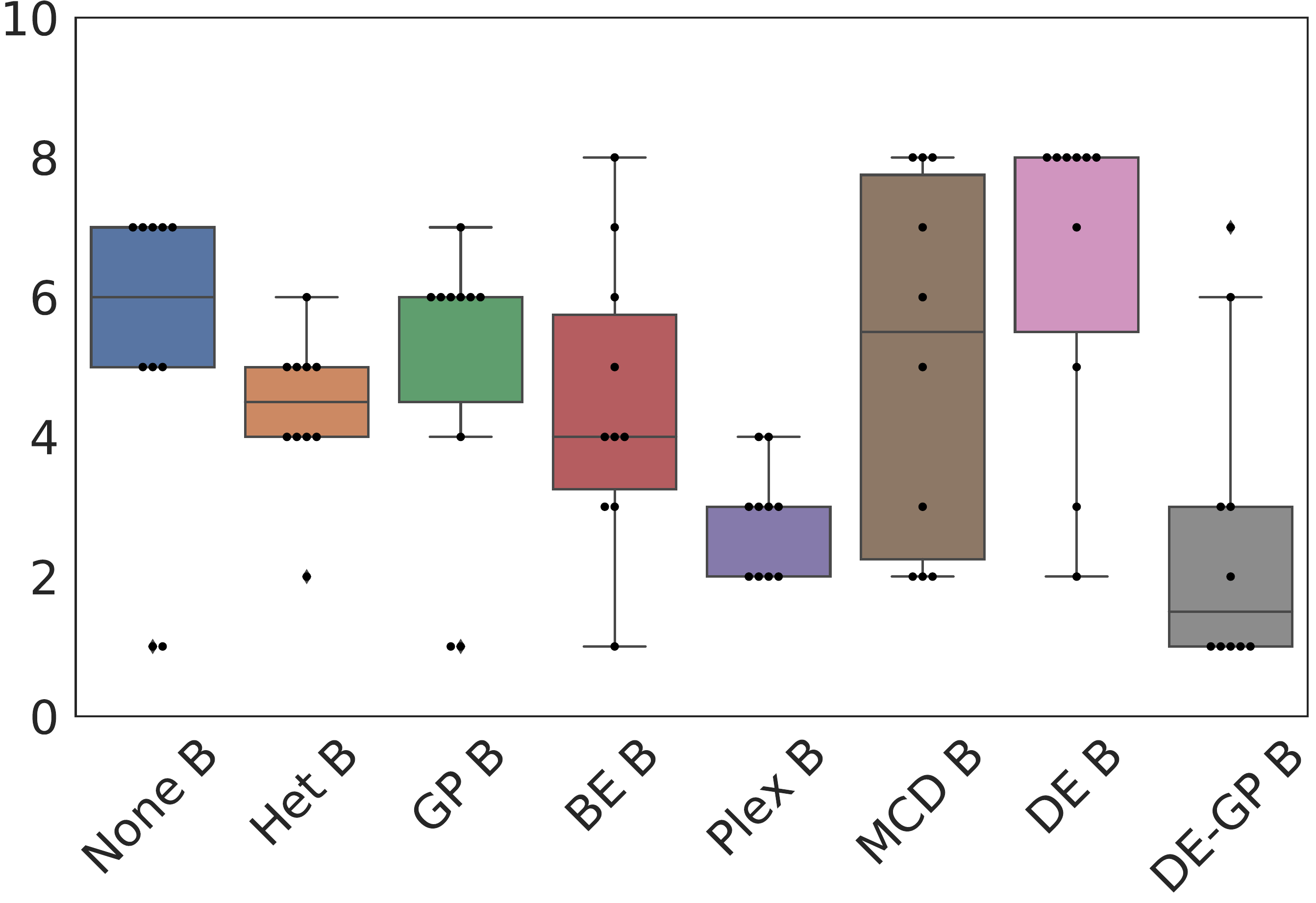}
\caption{Generalization, by method.}
\label{fig:t5-ood-generalization-methods}
\end{subfigure}
\hfill
\begin{subfigure}{.3\textwidth}
\centering
\includegraphics[width=\textwidth]{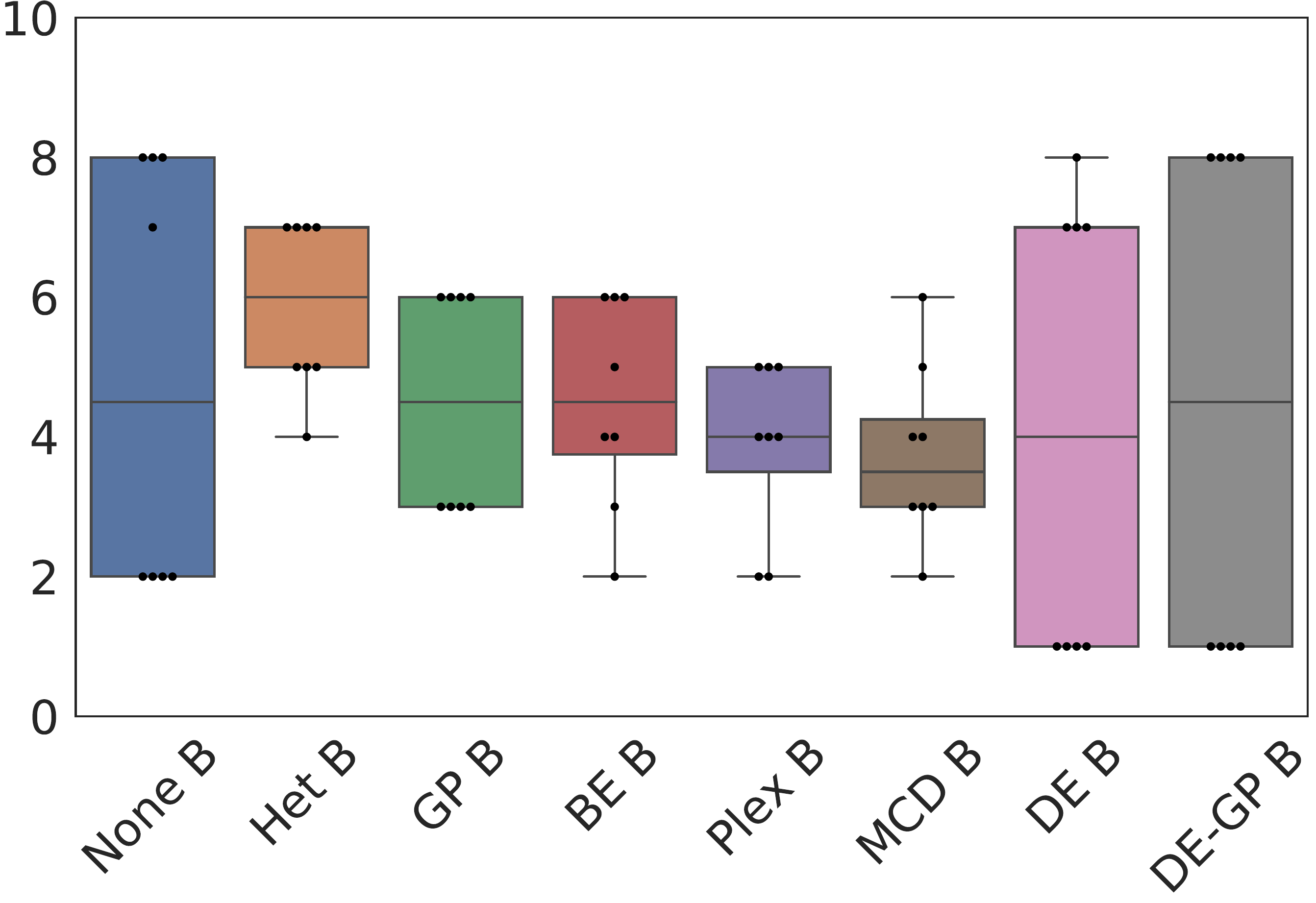}
\caption{Sel. Predict., by method.}
\label{fig:t5-ood-collaboration-methods}
\end{subfigure}
\hfill
\begin{subfigure}{.3\textwidth}
\centering
\includegraphics[width=\textwidth]{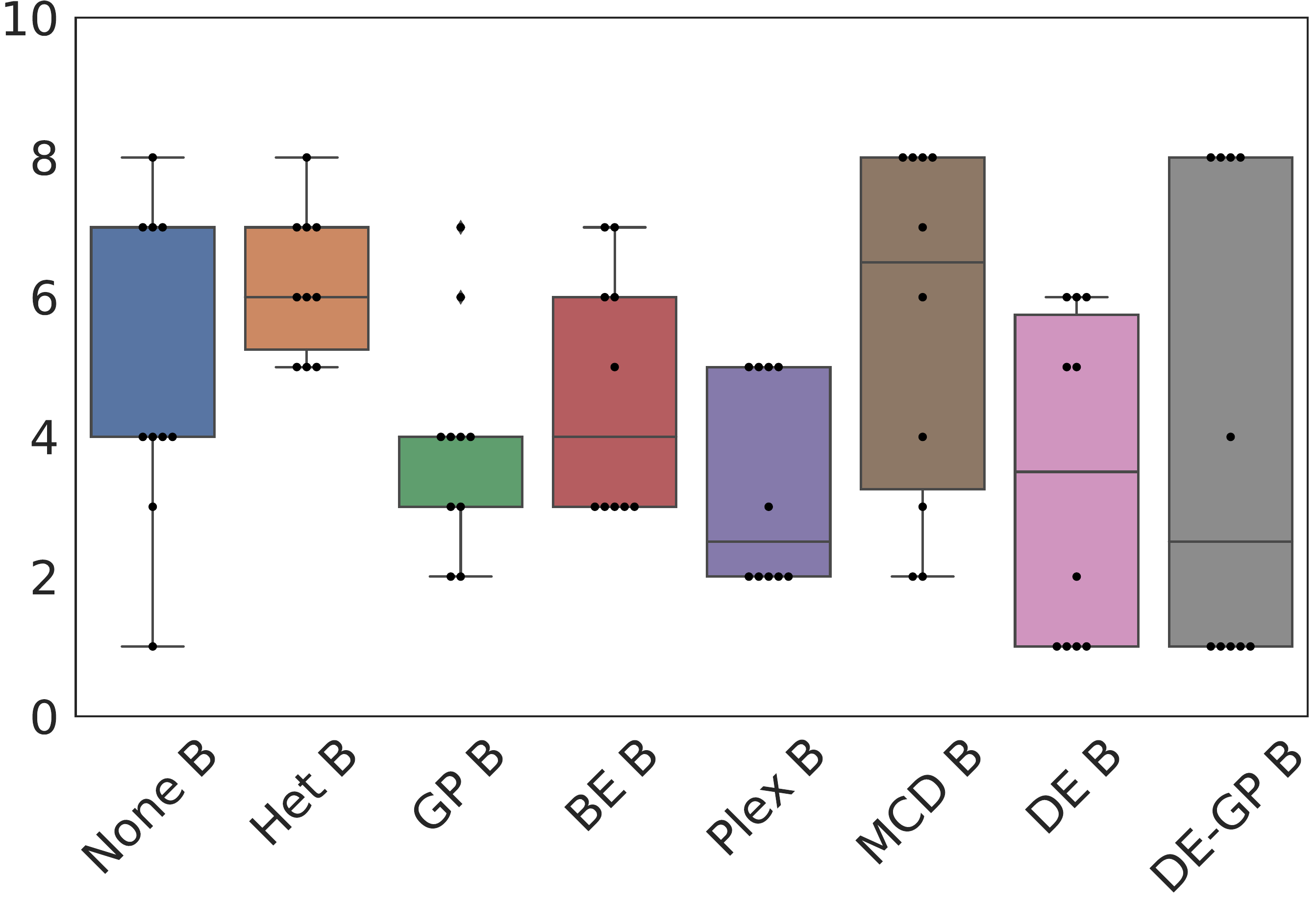}
\caption{Calibration, by method.}
\label{fig:t5-ood-calibration-methods}
\end{subfigure}
\\
\begin{subfigure}{.3\textwidth}
\centering
\includegraphics[width=\textwidth]{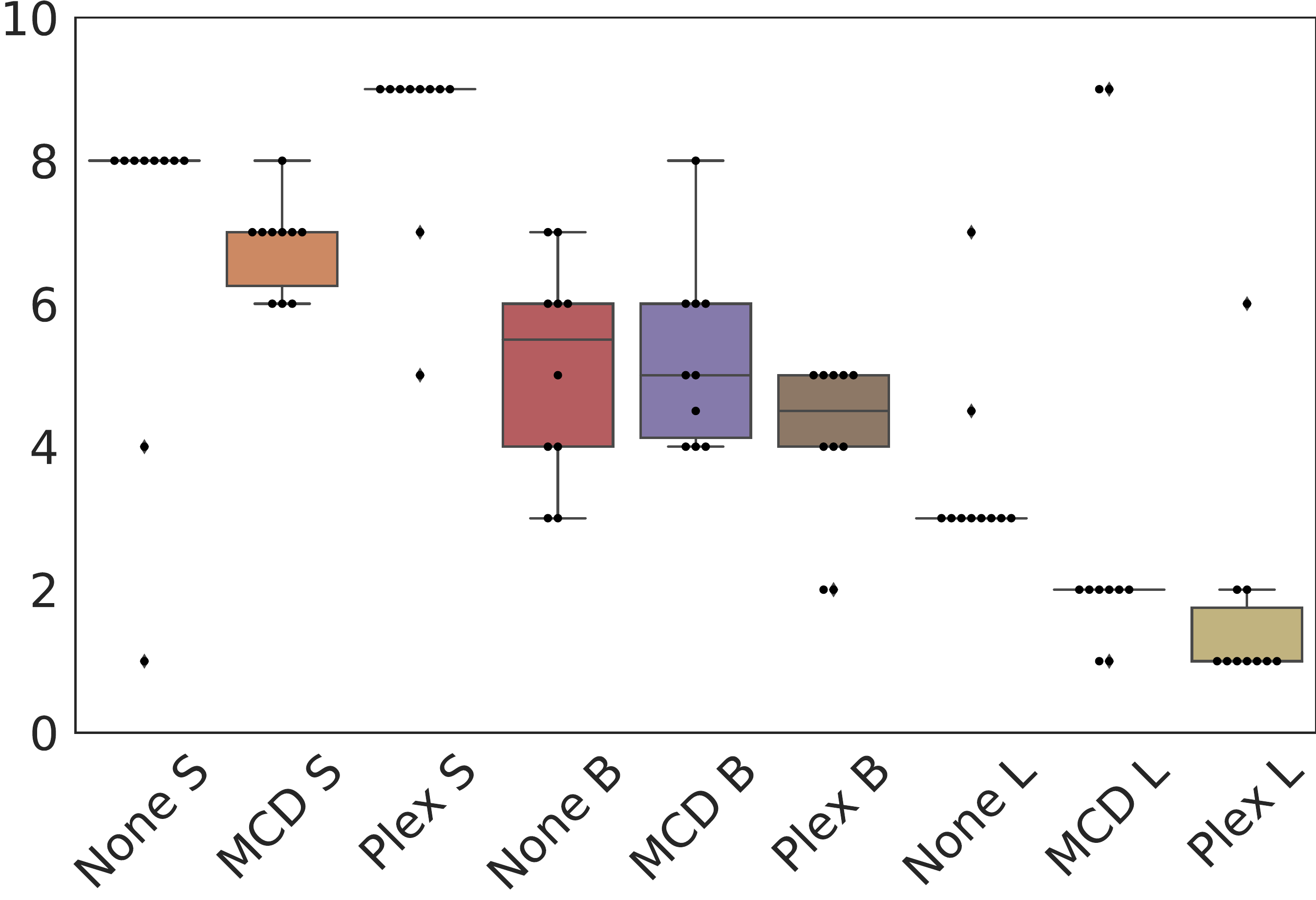}
\caption{Generalization, by size.}
\label{fig:t5-ood-generalization-size}
\end{subfigure}
\hfill
\begin{subfigure}{.3\textwidth}
\centering
\includegraphics[width=\textwidth]{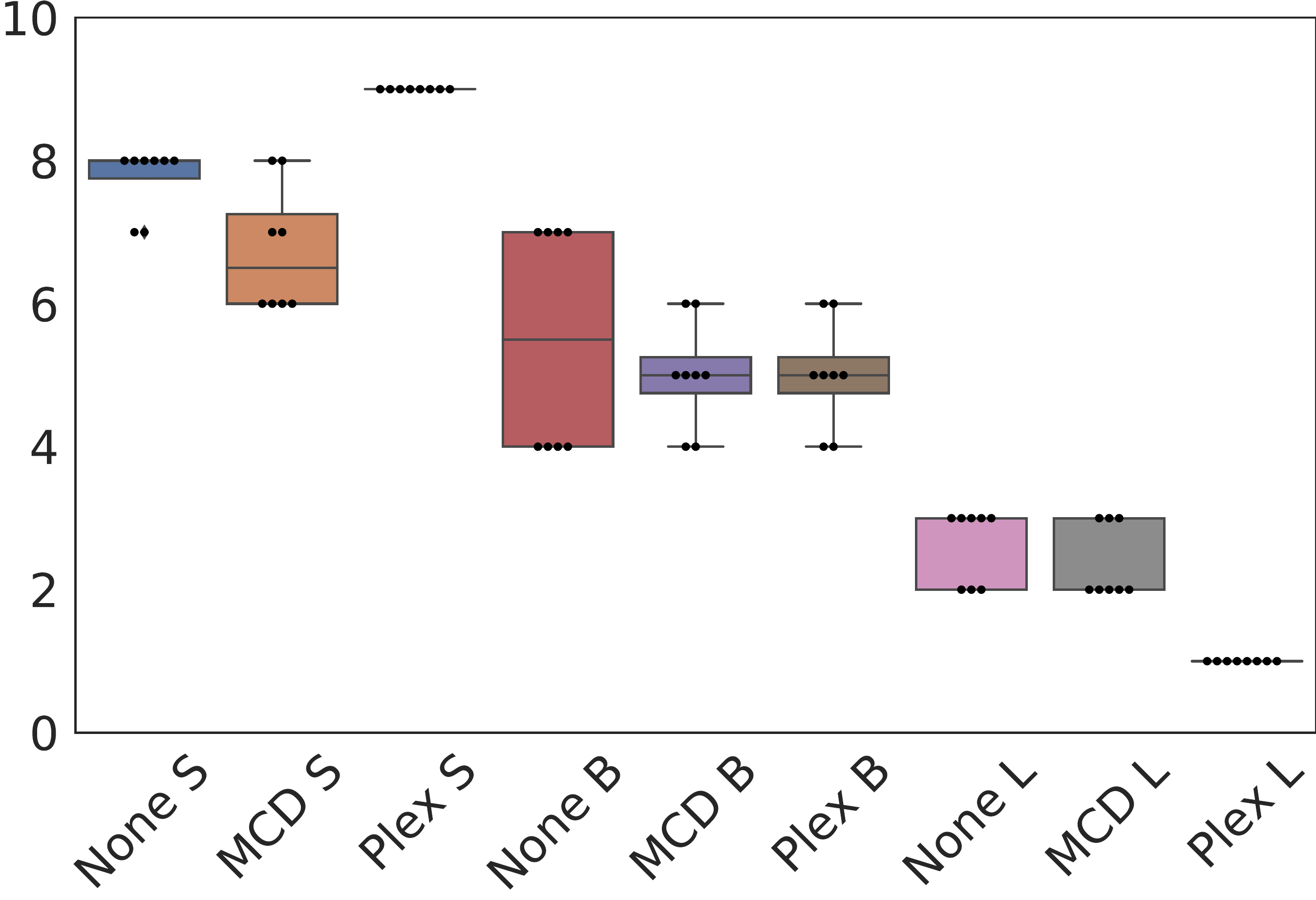}
\caption{Sel. Predict., by size.}
\label{fig:t5-ood-collaboration-size}
\end{subfigure}
\hfill
\begin{subfigure}{.3\textwidth}
\centering
\includegraphics[width=\textwidth]{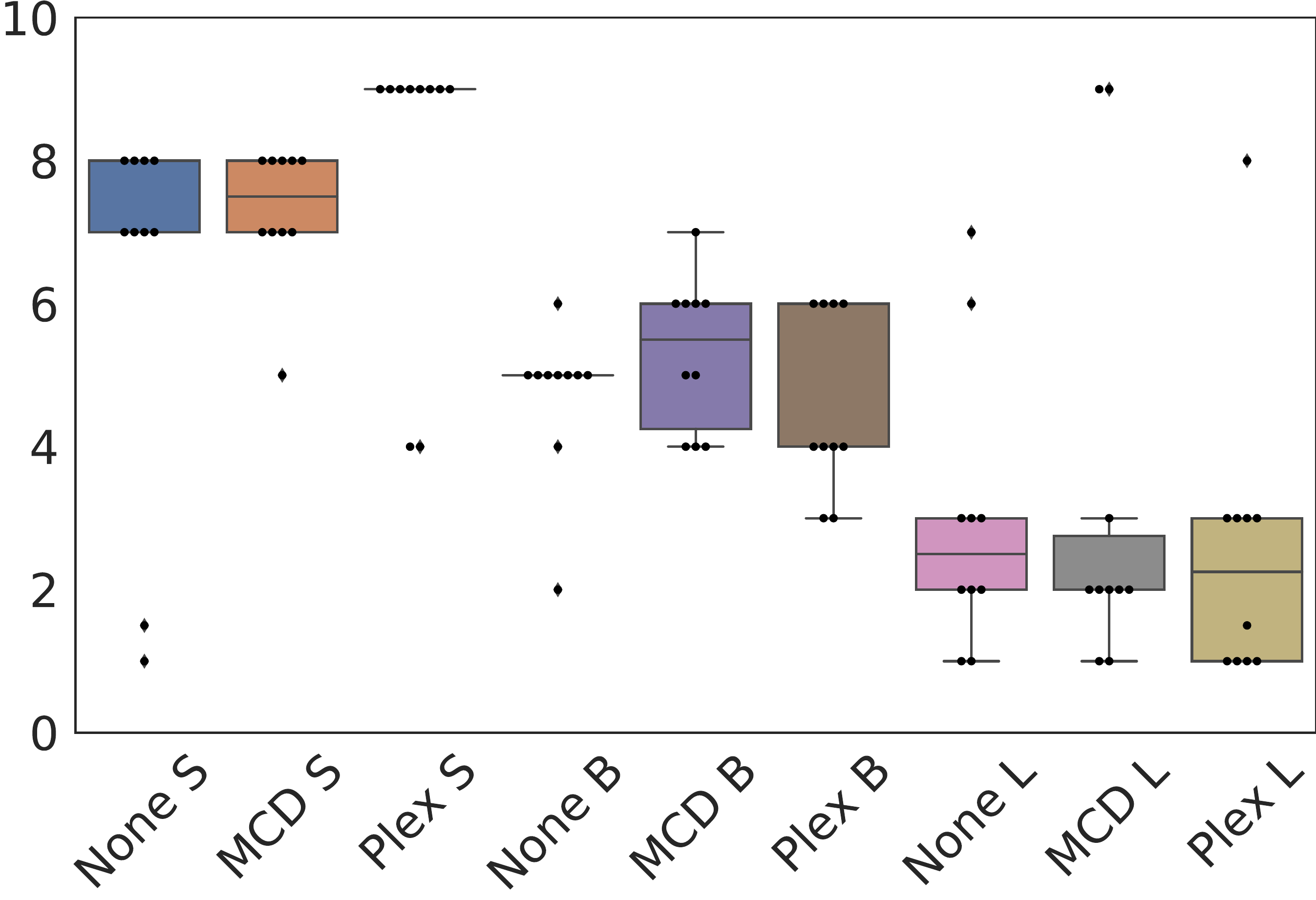}
\caption{Calibration, by size.}
\label{fig:t5-ood-calibration-size}
\end{subfigure}
\vspace{-1.5ex}
\caption{T5-Plex model's ranking performance in generalization under domain shift compared to different methods (\ref{fig:t5-ood-generalization-methods}-\ref{fig:t5-ood-calibration-methods}) and across architecture sizes (\ref{fig:t5-ood-generalization-size}-\ref{fig:t5-ood-calibration-size}).
}
\label{fig:t5-ood-rank-compare}
\end{figure*}

\paragraph{Language Style and Topical Drift} In natural language applications, it is common for a trained NLP model to be deployed to an environment that exhibits significant style and topical drift when compared to training data. For example, a NLI model trained on written literature is used to analyze in-person dialogues, or a toxic comment detection model trained on U.S. web forums is deployed to non-U.S. news websites \citep{williams2017broad, kivlichan2021measuring}. To understand the robustness of large uncertainty models with respect to these common language drifts, we evaluate the models' prediction performance on out-of-distribution splits of the NLI and Toxic Comment tasks (MNLI-mismatched and Civil Comments, respectively). Specifically, MNLI-mismatched represents a low-degree shift scenario where models trained on written and spoken genres (e.g., government report, fiction, phone conversations) are tested on different subtypes of similar genre (e.g., fundraising letters, in-person conversation, 9/11 calls). On the other hand, Civil Comments represents a high-degree shift scenario where models trained on web forum conversations before 2015 were tested on news websites comments from a different period (2015-2017). \Cref{fig:t5-ood-rank-compare} compares the performance of nine models based on T5$_{\texttt{base}}$, and also that of three representative efficient methods (None, MCD and Plex) across model scales (i.e., T5$_{\texttt{small}}$, T5$_{\texttt{base}}$, T5$_{\texttt{large}}$) (see Appendix Table \ref{fig:t5-rank-method}-\ref{fig:t5-rank-arch} for detailed results). As shown, the models' performance in out-of-distribution generalization correlates well with their in-domain performance, with DE-GP being the most competitive ensemble model, and BE / Plex the most competitive efficient method for MNLI, and MCD the most efficient method for Toxic Comments. Comparing between model scales, larger models in general lead to stronger out-of-distribution performance, with Plex being the most competitive method for moderate-to-large size models (T5$_{\texttt{base}}$, T5$_{\texttt{large}}$).

\subsubsection{Subpopulation shift} 
\label{sub:subpopulation-shift}

\begin{tldr}
Plex improves generalization under subpopulation shift relative to models omitting pretraining, efficient ensembling, or last layer changes, for both language and vision datasets.
\end{tldr}

We next study performance on shifts where data is composed of subpopulations which may be rare or unseen during training. A common assumption is that subpopulation shift data is drawn from a distribution of distributions: subpopulations are drawn from a population distribution, and each subpopulation has its own data-generating distribution \citep{santurkar2020breeds,yuan2022we}. Since data in practice is often generated by individuals or groups who may have different data distributions, and models ideally generalize to unseen individuals and groups, this setting can be used to measure a natural notion of reliability. We study Plex under multiple ways of partitioning data into subpopulations.


\begin{figure}[t]
\centering
\includegraphics[width=\textwidth]{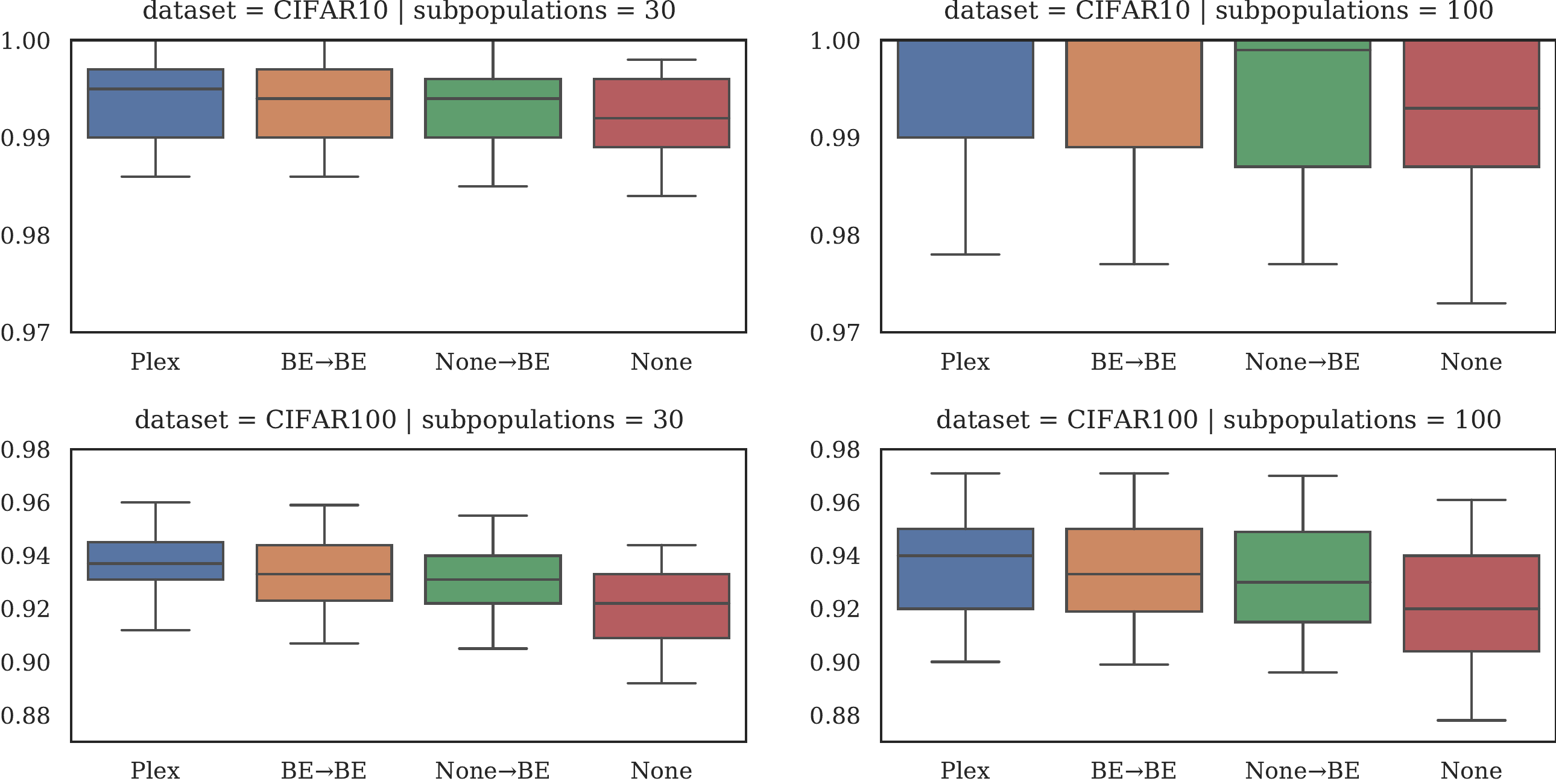}
\caption{%
Accuracy over subpopulations for CIFAR-10 and CIFAR-100. For each setting, we display the 5th, 25th, 50th, 75th, and 95th percentile accuracy for ViT-Plex L among the subpopulations. See \Cref{fig:radar} for comparison with no pretraining, which performs much worse.
}
\label{fig:subpopl_ablations}
\end{figure}

\paragraph{Semantic Partitioning.} One method for producing data with subpopulation shift is partitioning a standard dataset into subpopulations such that each subpopulation contains semantically similar examples. We leverage the Semantically Partitioned CIFAR10 and CIFAR100 datasets from \citet{yuan2022we} to study how Plex performs on new subpopulations. In particular, we measure classification accuracy with 30 and 100 subpopulations unseen during training, studying how accuracy varies among subpopulations.

From \Cref{fig:radar}, Plex greatly outperforms the baseline presented in \citet{yuan2022we} for tail subpopulation shift accuracy (we report 25th percentile among subpopulations). The baseline does not leverage large-scale pretraining or other changes Plex introduces. To better understand how Plex's ingredients lead to improvements in accuracy under subpopulation shift, we perform a series of ablations shown in \Cref{fig:subpopl_ablations}.
Comparing Plex to training with BatchEnsemble for both pretraining and finetuning (`BE→BE'), the latter produces a slight degradation in performance for both median and tail subpopulations, especially for CIFAR100, where performance is less saturated. This indicates that last layer changes offer some improvement but do not fully explain Plex's improvement over baseline in this setting. Looking at `None→BE', standard pretraining slightly degrades performance relative to BatchEnsemble pretraining, but using BatchEnsemble for finetuning still offers a significant boost over completely standard training (`None'); ensembling appears to be more important in this setting than last layer changes. Comparing Plex and `None', we see a significant difference in performance, but the difference between Plex's performance and that of the baseline presented in \Cref{fig:radar} remains much larger. This indicates that the difference in pretraining is an important driver of Plex's improvements on unseen and long-tail subpopulations, since the baseline does not pretrain on large-scale data.

\begin{figure*}[ht]
\begin{subfigure}{.3\textwidth}
\centering
\includegraphics[width=\textwidth]{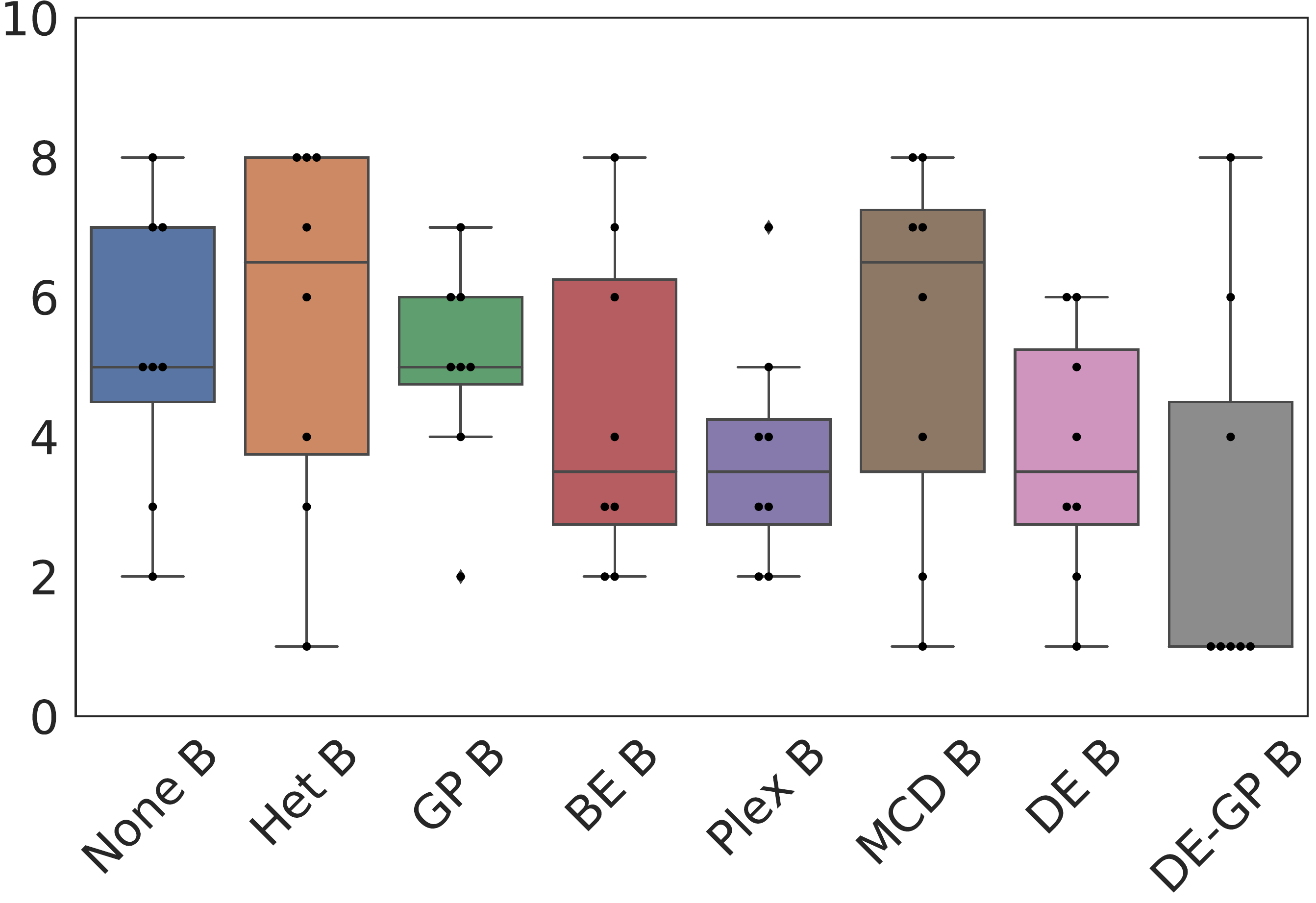}
\caption{Generalization, by method.}
\label{fig:t5-subpop-in-generalization-methods}
\end{subfigure}
\hfill
\begin{subfigure}{.3\textwidth}
\centering
\includegraphics[width=\textwidth]{figures/t5/collaboration-sub-population-uncertainty-method}
\caption{Sel. Predict., by method.}
\label{fig:t5-subpop-collaboration-methods}
\end{subfigure}
\hfill
\begin{subfigure}{.3\textwidth}
\centering
\includegraphics[width=\textwidth]{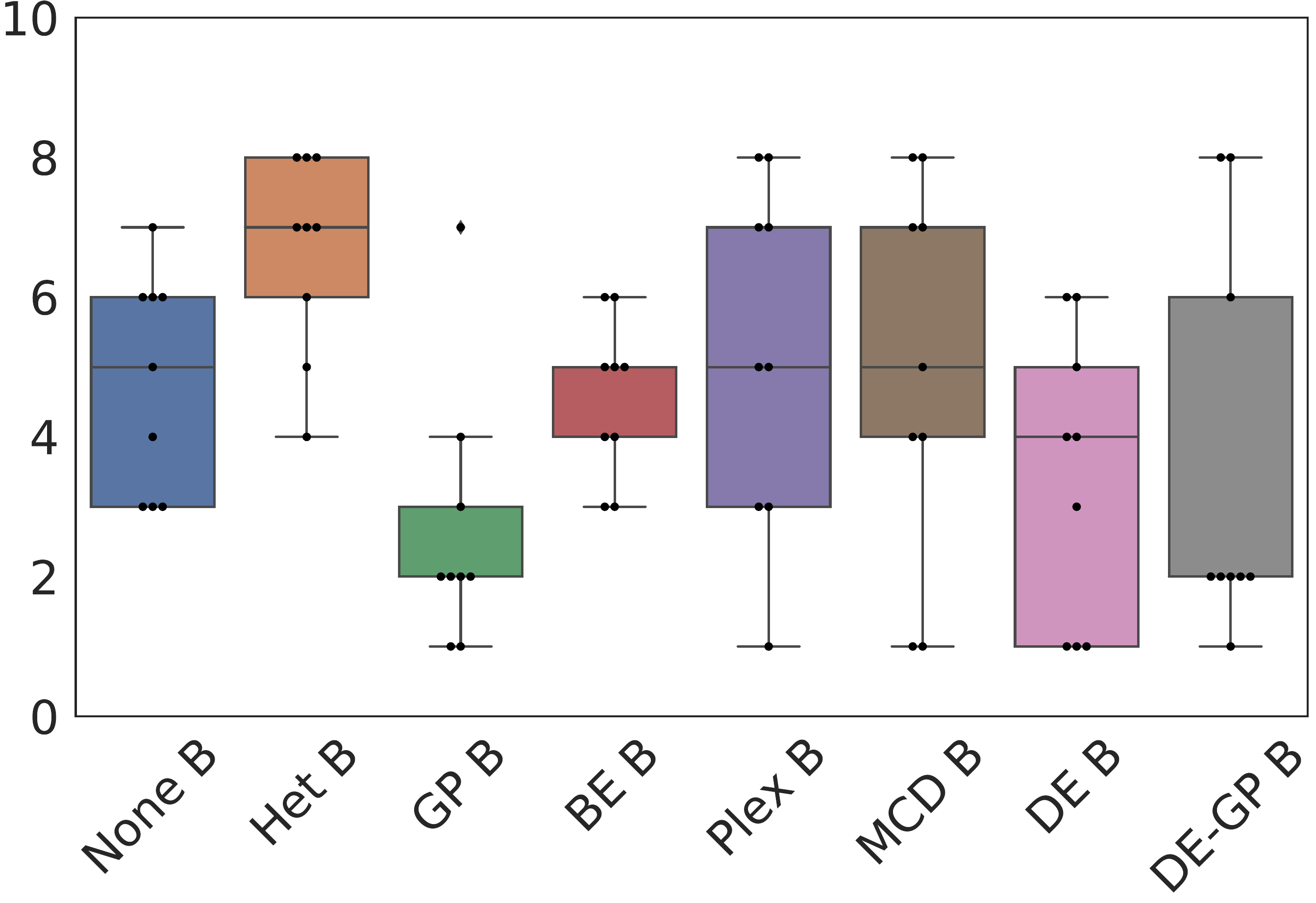}
\caption{Calibration, by method.}
\label{fig:t5-subpop-calibration-methods}
\end{subfigure}
\begin{subfigure}{.3\textwidth}
\centering
\includegraphics[width=\textwidth]{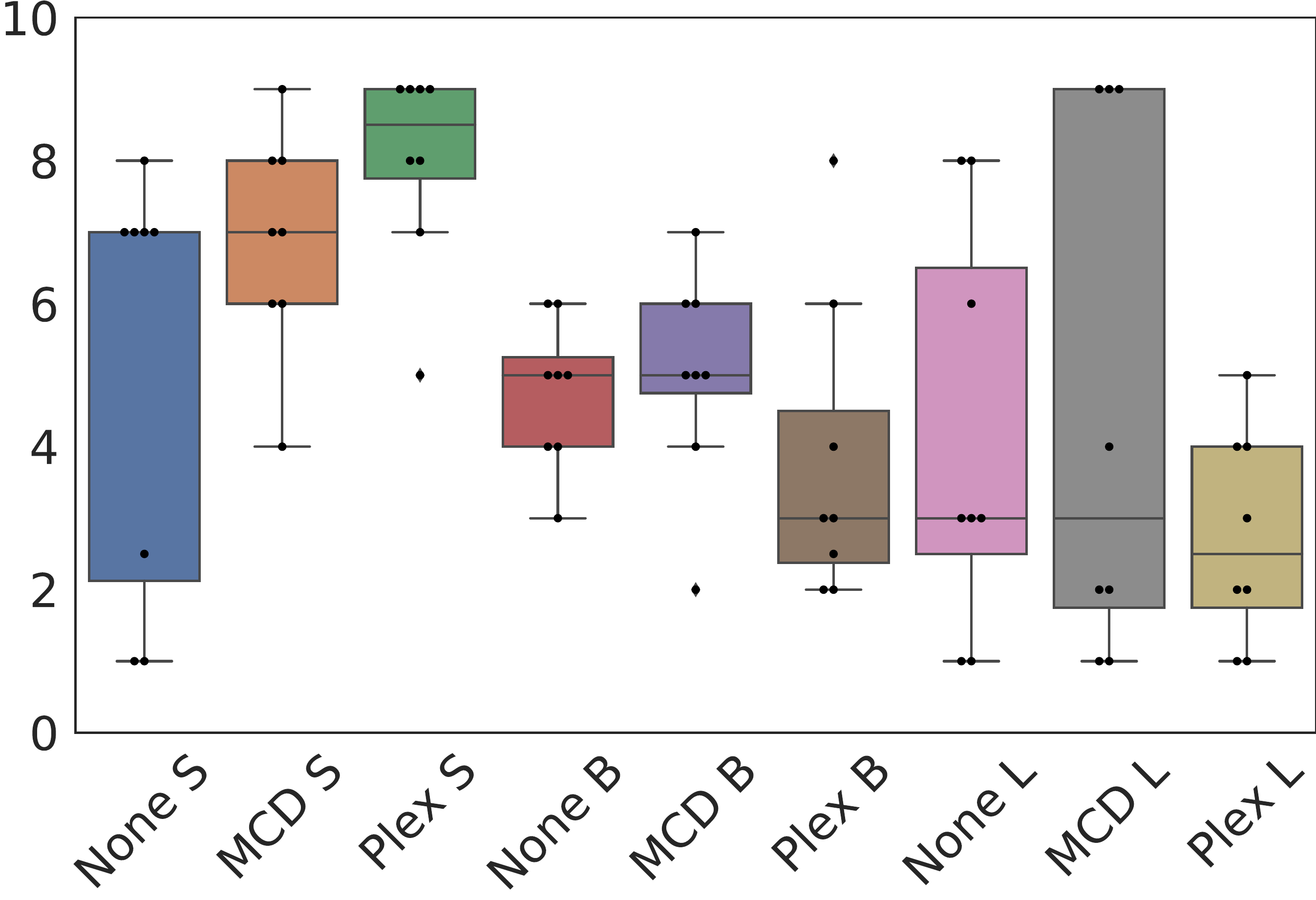}
\caption{Generalization, by size.}
\label{fig:t5-subpop-generalization-size}
\end{subfigure}
\hfill
\begin{subfigure}{.3\textwidth}
\centering
\includegraphics[width=\textwidth]{figures/t5/collaboration-sub-population-architecture-size}
\caption{Sel. Predict., by size.}
\label{fig:t5-subpop-collaboration-size}
\end{subfigure}
\hfill
\begin{subfigure}{.3\textwidth}
\centering
\includegraphics[width=\textwidth]{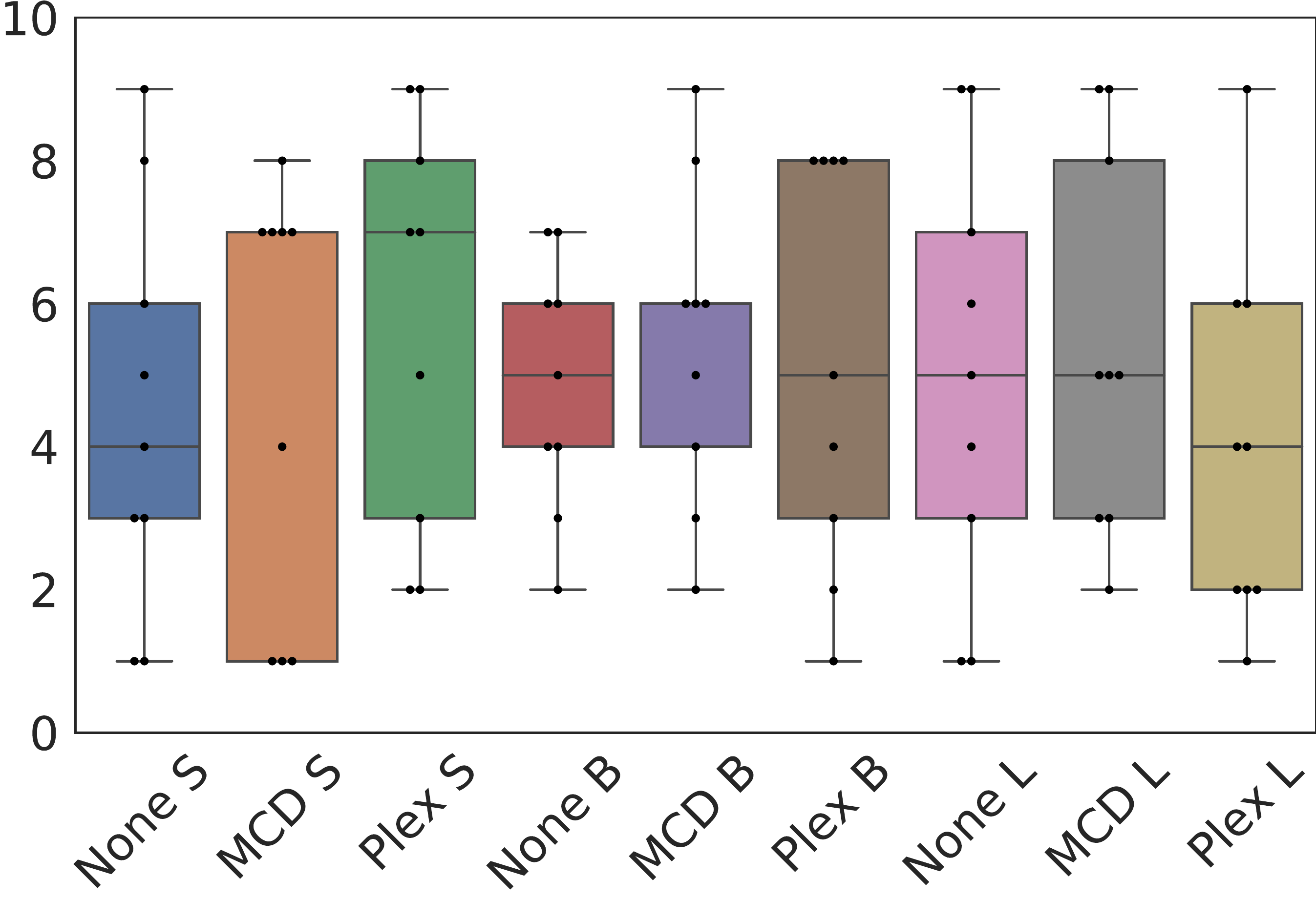}
\caption{Calibration, by size.}
\label{fig:t5-subpop-calibration-size}
\end{subfigure}
\vspace{-1.5ex}
\caption{T5-Plex model's ranking performance in subpopulations with spurious patterns compared to different methods (\ref{fig:t5-subpop-in-generalization-methods}-\ref{fig:t5-subpop-calibration-methods}) and across architecture sizes (\ref{fig:t5-subpop-generalization-size}-\ref{fig:t5-subpop-calibration-size}).
}
\label{fig:t5-subpop-rank-compare}
\end{figure*}

\paragraph{Spurious Correlations.} 
A well-observed issue in machine learning models is their tendency to learn decision rules that rely on spurious, non-causal patterns that exhibit strong statistical correlation with the outcome in the training data, i.e., shortcut learning \citep{geirhos2020shortcut}. In particular, previous theoretical studies show that a model's tendency to learn spurious patterns is rooted in the overparameterized model's tendency to flexibly adapt to the (biased) empirical distribution of the data, and cannot be fully addressed by increasing model size \citep{bommasani2021opportunities}. However, recent empirical studies also show that, when the data distribution exhibits suitable diversity such that it contains a small amount of counterexamples where the spurious pattern doesn't hold, large pretrained models can still lead to improved robustness \citep{tu2020empirical}. Therefore, we investigate if the large pretrained models' robustness indeed improves with model scale, and if the uncertainty methods bring additional gain on top of the None baselines.

For the language modality, we evaluate the performance of T5 models on two spurious-correlation subpopulations: HANS for natural language inference, and CivilComments-Identities for toxic comment detection \citep{mccoy2019right, borkan2019nuanced}. Specifically, HANS evaluates the NLI model's robustness against non-causal heuristic patterns that are empirically associated with sentence entailment (e.g., lexical overlap) \citep{mccoy2019right}. On the other hand, CivilComments-Identities contains comments that mention certain gender, sexual orientation, ethnic or religious identities, which empirically exhibits differential distribution of toxicity label with respect to their surface-level textual features. A toxic detection model that relies on these surface-level identity mentions can lead to unintended consequences, unjustly  reinforcing existing social stereotypes to disadvantaged identity groups \citep{borkan2019nuanced}.

\Cref{fig:t5-subpop-rank-compare} summarizes the performance of different uncertainty methods, and across three different scales (i.e., T5$_{\texttt{small}}$, T5$_{\texttt{base}}$, T5$_{\texttt{large}}$) (See Appendix \Cref{fig:t5-rank-arch,fig:t5-rank-method} for detailed results). Compared to the None baseline, the uncertainty methods (BE, DE, GP and their combinations) provide significant improvements across all aspects of model performance (i.e., generalization, selective prediction, and uncertainty calibration). For generalization and selective prediction performance, the ensemble of SNGP models (DE-GP) and Plex (BE-GP) perform the best among ensemble and non-ensemble methods, respectively. Furthermore, within each method class, the subpopulation generalization improves as the model scale increases, with Plex (BE+GP) being the most competitive method for moderate- and large-size models. Finally, the trend in uncertainty calibration is less consistent, we see that among all methods, GP and BE are on average the most calibrated ensemble and non-ensemble method, while the model scale does not seem to have a consistent impact to calibration performance.

\subsection{Adaptation}
\label{sub:adaptation}

In this section, we examine the models’ ability to adapt to new data. We focus on data-efficient learning (small data generalization), where the goal is to attain high reliability performance with only a small set of examples.
Our study focuses on ViT vision models, since adaptation has not yet been studied for T5 text models \citep{raffel2020exploring}. 


\input{active_learning.tex}

\subsubsection{Few-shot Learning}
\label{sub:few-shot-learning}


\begin{tldr}
Plex's BatchEnsemble representation improves few-shot and over increasing model and pretraining dataset scales.
Full-model training can work better than linear evaluation if the goal is to maximize performance from limited examples.
\end{tldr}

In few-shot learning, we examine how well representations learned during pretraining enable fast downstream adaptation. We use a linear evaluation protocol where we extract features from the models’ pre-logits layer and train a multinomial logistic regression model using L-BFGS. Unlike linear regression which is also popular
(e.g., \citet{dosovitskiy2020image}), logistic regression produces a categorical distribution, enabling one to measure the quality of the few-shot model's uncertainty estimates; we take advantage of this in \Cref{sub:few-shot-uncertainty}. 

%

\begin{figure*}[!tb]
\centering
\includegraphics[width=\textwidth]{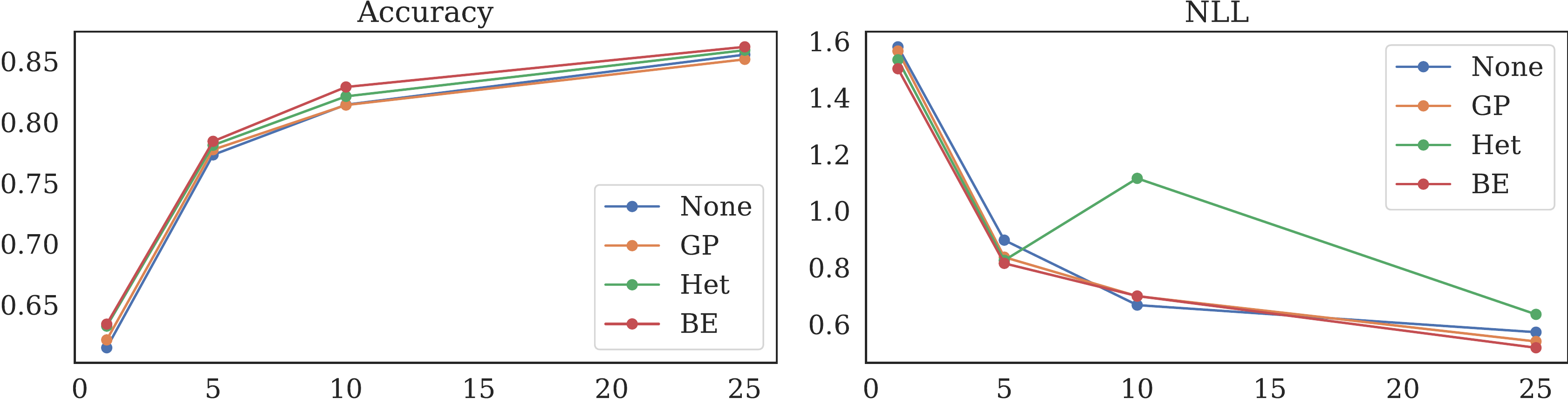} \\
\includegraphics[width=\textwidth]{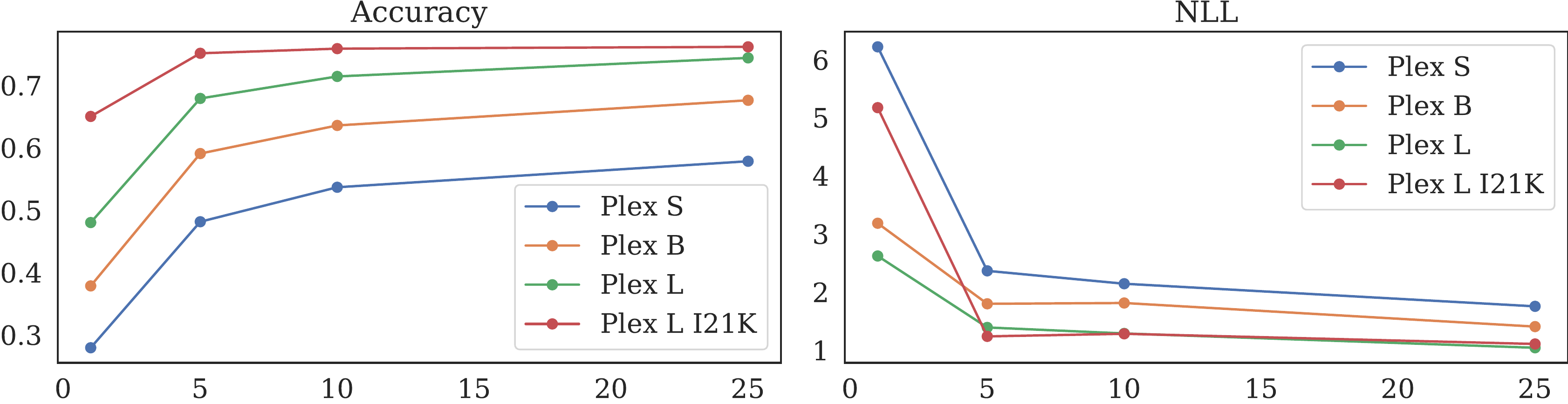}
\caption{%
Few-shot performance averaged over 9 image datasets. \textbf{(top)} Comparison of pretrained model choices. \textbf{(bottom)} Comparison of pretrained model and dataset scale, where the default dataset is JFT and I21K signifies ImageNet21K.
%
BatchEnsemble's representation slightly outperforms others.
Model and pretraining dataset size have the most significant effect, with larger doing better and ImageNet21K doing better than JFT.
}
\label{fig:datasets-fewshot}
\end{figure*}

\paragraph{Model choice and size}
\Cref{fig:datasets-fewshot} displays test set accuracy and negative log likelihood (NLL) averaged over 9 datasets: birds, caltech, cars, cifar100, col hist, dtd, imagenet, pets, and uc merced.
The choice of model translates to a consistent 1-2\% improvement in accuracy depending on the few-shot setting. The BatchEnsemble representation is among the top performing that gives high predictive performance.

The size of the model and pretraining dataset also has a noticeable effect on few-shot performance.
While ImageNet-21K is 10 times smaller than JFT, models pretrained on ImageNet-21K consistently outperform those pretrained on JFT across prediction and uncertainty tasks, with the exception of calibration. We hypothesize this is because ImageNet-21K is closer in distribution to the datasets we evaluate on average. It is also surprising that when finetuning over the full ImageNet dataset (\Cref{sub:robust-generalization}), the conclusion is reversed with JFT models outperforming ImageNet-21K. This suggests a transition point when the pretrained models are trained on a sufficiently high number of examples.

\begin{figure*}[!tb]
\centering
\includegraphics[width=\textwidth]{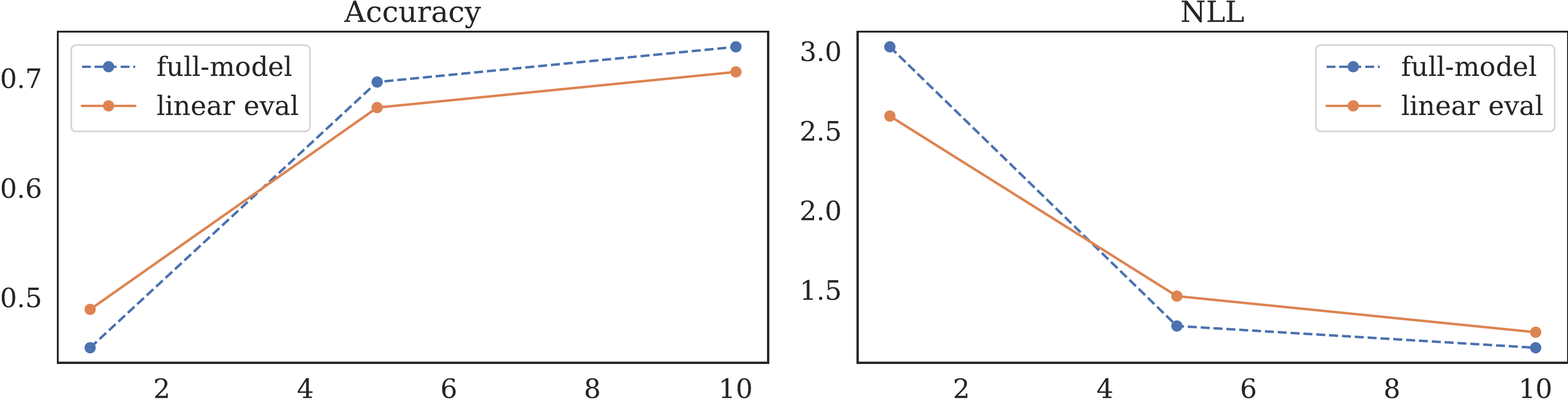}
\caption{%
Linear evaluation versus full-model training on ImageNet. Full-model training consistently outperforms linear evaluation at 5-shot and higher.
}
\label{fig:full-model-fewshot}
\end{figure*}


\paragraph{Full-model training}
Intuitively, linear evaluation is about ``representation learning'', assessing how well a pretrained model’s fixed representation layer generalizes to many different scenarios. This setup may not be optimal for real-world transfer if the goal is simply to maximize performance from limited examples, without constraints on the approach to doing so.

In \Cref{fig:full-model-fewshot}, we adopt the usual finetuning protocol and experiment with training the model’s full set of parameters given the pretrained checkpoint.
%
%
We set up a few-shot version of ImageNet1k for this study.\footnote{%
\url{https://www.tensorflow.org/datasets/catalog/imagenet2012_fewshot}
}
%
%
On 1-shot, linear evaluation outperforms full-model training which may not be surprising given the high possibility of overfitting.
Starting from 5-shot (and which is likely to hold for even lower shot settings), full-model training surpasses linear evaluation
with a consistent 2-3\% improvement in accuracy.



\subsubsection{Few-shot Uncertainty}
\label{sub:few-shot-uncertainty}

\begin{tldr}
Plex performs well on both few-shot and zero-shot open set recognition, particularly using larger model sizes and on ImageNet21K.
Few-shot calibration remains a challenging problem.
\end{tldr}

\begin{figure*}[!tb]
\centering
\includegraphics[width=\textwidth]{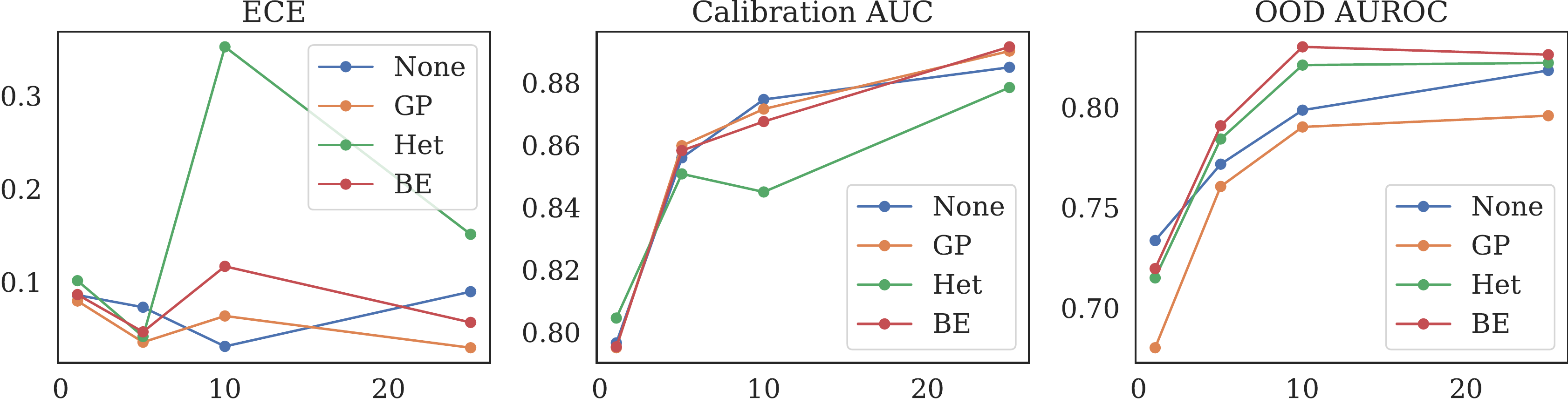}
\includegraphics[width=\textwidth]{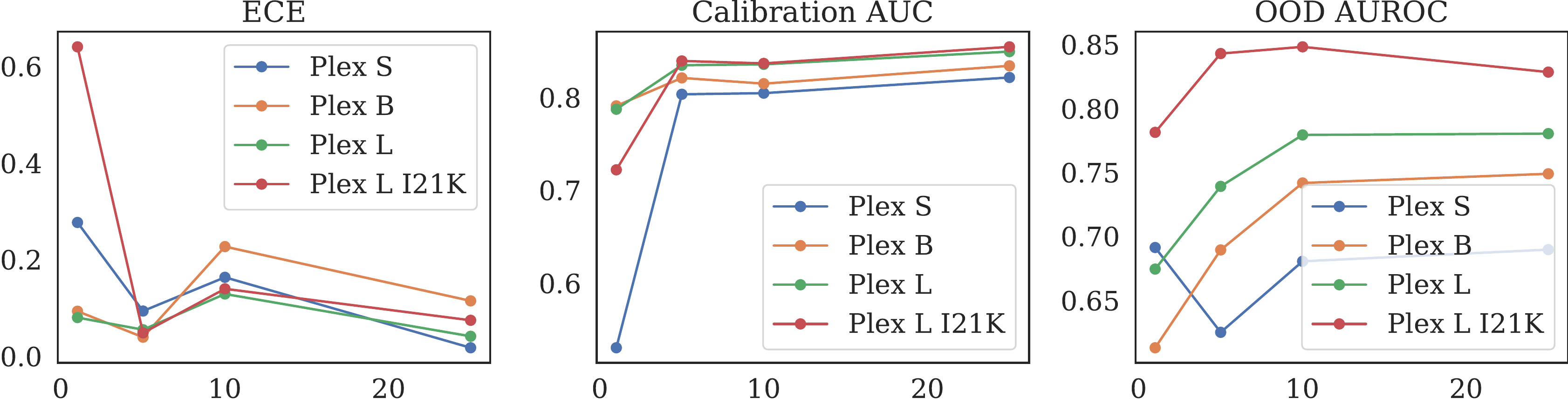}
\caption{%
Few-shot uncertainty performance averaged over 9 image datasets for calibration and selective prediction, and 2 datasets for open set recognition. 
\textbf{(top)} Comparison of pretrained model choices. \textbf{(bottom)} Comparison of pretrained model and dataset scale, where the default dataset is JFT and I21K signifies ImageNet21K.
Plex's BatchEnsemble representation works best on few-shot open set recognition, and with ImageNet21K as the pretraining dataset.
Calibration and selective prediction performance are similar across model and scaling choices; they remain challenging problems. 
}
\label{fig:uncertainty-few-shot}
\end{figure*}

Few-shot learning is an important and practical setting, for which the literature almost exclusively considers accuracy to measure performance. \emph{Few-shot uncertainty quantification} is also a key problem:
compared to finetuning, few-shot models are more sensitive to the examples they've seen and by extension, much less accurate. Thus few-shot models should be more likely to be uncertain, and the quality of their uncertainty estimates must be good in order to know when to refrain from making a prediction.

\paragraph{Few-shot uncertainty tasks}
In \Cref{fig:uncertainty-few-shot}, we measure uncertainty performance
on three metrics: ECE and Calibration AUROC for calibration, averaged over all 9 datasets used for few-shot learning, and OOD AUROC for open set recognition.
Open set recognition requires an OOD dataset in addition to a test split, so we only evaluate it for two datasets: CIFAR-100 with an OOD dataset of CIFAR-10; and ImageNet1K with an OOD dataset of Places365.
Calibration performance is roughly comparable across methods. BatchEnsemble has a noticeable improvement on open set recognition.
The scale of the pretraining dataset has no significant effect on calibration, where the trend is weaker than what's seen on calibration of finetuned models (\Cref{sub:calibration}). The choice of pretraining dataset does have a noticeable effect on open set recognition, where like in \Cref{sub:few-shot-learning}, Plex on ImageNet21K surprisingly does better than JFT models.



\begin{table*}[tb]
\centering
\small
\begin{tabular}{ccc|cc}
               & \multicolumn{2}{c}{Far-OOD}          & \multicolumn{2}{c}{Near-OOD}                  \\
               & C-10 vs SVHN & C-100 vs SVHN & C-10 vs C-100 & C-100 vs C-10 \\  \toprule
               \multicolumn{5}{c}{Mahalanobis} \\ \midrule
Plex S        & 0.596            & 0.449             & 0.670                 & 0.592                 \\
Plex B        & 0.797            & 0.643             & 0.733                 & 0.679                 \\
Plex L        & 0.717            & 0.586             & 0.716                 & 0.690                 \\
Plex L I21K & \underline{0.895}            & \textbf{0.956}             & \underline{0.864}                 & \underline{0.902}                 \\
\midrule
\multicolumn{5}{c}{Relative Mahalanobis} \\ \midrule
Plex S        & 0.843            & 0.814             & 0.805                 & 0.678           \\
Plex B        & 0.938            & 0.848             & 0.863                 & 0.769                 \\
Plex L        & 0.967            & 0.875             & 0.890                 & 0.821                 \\
Plex L I21K & \textbf{0.957}            & \underline{0.932}             & \textbf{0.919}                 & \textbf{0.915}                 \\
\bottomrule      
\end{tabular}
\caption{%
Zero-shot open set recognition with Plex, varying scale and detection method. 
}
\label{table:osr-zero-shot}
\end{table*}

\paragraph{Zero-shot open set recognition}
Given that the pretrained model itself provides rich and robust representations, we can use that as a feature extractor and conduct zero-shot open set recognition without \emph{any} training examples.
To do so, we can take advantage of Mahalanobis distance (Maha) \citep{lee2018simple} as a detection score: Maha measures the distance between the test input and the fitted training distribution in the embedding space. It operates on a fixed representation layer and does not require operating on softmax outputs with a newly trained last layer. 
The training distribution is fitted using a class conditional Gaussian $\gaussk, k=1, 2, \dots, K$ to each of the $K$ in-distribution classes based on the embedding $\vz$. We estimate the mean vectors and covariance matrix as: 
   $ \mathbf{\vmu}_k = \frac{1}{N_k} \sum_{i:y_i=k} \vz_i$, for $k=1, \dots, K,$
   and $\vSigma = \frac{1}{N}\sum_{k=1}^K \sum_{i:y_i=k}\left( \vz_i-\mathbf{\vmu}_k \right)(\vz_i-\mathbf{\vmu}_k)^T$.
Note that class-conditional means $\mathbf{\vmu}_k$ are independent for each classes, while the covariance matrix $\Sigma$ is shared by all classes to avoid under-fitting errors. 
For a test input, Mahalanobis distances from a test input to each of the fitted $K$ in-distribution Gaussian distributions $\gaussk$ is computed, and the minimum of the distances over all classes is used as the uncertainty score, 
$\text{MD}(\vz) = \min_{k} \{ \left(\vz-\mathbf{\vmu}_k\right)^T \vSigma^{-1} \left(\vz-\mathbf{\vmu}_k \right) \}.$

We also experiment with a Relative Mahalanobis distance variant (RMaha) \citep{ren2021simple}, a modified version of Mahalanobis distance which corrects for the background confounding effect using another Gaussian distribution fitted using entire training data ignoring class labels. 
The uncertainty score is defined as $\text{RMD}_k(\vz) \nonumber = \text{MD}(\vz)  - \text{MD}_0(\vz)$, where $\text{MD}_0(\vz)$ indicates the Mahalanobis distance to a distribution fitted to the entire training data not considering the class labels: $\mathcal{N}(\mathbf{\vmu}_0, \vSigma_0)$, where $\vmu_0 = \frac{1}{N} \sum_{i=1}^N \vz_i$ and $\vSigma_0 = \frac{1}{N}\sum_{i=1}^N \left( \vz_i-\vmu_0 \right)(\vz_i-\vmu_0)^T$.
This is a good proxy for the background distribution.


Zero-shot results are displayed in \Cref{table:osr-zero-shot}.
Interestingly, the zero-shot setting achieves performance close to that of the finetuned setting. For example, in the challenging near-OOD task CIFAR-100 vs CIFAR-10, the best zero-shot AUROC achieved by Plex L (I21K) is 0.915, while its performance with finetuning is 0.934, and the best model which uses JFT and finetuning achieves 0.954 (\Cref{table:osr-msp}).
For the far-OOD task CIFAR-100 vs SVHN, the best AUROC achieved by zero-shot is 0.933, and the best model which uses JFT and finetuning achievs 0.938.

The Relative Mahalanobis distance also significantly outperforms the raw Mahalanobis distance, suggesting that there are background features that confound the raw Mahalanobis distance. In fact, the raw Mahalanobis distance does not work well for most of the pretrained models, mostly having AUROC lower than 0.9 and varies depending on the model types.

\section{Conclusion}

We presented a framework for thinking about reliable deep learning, and provided a number of tasks and datasets for stress-testing the reliability of models through multiple tasks such as the ability to quantify its confidence in predictions, be robust to distribution shifts, and adapt quickly to new distributions. To improve reliability, we also developed Plex, pretrained large model extensions for vision (ViT-Plex) and language (T5-Plex) that significantly improve the reliability in deep learning. Our techniques are broadly applicable for other large models (cf. LaMDA \citep{thoppilan2022lamda}, PaLM \citep{chowdhery2022palm}) and wider suite of tasks, cf. BIG-bench \citep{srivastava2022beyond}.

\section*{Acknowledgements}
We thank
Ben Adlam, 
Dilip Krishnan,
Ed Chi, 
Neil Houlsby,
Rif A. Saurous,
and
Sharat Chikkerur 
for helpful discussions and feedback on earlier drafts of the paper. 
We also thank Tom Small and Ajay Nainani for assisting with visualizations.

\newpage
\bibliography{main}
\appendix
\newpage

\section{Author Contributions}
\label{appendix:contributions}

\begin{itemize}
    \item
Andreas Kirsch: Active Learning only -- joint with Joost van Amersfoort. Implemented Active Learning loop, various acquisition functions and datasets, and ran experiments on CIFAR-10/CIFAR-100. Advised on experimental setup for larger datasets, and wrote active learning section.
    \item
Balaji Lakshminarayanan: Helped with direction, narrative, getting resources, advising folks on experiment setup (open set recognition, few-shot adaptation, last layer), helped with writing (abstract, intro, contributions) and revising other sections. 
    \item
Clara Huiyi Hu: Implemented BE + GP in language domain, executed associated experiments and helped building post-analysis colabs. On the vision side, contributed to sweeping and debugging BE/GP variants. 
    \item
Dustin Tran: Came up with work's vision, lead overall writing and experiment design, coordinated individual leads and contributors. Designed and tuned Plex, various figures and plotting code, and core infra setup.
    \item
Du Phan: Lead and designed the infrastructure to evaluate uncertainty methods on the language domain. Implemented methods None, DE, MCD, GP, DE-GP, Het for language and performed most corresponding experiments across MNLI, ToxicComments, NaLUE tasks.
    \item
D. Sculley: Advisor on project, including its direction and the paper narrative.
    \item
Honglin Yuan: Initial experimental setup for CIFAR subpopulation shift experiments, including pipeline for converting data into format to be consumed by uncertainty baselines codebase.
    \item
Jasper Snoek: Helped with direction, narrative, getting resources, helped with writing, editing, aggregating results, various plots.
    \item
Jeremiah Liu: 
Lead for language domain efforts. Developed all tasks, datasets, evaluation metrics for language.
Implemented initial t5x infra and baseline (None) experiments. Work with eng co-lead Du and teammates Clara and Jie in finishing executing experiments and generating visualizations. Wrote the sections on language. On the vision side, implemented the ViT-GP model, conducted initial pretraining experiments. Assisted editing the section on SNGP method. 
    \item
Jie Ren: Lead work on open set recognition (vision and text), GPs, and assisted with few-shot. Helped with engineering work including building vision finetuning models, running vision and text experiments. Made the demo figures for vision and text. Helped with Mahalanobis distance-based active learning method. 
    \item
Joost van Amersfoort: See Andreas Kirsch (joint contribution). Co-wrote compute workflow tutorial for collaborators.
    \item
Karan Singhal: Vision subpopulation shift setup and experiments: integrated code for CIFAR-10/100 subpopulation evaluation into codebase, set up, ran, and tuned experiments with different ingredients (Det, BE, BE+HET, BE+GP). Ran ablations: omitting last-layer changes, omitting ensembling during pretraining, ImageNet-only pretraining, no pretraining. Created ablations figure and added discussion. Made detailed edits to paper.
    \item
Kehang Han: Responsible for finetuning experiments involving GP and few-shot experiments. Created Imagenet2012Fewshot TFDS (\url{https://www.tensorflow.org/datasets/catalog/imagenet2012_fewshot}). Primary author of \Cref{sub:few-shot-learning,sub:few-shot-uncertainty}. Helped with GP model ablations.
    \item
Kelly Buchanan: Contributed to early formation of project and codebase, particularly around tasks involving structured output evaluation and uncertainty.
    \item
Kevin Murphy: Advisor on project, including its direction and the paper narrative.
    \item
Mark Collier: Responsible for most experimental results involving heteroscedastic model e.g. Het upstream, Het$\to$Het, None$\to$ Het, BE$\to$BE$+$Het. Primary author of section \Cref{app:heteroscedastic}. Implemented CIFAR-10H and ImageNet ReaL-H label uncertainty tasks. Primary author of \Cref{sub:label-uncertainty}. 
 Ran sensitivity analysis (together with Mike) to the number of JFT pretraining epochs and volume of pretraining data (JFT4B).
    \item
Mike Dusenberry: Engineering lead. Designed, built, and owned infrastructure for Plex, including reproducible training scripts, utilities, experiment scaling components, and open-sourcing efforts. Trained and tuned None ViT baselines (e.g., L) to match existing results. Scaled up experiments to JFT-4B. Ablated JFT dataset size and number of epochs with Mark Collier. Ran experiments and wrote \Cref{sub:covariate-shift} for generalization to covariate distribution shifts. Advised and worked on infra and experiments for the active learning section.
Co-wrote a compute workflow tutorial for collaborators. Created a demo notebook for ViT-Plex. Open-sourced code and model checkpoints.
    \item
Neil Band: Co-led design of experiments and narrative for integration of RETINA experiments, including selective prediction and distribution generalization. Led RETINA implementation and experiments. Co-wrote section on selective prediction and compute workflow tutorial for collaborators. Led open-sourcing of RETINA distribution shift tasks, baseline model checkpoints, evaluation, and plotting utilities. 
    \item
Nithum Thain: Contributed to experiments analyzing Deep Ensembles and their tradeoff with compute.
    \item
Rodolphe Jenatton: Developed the code to evaluate deep ensembles in \url{https://github.com/google/uncertainty-baselines} and ran associated experiments (together with Nithum Thain). Developed the code to evaluate models and ensembles in \url{https://github.com/google-research/robustness_metrics}. Developed the code to wrap the \texttt{pjit}-based \url{https://github.com/google-research/vmoe} in \url{https://github.com/google/uncertainty-baselines}, enabling the evaluation of sparse MoE models, namely \textsc{E}$^3$, V-MoE and ensembles thereof—ran corresponding experiments. Contributed to the shaping of the narrative of the ensembling section (together with Zelda Mariet). Helped with the computation of the FLOPs. Improved finetuning numbers by suggesting use of longer finetuning schedule.
    \item
Tim G. J. Rudner: Co-led design of experiments and narrative for integration of RETINA experiments, including selective prediction and distribution generalization. Contributed to all RETINA experiments.
Led writing of section on selective prediction and contributed to writing of section on calibration AUROC.
    \item
Yarin Gal: Advisor on project, including its direction and the paper narrative.
    \item
Zachary Nado: Designed and contributed to core infrastructure that was the major platform to iterate on our models, tasks, and other results.
   \item
Zelda Mariet: Ran BatchEnsemble experiments, ran ensemble diversity analysis, drew up vision for the ensembling analysis \Cref{sub:ensembles}, proposed and ran the upstream vs downstream comparison (\Cref{sub:effect}). Wrote the code for experimental analysis, hyperparameter tuning, visualization and summarization. Code refactoring for the overall project.
    \item
Zi Wang: Led the 
active learning effort including initial exploration as an engagement with internal teams as a close collaboration between Google Reliable Deep Learning and the Oxford OATML group (with Mike, Andreas, Joost, and Dustin). Contributed to refactoring BE and None to facilitate better integration with AL. Contributed to active learning code (debugging, adding new features etc). Extensively experimented with AL on RDL models, conducted analyses and obtained/wrote final results. Added discussions on the relations between prior learning and pretraining.
    \item
Zoubin Ghahramani: Advisor on project, including its direction and the paper narrative.
\end{itemize}

\section{Details behind Key Figures}
\label{sec:reliability:score}

For the task radar plots of \Cref{fig:radar}, we list references used for task-specific state-of-the-art in \Cref{table:radar_plot_sota}.

\begin{table*}[!tbh]
\centering
\begin{tabular}{rrl}
\toprule
Task & Task-specific SOTA \\
\midrule
\textbf{Text} & & \\
MNLI & T5 11B~\citep{raffel2020exploring}\\
MNLI Mismatched (Subpop Shift) & T5 11B~\citep{raffel2020exploring} \\
MNLI Hans All (OOD) & RoBERTa base~\citep{tu2020empirical} \\
WikipediaTalk & Deep Ensembles~\citep{kivlichan2021measuring} \\
\midrule
\textbf{Vision} & & \\
OOD Places 365 & \citep{hendrycks2019scaling} \\
AL Accuracy CIFAR10 & \citep{yi2022} \\
AL Accuracy CIFAR100 & \citep{citovsky2021batch} \\
AL Accuracy ImageNet & \citep{emam2021active} \\
Accuracy RETINA & \citep{band2021benchmarking} \\
Selective Prediction RETINA & \citep{band2021benchmarking} \\
\bottomrule
\end{tabular}
\caption{References for the task-specific state-of-the-art used in \Cref{fig:radar}.}
\label{table:radar_plot_sota}
\end{table*}

For the model scaling plot of \Cref{fig:scale}, we aggregate all task metrics under a single scalar between 0 and 100. In order to do this, we normalize all metrics to be between 0 and 100; we then compute an unweighted average. Most metrics are already bounded between 0 and 100: for example, accuracy, expected calibration error (we do $100-\text{ECE}$ so higher is better), Calibration AUROC, and OOD AUROC. The one exception are scoring rules such as log-loss and Brier score. Because the output distributions are discrete, log-loss has a lower bound of 0 and an upper bound given by the highest entropy distribution (uniform). Therefore we rescale scoring rule values based on their lower and upper bounds so that they're now between 0 and 100 and so that higher is better.

\section{Additional Details on Tasks \& Datasets}
\label{appendix:datasets}

\begin{table*}[htb]
\centering
\begin{tabular}{rrl}
\toprule
Task & & Metric \\
\midrule
Calibration & $\downarrow$ & Expected Calibration Error (ECE) \\
& $\uparrow$ & Calibration AUROC \\
Selective Prediction & $\uparrow$ & Referral Rate Curve \\
& $\uparrow$ & Oracle Collaborative Accuracy \\
Open Set Recognition & $\uparrow$ & AUROC \\
Label Uncertainty & $\downarrow$ & KL Divergence \\
\midrule
In-Dist. Generalization & $\uparrow$ & Accuracy \\
& $\downarrow$ & Negative Log-Likelihood (NLL) \\
& $\downarrow$ & Brier Score \\
& $\uparrow$ & (Binary Classification) AUROC \\
& $\uparrow$ & (Binary Classification) AUPRC \\
Covariate Shift & $\uparrow$ & Accuracy \\
& $\downarrow$ & Negative Log-Likelihood \\
& $\downarrow$ & Brier Score \\
& $\uparrow$ & (Binary Classification) AUROC \\
& $\uparrow$ & (Binary Classification) AUPRC \\
Subpopulation Shift & $\uparrow$ & Accuracy \\
& $\downarrow$ & Negative Log-Likelihood (NLL) \\
& $\downarrow$ & Brier Score \\
& $\uparrow$ & (Binary Classification) AUROC \\
& $\uparrow$ & (Binary Classification) AUPRC \\
\midrule
Few-Shot Learning & $\uparrow$ & Accuracy \\
& $\downarrow$ & Negative Log-Likelihood (NLL) \\
Few-Shot Uncertainty & $\downarrow$ &  Expected Calibration Error \\
& $\uparrow$ & Calibration AUROC \\
& $\uparrow$ & (Zero-Shot and Few-Shot) AUROC \\
Active Learning & $\uparrow$ & Accuracy vs \# Labels Curve \\
\bottomrule
\end{tabular}
\caption{%
Metrics used for each task. $\uparrow$ / $\downarrow$ mean higher / lower is better.
}
\label{table:metrics}
\end{table*}

We summarize the list of metrics in \Cref{table:metrics}.




For out-of-distribution evaluation in vision, we use the following datasets.

 \begin{enumerate}
     \item 
\emph{Covariate shift}.
\begin{itemize}
    \item 
CIFAR-10: CIFAR-10-C \citep{hendrycks_benchmarking_2019}.
  \item 
CIFAR-100: CIFAR-100-C \citep{hendrycks_benchmarking_2019}.
  \item  ImageNet1K: ImageNet-A \citep{hendrycks2021natural}, ImageNet-C \citep{hendrycks_benchmarking_2019}, ImageNetV2 \citep{recht2019imagenet}, ImageNet-Vid-Robust, YTBB Robust \citep{shankar2021image}, ObjectNet \citep{barbu2019objectnet}, and ImageNet-R \citep{hendrycks2021many}.
  \item  RETINA: RETINA's Country Shift dataset \citep{band2021benchmarking}.
We train models on images of retinas obtained from patients in the United States (EyePACS) and evaluate trained models on images of retinas obtained from patients in India using different collection equipment (APTOS).
  \end{itemize}

 \item
\emph{Semantic (class) shift}.
\begin{itemize}[leftmargin=1em]
    \item 
CIFAR-10: CIFAR-100, SVHN.
  \item 
CIFAR-100: CIFAR-10, SVHN.
  \item 
ImageNet1K: Places365.
  \item 
RETINA: RETINA’s Severity Shift dataset \citep{band2021benchmarking}.
We train models on images of retinas exhibiting no worse than mild diabetic retinopathy, and consider a shifted evaluation dataset with images of moderate diabetic retinopathy or worse.
The evaluation data contains features not contained in the training images, such as vitreous hemorrhages.
The motivation for this shift is that images of retinas with more severe retinopathy are relatively scarce and that it is likely for a model to be trained only on more widely-available images of retinas exhibiting mild cases of diabetic retinopathy.
\end{itemize}

 \item
\emph{Label uncertainty}.
 We use CIFAR-10H \citep{peterson2019human} which captures human uncertainty over labels for CIFAR-10 dataset. 
 We also construct a larger-scale variant, which we call ImageNet ReaL-H. Individual human ratings were recollected for the original ImageNet test set, available as raw data from ImageNet ReaL \citep{beyer2020we}, and we use them to newly construct soft label targets.
 \item 
 \emph{Subpopulation shift}.
We use Semantically Partitioned CIFAR-10/100 \citep{yuan2022we} for vision subpopulation shift. CIFAR-10/100 test data is partitioned into semantically similar subpopulations, where each subpopulation has its own data-generating distribution sampled from a meta subpopulation distribution. We aim to improve predictive performance on tail subpopulations.
 \end{enumerate}

For covariate and semantic shift in language, we use:
\begin{itemize}
    \item 
the MNLI-mismatched \citep{williams2017broad} data as the OOD set for NLI, which contains sentence pairs from 5 genres that are distinct from those in MNLI training data. 
    \item 
the CivilComments corpus \citep{borkan2019nuanced} as the OOD set for toxic comment prediction, which consists of one million public comments appearing on approximately 50 English-language news sites across the world. 
\end{itemize}

We also analyze language subpopulation shift where long-tail groups from the in-distribution set are a desired area for generalization.
\begin{itemize}[leftmargin=1em,itemsep=0em]
    \item 
HANS \citep{mccoy2019right} eval datasets for NLI, which contains template-generated examples attacking the surface-level heuristics that the neural models are found to rely on when predicting entailment relationships.
    \item 
CivilCommentsIdentity \citep{borkan2019nuanced} for toxic comments, which is a subset of CivilComments that has explicit mention of social identities (e.g., muslim, LGBTQ, etc) that the model are often found to generate mispredictions.
    \item 
NaLUE-tail dataset for CLU, which is a subset of NaLUE corresponding to utterances from 28 low-frequency intents categories. 
\end{itemize}

\section{Details of Plex ingredients}
\label{appendix:ingredients}

\paragraph{BatchEnsemble (BE)} BatchEnsembles \citep{wen_batchensemble_2020-1} approximate deep ensembles \citep{deep-ensembles}, but reduce their computational and memory costs by sharing weights across the ensemble members. The weight matrix $\mW_i$ of any given ensemble member $i$ is written as the Hadamard product of a shared weight matrix $\mW_0$ and a local rank-1 matrix $r_i  s_i^\top$: 
\begin{align}
    \mW_i = \mW_0 \circ  r_i s_i^\top.
\end{align}
The vectors r and s are commonly referred to as fast weights.

Unless otherwise stated, Plex applies BE to all layers in the last 2 residual blocks of the network. This idea follows work for mixture of experts \citep{riquelme2021scaling}.

\paragraph{Spectral-normalized Neural Gaussian Process (SNGP)} Unlike ensemble approaches, SNGP proposed by \citet{liu2020simple} focuses on improving the uncertainty quality of a neural network given a fixed representation (a.k.a.~\emph{deterministic uncertainty quantification} setting \citep{van2020uncertainty}. 
When applied to a DNN without pretraining, SNGP enhances the DNN uncertainty property by applying spectral normalization to the hidden weights, and replaces the output layer from a dense layer to a random-feature Gaussian process (GP) layer.  That is, given hidden representations $h(\vx)$, the GP layer enables scalable computation of a GP posterior by applying a random feature approximation $\phi$ to the predictive function and then a Laplace approximation to the predictive variance:
\begin{align*}
 g(\vx) &\sim N(\logit(\vx), \var(\vx)) \\
\logit(\vx)	&= \phi(x)^\top\evbeta, 
\quad	\textrm{where} \\  
\phi(\vx) &= \cos(\mW h(\vx) + \vb) \\
\var(x)	&= \phi(\vx)^\top(I + \Phi^T\Phi)^{-1}\phi(\vx),   
\end{align*}
where $(\mW, \vb)$ are frozen random weights of the random feature embedding $\phi(\vx) = \cos(\mW h(\vx) + \vb)$, and $\Phi^\top\Phi= \sum_i\phi(\vx_i)\phi(\vx_i)^\top$ is the covariance of the random feature embedding estimated using the training data.

\citet{liu2020simple,liu2022simple} show that this combined technique improves the model's awareness of the semantic distance between the test and train examples on the data manifold, leading to improved performance in calibration and out-of-domain detection. When applied to a large pretrained DNN, we find it sufficient to only use the last-layer Gaussian process (i.e., omit the spectral normalization regularization), as the pretrained embedding has already provided a semantic-distance-aware representation of the data.

\paragraph{Heteroscedastic last layer (Het)}
Heteroscedastic last layers are designed to model input-dependent label noise/label uncertainty (a.k.a. aleatoric uncertainty \citep{kendall2017uncertainties}) that is present in the data. We use the Heteroscedastic (Het) last layer introduced by  \citet{collier2020simple,collier2021correlated} who place a multivariate Gaussian distribution over the logits in a standard DNN classifier. A low-rank approximation to the $K \times K$ covariance matrix ($K$ = number of classes/outputs) is made when $K$ is large and \citep{collier2021correlated} further develop a parameter efficient version of the method with parameterization inspired by BE to enable scaling to tens of thousands of classes. 

Among the above methods, None, Het, GP, and BE are both memory and compute efficient, since they only require a single forward pass from a single model to compute the output distribution. We also experiment with deep ensembles (\Cref{sub:ensembles}), which is the most expensive in both memory and compute, since they require forward passes from multiple trained models, as well as Monte Carlo Dropout \citep{gal2016dropout}.


\section{Additional Results for Open Set Recognition}
\label{appendix:osr}

\paragraph{Comparing OOD detection methods across all the tasks and model types.}
While we use MSP as the basic uncertainty score in the main text, we also would like to discuss other alternative uncertainty scores, and how is their performance improved by the Plex models. We study the following 4 more uncertainty scores, 
\begin{itemize}
    \item Mahalanobis distance (Maha) \citep{lee2018simple} which measures the distance between the test input and the fitted training distribution in the embedding space. The training distribution is fitted using a class conditional Gaussian. The uncertainty score is the distance. 
    \item Relative Mahalanobis distance \citep{ren2021simple} is a modified version of Mahalanobis distance which corrects for the background confounding effect using another Gaussian distribution fitted using entire training data ignoring class labels. 
    \item Entropy of the softmax probability. High entropy suggests high uncertainty. So the uncertainty score is entropy. 
    \item Maximum over the logits (MaxLogit) \citep{hendrycks2019scaling} which uses the maximum of the un-normalized logits as the confidence score. Then the uncertainty score is then the negative of the MaxLogit. 
\end{itemize}

We study the performance of five different OOD detection methods, MSP, Maha, RMaha, Entropy and MaxLogit, across the various tasks and model types. As shown in \Cref{table:osr-methods}, Mahalanobis distance outperforms all other scores, and Relative Mahalanobis distance performs the second best.

Despite the good performance, one drawback of Mahalanobis distance based method is that its computational time is linear to the number of classes. For the ImageNet2012 vs Places365 task, the in-distribution data ImageNet2012 has 1,000 classes where by definition of Mahalanobis distance method, we need to fit 1K Gaussians for each of the classes and compute the distance between the test input to each of the fitted Gaussian. It becomes very time consuming and not scalable. Therefore in that task, we study the lightweight OOD scores, including MSP, Entropy, and MaxLogit. Interestingly, we noticed that MaxLogit is the best among the three methods for single models (None, GP, None$\to$GP) except for Het and None$\to$Het, while Entropy is the best for ensemble models (BE, DE based models).

Overall we suggest using Mahalanobis distance based methods to achieve the best performance, but in case there is a computation budget, MaxLogit is a good choice for single models and Entropy is a good choice for ensemble models.

\begin{table*}[tb]
\centering
\small
\begin{tabular}{cccccc|ccc}
                & \multicolumn{5}{c}{C-100 vs C-10}            & \multicolumn{3}{c}{ImageNet vs Places} \\
                & MSP   & Entropy & MaxLogits & Maha           & RMaha & MSP    & Entropy         & MaxLogits      \\ \toprule
None             & 0.929 & 0.944   & 0.953     & \textbf{0.982} & 0.965 & 0.828  & 0.855           & \textbf{0.880} \\
None I21K        & 0.924 & 0.934   & 0.940     & \textbf{0.964} & 0.943 & 0.838  & 0.861           & \textbf{0.887} \\
None$\to$GP      & 0.927 & 0.940   & 0.948     & \textbf{0.975} & 0.956 & 0.824  & 0.851           & \textbf{0.877} \\
None$\to$Het     & 0.948 & 0.962   & 0.948     & \textbf{0.985} & 0.970 & 0.827  & \textbf{0.854}  & 0.827          \\
None$\to$DE L & 0.940 & 0.953   & 0.940     & \textbf{0.982} & 0.965 & 0.831  & \textbf{0.860}  & 0.831          \\
None$\to$BE      & 0.924 & 0.939   & 0.924     & \textbf{0.980} & 0.962 & 0.822  & \textbf{0.850}  & 0.822          \\
GP              & 0.927 & 0.936   & 0.945     & \textbf{0.971} & 0.945 & 0.828  & 0.854           & \textbf{0.873} \\
Het             & 0.948 & 0.964   & 0.948     & \textbf{0.981} & 0.979 & 0.828  & \textbf{0.856}  & 0.828          \\
DE L         & 0.946 & 0.961   & 0.946     & \textbf{0.987} & 0.976 & 0.834  & \textbf{0.863}  & 0.834          \\
BE L         & 0.940 & 0.954   & 0.940     & \textbf{0.985} & 0.971 & 0.827  & \textbf{0.855}  & 0.827          \\
BE L (I21K)  & 0.934 & 0.943   & 0.934     & \textbf{0.975} & 0.959 & 0.842  & \textbf{0.864}  & 0.842          \\
BE$\to$BE$+$Het   & 0.954 & 0.969   & 0.954     & \textbf{0.984} & 0.980 & 0.831  & \textbf{0.857}  & 0.831          \\
BE B         & 0.906 & 0.927   & 0.906     & \textbf{0.966} & 0.959 & 0.801  & \textbf{0.831}  & 0.801          \\
BE S         & 0.864 & 0.889   & 0.864     & \textbf{0.923} & 0.916 & 0.773  & \textbf{0.803}  & 0.773     \\ \bottomrule     
\end{tabular}
\caption{%
The AUROC based on the five OOD measures on the most challenging near-OOD tasks CIFAR-100 vs CIFAR-10 and ImageNet2012 vs Places365.
}
\label{table:osr-methods}
\end{table*}

\section{Additional Results for Selective Prediction}
\label{appendix:selective-prediction}

We performed model selection on pretrained models finetuned with different hyperparameters using the area under the in-distribution selective-prediction accuracy curve.
We evaluated the predictive performance of the finetuned models on in-distribution and distributionally shifted data.
In particular, we computed models’ predictive accuracy, negative log-likelihood, expected calibration error, and the area under the selective prediction curve obtained by computing the area under the ROC curve for selective prediction referral rates from 0\% to 99\% using different evaluation metrics (accuracy and, where applicable, AUROC and AUPRC).

\begin{table*}[!tb]
\centering
\begin{tabular}{lcccccc}
\toprule
& SoTA   & 
SoTA   &
Plex L  &
Plex L  & 
None L & 
None L \\
&  (Ensemble) & 
 (Single Model) &
 (JFT) &
(I21K) & 
(JFT) & 
 (I21K)\\
\toprule
In-Distribution &
\textbf{95.0\%} &
94.4\% &
92.7\% &
92.3\% &
94.3\% &
94.1\% \\
Out-of-Distribution &
77.9\% &
78.3\% &
\textbf{84.8\%} &
78.9\% &
79.5\% &
78.3\% \\
\bottomrule
\end{tabular}
\caption{%
    Selective prediction performance, using AUROC as the evaluation metric, on the RETINA Country Shift task (for in and out-of-distribution evaluation). 
    For in-distribution, the SoTA (Ensemble) outperforms Plex L pretrained on both ImageNet-21K or JFT.
    For out-of-distribution, Plex L pretrained on either ImageNet-21K or JFT outperforms state-of-the-art.
}
\label{table:retina-sel-pred-auroc-country}
\end{table*}


\begin{figure*}[!tb]
\centering
\begin{subfigure}{\linewidth}
    \includegraphics[width=\linewidth]{figures/retina/legend.pdf}
\end{subfigure}
\hspace*{10pt}\begin{subfigure}[l]{0.45\linewidth}
    \includegraphics[width=\linewidth]{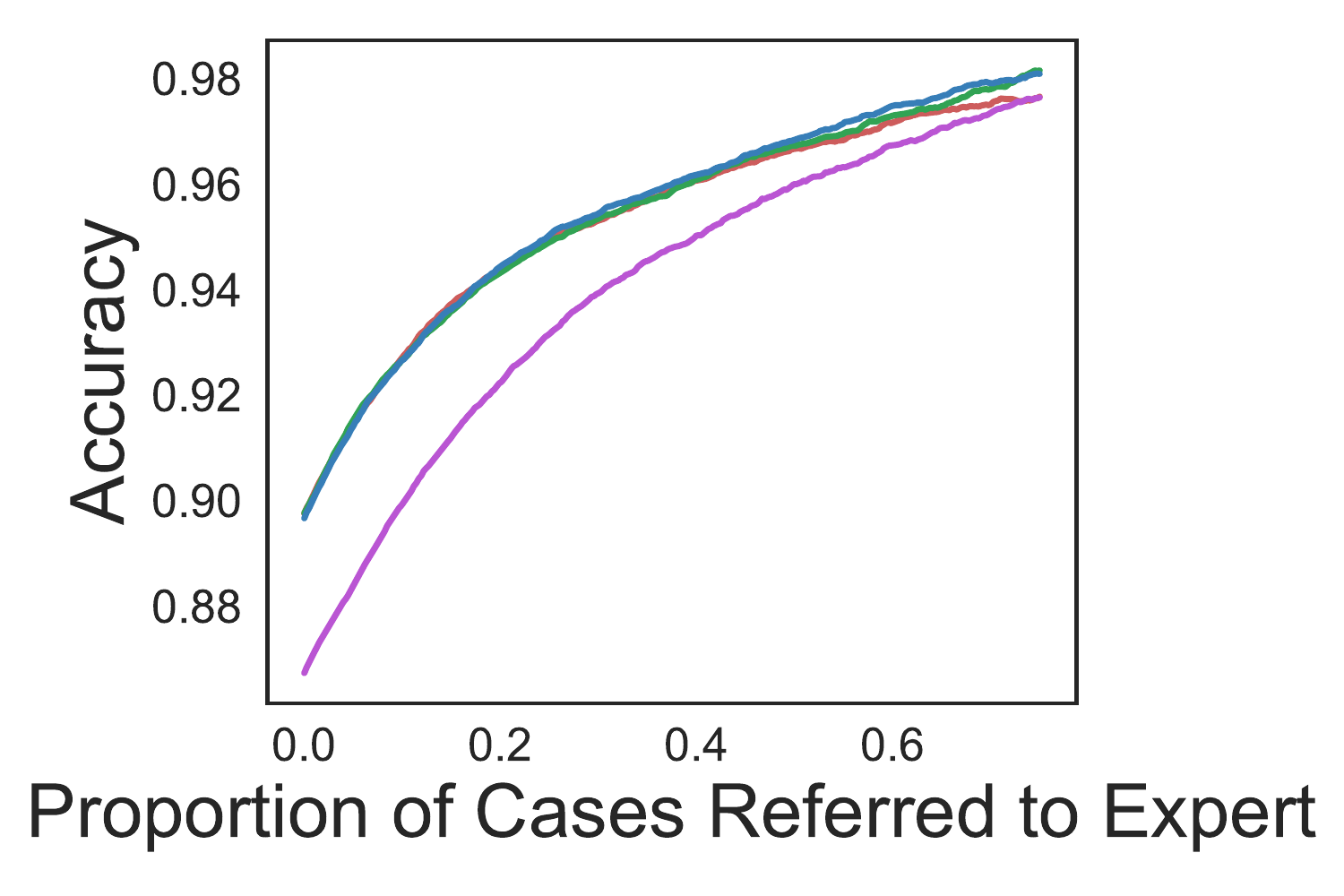}
    \caption{
        \textbf{Country Shift: In-Distribution\\$~$}
    }
\end{subfigure}
\hspace*{10pt}
\begin{subfigure}[l]{0.45\linewidth}
    \includegraphics[width=\linewidth]{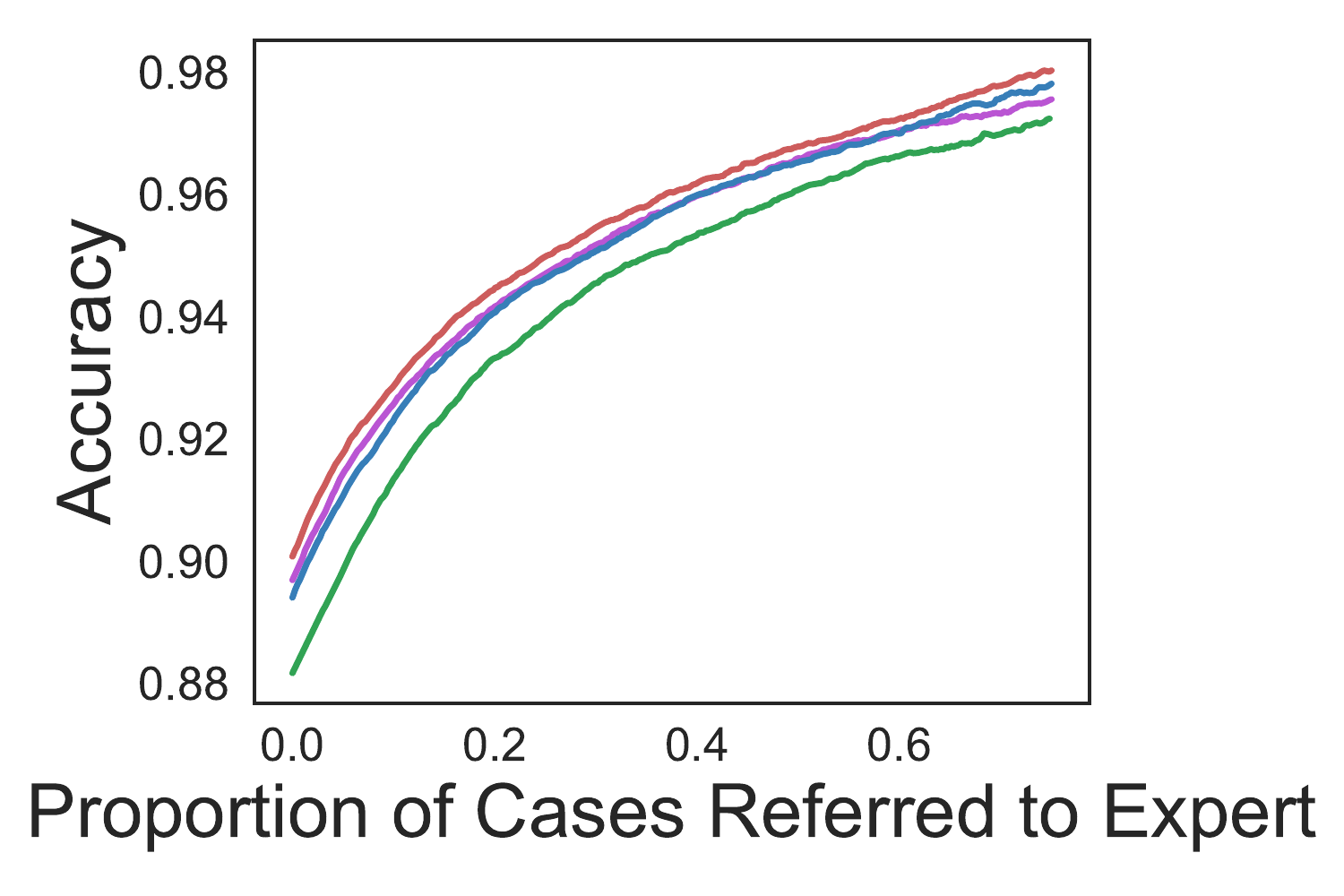}
    \caption{
        \textbf{Severity Shift: In-Distribution\\$~$}
    }
\end{subfigure}
\vspace{-3ex}
\caption{%
    \textbf{Selective Prediction on RETINA (In-Distribution).}
    \textbf{(left)} 
    The in-distribution Country Shift task is concerned with diagnosing diabetic retinopathy.
    \textbf{(right)}
    The in-distribution Severity Shift task is concerned with diagnosing diabetic retinopathy 
    amongst images containing at worst moderate cases of diabetic retinopathy.
    Accuracy is used as the evaluation metric and predictive entropy is used as the per-example uncertainty estimate.
}
\label{fig:retina_sel_pred_id}
\end{figure*}

\section{Extensions to Sparse Mixtures of Experts}\label{sec:sparse_moes}

In this section, we discuss extensions of some approaches of this paper in the light of recent advances in sparse mixtures of experts models (sparse MoEs).

\subsection{Background}
Sparse MoEs constitute a family of models that rely on \textit{conditional computation}~\citep{bengio2013estimating, bengio2015conditional} to combine multiple submodels---referred to as \textit{experts}---in an input-dependent fashion.
Sparse MoEs have successfully been applied both in NLP~\citep{shazeer2017outrageously,lepikhin2020gshard,fedus2021switch} and more recently in computer vision~\citep{riquelme2021scaling, lou2021moe}. The goal of conditional computation is to grow the number of parameters of the model while maintaining its training and inference costs constant, by only activating a particular subset of experts given an input. This is in contrast with classical neural networks that use all their parameters for each input.

In the context of this paper, we focus on~\citet{allingham2021sparse} that studied the robustness and uncertainty estimates of several extensions of ViT endowed with conditional computation, namely:
\begin{itemize}
    \item \textbf{V-MoE}: The authors of \citet{riquelme2021scaling} extended ViT to sparse MoEs by placing experts at the level of the MLPs of the transformer architecture. Since ViT operates on image patches, the conditional computation of V-MoEs also operates on image patches. More specifically, the image patches are sparsely routed to $K$ out of $E$ available experts (in practice, $E=32$). The experts are selected by computing some gating weights in combination with a top-$K$ strategy, allowing for an end-to-end training procedure.
    When only a single expert is selected ($K=1$), the training and inference costs of V-MoEs almost match those of a standard ViT model~\citep{riquelme2021scaling}, while leading to improved predictive performance at both pretraining and finetuning time. 
    
    In the next section, we will use the acronym \textsc{MoE} to refer to V-MoE.
    \item \textbf{Deep ensembles of V-MoEs}: \citet{allingham2021sparse} explored the combination of the static ensembling of deep ensembles and the conditional computation of V-MoEs. They found that both effects are complementary. In particular, over tasks where either V-MoEs or deep ensembles are known to perform well (e.g., respectively, few-shot classification and OOD detection), the combined approach inherits from the best of both worlds, while matching the cost of standard deep ensembles.
    
    In what follows, we will use the acronym $[\textsc{MoE}]_4$  to refer to a deep ensemble formed by four \textsc{MoE}s where both the upstream \textit{and} downstream models were varied. In the situation where only the downstream models were varied for a single fixed pretrained model, we use the notation $\textsc{MoE}\rightarrow[\textsc{MoE}]_4$. We will use the same convention for the model without any changes, None. While the rest of the paper uses ensembles of size 3, we consider in this section ensembles of size 4 to be faithful to the setting of  \citet{allingham2021sparse}.
    \item \textbf{\textsc{E}$^3$}: The authors of \citet{allingham2021sparse} further designed an efficient ensemble approach, referred to as \textsc{E}$^3$, to overcome the computational burden of the naive ensembling of V-MoEs described above. In a nutshell, \textsc{E}$^3$ jointly learns an ensemble of smaller sparse MoEs where all layers not equipped with experts (e.g., attention layers) are shared across the ensemble members. Interestingly, \textsc{E}$^3$ features some of the extensions of Plex due to some of its conceptual similarities with BE. \citet{allingham2021sparse} observed that \textsc{E}$^3$ tends to lie either on, or close to, the Pareto frontiers of several metrics---e.g., NLL, Brier score and few-shot classification error---versus the computational cost of the models, as measured by FLOPs.
\end{itemize}

\subsection{Results}

We present the results in Table~\ref{table:moe_comparison}. They extend the evaluation from \citet{allingham2021sparse} to the additional metrics of this paper related to the computer vision tasks. We summarize the results with the prediction, uncertainty and adaptation scores.

In agreement with the conclusions reported in \citet{allingham2021sparse}, we can see that $[\textsc{MoE}]_4$ outperforms\footnote{A closer inspection at the results shows that the prediction score of $[\textsc{MoE}]_4$ is slightly worse than that of $[\textsc{None}]_4$ because of isolated, slightly worse performance on CIFAR-10.} the standard deep ensemble $[\textsc{None}]_4$ while having the same computational cost. This confirms the complementary of static ensembling with the adaptivity of sparse MoEs within the broader evaluation of this paper. More generally, combining ensembling and sparse MoEs seems to be a promising direction to improve the reliability of models.

As observed in \citet{riquelme2021scaling}, \textsc{MoE} has strong performance in fewshot learning tasks, which is highlighted here by the adaptation scores of \textsc{MoE} and $[\textsc{MoE}]_4$.
Moreover, \citet{allingham2021sparse} further observed that \textsc{MoE} did not perform well in terms of calibration, as illustrated by its lower uncertainty score compared with \textsc{None}.
The efficient ensemble \textsc{E}$^3$ has performance comparable to deep ensembles formed from a single pretrained model (i.e., $\textsc{None}\rightarrow[\textsc{None}]_4$ and $\textsc{MoE}\rightarrow[\textsc{MoE}]_4$), while being substantially cheaper. This conclusion echoes the take-home messages from \citet{allingham2021sparse}.

\begin{table}[t]
\centering
\resizebox{\textwidth}{!}{
\begin{tabular}{ccccc}
\toprule
{} & \textsc{Score} & \textsc{Score prediction} & \textsc{Score uncertainty} & \textsc{Score adaptation} \\
\midrule
\textsc{None}                                       &          85.99 &                     91.13 &                      92.71 &                     84.65 \\
\textsc{MoE}                                      &          86.76 &                     90.52 &                      91.97 &                     85.75 \\
\textsc{E}$^3$                            &            $-$ &                     91.44 &                      92.77 &                       $-$ \\
\midrule
$\textsc{None}\rightarrow[\textsc{None}]_4$ &            $-$ &                     91.41 &                      92.83 &                       $-$ \\
$\textsc{MoE}\rightarrow[\textsc{MoE}]_4$ &            $-$ &                     91.07 &                      92.82 &                       $-$ \\
\midrule
$[\textsc{None}]_4$                        &          88.54 &                     91.99 &                      93.44 &                     87.59 \\
$[\textsc{MoE}]_4$                        &          88.66 &                     91.53 &                      93.51 &                     87.77 \\
\bottomrule
\end{tabular}
}
\caption{Results of sparse MoE variants developed in \citet{allingham2021sparse}. \textsc{E}$^3$, $\textsc{None}\rightarrow[\textsc{None}]_4$ and $\textsc{MoE}\rightarrow[\textsc{MoE}]_4$ do not have adaptation scores since those approaches are applied only at finetuning time.}
\label{table:moe_comparison}
\end{table}

\section{Analysis of Heteroscedastic Last Layer}
\label{app:heteroscedastic}
We add a heteroscedastic output layer to the base deterministic model and the BE base model. We assess whether enabling the network to model input-dependent (heteroscedastic) label noise results in better performance on datasets known to have noisy labels.

\subsection{Heteroscedastic improves pretraining performance on JFT.}

\begin{figure}[!tb]
    \centering
    \begin{subfigure}{.49\textwidth}
        \centering
        \includegraphics[width=\textwidth]{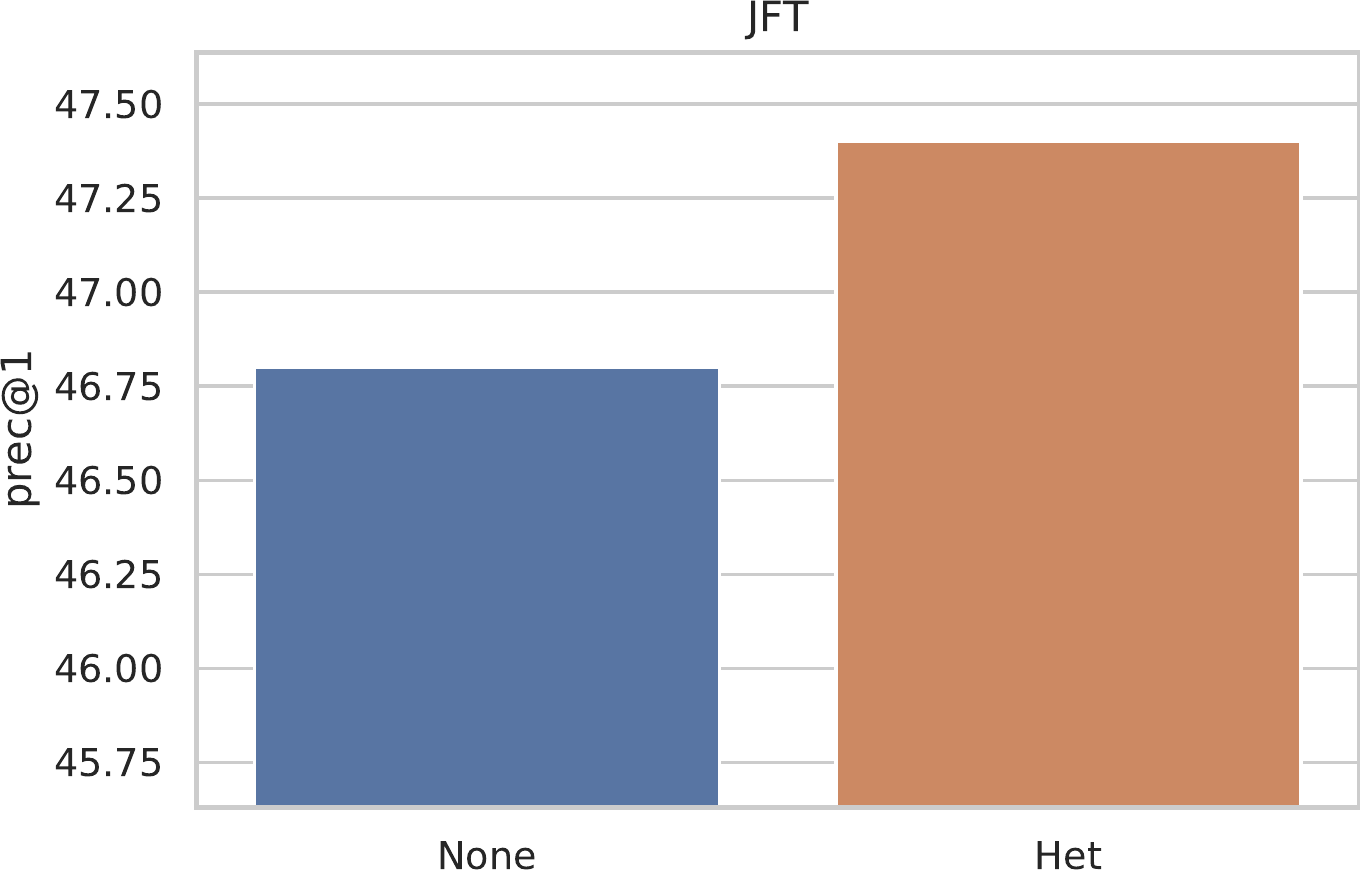}
        \caption{Accuracy
        }
    \end{subfigure}
    \begin{subfigure}{.49\textwidth}
        \centering
        \includegraphics[width=\textwidth]{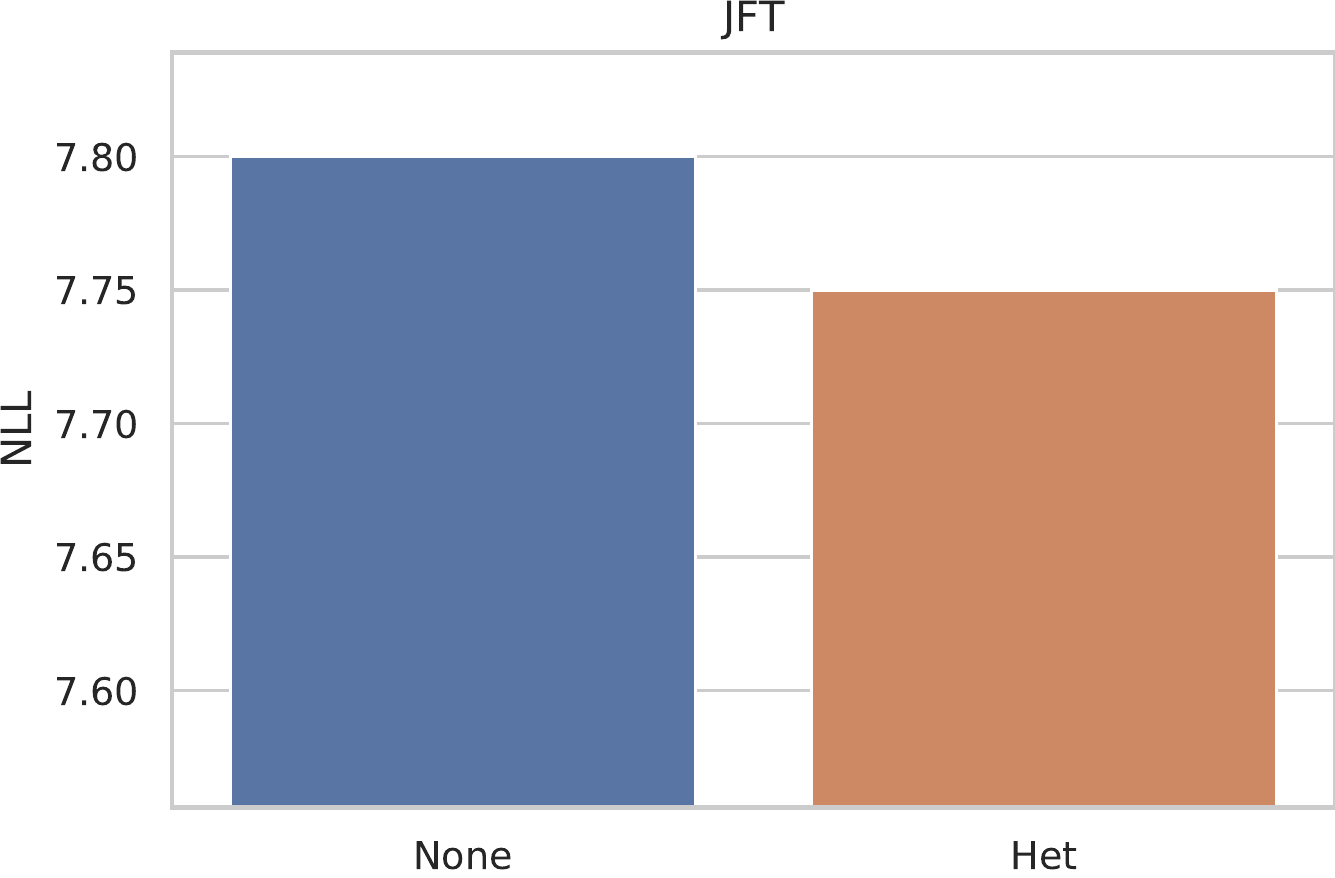}
        \caption{NLL}
    \end{subfigure}
    \caption{
        Performance on JFT for deterministic and heteroscedastic models.
    }
    \label{fig:jft_het}
\end{figure}

\begin{figure}[!tb]
    \centering
    \begin{subfigure}{.49\textwidth}
        \centering
        \includegraphics[width=\textwidth]{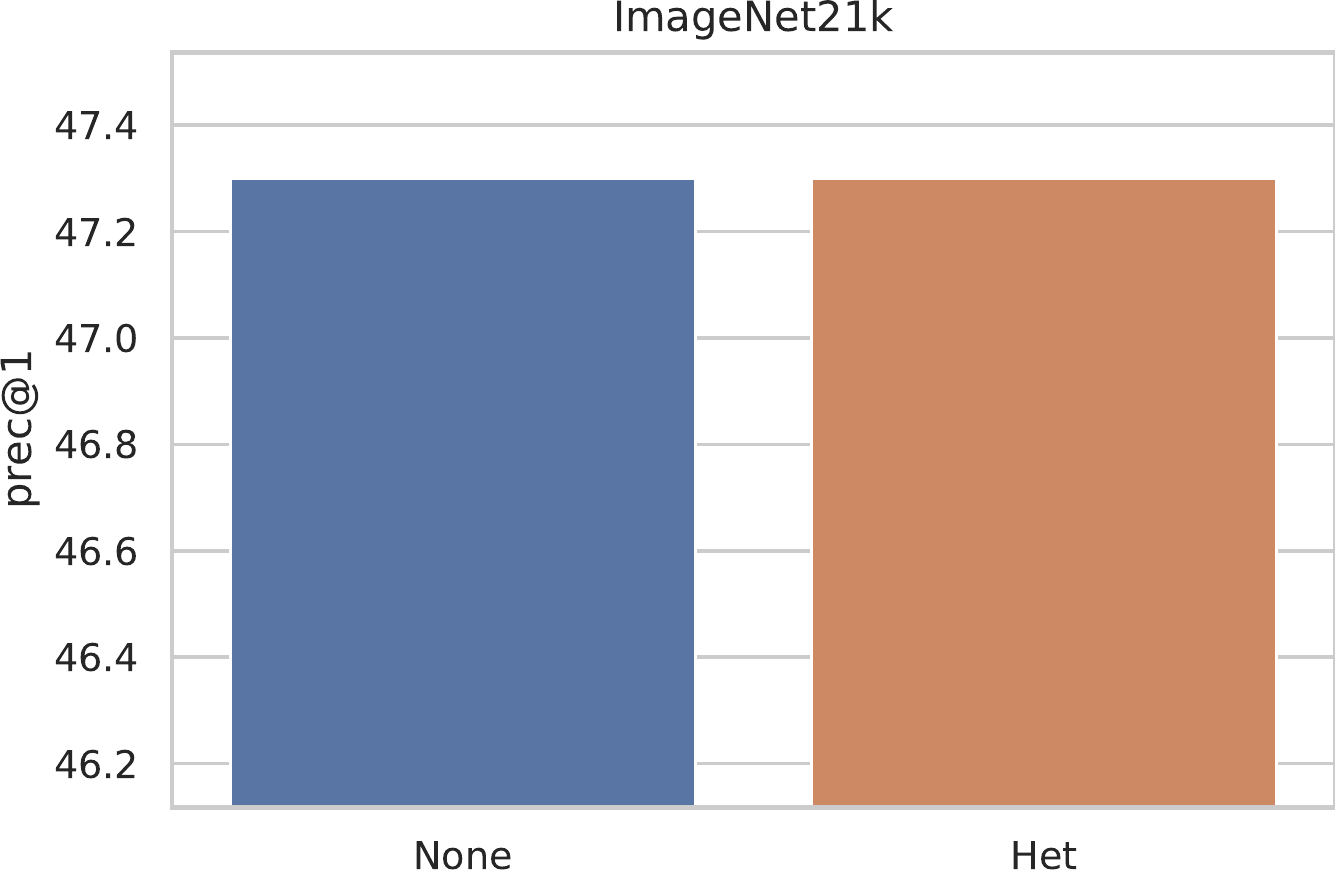}
        \caption{Accuracy
        }
    \end{subfigure}
    \begin{subfigure}{.49\textwidth}
        \centering
        \includegraphics[width=\textwidth]{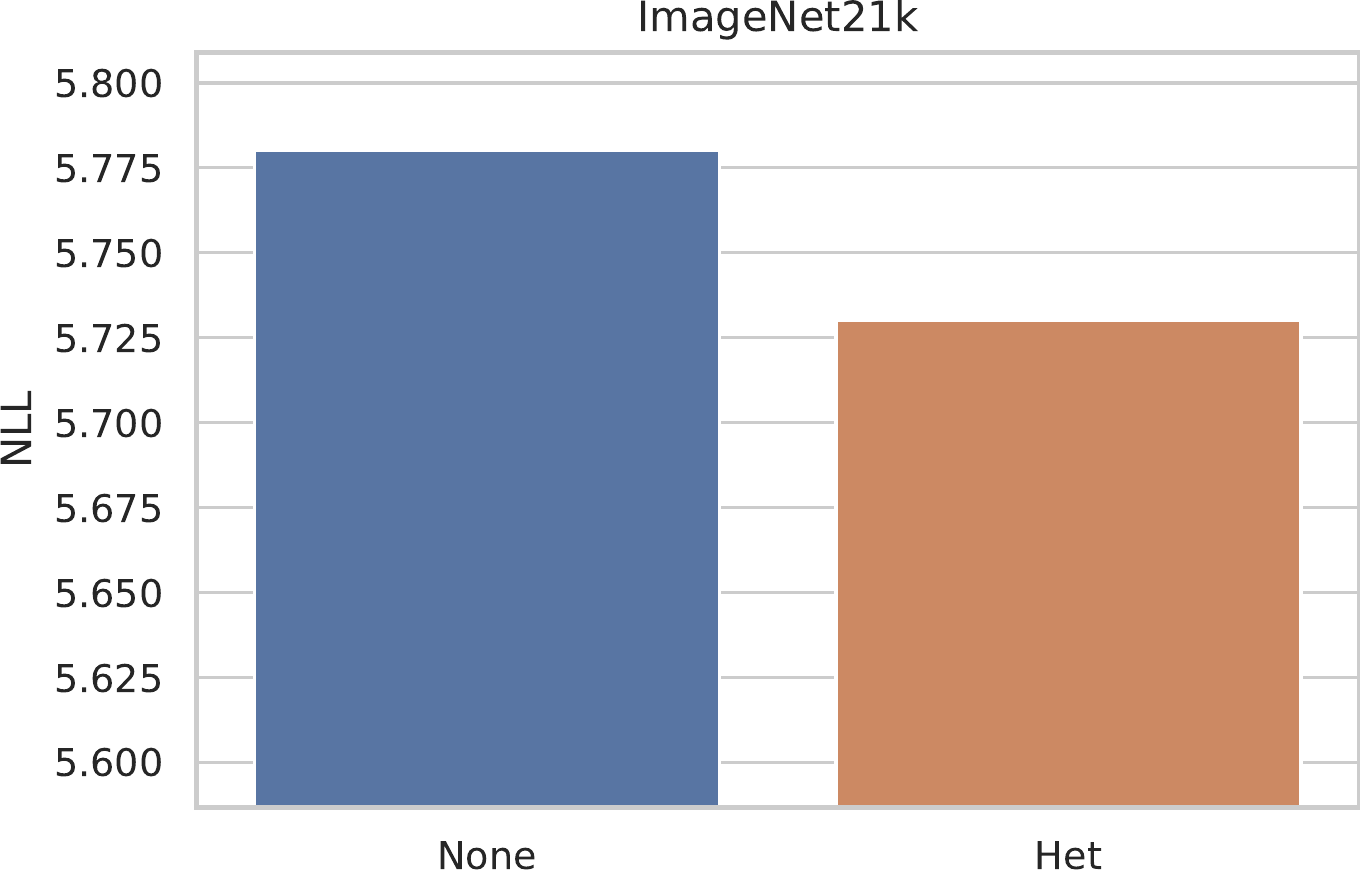}
        \caption{NLL}
    \end{subfigure}
    \caption{
        Performance on ImageNet21K for deterministic and heteroscedastic models.
    }
    \label{fig:inet21k_het}
\end{figure}

In \Cref{fig:jft_het} and \Cref{fig:inet21k_het} the heteroscedastic model, when applied on top of the base deterministic model, provides performance gains on the JFT dataset while performance is neutral on ImageNet-21K.

\begin{figure}[!tb]
    \centering
    \begin{subfigure}{.32\textwidth}
        \centering
        \includegraphics[width=\textwidth]{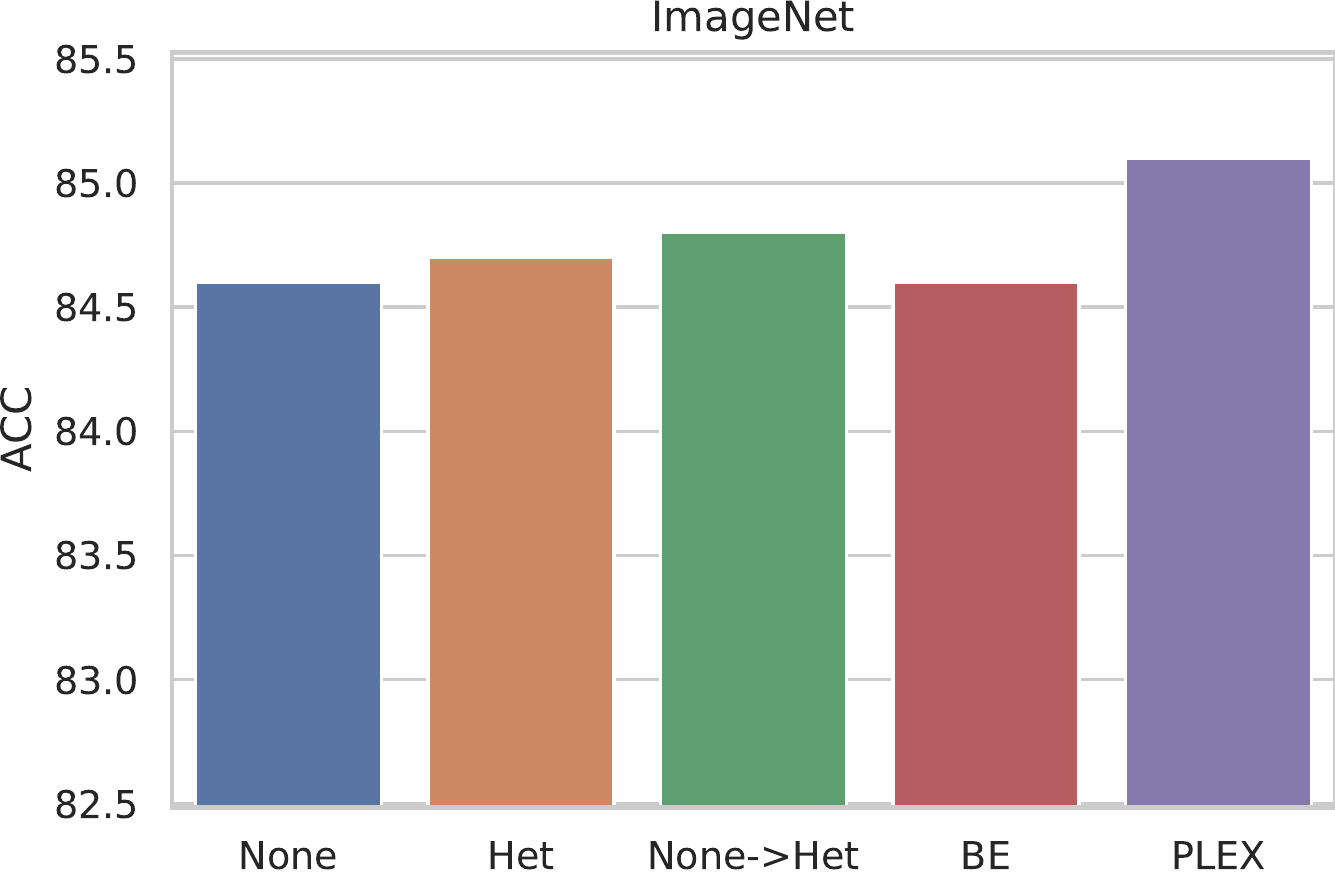}
        \caption{Accuracy}
    \end{subfigure}
    \begin{subfigure}{.32\textwidth}
        \centering
        \includegraphics[width=\textwidth]{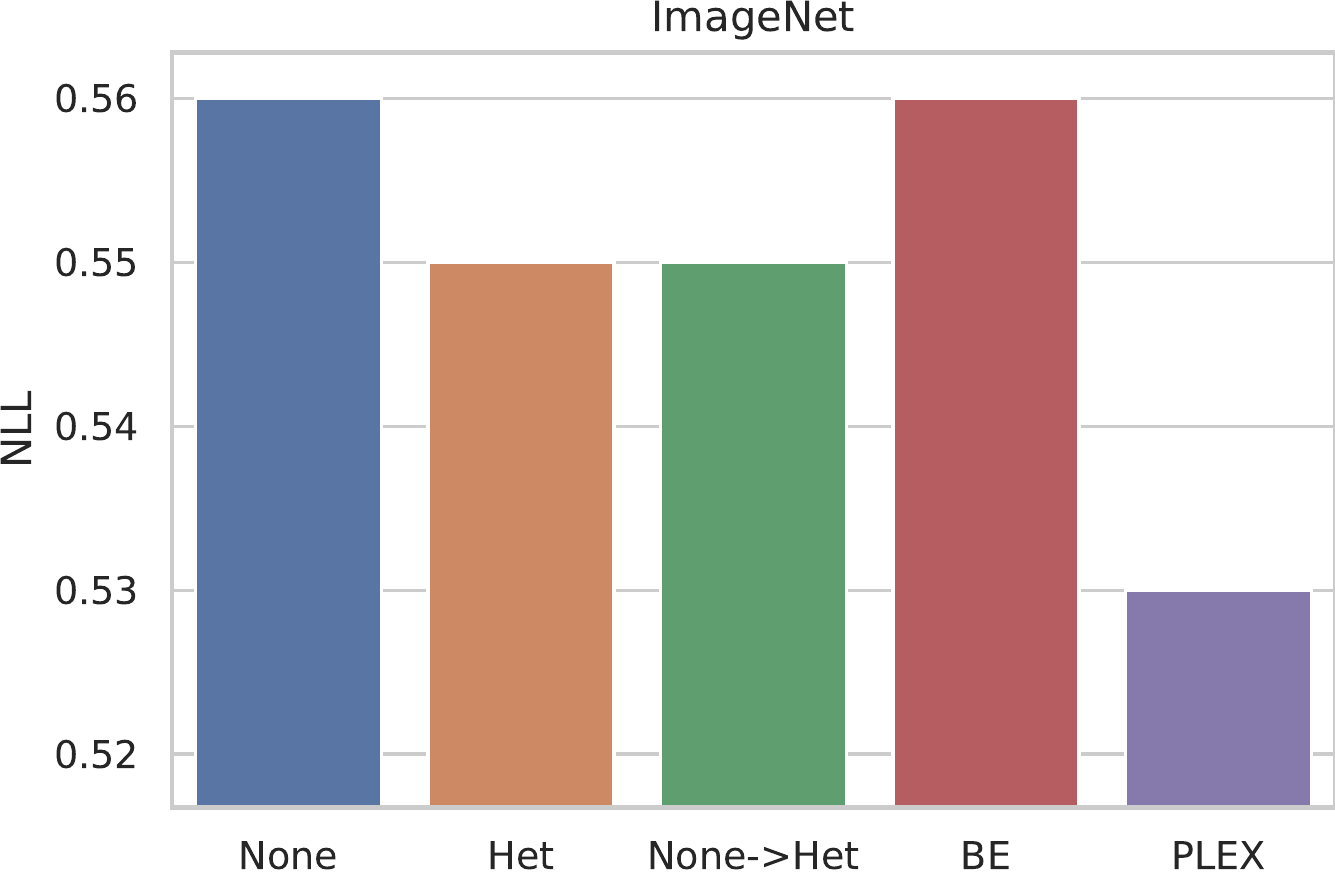}
        \caption{NLL}
    \end{subfigure}
    \begin{subfigure}{.32\textwidth}
        \centering
        \includegraphics[width=\textwidth]{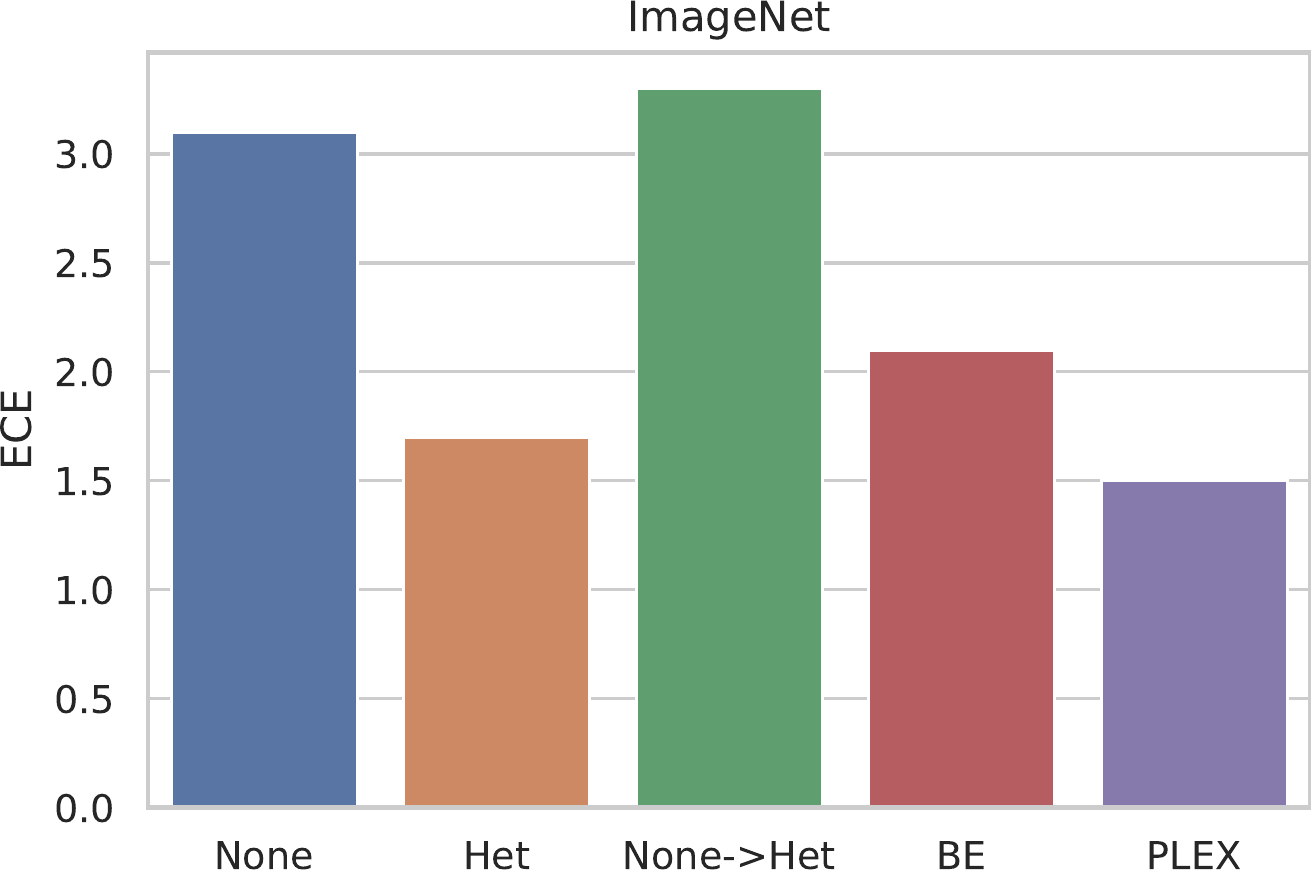}
        \caption{ECE}
    \end{subfigure}
    \caption{
        Performance on ImageNet1K for deterministic, heteroscedastic (upstream and downstream), deterministic upstream$\to$heteroscedastic downstream, BatchEnsemble L (upstream and downstream) and BatchEnsemble L upstream$\to$BatchEnsemble L + heteroscedastic downstream models.
    }
    \label{fig:inet_het}
\end{figure}

\begin{figure}[!tb]
    \centering
    \begin{subfigure}{.32\textwidth}
        \centering
        \includegraphics[width=\textwidth]{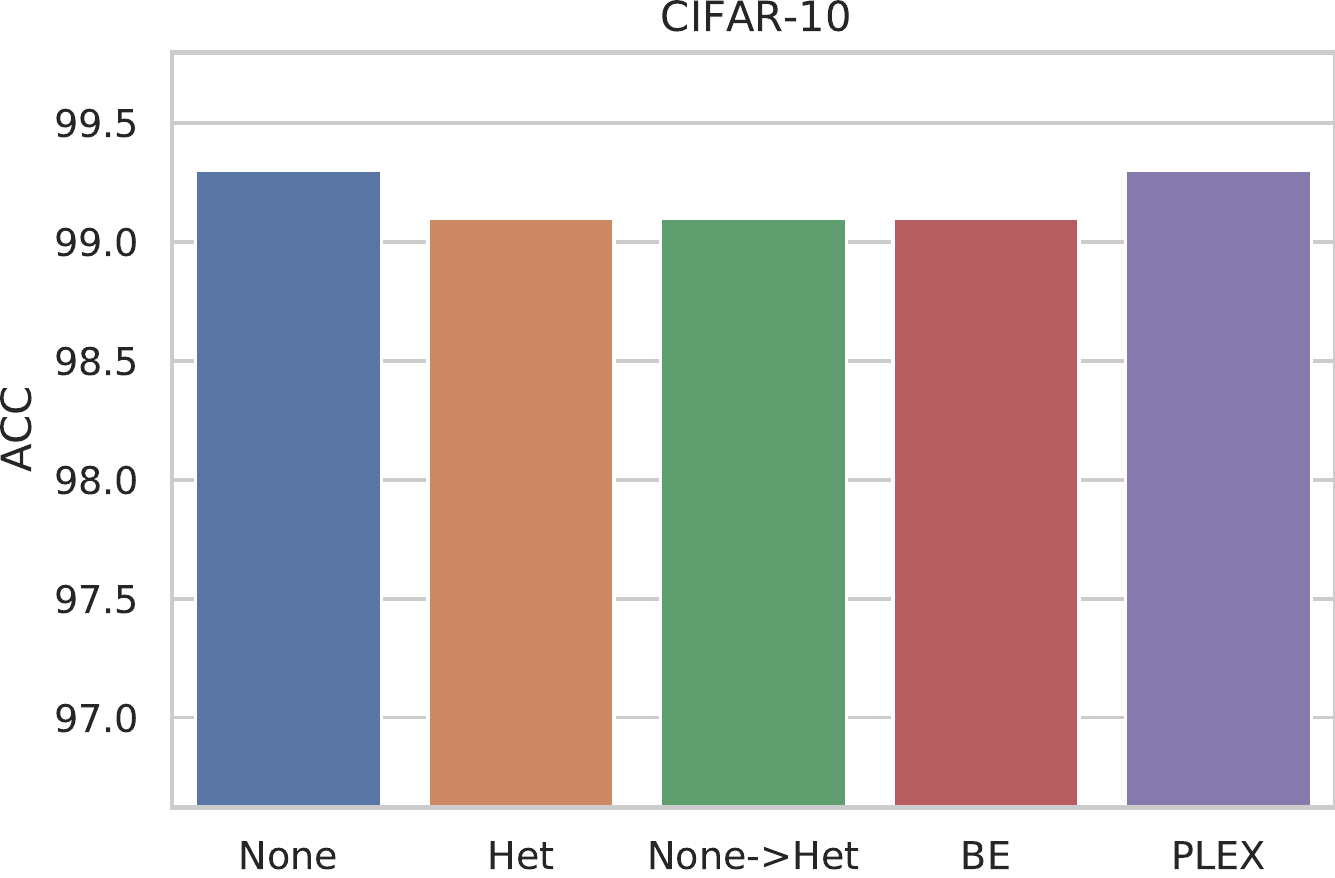}
        \caption{Accuracy}
    \end{subfigure}
    \begin{subfigure}{.32\textwidth}
        \centering
        \includegraphics[width=\textwidth]{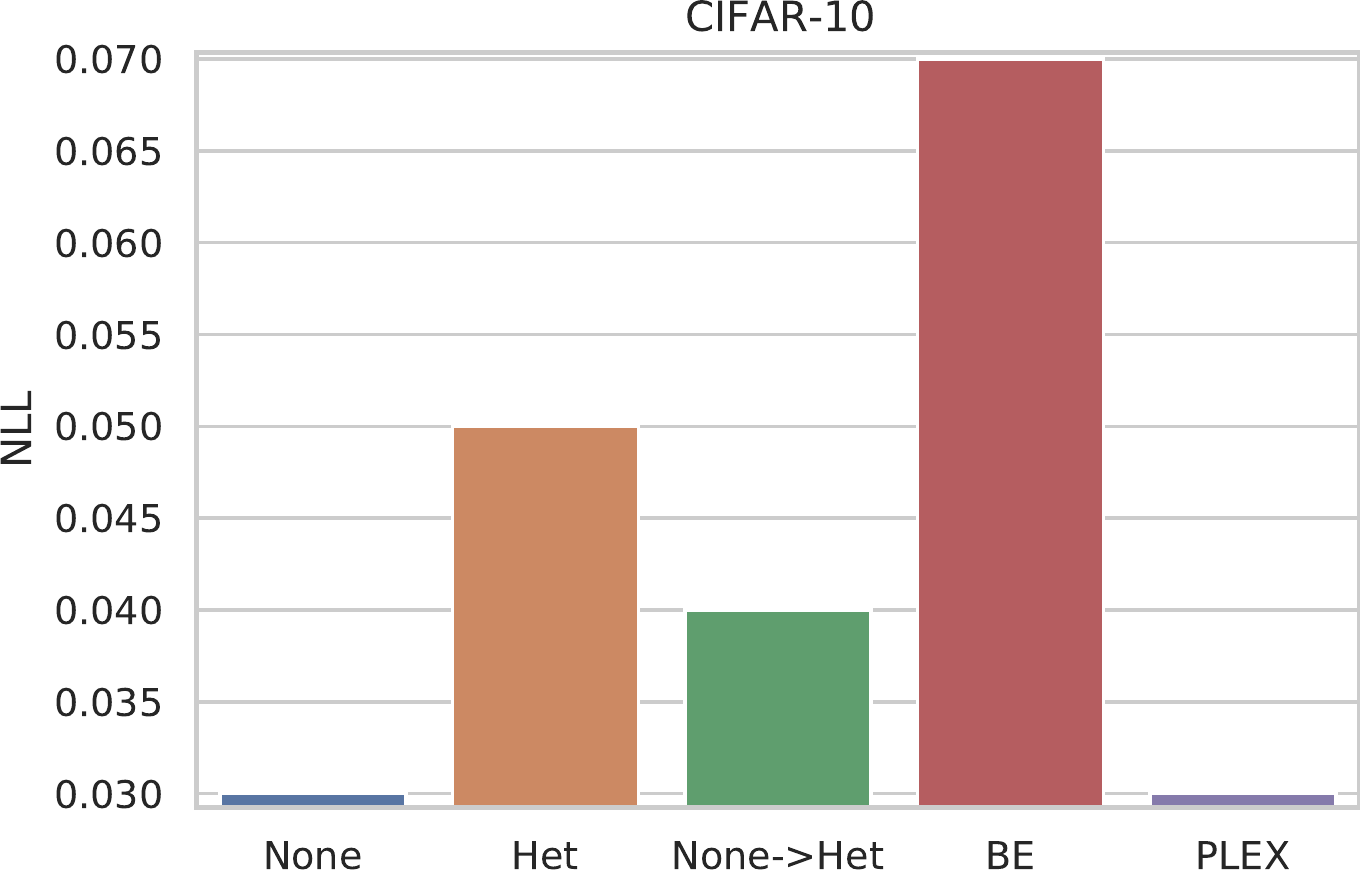}
        \caption{NLL}
    \end{subfigure}
    \begin{subfigure}{.32\textwidth}
        \centering
        \includegraphics[width=\textwidth]{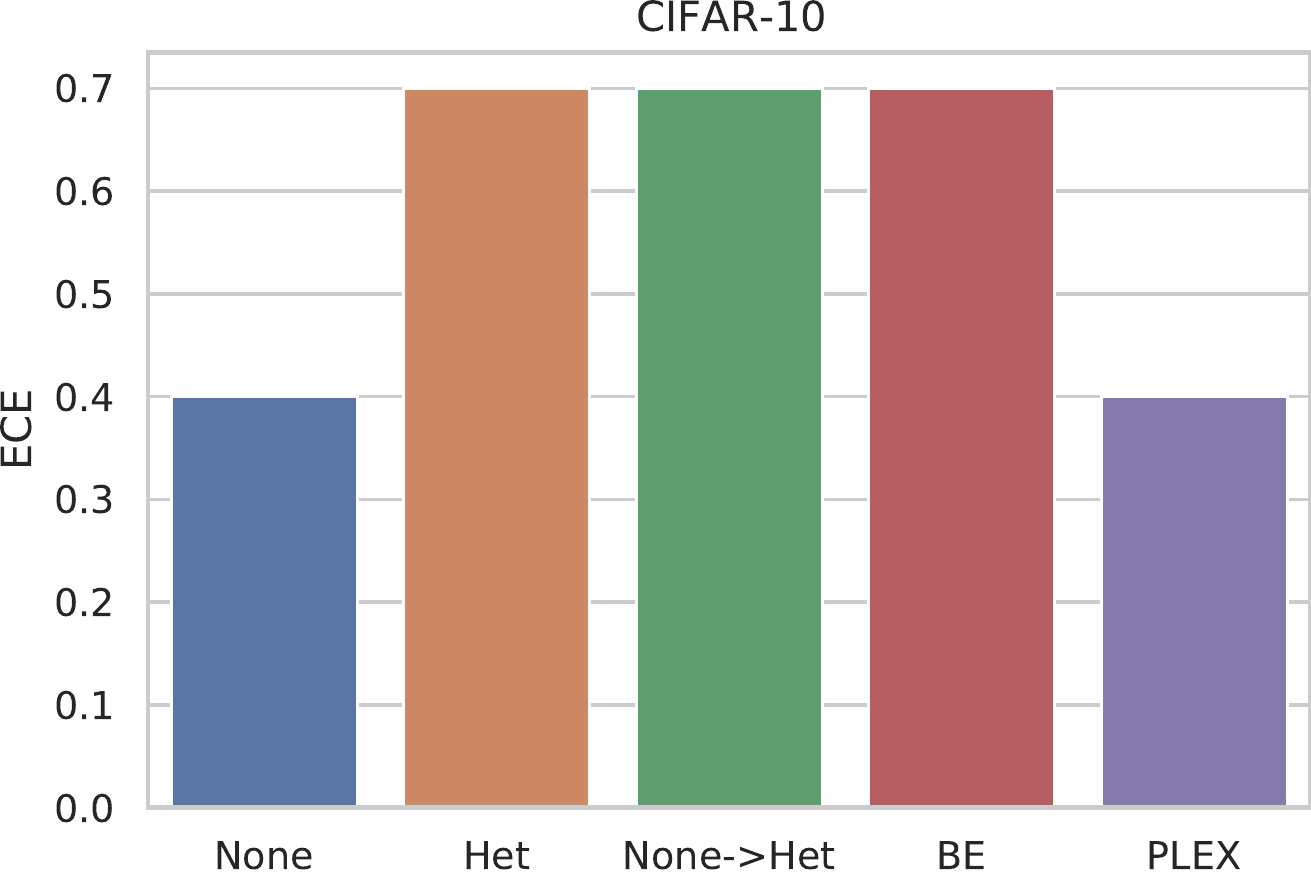}
        \caption{ECE}
    \end{subfigure}
    \caption{
        Performance on CIFAR-10 for deterministic, heteroscedastic (upstream and downstream), deterministic upstream$\to$heteroscedastic downstream, BatchEnsemble L (upstream and downstream) and BatchEnsemble L upstream$\to$BatchEnsemble L + heteroscedastic downstream models.
    }
    \label{fig:cifar10_het}
\end{figure}

\begin{figure}[!tb]
    \centering
    \begin{subfigure}{.32\textwidth}
        \centering
        \includegraphics[width=\textwidth]{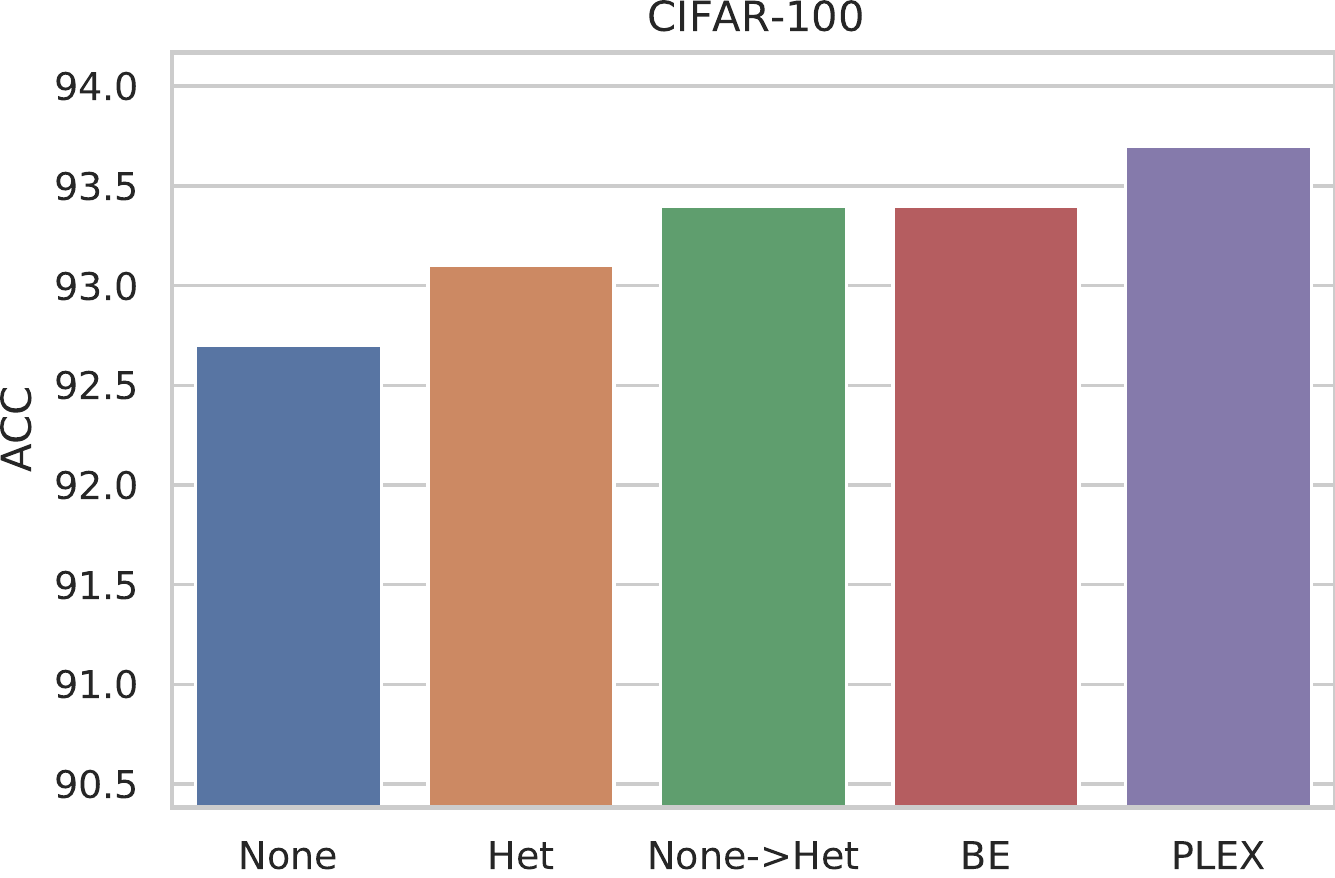}
        \caption{Accuracy}
    \end{subfigure}
    \begin{subfigure}{.32\textwidth}
        \centering
        \includegraphics[width=\textwidth]{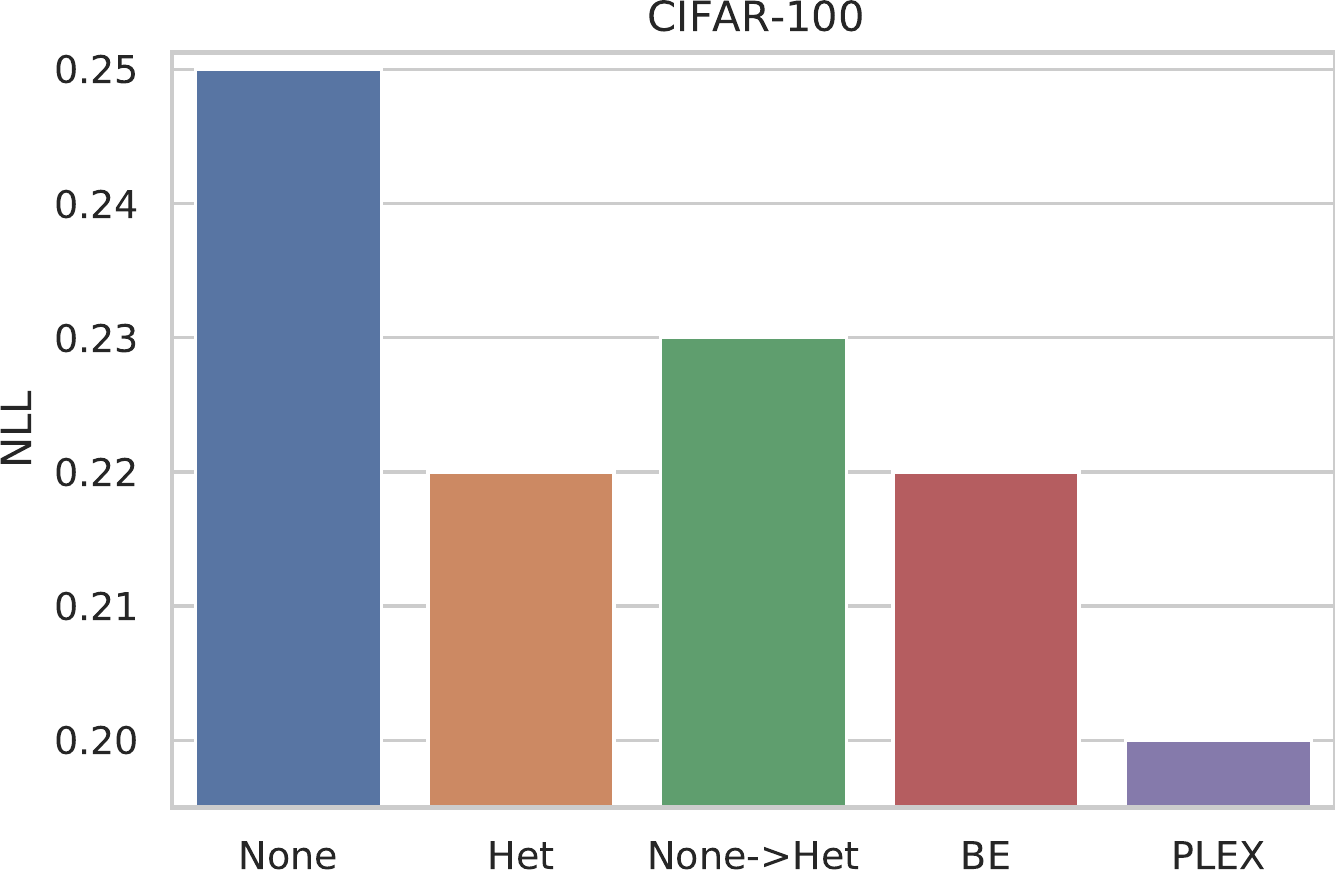}
        \caption{NLL}
    \end{subfigure}
    \begin{subfigure}{.32\textwidth}
        \centering
        \includegraphics[width=\textwidth]{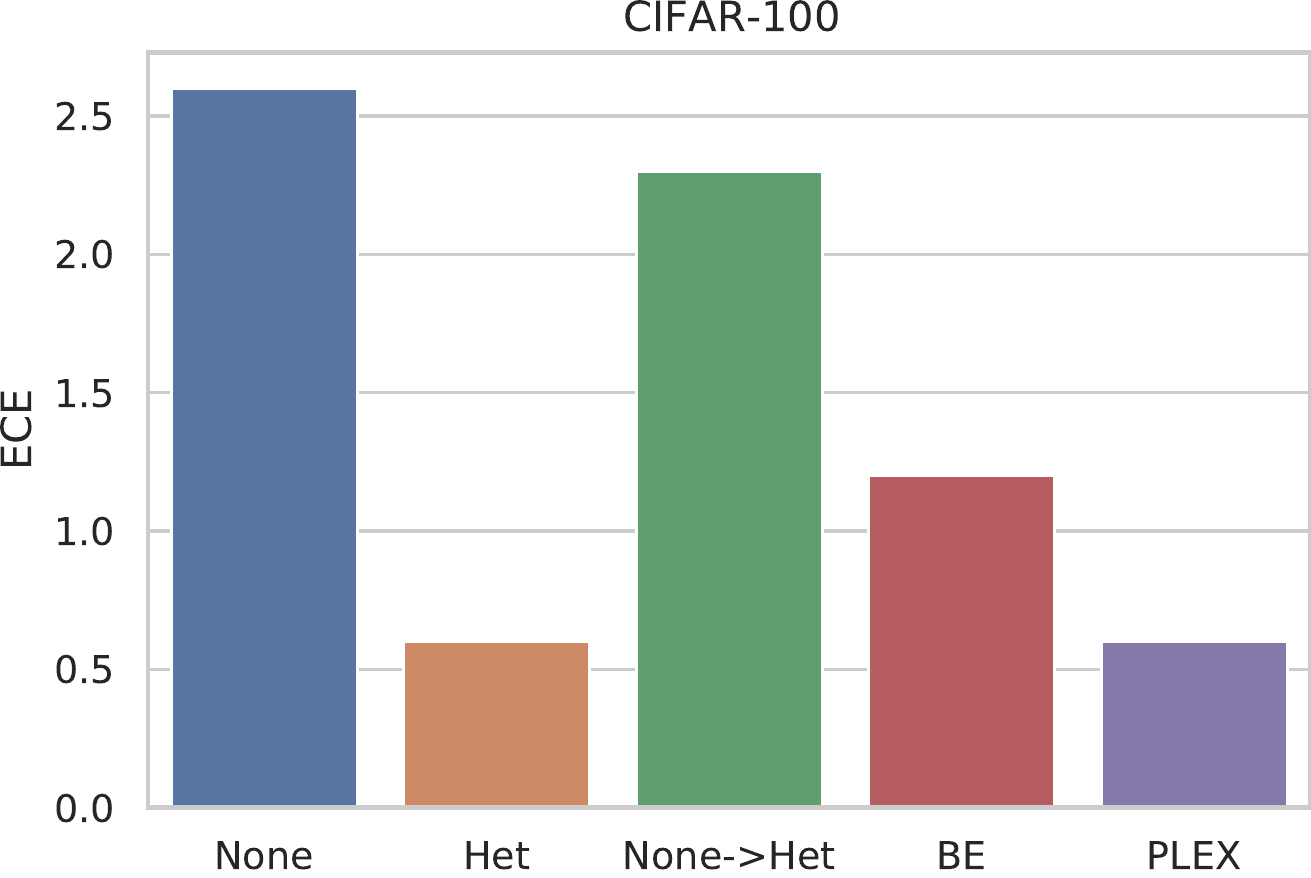}
        \caption{ECE}
    \end{subfigure}
    \caption{
        Performance on CIFAR-100 for deterministic, heteroscedastic (upstream and downstream), deterministic upstream$\to$heteroscedastic downstream, BatchEnsemble L (upstream and downstream) and BatchEnsemble L upstream$\to$BatchEnsemble L + heteroscedastic downstream models.
    }
    \label{fig:cifar100_het}
\end{figure}

\subsection{Impact of heteroscedastic last layer downstream}

In \Cref{fig:inet_het}, \Cref{fig:cifar10_het} and \Cref{fig:cifar100_het} we look at the core in-distribution performance metrics for downstream vision datasets. Het$\to$Het refers to a model trained with a heteroscedastic output layer upstream on JFT and finetuned on the target dataset with a heteroscedastic head. None$\to$Het is similar but where the cheaper and simpler baseline deterministic model without a heteroscedastic head is used upstream. BE$\to$BE + Het refers to pretraining using a Batch Ensemble model upstream on JFT and finetuning with a BE model with a heteroscedastic head downstream.

\paragraph{On all downstream datasets None$\to$Het performs similarly or better than Het$\to$Het.}
The outperformance of a deterministically pre-trained model finetuned with a heteroscedastic last layer, indicates that the gains from a heteroscedastic model downstream can be realized without heteroscedastic upstream pretraining. This is a surprising result given that the heteroscedastic model improves JFT performance upstream. However this result suggests that when the representations are transferred from this upstream model the Deterministic model is sufficient. Given that heteroscedastic model is more expensive than the deterministic model and requires tuning the temperature and the rank for the low-rank approximation hyperparameters this is an encouraging result.

\paragraph{Plex: heteroscedastic last layer helps downstream.}
None$\to$Het outperforms None on all metrics for ImageNet and CIFAR-100. Performance on CIFAR-10 is saturated and similar. This demonstrates that applying the heteroscedastic head on the downstream model improves downstream performance. Further, see \Cref{fig:ranking} which shows that on a wide variety of reliability metrics beyond in-distribution accuracy, NLL and ECE None$\to$Het outperforms None. We will also see later that the heteroscedastic method models label uncertainty better than baseline models.
Heteroscedastic can be combined with BE upstream pretraining for excellent results. The results in the above table and the more comprehensive summary results in \Cref{fig:ranking} show that the heteroscedastic head can be combined with a BE model to give the best performance single model on vision tasks (this is \textsc{Plex}). BE$\to$BE$+$Het (\textsc{Plex}) combines the epistemic uncertainty model capability of Batch Ensembles with the label uncertainty modeling of the heteroscedastic method. \textsc{Plex} approaches Deep Ensembles in terms of overall reliability, despite being substantially cheaper. Therefore we can recommend this model as our best single model for use by practitioners.

\section{Summarization of Language Results}
\label{appendix:language}

\begin{figure*}[!tb]
\begin{subfigure}{.24\textwidth}
\centering
\includegraphics[width=\textwidth]{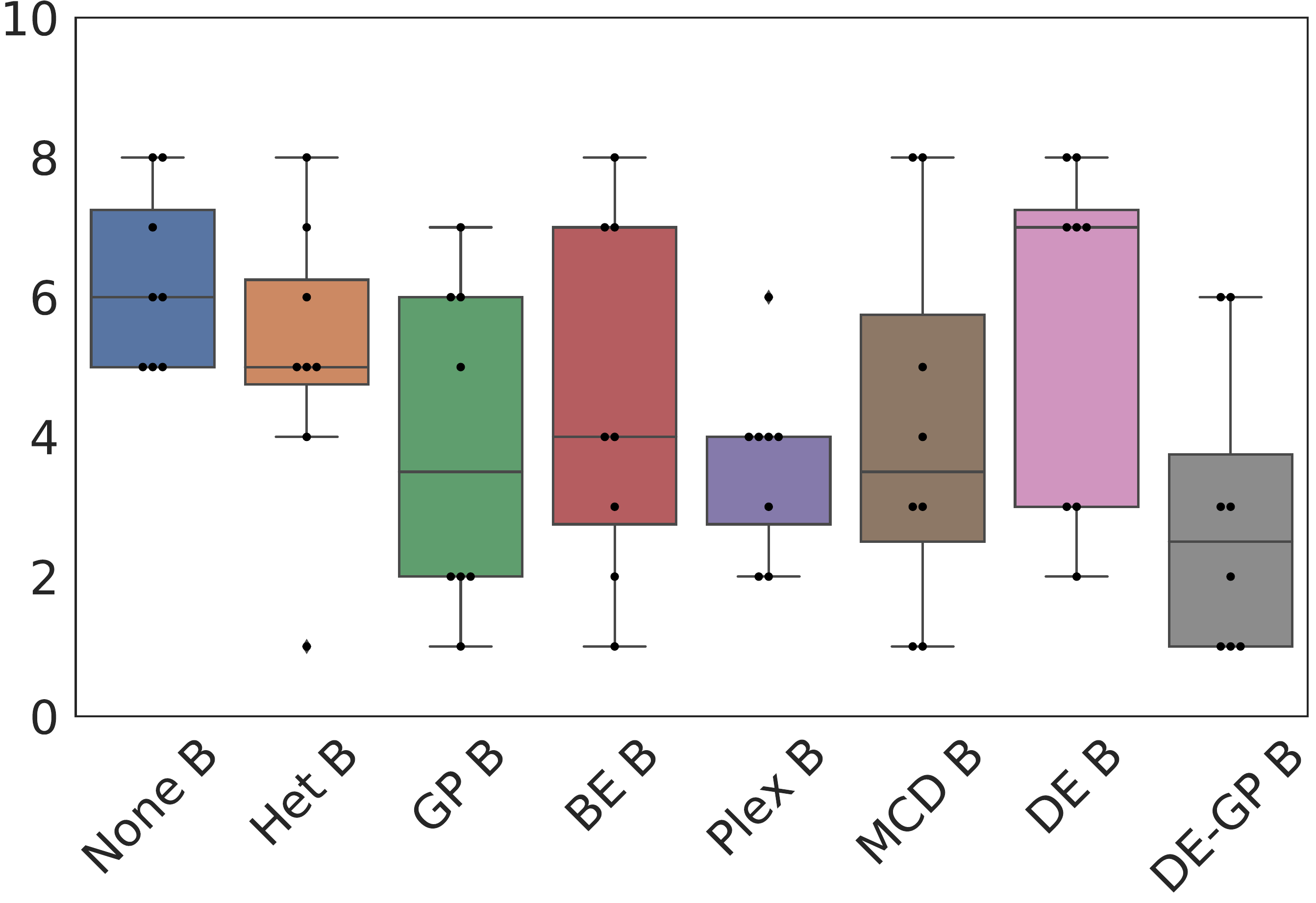}
\caption{\scriptsize{Generalization, IND.}}
\end{subfigure}
\hfill
\begin{subfigure}{.24\textwidth}
\centering
\includegraphics[width=\textwidth]{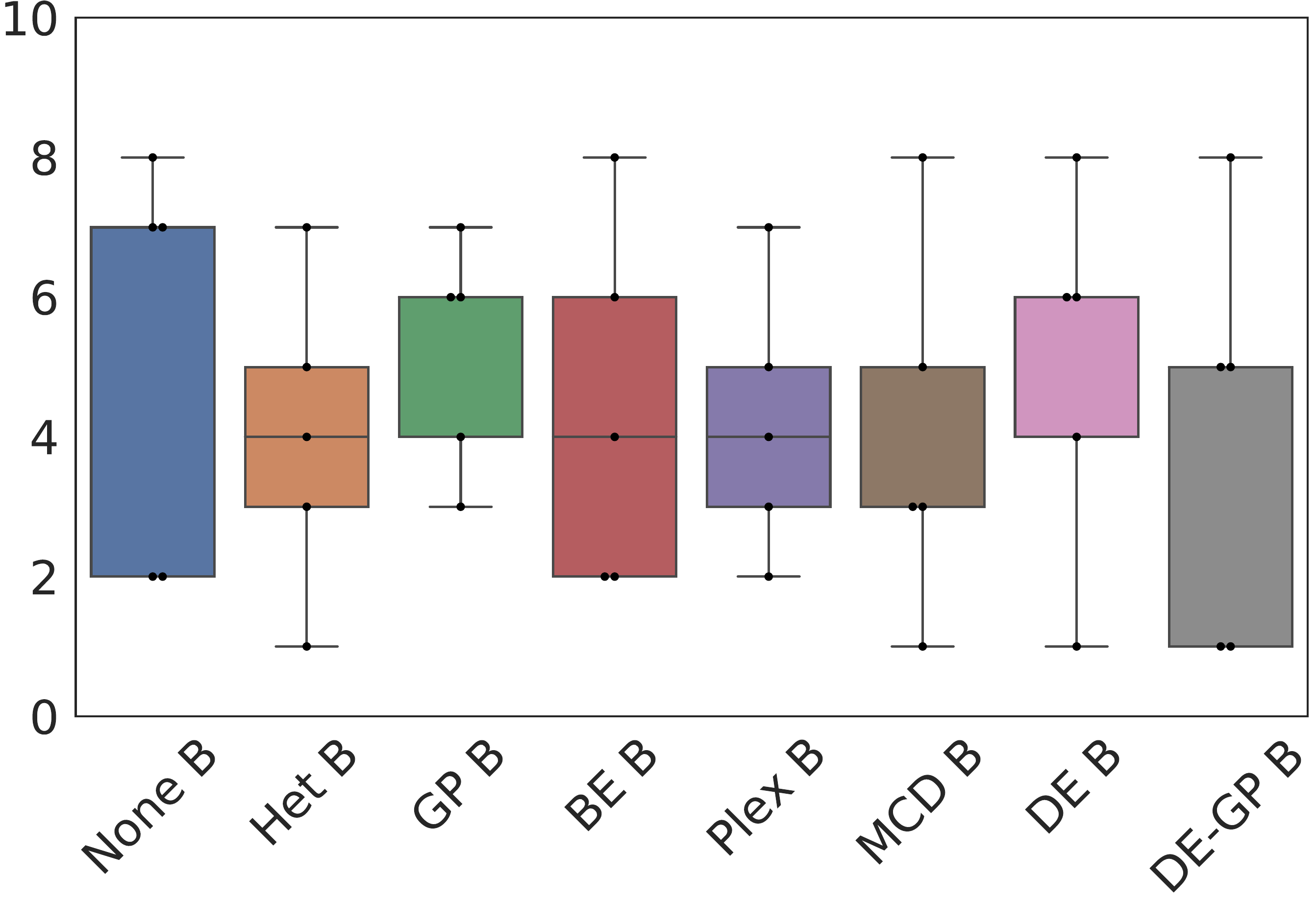}
\caption{\scriptsize{Generalization., OOD.}}
\end{subfigure}
\hfill
\begin{subfigure}{.24\textwidth}
\centering
\includegraphics[width=\textwidth]{figures/t5/generalization-sub-population-uncertainty-method}
\caption{\scriptsize{Generalization, SUB.}}
\label{fig:t5-rank-sub-generalization-size}
\end{subfigure}
\hfill
\begin{subfigure}{.24\textwidth}
\centering
\includegraphics[width=\textwidth]{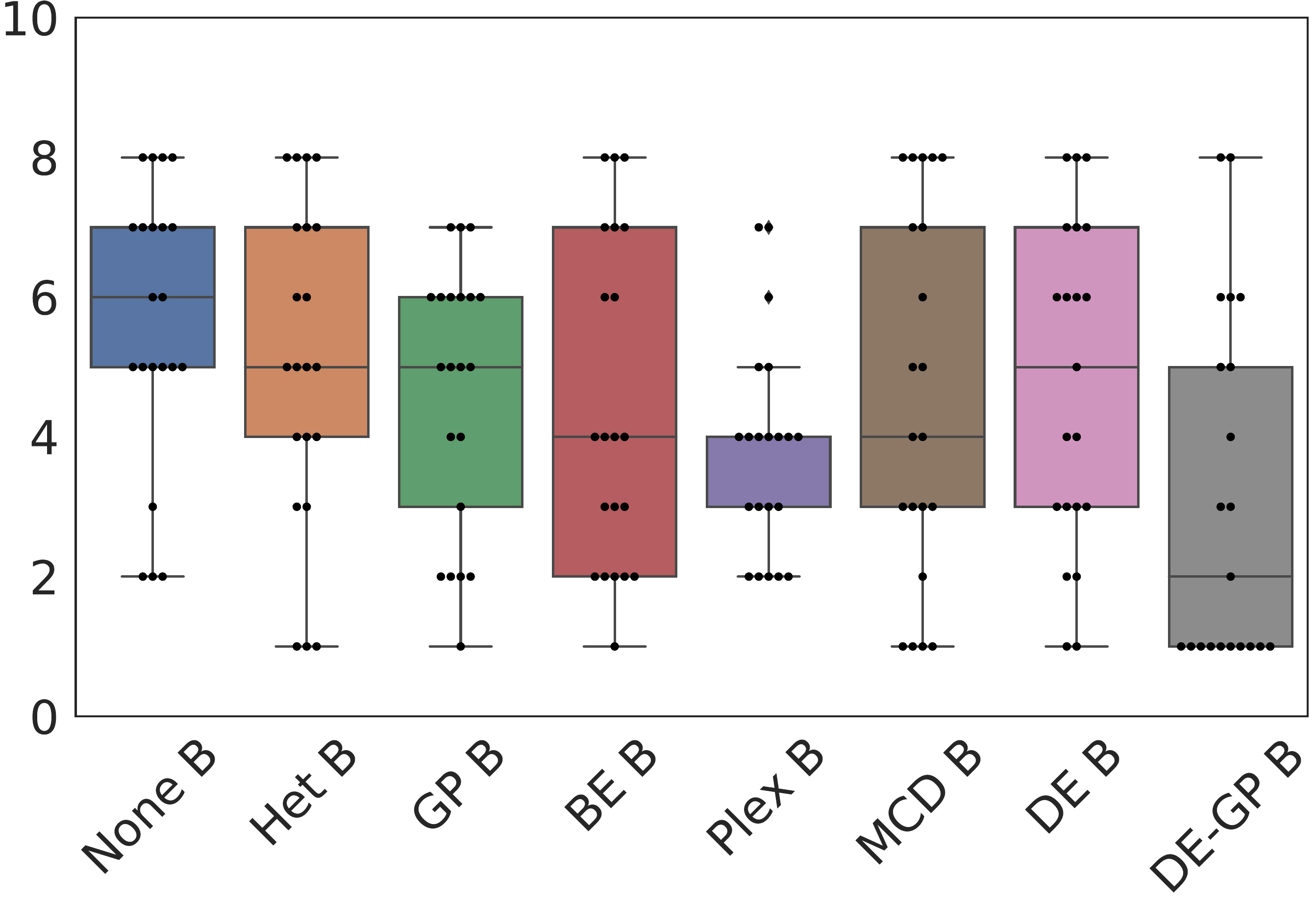}
\caption{\scriptsize{Generalization, ALL.}}
\end{subfigure}
\begin{subfigure}{.24\textwidth}
\centering
\includegraphics[width=\textwidth]{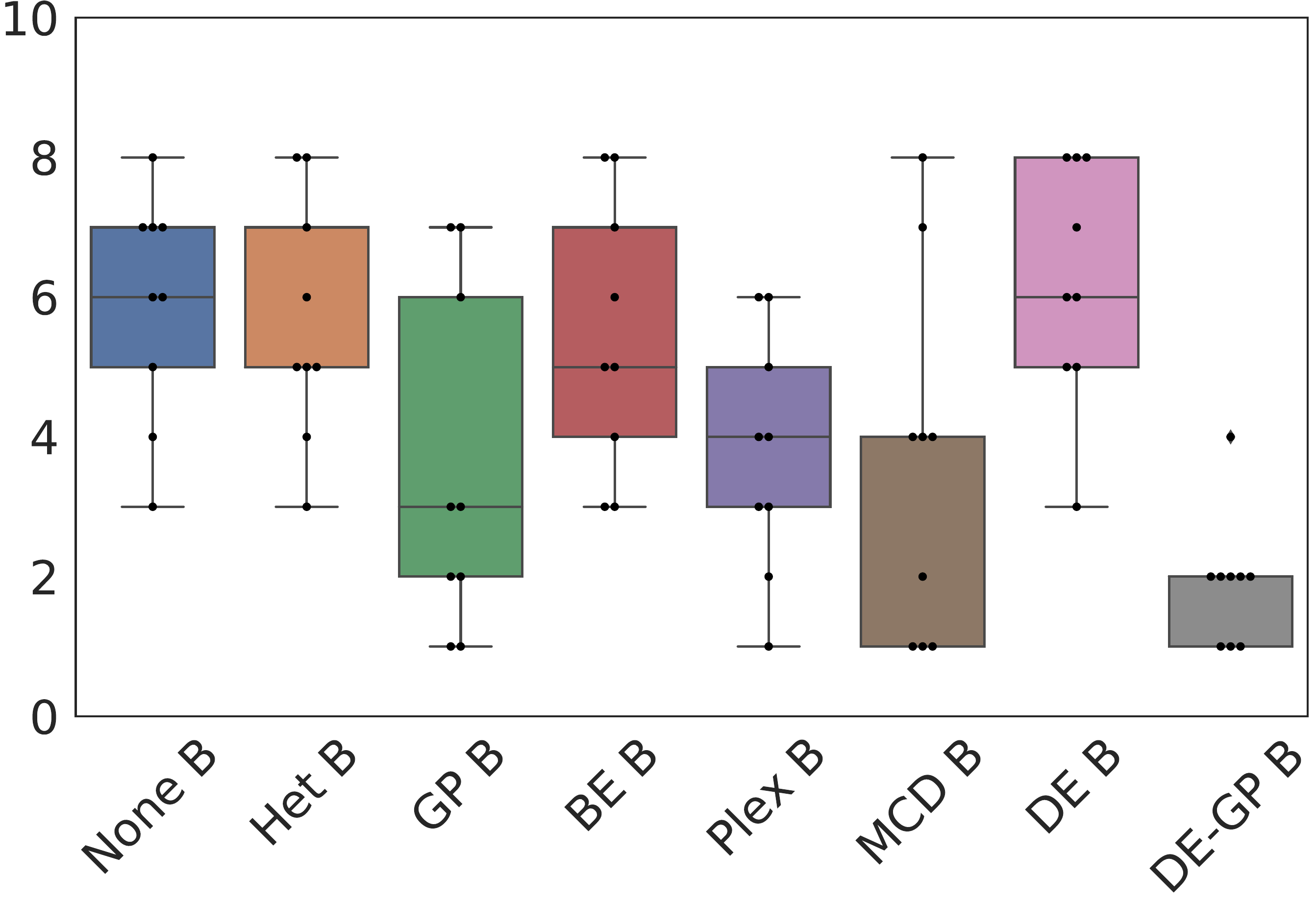}
\caption{\scriptsize{Calibration, IND.}}
\label{fig:t5-rank-ind-calibration-size}
\end{subfigure}
\hfill
\begin{subfigure}{.24\textwidth}
\centering
\includegraphics[width=\textwidth]{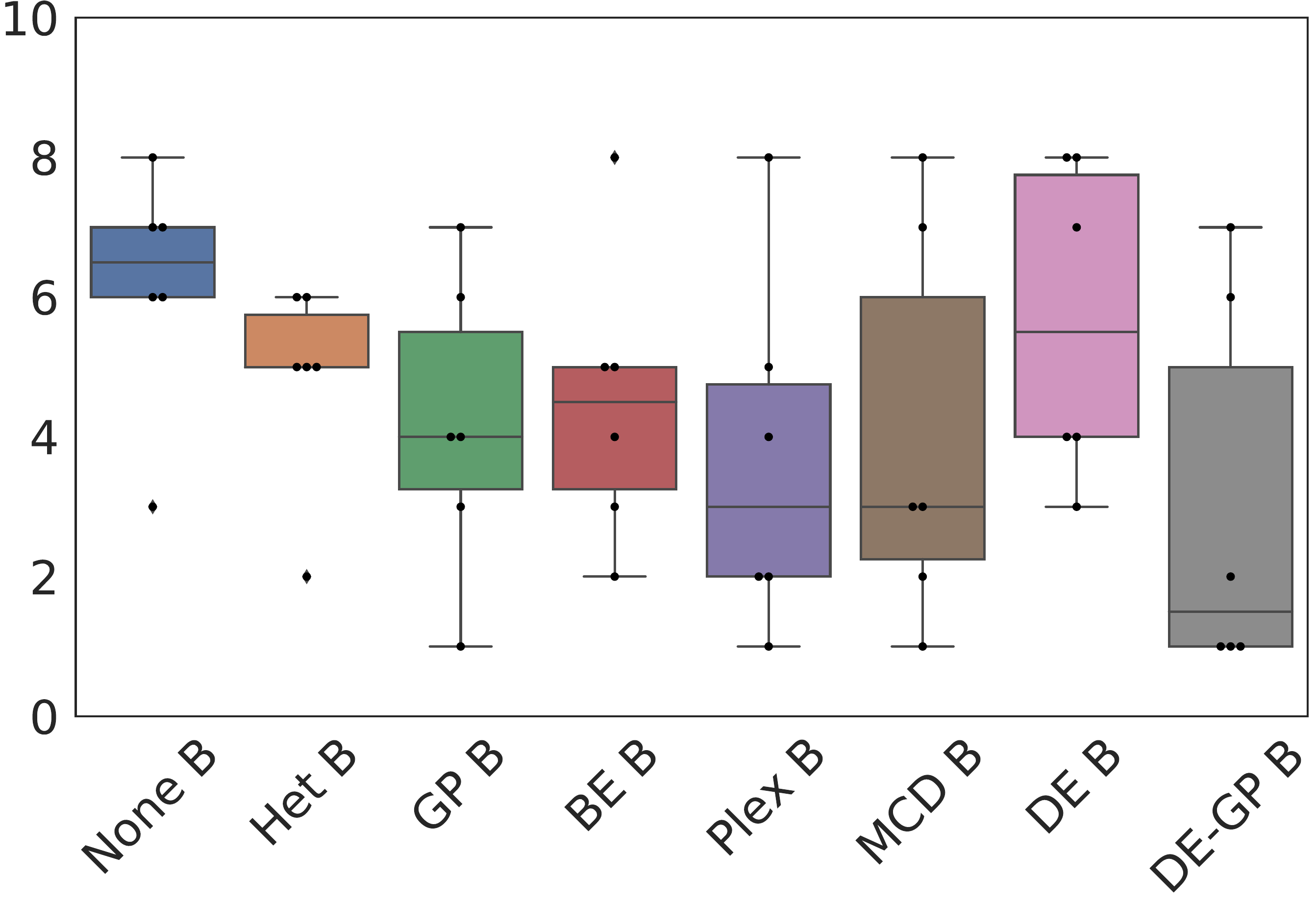}
\caption{\scriptsize{Calibration, OOD.}}
\label{fig:t5-rank-ood-calibration-size}
\end{subfigure}
\hfill
\begin{subfigure}{.24\textwidth}
\centering
\includegraphics[width=\textwidth]{figures/t5/uncertainty-sub-population-uncertainty-method}
\caption{\scriptsize{Calibration, SUB.}}
\label{fig:t5-rank-sub-calibration-size}
\end{subfigure}
\hfill
\begin{subfigure}{.24\textwidth}
\centering
\includegraphics[width=\textwidth]{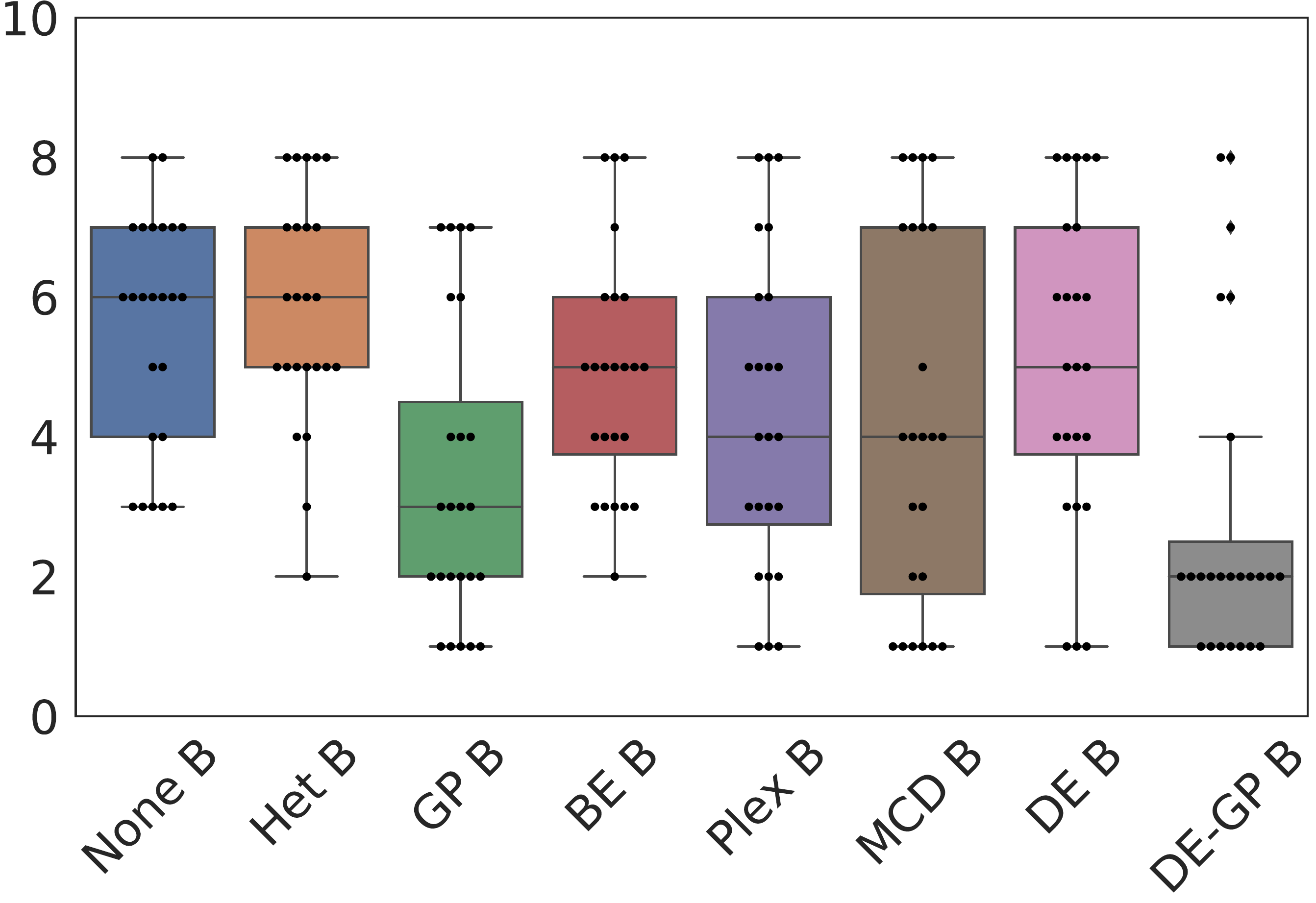}
\caption{\scriptsize{Calibration, ALL.}}
\label{fig:t5-rank-all-calibration-size}
\end{subfigure}
\begin{subfigure}{.24\textwidth}
\centering
\includegraphics[width=\textwidth]{figures/t5/collaboration-in-domain-uncertainty-method}
\caption{\scriptsize{Selective Pred., IND.}}
\end{subfigure}
\hfill
\begin{subfigure}{.24\textwidth}
\centering
\includegraphics[width=\textwidth]{figures/t5/collaboration-ood-generalization-uncertainty-method}
\caption{\scriptsize{Selective Pred., OOD.}}
\end{subfigure}
\hfill
\begin{subfigure}{.24\textwidth}
\centering
\includegraphics[width=\textwidth]{figures/t5/collaboration-sub-population-uncertainty-method}
\caption{\scriptsize{Selective Pred., SUB.}}
\end{subfigure}
\hfill
\begin{subfigure}{.24\textwidth}
\centering
\includegraphics[width=\textwidth]{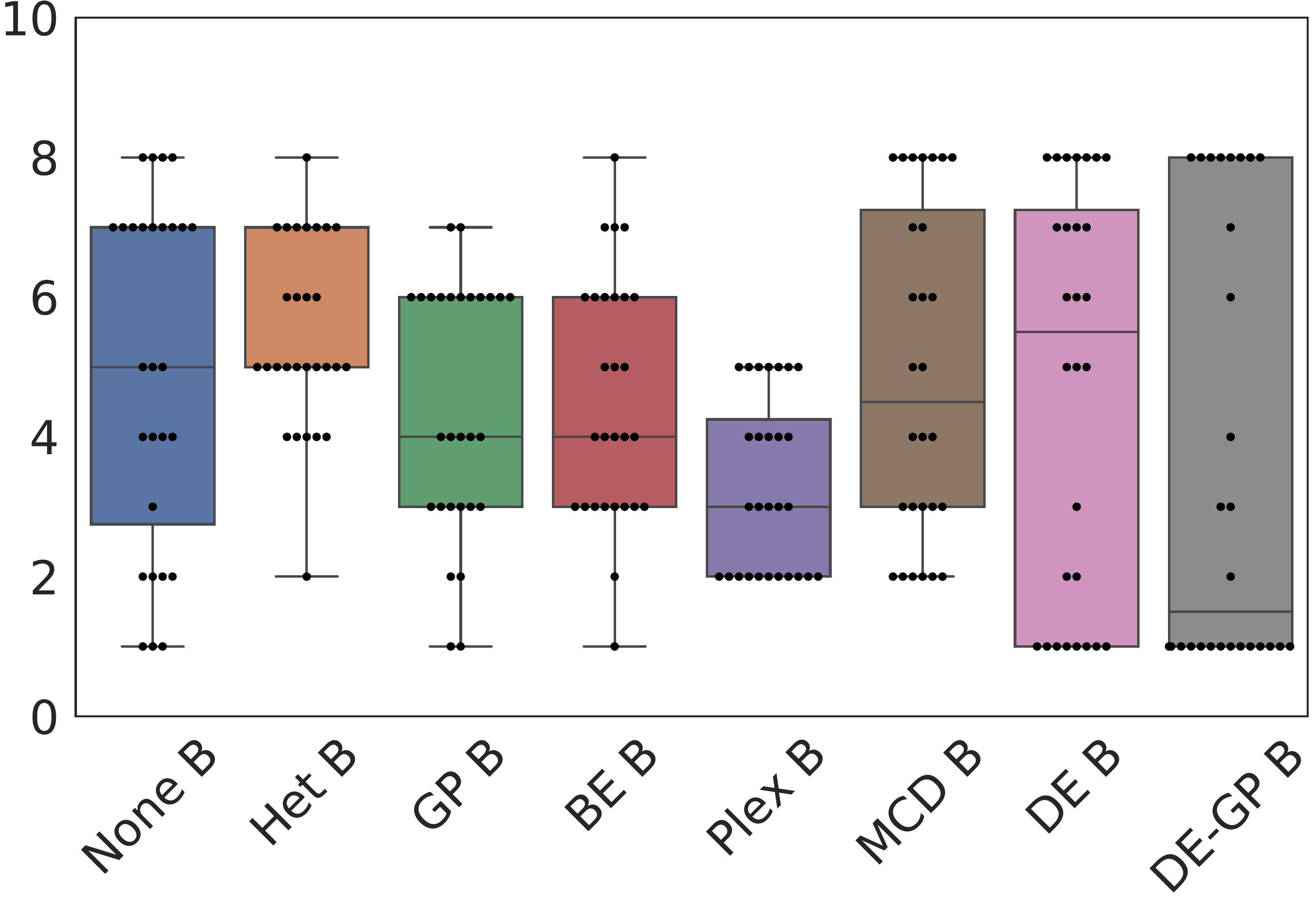}
\caption{\scriptsize{Selective Pred., ALL.}}
\end{subfigure}
\vspace{-1.5ex}
\caption{T5-Plex model's ranking comparison between different uncertainty methods and across different evaluation datasets. IND: in-domain. OOD: out-of-domain. SUB: subpopulation shift. ALL: aggregated performance across all datasets.}
\label{fig:t5-rank-method}
\end{figure*}

We first compare the performance across types of uncertainty methods, fixing the architecture size to T5-base. 
We compare performances in prediction, uncertainty calibration, and human-model collaboration, across all datasets (MNLI, NaLUE and Toxic Comments) and all splits (In-domain, OOD, and tail-population). 
\Cref{fig:t5-arch-compare} reports the full results, and \Cref{fig:t5-rank-method} summarizes the rankings of uncertainty methods under each type of population shift (in-domain v.s. OOD v.s. tail-population). 
Among all methods, DE$+$GP, Plex (i.e., BE$+$GP), BE, and MC Dropout tend to have the strongest performance. In particular, DE$+$GP almost always dominates the other methods on MNLI and NaLUE, and remains competitive in the case of label imbalance (i.e., Toxic Comments). However, DE$+$GP is an expensive method that costs x10 more in memory and compute and therefore is not competitive in scale (a more thorough analysis is in \Cref{sub:ensembles}). On the other hand, among the more efficient, single-model methods, BE and Plex perform well on MNLI and NaLUE (notably, outperform the most expensive DE), while MCD stands out in the Toxic Comments. The above observations suggest that, when the training example has a simple distribution, quantifying output-layer uncertainty alone is sufficient to attain strong performance. However, when there are pathologies in the data distribution (e.g., extreme label imbalance), quantifying the uncertainty within the model's intermediate representations (e.g., via some form of perturbation like BE) becomes important.

\begin{figure*}[!tb]
\begin{subfigure}{.24\textwidth}
\centering
\includegraphics[width=\textwidth]{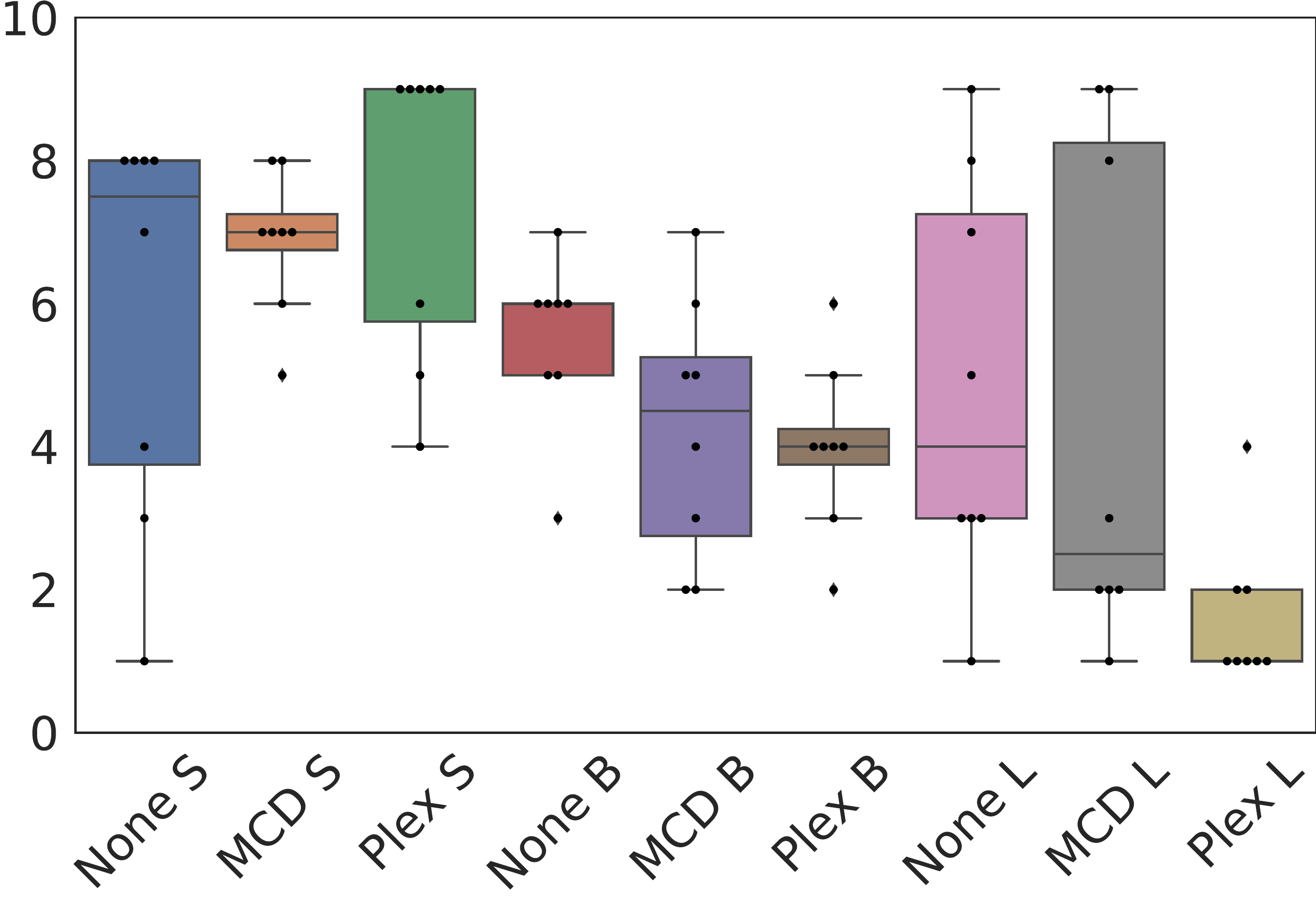}
\caption{\scriptsize{Generalization, IND.}}
\end{subfigure}
\hfill
\begin{subfigure}{.24\textwidth}
\centering
\includegraphics[width=\textwidth]{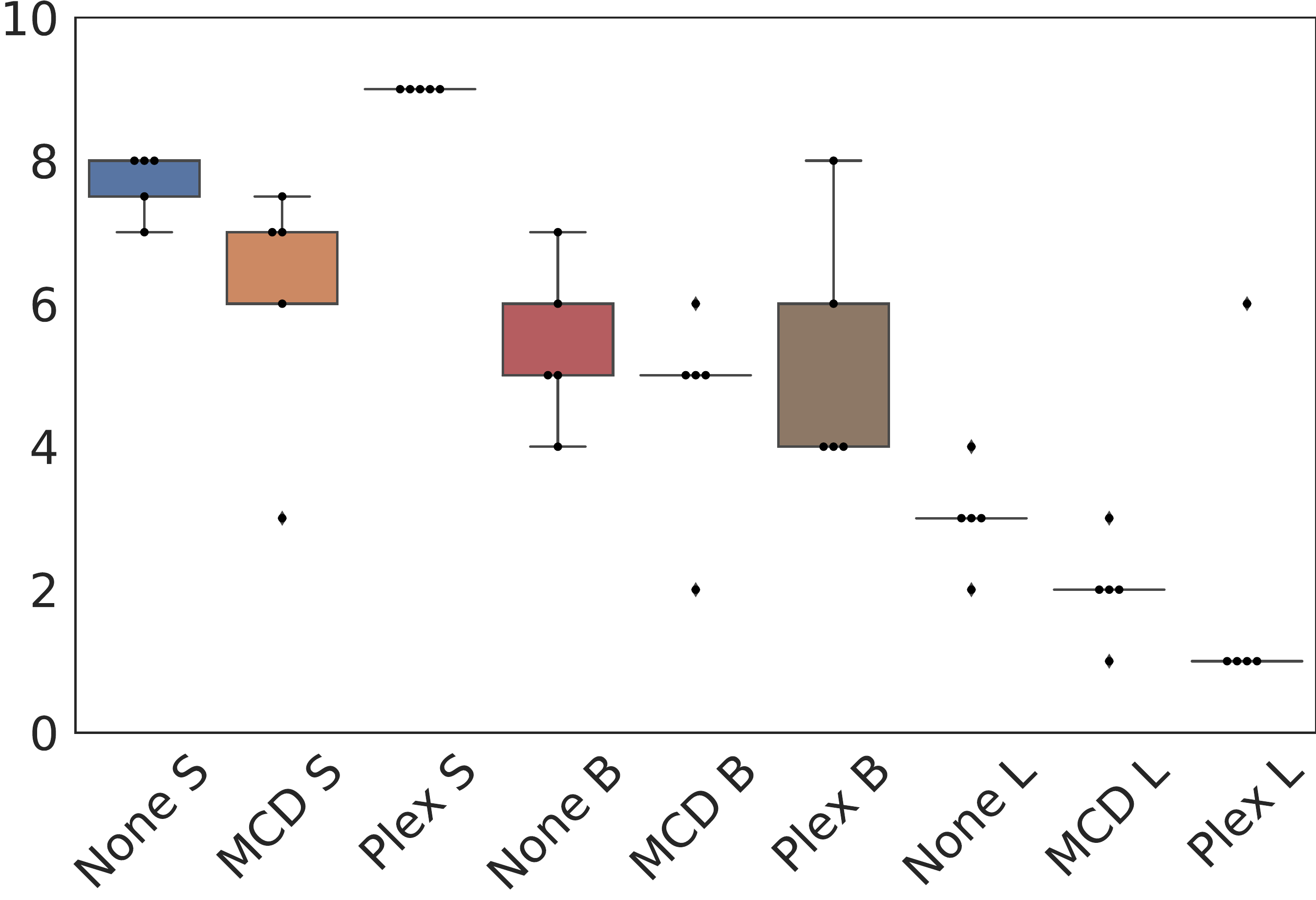}
\caption{\scriptsize{Generalization., OOD.}}
\end{subfigure}
\hfill
\begin{subfigure}{.24\textwidth}
\centering
\includegraphics[width=\textwidth]{figures/t5/generalization-sub-population-architecture-size}
\caption{\scriptsize{Generalization, SUB.}}
\end{subfigure}
\hfill
\begin{subfigure}{.24\textwidth}
\centering
\includegraphics[width=\textwidth]{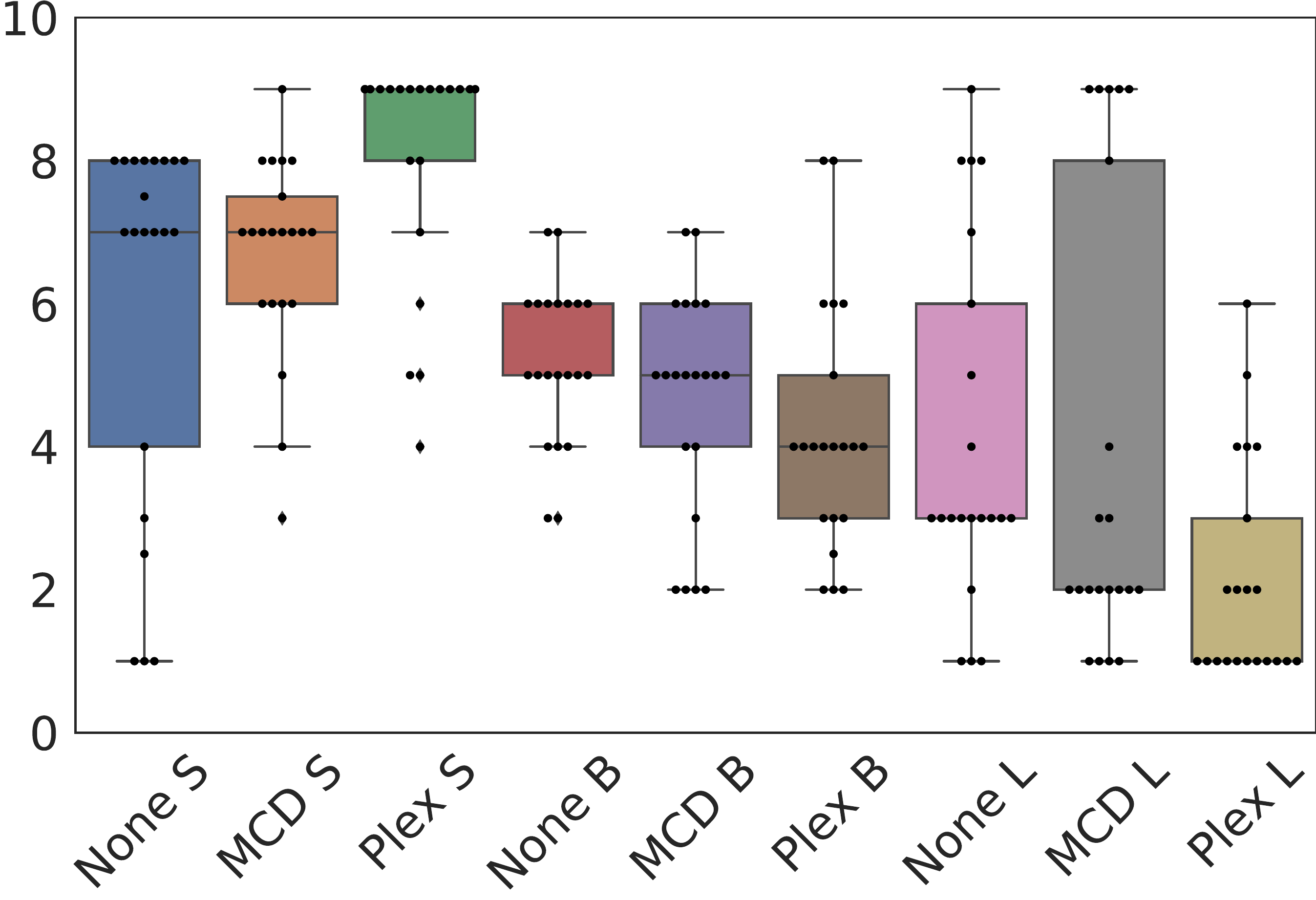}
\caption{\scriptsize{Generalization, ALL.}}
\end{subfigure}
\begin{subfigure}{.24\textwidth}
\centering
\includegraphics[width=\textwidth]{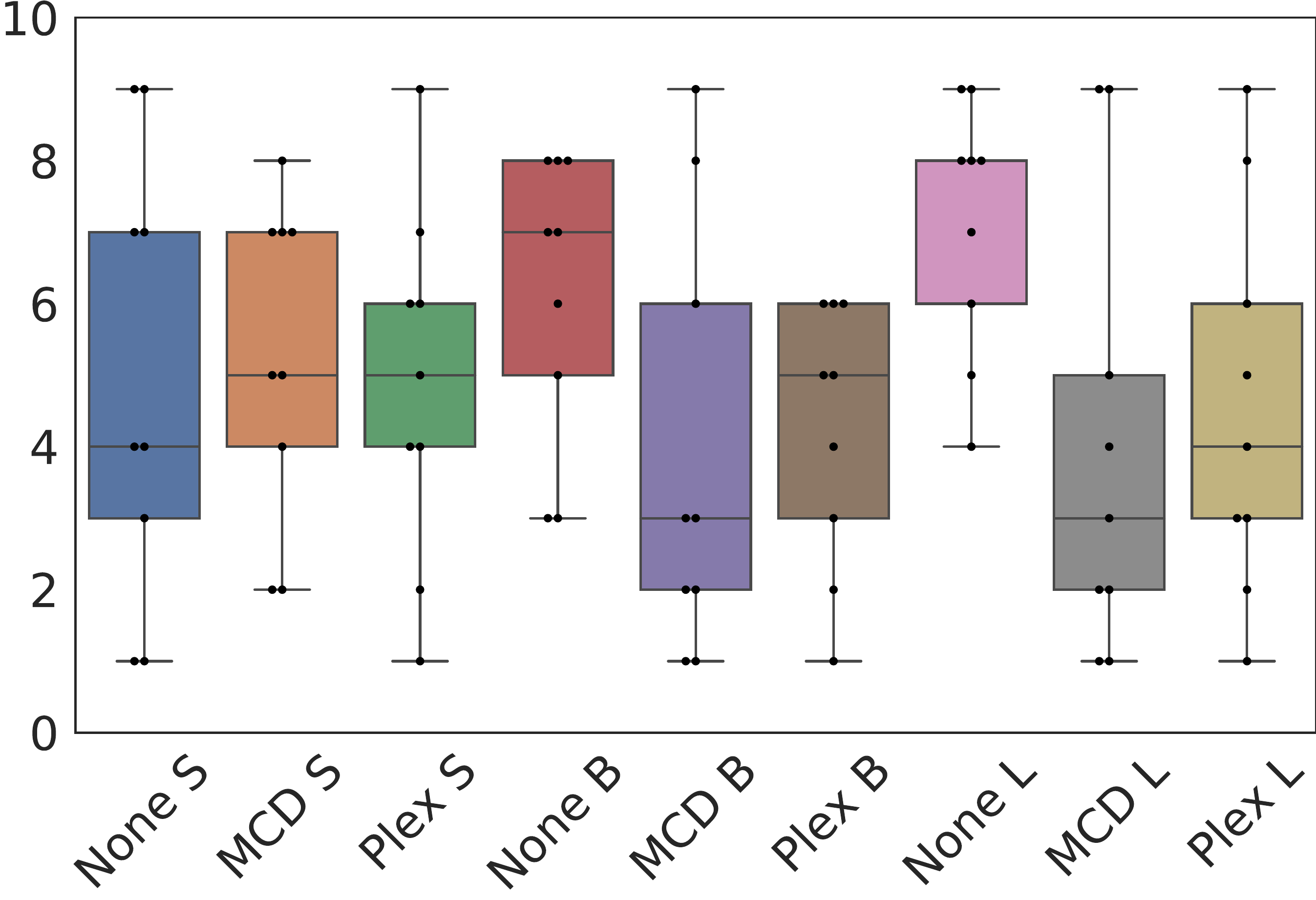}
\caption{\scriptsize{Calibration, IND.}}
\end{subfigure}
\hfill
\begin{subfigure}{.24\textwidth}
\centering
\includegraphics[width=\textwidth]{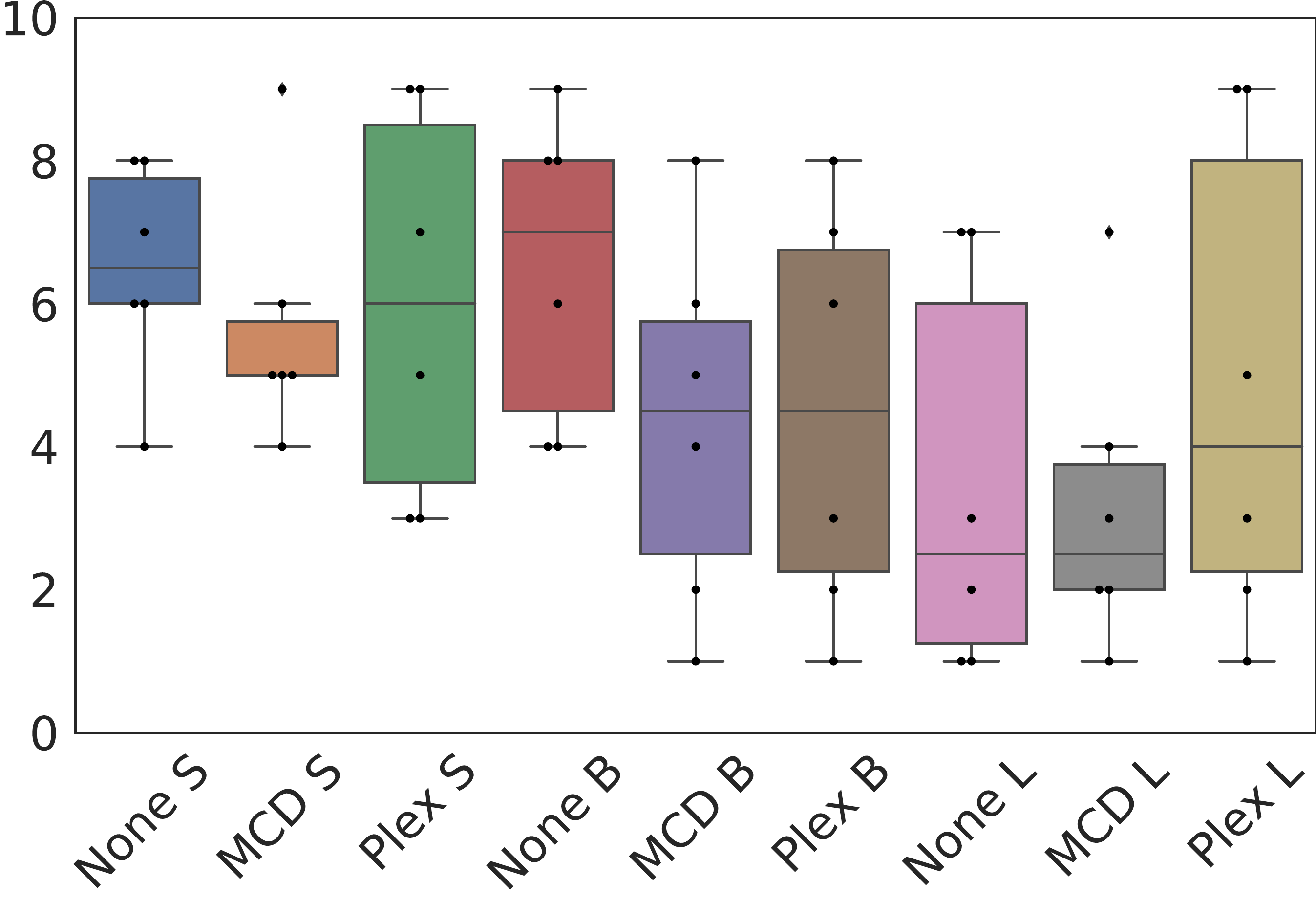}
\caption{\scriptsize{Calibration, OOD.}}
\end{subfigure}
\hfill
\begin{subfigure}{.24\textwidth}
\centering
\includegraphics[width=\textwidth]{figures/t5/uncertainty-sub-population-architecture-size}
\caption{\scriptsize{Calibration, SUB.}}
\end{subfigure}
\hfill
\begin{subfigure}{.24\textwidth}
\centering
\includegraphics[width=\textwidth]{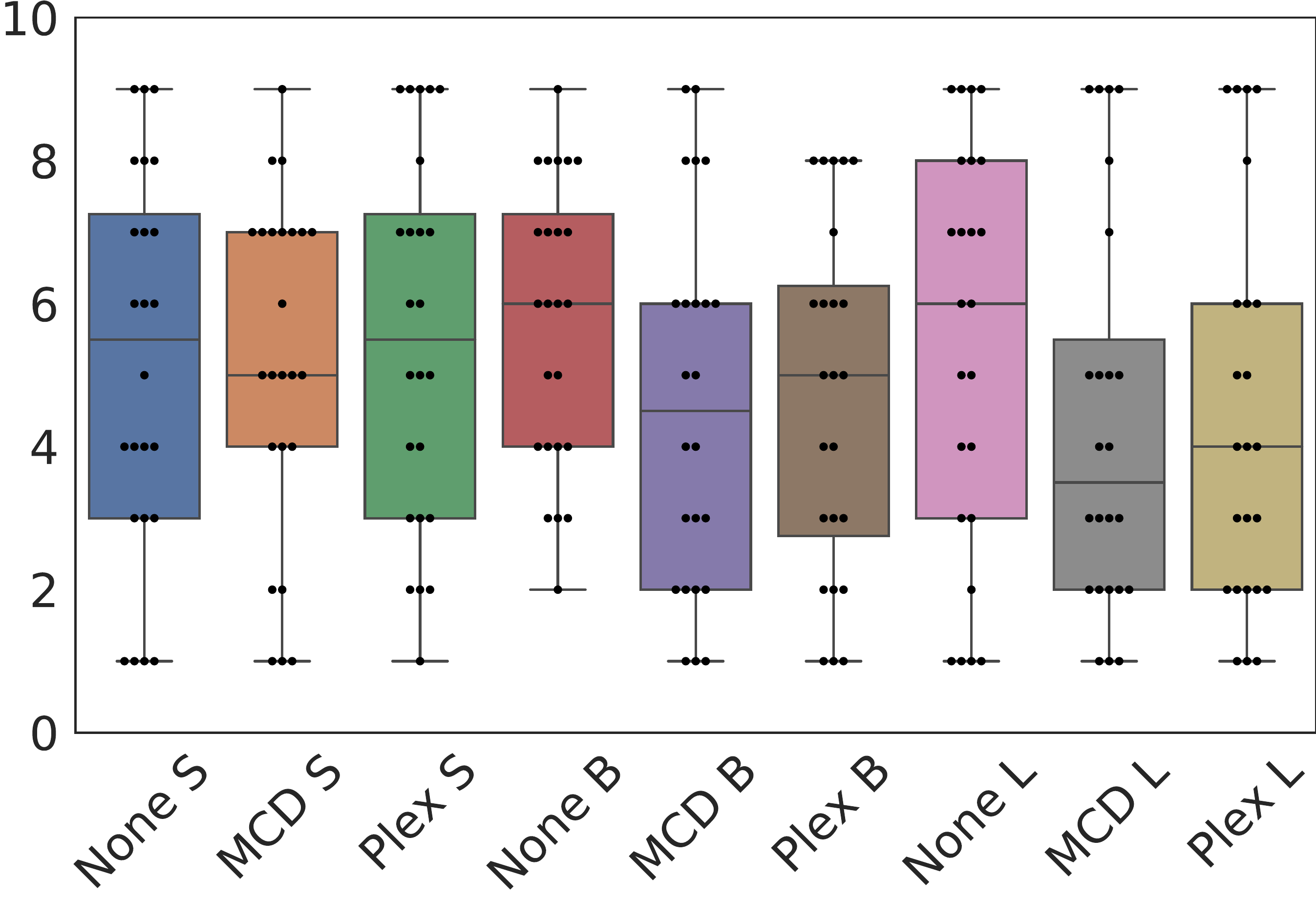}
\caption{\scriptsize{Calibration, ALL.}}
\end{subfigure}
\begin{subfigure}{.24\textwidth}
\centering
\includegraphics[width=\textwidth]{figures/t5/collaboration-in-domain-architecture-size}
\caption{\scriptsize{Selective Pred., IND.}}
\end{subfigure}
\hfill
\begin{subfigure}{.24\textwidth}
\centering
\includegraphics[width=\textwidth]{figures/t5/collaboration-ood-generalization-architecture-size}
\caption{\scriptsize{Selective Pred., OOD.}}
\end{subfigure}
\hfill
\begin{subfigure}{.24\textwidth}
\centering
\includegraphics[width=\textwidth]{figures/t5/collaboration-sub-population-architecture-size}
\caption{\scriptsize{Selective Pred., SUB.}}
\end{subfigure}
\hfill
\begin{subfigure}{.24\textwidth}
\centering
\includegraphics[width=\textwidth]{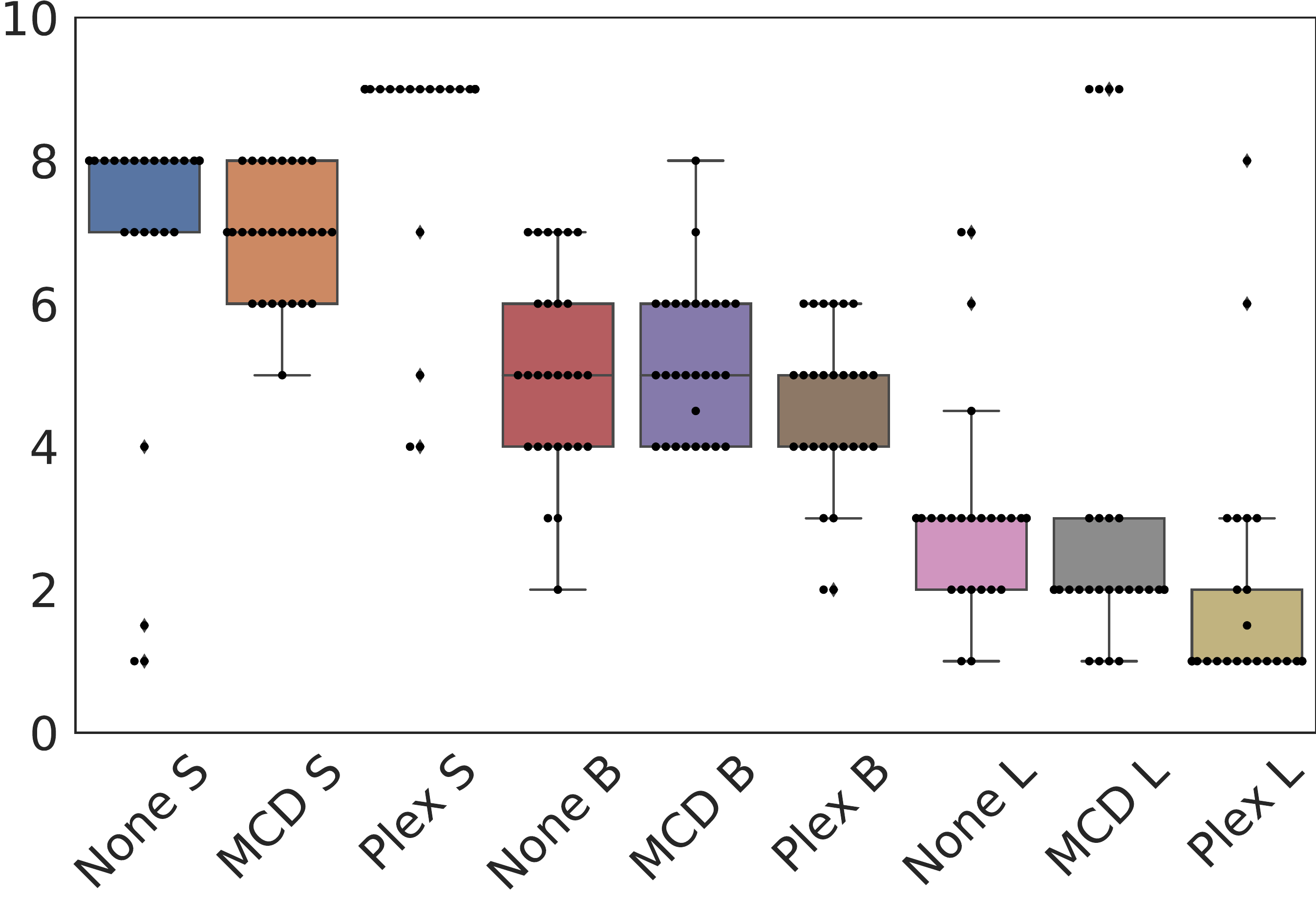}
\caption{\scriptsize{Selective Pred., ALL.}}
\end{subfigure}
\vspace{-1.5ex}
\caption{T5-Plex model's ranking comparison between architecture sizes and across different evaluation datasets. IND: in-domain. OOD: out-of-domain. SUB: subpopulation shift. ALL: aggregated performance across all datasets.}
\label{fig:t5-rank-arch}
\end{figure*}

We then investigate how a model's uncertainty performance is impacted by the architecture size. 
For model size scaling, we evaluate Plex, None, and MC Dropout, the three best-performing and efficient methods in the previous study. We evaluate the performance of each method under three progressively larger architectures: T5 S, T5 B, and T5 L, and observe how the behavior changes across the method and with respect to the architecture size.
\Cref{fig:t5-arch-compare} reports the full results, and \Cref{fig:t5-rank-arch} summarizes the rankings of uncertainty methods organized by the sizes of the architecture. 
As shown, comparing across architecture sizes, we see a larger architecture almost always leads to stronger performance in collaborative performance. This trend remains largely consistent even when  out-of-distribution. 



\begin{table}[!tb]
\tiny
\centering
\begin{tabular}{ccc|cccccccc}
\toprule
               &               &               &  None B &  Het B &  GP B &  BE B &  Plex B &  MCD B &  DE B &  DE-GP B \\
Task & Split & Score &         &        &       &       &         &        &       &          \\
\midrule
\multirow{9}{*}{MNLI}  &  \multirow{3}{*}{In-domain}  &  calibration  & 0.381 & 0.384 & 0.372 & \cellcolor{black!50}{0.401} & 0.388 & \cellcolor{black!90}{\color{white} 0.416} & 0.383 & \cellcolor{black!10}{0.4}\\
                &                 &  generalization  & 0.938 & \cellcolor{black!50}{0.949} & 0.944 & \cellcolor{black!50}{0.949} & 0.948 & 0.946 & 0.938 & \cellcolor{black!90}{\color{white} 0.95}\\
                &                 &  select. pred.  & 0.961 & 0.971 & 0.968 & \cellcolor{black!50}{0.973} & \cellcolor{black!50}{0.973} & \cellcolor{black!50}{0.973} & 0.961 & \cellcolor{black!90}{\color{white} 0.975}\\
\cline{2-11}
                &  \multirow{3}{*}{OOD}  &  calibration  & 0.391 & 0.401 & 0.393 & \cellcolor{black!50}{0.413} & 0.394 & \cellcolor{black!10}{0.41} & 0.388 & \cellcolor{black!90}{\color{white} 0.416}\\
                &                 &  generalization  & 0.937 & \cellcolor{black!10}{0.948} & 0.941 & \cellcolor{black!50}{0.949} & \cellcolor{black!10}{0.948} & 0.946 & 0.938 & \cellcolor{black!90}{\color{white} 0.95}\\
                &                 &  select. pred.  & 0.959 & 0.971 & 0.966 & \cellcolor{black!50}{0.973} & \cellcolor{black!50}{0.973} & 0.972 & 0.96 & \cellcolor{black!90}{\color{white} 0.975}\\
\cline{2-11}
                &  \multirow{3}{*}{Subpopulation}  &  calibration  & \cellcolor{black!10}{0.451} & 0.434 & \cellcolor{black!50}{0.454} & 0.443 & 0.401 & \cellcolor{black!90}{\color{white} 0.474} & 0.443 & 0.418\\
                &                 &  generalization  & 0.749 & 0.766 & 0.762 & \cellcolor{black!50}{0.791} & \cellcolor{black!10}{0.788} & 0.739 & 0.764 & \cellcolor{black!90}{\color{white} 0.798}\\
                &                 &  select. pred.  & 0.811 & 0.824 & 0.842 & \cellcolor{black!10}{0.87} & \cellcolor{black!50}{0.871} & 0.831 & 0.826 & \cellcolor{black!90}{\color{white} 0.885}\\
\cline{1-11}
\cline{2-11}
\multirow{8}{*}{NaLUE}  &  \multirow{3}{*}{In-domain}  &  calibration  & \cellcolor{black!50}{0.498} & 0.484 & \cellcolor{black!90}{\color{white} 0.512} & 0.464 & 0.486 & 0.471 & 0.487 & \cellcolor{black!10}{0.494}\\
                &                 &  generalization  & 0.939 & 0.932 & \cellcolor{black!10}{0.94} & 0.935 & \cellcolor{black!10}{0.94} & 0.938 & \cellcolor{black!50}{0.942} & \cellcolor{black!90}{\color{white} 0.944}\\
                &                 &  select. pred.  & 0.936 & 0.935 & 0.932 & \cellcolor{black!90}{\color{white} 0.938} & \cellcolor{black!90}{\color{white} 0.938} & 0.932 & 0.937 & \cellcolor{black!90}{\color{white} 0.938}\\
\cline{2-11}
                &  OOS, Near  &  detection  & 0.706 & 0.673 & 0.719 & \cellcolor{black!50}{0.768} & 0.716 & \cellcolor{black!10}{0.766} & 0.721 & \cellcolor{black!90}{\color{white} 0.771}\\
                &  OOS, Standard  &  detection  & 0.964 & 0.957 & 0.992 & \cellcolor{black!90}{\color{white} 0.998} & 0.991 & \cellcolor{black!10}{0.994} & 0.973 & \cellcolor{black!90}{\color{white} 0.998}\\ \cline{2-11}
                &  \multirow{3}{*}{Subpopulation}  &  calibration  & \cellcolor{black!10}{0.518} & 0.511 & \cellcolor{black!90}{\color{white} 0.553} & 0.514 & \cellcolor{black!50}{0.519} & 0.466 & 0.513 & 0.514\\
                &                 &  generalization  & 0.866 & 0.846 & 0.869 & 0.87 & \cellcolor{black!50}{0.873} & 0.858 & \cellcolor{black!10}{0.871} & \cellcolor{black!90}{\color{white} 0.882}\\
                &                 &  select. pred.  & \cellcolor{black!90}{\color{white} 0.862} & 0.82 & 0.828 & 0.845 & \cellcolor{black!10}{0.856} & 0.829 & \cellcolor{black!50}{0.861} & 0.851\\
\cline{1-11}
\cline{2-11}
\multirow{9}{*}{\shortstack{Toxic\\Comments}}  &  \multirow{3}{*}{In-domain}  &  calibration  & 0.459 & \cellcolor{black!10}{0.462} & 0.46 & 0.461 & \cellcolor{black!90}{\color{white} 0.471} & 0.442 & 0.459 & \cellcolor{black!50}{0.465}\\
                &                 &  generalization  & 0.888 & 0.89 & \cellcolor{black!50}{0.899} & 0.889 & \cellcolor{black!10}{0.895} & \cellcolor{black!90}{\color{white} 0.904} & 0.885 & 0.892\\
                &                 &  select. pred.  & 0.936 & 0.938 & \cellcolor{black!50}{0.94} & 0.938 & \cellcolor{black!50}{0.94} & \cellcolor{black!90}{\color{white} 0.941} & 0.936 & 0.939\\
\cline{2-11}
                &  \multirow{3}{*}{OOD}  &  calibration  & 0.425 & \cellcolor{black!10}{0.427} & \cellcolor{black!50}{0.438} & 0.423 & \cellcolor{black!90}{\color{white} 0.447} & 0.413 & 0.426 & 0.421\\
                &                 &  generalization  & \cellcolor{black!50}{0.82} & 0.817 & \cellcolor{black!10}{0.818} & 0.81 & 0.816 & \cellcolor{black!90}{\color{white} 0.831} & 0.817 & \cellcolor{black!10}{0.818}\\
                &                 &  select. pred.  & \cellcolor{black!50}{0.86} & 0.857 & 0.855 & 0.85 & 0.855 & \cellcolor{black!90}{\color{white} 0.862} & \cellcolor{black!50}{0.86} & 0.852\\
\cline{2-11}
                &  \multirow{3}{*}{Subpopulation}  &  calibration  & 0.415 & 0.405 & \cellcolor{black!50}{0.421} & 0.412 & \cellcolor{black!90}{\color{white} 0.428} & 0.405 & \cellcolor{black!10}{0.416} & 0.4\\
                &                 &  generalization  & \cellcolor{black!50}{0.806} & 0.803 & \cellcolor{black!10}{0.804} & 0.795 & 0.801 & \cellcolor{black!90}{\color{white} 0.814} & 0.801 & 0.803\\
                &                 &  select. pred.  & \cellcolor{black!50}{0.831} & 0.828 & 0.828 & 0.821 & 0.826 & \cellcolor{black!90}{\color{white} 0.835} & \cellcolor{black!10}{0.83} & 0.823\\
\bottomrule
\end{tabular}
\caption{Comparison of method performance between uncertainty methods. Black: Best. Dark Grey: Second. Light Grey: Third.}
\label{fig:t5-method-compare}
\end{table}

\begin{table}[!tb]
\tiny
\centering
\begin{tabular}{ccc|ccccccccc}
\toprule
               &               &               &  None S &  MCD S &  Plex S &  None B &  MCD B &  Plex B &  None L &  MCD L &  Plex L \\
Task & Split & Score &         &        &         &         &        &         &         &        &         \\
\midrule
\multirow{9}{*}{MNLI}  &  \multirow{3}{*}{In-domain}  &  calibration  & 0.399 & \cellcolor{black!50}{0.406} & 0.364 & 0.381 & \cellcolor{black!90}{\color{white} 0.416} & 0.388 & 0.39 & \cellcolor{black!10}{0.404} & 0.394\\
                &                 &  generalization  & 0.924 & 0.927 & 0.913 & 0.938 & 0.946 & 0.948 & \cellcolor{black!10}{0.963} & \cellcolor{black!50}{0.964} & \cellcolor{black!90}{\color{white} 0.965}\\
                &                 &  select. pred.  & 0.953 & 0.959 & 0.942 & 0.961 & 0.973 & 0.973 & \cellcolor{black!10}{0.982} & \cellcolor{black!90}{\color{white} 0.985} & \cellcolor{black!90}{\color{white} 0.985}\\
\cline{2-12}
                &  \multirow{3}{*}{OOD}  &  calibration  & 0.398 & 0.396 & 0.367 & 0.391 & \cellcolor{black!10}{0.41} & 0.394 & \cellcolor{black!90}{\color{white} 0.418} & \cellcolor{black!50}{0.411} & 0.406\\
                &                 &  generalization  & 0.924 & 0.93 & 0.911 & 0.937 & 0.946 & 0.948 & \cellcolor{black!10}{0.963} & \cellcolor{black!50}{0.965} & \cellcolor{black!90}{\color{white} 0.967}\\
                &                 &  select. pred.  & 0.953 & 0.96 & 0.94 & 0.959 & 0.972 & 0.973 & \cellcolor{black!10}{0.983} & \cellcolor{black!50}{0.986} & \cellcolor{black!90}{\color{white} 0.987}\\
\cline{2-12}
                &  \multirow{3}{*}{Subpopulation}  &  calibration  & \cellcolor{black!50}{0.555} & \cellcolor{black!90}{\color{white} 0.56} & \cellcolor{black!10}{0.514} & 0.451 & 0.474 & 0.401 & 0.417 & 0.45 & 0.447\\
                &                 &  generalization  & 0.648 & 0.619 & 0.59 & 0.749 & 0.739 & 0.788 & \cellcolor{black!90}{\color{white} 0.817} & \cellcolor{black!50}{0.807} & \cellcolor{black!10}{0.803}\\
                &                 &  select. pred.  & 0.747 & 0.717 & 0.677 & 0.811 & 0.831 & 0.871 & \cellcolor{black!10}{0.875} & \cellcolor{black!90}{\color{white} 0.896} & \cellcolor{black!50}{0.876}\\
\cline{1-12}
\cline{2-12}
\multirow{8}{*}{NaLUE}  &  \multirow{3}{*}{In-domain}  &  calibration  & \cellcolor{black!90}{\color{white} 0.507} & 0.48 & \cellcolor{black!50}{0.498} & \cellcolor{black!50}{0.498} & 0.471 & 0.486 & 0.486 & 0.453 & 0.496\\
                &                 &  generalization  & \cellcolor{black!50}{0.942} & 0.932 & 0.937 & 0.939 & 0.938 & \cellcolor{black!10}{0.94} & 0.931 & 0.928 & \cellcolor{black!90}{\color{white} 0.944}\\
                &                 &  select. pred.  & \cellcolor{black!50}{0.937} & 0.926 & 0.929 & \cellcolor{black!10}{0.936} & 0.932 & \cellcolor{black!90}{\color{white} 0.938} & 0.93 & 0.922 & 0.935\\
\cline{2-12}
                &  OOS, Near  &  detection  & 0.71 & \cellcolor{black!10}{0.756} & 0.689 & 0.706 & \cellcolor{black!50}{0.766} & 0.716 & 0.692 & 0.733 & \cellcolor{black!90}{\color{white} 0.781}\\
                &  OOS, Standard  &  detection  & 0.968 & \cellcolor{black!10}{0.992} & \cellcolor{black!90}{\color{white} 0.999} & 0.964 & \cellcolor{black!50}{0.994} & 0.991 & 0.956 & 0.991 & 0.991\\ \cline{2-12}
                &  \multirow{3}{*}{Subpopulation}  &  calibration  & \cellcolor{black!90}{\color{white} 0.528} & 0.462 & 0.499 & \cellcolor{black!10}{0.518} & 0.466 & \cellcolor{black!50}{0.519} & \cellcolor{black!10}{0.518} & 0.466 & 0.492\\
                &                 &  generalization  & \cellcolor{black!90}{\color{white} 0.878} & 0.854 & 0.851 & 0.866 & 0.858 & \cellcolor{black!50}{0.873} & 0.843 & 0.83 & \cellcolor{black!10}{0.871}\\
                &                 &  select. pred.  & \cellcolor{black!90}{\color{white} 0.864} & 0.836 & 0.84 & \cellcolor{black!50}{0.862} & 0.829 & \cellcolor{black!10}{0.856} & 0.835 & 0.801 & 0.835\\
\cline{1-12}
\cline{2-12}
\multirow{9}{*}{\shortstack{Toxic\\Comments}}  &  \multirow{3}{*}{In-domain}  &  calibration  & 0.455 & 0.445 & \cellcolor{black!90}{\color{white} 0.478} & \cellcolor{black!10}{0.459} & 0.442 & \cellcolor{black!50}{0.471} & 0.448 & 0.451 & 0.436\\
                &                 &  generalization  & 0.879 & \cellcolor{black!10}{0.898} & 0.863 & 0.888 & \cellcolor{black!50}{0.904} & 0.895 & 0.886 & \cellcolor{black!90}{\color{white} 0.906} & 0.89\\
                &                 &  select. pred.  & 0.932 & 0.938 & 0.919 & 0.936 & \cellcolor{black!10}{0.941} & 0.94 & 0.936 & \cellcolor{black!90}{\color{white} 0.944} & \cellcolor{black!50}{0.942}\\
\cline{2-12}
                &  \multirow{3}{*}{OOD}  &  calibration  & 0.423 & 0.412 & 0.425 & 0.425 & 0.413 & \cellcolor{black!50}{0.447} & \cellcolor{black!10}{0.432} & 0.417 & \cellcolor{black!90}{\color{white} 0.459}\\
                &                 &  generalization  & 0.81 & \cellcolor{black!10}{0.823} & 0.807 & 0.82 & \cellcolor{black!50}{0.831} & 0.816 & \cellcolor{black!10}{0.823} & \cellcolor{black!90}{\color{white} 0.837} & 0.816\\
                &                 &  select. pred.  & 0.85 & 0.851 & 0.838 & 0.86 & 0.862 & 0.855 & \cellcolor{black!50}{0.865} & \cellcolor{black!90}{\color{white} 0.869} & \cellcolor{black!10}{0.863}\\
\cline{2-12}
                &  \multirow{3}{*}{Subpopulation}  &  calibration  & 0.409 & 0.404 & 0.412 & 0.415 & 0.405 & \cellcolor{black!50}{0.428} & \cellcolor{black!10}{0.426} & 0.403 & \cellcolor{black!90}{\color{white} 0.46}\\
                &                 &  generalization  & 0.795 & 0.806 & 0.786 & 0.806 & \cellcolor{black!50}{0.814} & 0.801 & \cellcolor{black!10}{0.809} & \cellcolor{black!90}{\color{white} 0.822} & 0.805\\
                &                 &  select. pred.  & 0.82 & 0.824 & 0.803 & 0.831 & 0.835 & 0.826 & \cellcolor{black!10}{0.837} & \cellcolor{black!90}{\color{white} 0.842} & \cellcolor{black!50}{0.838}\\
\bottomrule
\end{tabular}
\caption{Comparison of method performance between architecture sizes. Black: Best. Dark Grey: Second. Light Grey: Third.}
\label{fig:t5-arch-compare}
\end{table}



\end{document}

%% file: math_commands.tex
\usepackage{amsmath,amsfonts,bm}

\def\eqref#1{equation~\ref{#1}}

\def\1{\bm{1}}

\def\vmu{{\bm{\mu}}}

\def\vb{{\bm{b}}}

\def\vx{{\bm{x}}}

\def\vz{{\bm{z}}}

\def\evbeta{{\beta}}

\def\mW{{\bm{W}}}

\DeclareMathAlphabet{\mathsfit}{\encodingdefault}{\sfdefault}{m}{sl}
\SetMathAlphabet{\mathsfit}{bold}{\encodingdefault}{\sfdefault}{bx}{n}

\def\vSigma{{\bm{\Sigma}}}
\newcommand{\gaussk}{\mathcal{N}(\mathbf{\vmu}_k, \vSigma)}

\newcommand{\logit}{\mathrm{logit}}
\newcommand{\var}{\mathrm{var}}

\usepackage{todonotes}
\usepackage[nameinlink]{cleveref}
\creflabelformat{equation}{#2\textup{#1}#3}  

\usepackage{framed}
\newenvironment{tldr}[1][]
{
\begin{framed}
\noindent
{\footnotesize\textbf{TL;DR}}{\small~#1}
}
{ 
\end{framed}
}

\makeatletter
\def\blfootnote{\xdef\@thefnmark{}\@footnotetext}
\makeatother

%% file: active_learning.tex
\subsubsection{Active Learning}
\label{sub:active-learning}

\begin{tldr}
Large pretrained models are good active learners. Plex not only provides a significant boost in initial performance, but it also finds examples in order to adapt and improve at a faster rate than active learning models without pretraining.
\end{tldr}

The goal in active learning (AL) \citep{cohn1996active,settles2009active} is to maximize label efficiency for machine learning models where label annotations are scarce. The success of AL has been demonstrated on a range of real-world problems where labels are expensive to acquire, e.g. computer vision \citep{gal2017deep,citovsky2021batch},
natural language \citep{thompson1999active, siddhant2018deep}, 
speech \citep{hakkani2002active, riccardi2005active}
and robotics \citep{martinez2007active, wang2018active}. 
Here, we investigate Plex for AL on the image domain and focus on four datasets: CIFAR-10 and CIFAR-100 which are standard in active learning research \citep{tran2019bayesian,hu2018active,song2019combining}; 
and we scale active learning to a larger dataset, ImageNet, which is less commonly evaluated \citep{emam2021active, beluch2018power}. 
Finally, we evaluate active learning on Places365 (roughly 1.8 million examples in the training pool) which is also less explored.

We adopt a standard setup of AL for multi-class single-label image datasets (\Cref{ssec:datasets}), where we assume an initial model, a training pool of unlabeled images, and a budget of total number of labels to acquire. AL operates in a loop where in each round, labels are acquired for examples with the highest acquisition score from the training pool, the model is subsequently finetuned on the image-label pairs observed so far, and the training pool is updated to remove the newly labeled images. Once we exhaust the budget on label acquisition, the cycle of AL stops and the final model is obtained.




For the label acquisition strategy, we use \emph{margin sampling} (Margin) as a representative AL approach \citep{scheffer2001active, roth2006margin}, and it has been found competitive on AL tasks with deep learning models \citep{citovsky2021batch}. Margin uses the difference between the highest and second highest predicted probabilities \citep{scheffer2001active} to score informativeness: the smaller the difference, the more uncertain the model is about an example and the more informative it is expected to be.
As a baseline, we compare to uniformly randomly sampling from the training pool (Uniform). 

To better simulate realistic label acquisition settings with parallel annotators, we adopt the batch active learning setup \citep{settles2009active} for both Margin and Uniform. For Margin, the examples with the top-$K$ margin scores are chosen for each acquisition round. For Uniform, we randomly select $K$ examples without replacement.


\begin{figure}[!tb]
    \centering
    \begin{subfigure}{.49\textwidth}
        \centering
        \includegraphics[width=\textwidth]{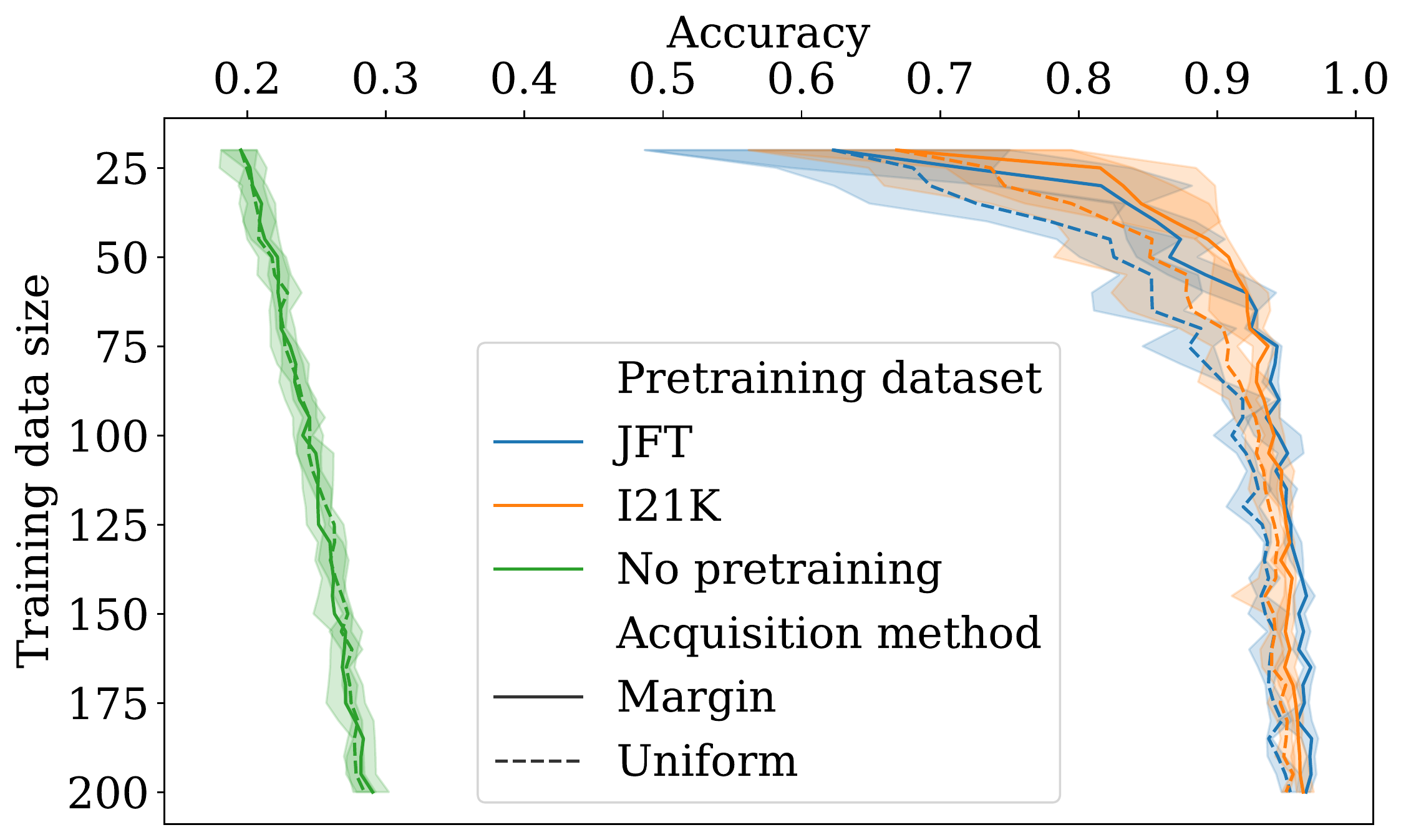}
        \caption{CIFAR-10}
    \end{subfigure}
    \begin{subfigure}{.49\textwidth}
        \centering
        \includegraphics[width=\textwidth]{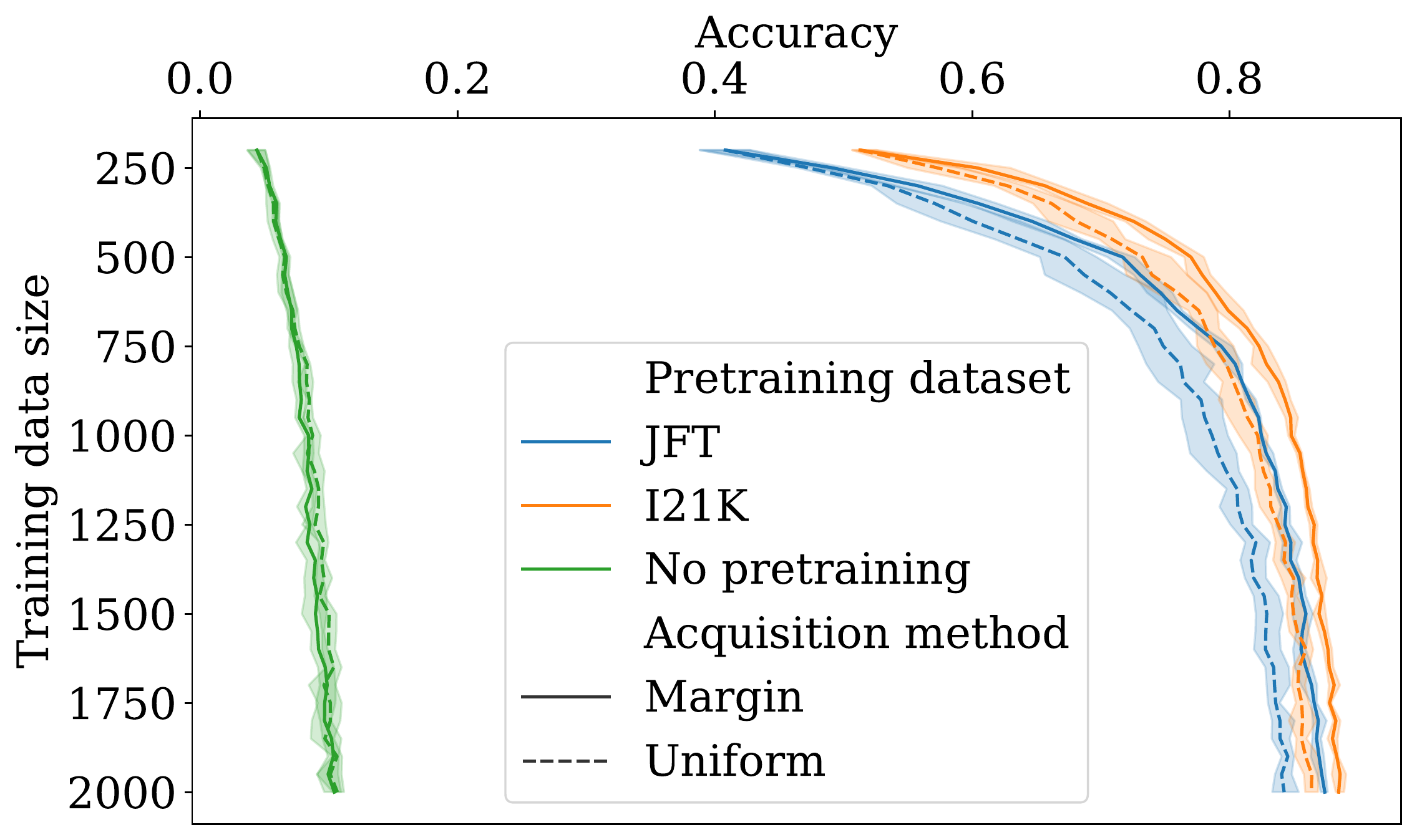}
        \caption{CIFAR-100}
    \end{subfigure}
    \begin{subfigure}{.49\textwidth}
        \centering
        \includegraphics[width=\textwidth]{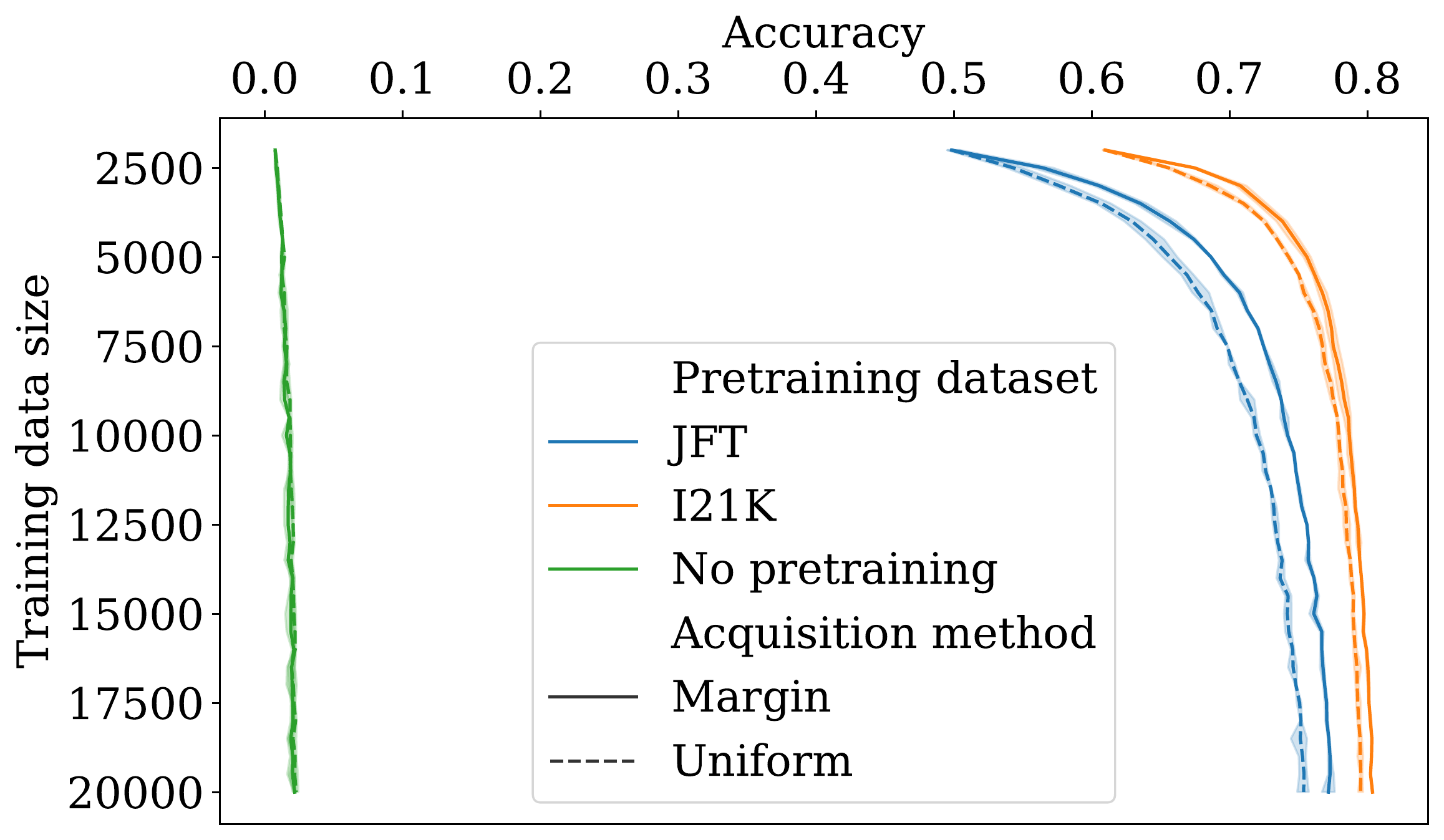}
        \caption{ImageNet}
    \end{subfigure}
    \begin{subfigure}{.49\textwidth}
        \centering
        \includegraphics[width=\textwidth]{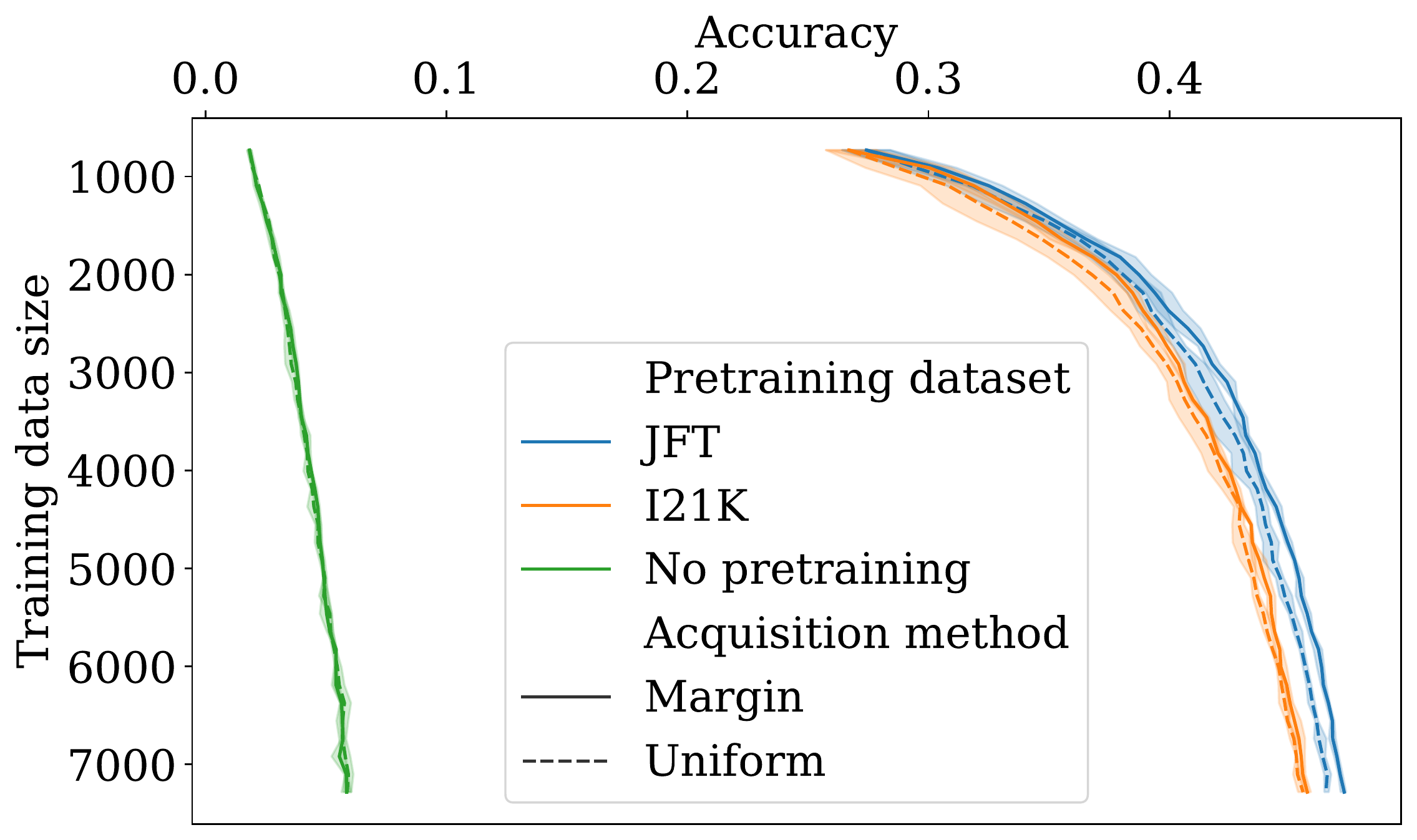}
        \caption{Places-365}
    \end{subfigure}
    \vspace{-1ex}
    \caption{
        Active learning on (a) CIFAR-10, (b) CIFAR-100, (c) ImageNet2012 and (d) Places365.
        Plex’s large scale pretraining and uncertainty estimates enables active learning on large datasets as well as fast adaptation over the number of labeled images. With Plex models pretrained on JFT, Margin requires roughly 30\% less labeled data than Uniform in order to reach the best accuracy of Uniform; and with Plex models pretrained on I21K, Margin is roughly 20\% more label efficient than Uniform.
    }
    \label{fig:active-learning}
\end{figure}

\Cref{fig:active-learning} displays the AL results on CIFAR-10, CIFAR-100, ImageNet and Places365. We compare Margin and Uniform with Plex models pretrained on ImageNet21K or JFT, or without any pretraining, i.e. randomly initialized Plex models. We set initial data size to be $2\times$ number of classes, max training set size to be $20\times$ number of classes, and acquisition batch size to be $0.5\times$ number of classes.

\Cref{fig:active-learning} shows that AL with pretrained models significantly outperform models without pretraining across all tasks. We observed two notable effects. First, pretraining has a significant initial boost in performance, where for example Plex starts at 20\% for CIFAR-10 without pretraining and 50-60\% with pretraining. Second, pretraining results in faster adaptation in terms of the accuracy gain for each labelled example: for example, it takes roughly 200 examples on CIFAR-10 to go from 20-30\% whereas it takes roughly 10 examples for pretrained Plex to go from 50-60\%.

Pretrained models also enables better performance from better AL acquisition methods. Without pretraining, we observe no gain in performance using Margin comparing to Uniform. However, with pretrained models, Margin almost always outperforms Uniform. Notably, for Plex models pretrained on JFT, Margin requires 37.5\%, 31.6\%, 35.9\% and 10.3\% less labeled data than Uniform respectively on CIFAR-10, CIFAR-100, ImageNet and Places365 in order to achieve the best accuracy obtained by Uniform. 


For CIFAR-100 and ImageNet2012, AL with models pretrained on I21K outperforms that on JFT. We hypothesize that CIFAR-100 and ImageNet are more ``in-distribution'' in I21K than JFT (\Cref{sub:few-shot-learning} also shows this pattern). However, Places365 is an OOD dataset for both I21K and JFT, where JFT is much larger than I21K. Likely due to better representations pretrained on a larger and more diverse dataset, AL with Plex models pretrained on JFT performs better than I21K. On a related note, \citet{evci2022head2toe} observed how domain shift can impact finetuning and hypothesized that the ability to leverage pretrained representations contributes to the effectiveness of finetuning; and \citet{tamkin2022active} observed related findings for pretrained models focused on task ambiguity.